\documentclass[oneside,11pt]{article}

\usepackage{url,mathrsfs,algorithm}
\usepackage{amsmath,wrapfig,color}
\usepackage{amsfonts}
\usepackage{graphicx}
\usepackage{subfigure}
\newtheorem{theorem}{Theorem}

\begin{document}
\title{\huge Linearized GMM Kernels and Normalized Random Fourier Features}

\author{ \bf{Ping Li} \\
         Department of Statistics and Biostatistics\\
         Department of Computer Science\\
       Rutgers University\\
          Piscataway, NJ 08854, USA\\
       \texttt{pingli@stat.rutgers.edu}\\
}

\date{}

\maketitle

\begin{abstract}

\noindent The method of ``random Fourier features (RFF)'' has become a popular tool for approximating the ``radial basis function (RBF)'' kernel. The variance of RFF is actually large. Interestingly, the variance can be substantially reduced by a simple normalization step as we theoretically demonstrate. We name the improved scheme as the ``normalized RFF (NRFF)'', and we provide a technical proof of the theoretical variance of NRFF, as validated by simulations.

\vspace{0.08in}

\noindent We also propose the ``generalized min-max (GMM)'' kernel as a measure of data similarity, where  data vectors can have both   positive and  negative entries. GMM is  positive definite  as there is an associated hashing method named ``generalized consistent weighted sampling (GCWS)'' which linearizes this nonlinear  kernel.  We provide an extensive empirical evaluation of the RBF kernel  and the  GMM kernel on more than 50 publicly available datasets. For a majority of the datasets in our experiments, the (tuning-free) GMM kernel outperforms the best-tuned RBF kernel.

\vspace{0.08in}

\noindent We  conduct extensive experiments for comparing the linearized RBF kernel using NRFF hashing with the linearized GMM kernel using GCWS hashing. We observe that, to reach a comparable classification accuracy, GCWS typically requires substantially fewer samples than NRFF, even on datasets where the original RBF kernel outperforms the original GMM kernel. As the costs of training, storage, transmission, and processing  are  proportional to the sample size, our experiments  demonstrate that GCWS would be a  more practical scheme for large-scale  learning.

\vspace{0.08in}

\noindent The empirical success of GCWS (compared to NRFF) can also be explained from a theoretical perspective. Firstly, the relative variance (normalized by the squared expectation) of GCWS is substantially smaller than that of NRFF, except for the very high similarity region (where the variances of both methods are close to zero). Secondly, if we make a gentle model assumption on the data, we can show analytically that GCWS exhibits much smaller variance than NRFF for estimating the same object (e.g., the RBF kernel), except for the very high similarity region.

\vspace{0.08in}

\noindent Inspired by this work, \cite{Report:Li_epGMM17} developed ``tunable GMM kernels'' which in many datasets considerably improve the (tuning-free) GMM kernel. In fact,  kernel SVMs with tunable GMM kernels can be  strong competitors to deep nets and boosted trees. \cite{Report:Li_GInt17} compared GMM with the normalized GMM kernel and the intersection kernel. \cite{Report:Li_GMM_Nys16} reported the experiments for linearizing GMM with the Nystrom method. \cite{Report:Li_GMM_Theory16} developed a theoretical framework for analyzing the convergence property of the GMM kernel using classical statistics, by making  model assumptions.

\vspace{0.08in}

\noindent We expect that GMM and GCWS (and their variants) will be adopted in practice for large-scale statistical  learning  and efficient near neighbor search (as GCWS generates discrete hash values).

\end{abstract}

\section{Introduction}

It is  popular in machine learning practice to use linear   algorithms such as  logistic regression or linear SVM. It is  known that one can often improve the performance of linear methods by using nonlinear algorithms such as kernel SVMs, if the computational/storage burden can be resolved. In this paper, we introduce an effective  measure of data similarity termed ``generalized min-max (GMM)'' kernel and the associated hashing method named ``generalized consistent weighted sampling (GCWS)'', which efficiently converts this nonlinear kernel into linear kernel. Moreover, we will also introduce what we call ``normalized random Fourier features (NRFF)'' and compare it with GCWS.

\vspace{0.05in}

We start the introduction with the basic linear kernel. Consider two data vectors $u,v\in\mathbb{R}^D$. It is common to use the  normalized linear kernel (i.e., the correlation):
\begin{align}\label{eqn_rho}
\rho = \rho(u,v) = \frac{\sum_{i=1}^D u_i v_i }{\sqrt{\sum_{i=1}^D u_i^2}\sqrt{\sum_{i=1}^Dv_i^2}}
\end{align}
This normalization step is in general a recommended practice. For example, when using LIBLINEAR or LIBSVM packages~\cite{Article:Fan_JMLR08}, it is often suggested  to  first normalize the input   data vectors to  unit $l_2$ norm. In addition to packages such as LIBLINEAR which implement batch linear algorithms, methods based on stochastic gradient descent (SGD)  become increasingly important especially for truly large-scale industrial applications~\cite{URL:Bottou_SGD}.

In this paper, the proposed GMM kernel is defined on general data types which can have both negative and positive entries. The basic idea is to first transform the original data into nonnegative data and then compute the  min-max kernel~\cite{Report:Manasse_CWS10,Proc:Ioffe_ICDM10,Proc:Li_KDD15} on the transformed data.

\subsection{Data Transformation}

Consider the original data vector $u_i$, $i=1$ to $D$. We define the following transformation, depending on whether an entry $u_i$ is positive or negative:\footnote{
This transformation can be  generalized by considering a ``center vector'' $\mu_i$, $i=1$ to $D$, such  that
\begin{align}\notag
 \left\{\begin{array}{cc}
\tilde{u}_{2i-1} = u_i - \mu_i,\hspace{0.1in} \tilde{u}_{2i} = 0&\text{if } \ u_i >\mu_i\\
\tilde{u}_{2i-1} = 0,\hspace{0.1in} \tilde{u}_{2i} =  -u_i+\mu_i &\text{if } \ u_i \leq \mu_i
\end{array}\right.
\end{align}
In this paper, we always use $\mu_i=0,\ \forall i$. Note that the same center vector $\mu$ should be used for all data vectors.
}
\begin{align}\label{eqn_transform}
 \left\{\begin{array}{cc}
\tilde{u}_{2i-1} = u_i,\hspace{0.1in} \tilde{u}_{2i} = 0&\text{if } \ u_i >0\\
\tilde{u}_{2i-1} = 0,\hspace{0.1in} \tilde{u}_{2i} =  -u_i &\text{if } \ u_i \leq 0
\end{array}\right.
\end{align}
For example, when $D=2$ and $u = [-5\ \ 3]$, the transformed data vector becomes $\tilde{u} = [0\ \ 5\ \ 3\ \ 0]$.

\subsection{Generalized Min-Max (GMM) Kernel}

Given two data vectors $u, v\in\mathbb{R}^D$, we first transform them into $\tilde{u}, \tilde{v}\in\mathbb{R}^{2D}$  according to (\ref{eqn_transform}). Then the generalized min-max (GMM) similarity is defined as
\begin{align}\label{eqn_GMM}
&GMM(u,v) = \frac{\sum_{i=1}^{2D} \min(\tilde{u}_i,\ \tilde{v}_i)}{\sum_{i=1}^{2D} \max(\tilde{u}_i,\ \tilde{v}_i)}
\end{align}
We will show in Section~\ref{sec_kernel} that GMM is indeed an effective measure of data similarity through an extensive experimental study on kernel SVM classification.

It is generally  nontrivial to scale nonlinear kernels for  large data~\cite{Book:Bottou_07}. In a sense, it is not  practically meaningful to discuss nonlinear kernels without knowing how to compute them efficiently (e.g., via hashing).  In this paper, we focus on the generalized consistent weighted sampling (GCWS).

\vspace{-0.1in}
\subsection{Generalized Consistent Weighted Sampling (GCWS)}

Algorithm~\ref{alg_GCWS} summarizes the ``generalized consistent weighted sampling'' (GCWS).  Given two  data vectors  $u$ and $v$, we transform them into nonnegative vectors  $\tilde{u}$ and $\tilde{v}$ as in (\ref{eqn_transform}). We then  apply the original ``consistent weighted sampling'' (CWS)~\cite{Report:Manasse_CWS10,Proc:Ioffe_ICDM10} to generate  random tuples:
\begin{align}
\left(i^*_{\tilde{u},j}, t^*_{\tilde{u},j}\right)\ \text{ and }\  \left(i^*_{\tilde{v},j}, t^*_{\tilde{v},j}\right),\  \  j = 1, 2, ..., k
\end{align}
where $i^*\in[1,\ 2D]$  and $t^*$ is unbounded. Following~\cite{Report:Manasse_CWS10,Proc:Ioffe_ICDM10}, we have the basic probability result.
\begin{theorem}
\begin{align}\label{eqn_GCWS_Prob}
\mathbf{Pr}\left\{\left(i^*_{\tilde{u},j}, t^*_{\tilde{u},j}\right) =  \left(i^*_{\tilde{v},j}, t^*_{\tilde{v},j}\right)\right\} = GMM({u},{v})
\end{align}
\end{theorem}

\begin{algorithm}{
\textbf{Input:} Data vector $u$ = ($i=1$ to $D$)

\textbf{Transform:} Generate vector $\tilde{u}$ in $2D$-dim by (\ref{eqn_transform})

\textbf{Output:} Consistent uniform sample ($i^*$, $t^*$)

\vspace{0.08in}

For $i$ from 1 to $2D$

\hspace{0.25in}$r_i\sim Gamma(2, 1)$, \ $c_i\sim Gamma(2, 1)$,  $\beta_i\sim Uniform(0, 1)$

\hspace{0.2in} $t_i\leftarrow \lfloor \frac{\log \tilde{u}_i }{r_i}+\beta_i\rfloor$, \ \  $a_i\leftarrow \log(c_i)- r_i(t_i+1-\beta_i)$

End For

$i^* \leftarrow arg\min_i \ a_i$,\hspace{0.3in}  $t^* \leftarrow t_{i^*}$
}\caption{Generalized Consistent Weighted Sampling (GCWS). Note that we slightly re-write the expression for $a_i$ compared to \cite{Proc:Ioffe_ICDM10}.}
\label{alg_GCWS}
\end{algorithm}

With $k$ samples, we  can simply use the averaged  indicator to estimate $GMM(u,v)$. By property of the binomial distribution, we know the expectation ($E$) and variance  ($Var$) are
\begin{align}\label{eqn_GMM_E}
&E\left[1\{i^*_{\tilde{u},j} = i^*_{\tilde{v},j} \ \text{and} \ t^*_{\tilde{u},j} = t^*_{\tilde{v},j}\}\right] = GMM(u,v),\\\label{eqn_GMM_Var}
&Var\left[1\{i^*_{\tilde{u},j} = i^*_{\tilde{v},j} \ \text{and} \ t^*_{\tilde{u},j} = t^*_{\tilde{v},j}\}\right] =(1- GMM(u,v))GMM(u,v)
\end{align}
The estimation  variance, given $k$ samples,  will be  $\frac{1}{k}(1- GMM)GMM$, which  vanishes as GMM approaches 0 or 1,   or as the sample size $k\rightarrow\infty$.

\subsection{0-bit GCWS for Linearizing GMM Kernel SVM}

The so-called ``0-bit'' GCWS idea is that, based on intensive empirical observations~\cite{Proc:Li_KDD15}, one can safely ignore $t^*$ (which is unbounded) and simply use
\begin{align}\label{eqn_0bitGCWS}
\mathbf{Pr}\left\{i^*_{\tilde{u},j} =  i^*_{\tilde{v},j}\right\} \approx  GMM({u},{v})
\end{align}
For each data vector $u$, we obtain $k$ random samples $i^*_{\tilde{u},j}$, $j=1$ to $k$. We store only the lowest $b$ bits of $i^*$, based on the idea of~\cite{Proc:HashLearning_NIPS11}. We need to view those $k$ integers as locations (of the nonzeros) instead of numerical values. For example, when $b=2$, we should view $i^*$ as a  vector of length $2^b=4$. If $i^*=3$, then we code it as $[1\ 0\ 0\ 0]$; if $i^*=0$, we code it as $[0\ 0\ 0\ 1]$. We can concatenate all $k$ such vectors into a binary vector of length $2^b\times k$, with exactly $k$ 1's.

For linear methods, the computational cost is largely determined by the number of nonzeros in each data vector, i.e., the $k$ in our case. For the other parameter $b$,  we recommend to use $b\geq4$.

\vspace{0.08in}

The natural competitor of the GMM kernel is the RBF (radial basis function) kernel, and the competitor of the GCWS hashing method is the RFF (random Fourier feature) algorithm.

\section{ RBF Kernel and Normalized Random Fourier Features (NRFF)}

The radial basis function (RBF) kernel is widely used in machine learning and beyond.  In this study, for convenience (e.g.,  parameter tuning), we recommend the following version:
\begin{align}
RBF(u,v;\gamma) = e^{-\gamma(1-\rho)}
\end{align}
where $\rho=\rho(u,v)$ is the correlation defined in (\ref{eqn_rho}) and $\gamma>0$ is a crucial tuning parameter. Based on Bochner's Theorem~\cite{Book:Rudin_90}, it is known~\cite{Proc:Rahimi_NIPS07} that, if we sample $w\sim uniform(0,2\pi)$, $r_{i}\sim N(0,1)$ i.i.d., and let  $x = \sum_{i=1}^D u_i r_{ij}$, $y = \sum_{i=1}^D v_i r_{ij}$, where $\|u\|_2=\|v\|_2=1$, then we have
\begin{align}\label{eqn_RFF}
E\left(\sqrt{2}\cos(\sqrt{\gamma} x+w)\sqrt{2}\cos(\sqrt{\gamma} y+w)\right)
= e^{-\gamma(1-\rho)}
\end{align}
This provides a nice mechanism for linearizing the RBF kernel and the RFF method has become  popular in machine learning, computer vision, and beyond, e.g.,~\cite{Proc:BinaryRFF_NIPS09,Proc:Yang_NIPS12,Proc:Affandi_NIPS13,Proc:Hern_NIPS14,Proc:Dai_NIPS14,Proc:Yen_NIPS14,Proc:Hsieh_NIPS14,Proc:Shah_NIPS15,Chwialkowski_NIPS15,Richard_NIPS15}.


\begin{theorem}\label{thm_RFF}
Given $x\sim N(0,1)$, $y\sim N(0,1)$,  $E(xy) = \rho$, and $w\sim uniform(0,2\pi)$, we have
\begin{align}\label{eqn_RFF_E}
&E\left[\sqrt{2}\cos(\sqrt{\gamma}x+w)\sqrt{2}\cos(\sqrt{\gamma}y+w)\right]=e^{-\gamma(1-\rho)}\\
&E\left[\cos(\sqrt{\gamma}x)\cos(\sqrt{\gamma}y)\right]=\frac{1}{2}e^{-\gamma(1-\rho)} + \frac{1}{2}e^{-\gamma(1+\rho)}\\
\label{eqn_RFF_Var}
&Var\left[\sqrt{2}\cos(\sqrt{\gamma}x+w)\sqrt{2}\cos(\sqrt{\gamma}y+w)\right] = \frac{1}{2} + \frac{1}{2}\left(1-e^{-2\gamma(1-\rho)}\right)^2
\end{align}
\end{theorem}

\vspace{0.08in}

\noindent The proof for   (\ref{eqn_RFF_Var}) can also be found in~\cite{Proc:Sutherland_UAI15}. One can  see that the variance of RFF can be large.  Interestingly, the variance can be substantially reduced if we normalize the  hashed data, a procedure which we call ``normalized RFF (NRFF)''. The theoretical results are presented in Theorem~\ref{thm_NRFF}.

\begin{theorem}\label{thm_NRFF}
Consider $k$ iid samples ($x_j, y_j, w_j$) where $x_j\sim N(0,1)$, $y_j\sim N(0,1)$,  $E(x_jy_j) = \rho$, $w_j\sim uniform(0,2\pi)$, $j =1, 2, ..., k$. Let $X_j = \sqrt{2}\cos\left(\sqrt{\gamma}x_j + w_j\right)$ and $Y_j = \sqrt{2}\cos\left(\sqrt{\gamma}y_j + w_j\right)$. As $k\rightarrow\infty$, the following asymptotic normality holds:
\begin{align}\label{eqn_NRFF}
&\sqrt{k}\left(\frac{\sum_{j=1}^k X_j Y_j}{\sqrt{\sum_{j=1}^k X_j^2}\sqrt{\sum_{j=1}^k Y_j^2}} - e^{-\gamma(1-\rho)}\right)\overset{D}{\Longrightarrow}
N\left(0,V_{n,\rho,\gamma}\right)\\\notag
\text{where}\hspace{0.2in}&\\\label{eqn_NRFF_Var}
&V_{n,\rho,\gamma} = V_{\rho,\gamma}- \frac{1}{4}e^{-2\gamma(1-\rho)} \left[3-e^{-4\gamma(1-\rho)}  \right]\\
&V_{\rho,\gamma} = \frac{1}{2}+\frac{1}{2}\left(1-e^{-2\gamma(1-\rho)}\right)^2
\end{align}
\end{theorem}

\begin{figure}[h!]
\begin{center}
\includegraphics[width=3in]{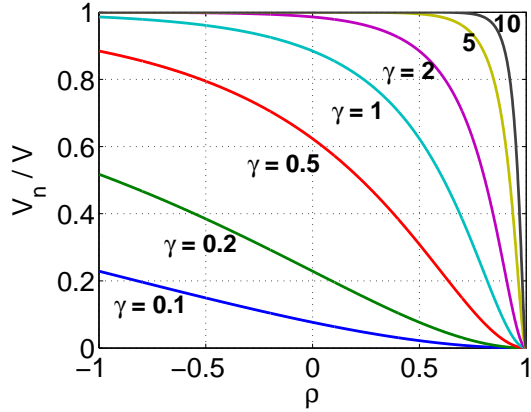}
\end{center}
\vspace{-0.3in}
\caption{The ratio $\frac{V_{n,\rho,\gamma}}{V_{\gamma,\gamma}}$ from Theorem~\ref{thm_NRFF} for visualizing the  improvement due to normalization.}\label{fig_Vn/V}
\end{figure}

Obviously, $V_{n,\rho,\gamma} < V_{\rho,\gamma}$ (in particular, $V_{n,\rho,\gamma}=0$ at $\rho=1$), i.e., the variance of the normalized RFF is  (much) smaller than that of the original RFF. Figure~\ref{fig_Vn/V} plots  $\frac{V_{n,\rho,\gamma}}{V_{\gamma,\gamma}}$ to visualize the  improvement due to normalization, which is most significant when $\rho$ is close to 1.

\begin{figure}[h!]
\begin{center}
\mbox{
\includegraphics[width=2.1in]{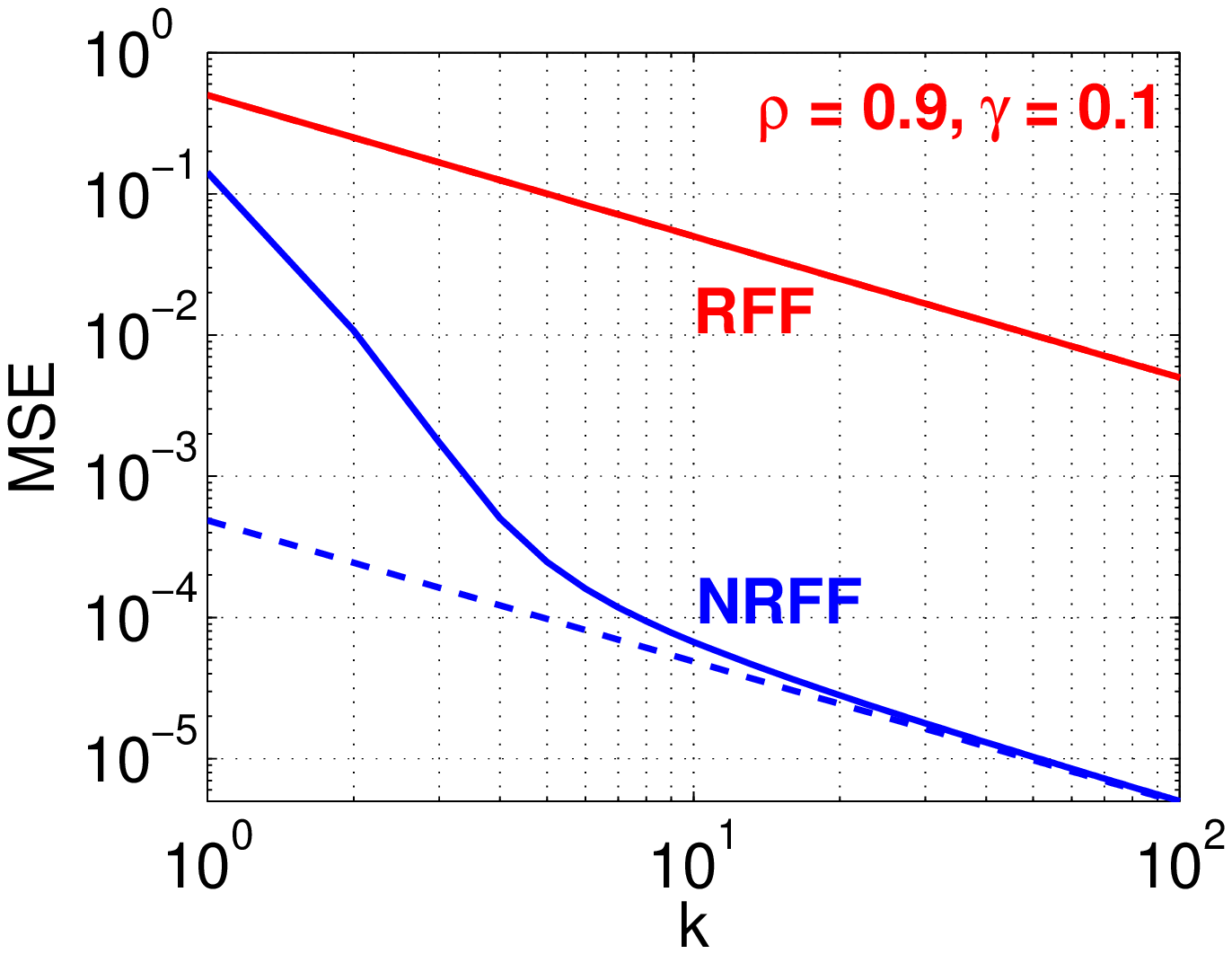}\hspace{-0.1in}
\includegraphics[width=2.1in]{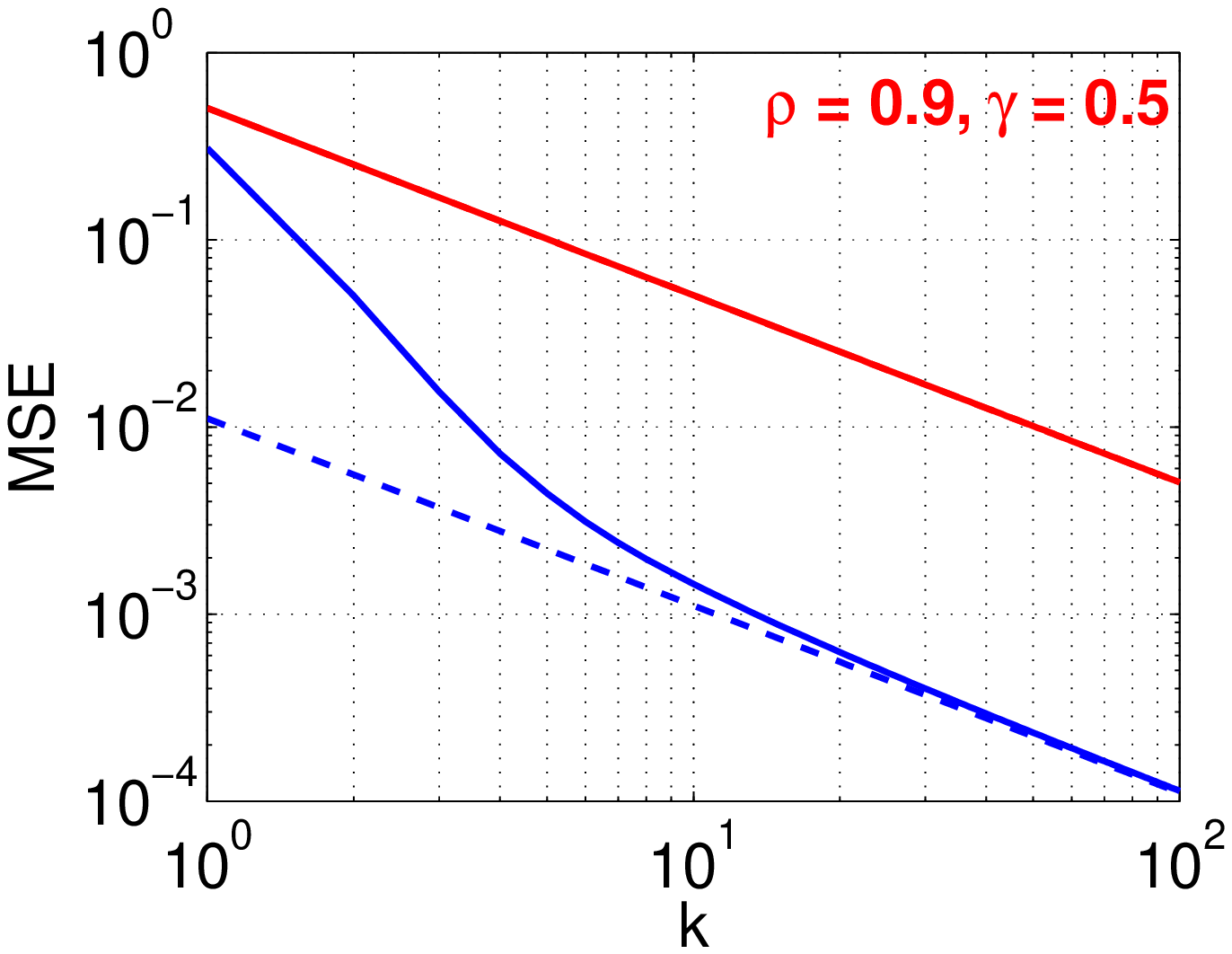}\hspace{-0.1in}
\includegraphics[width=2.1in]{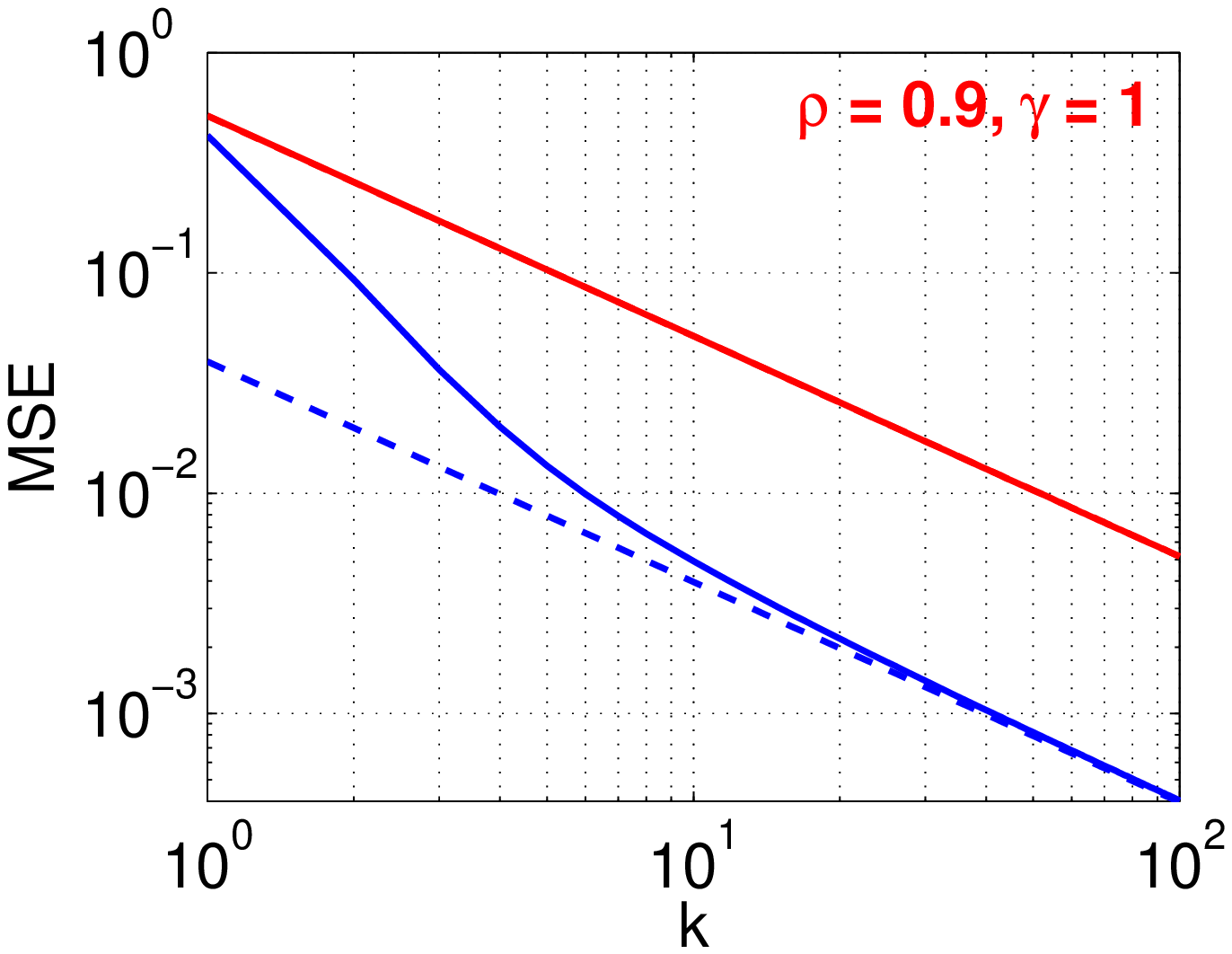}
}

\mbox{
\includegraphics[width=2.1in]{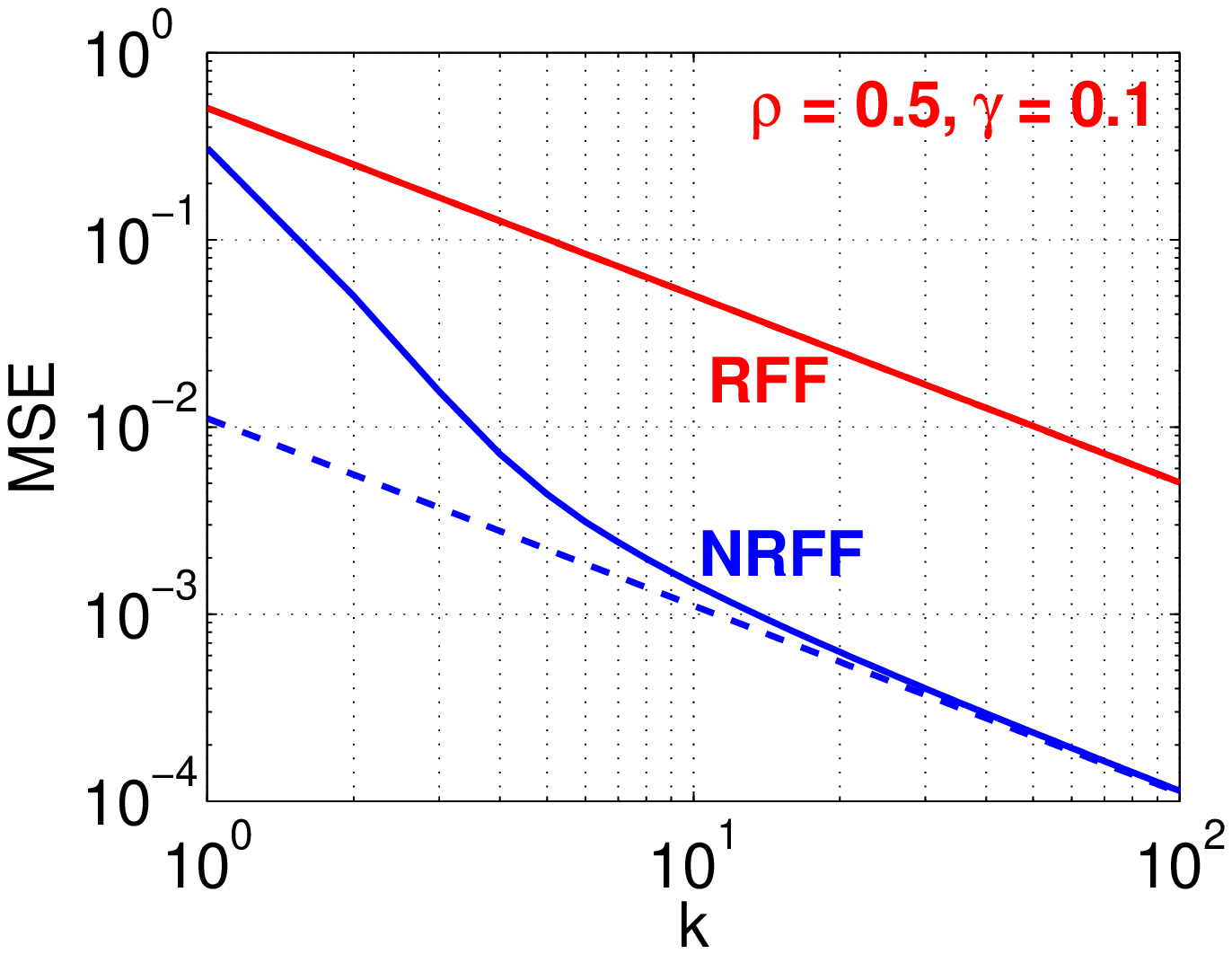}\hspace{-0.1in}
\includegraphics[width=2.1in]{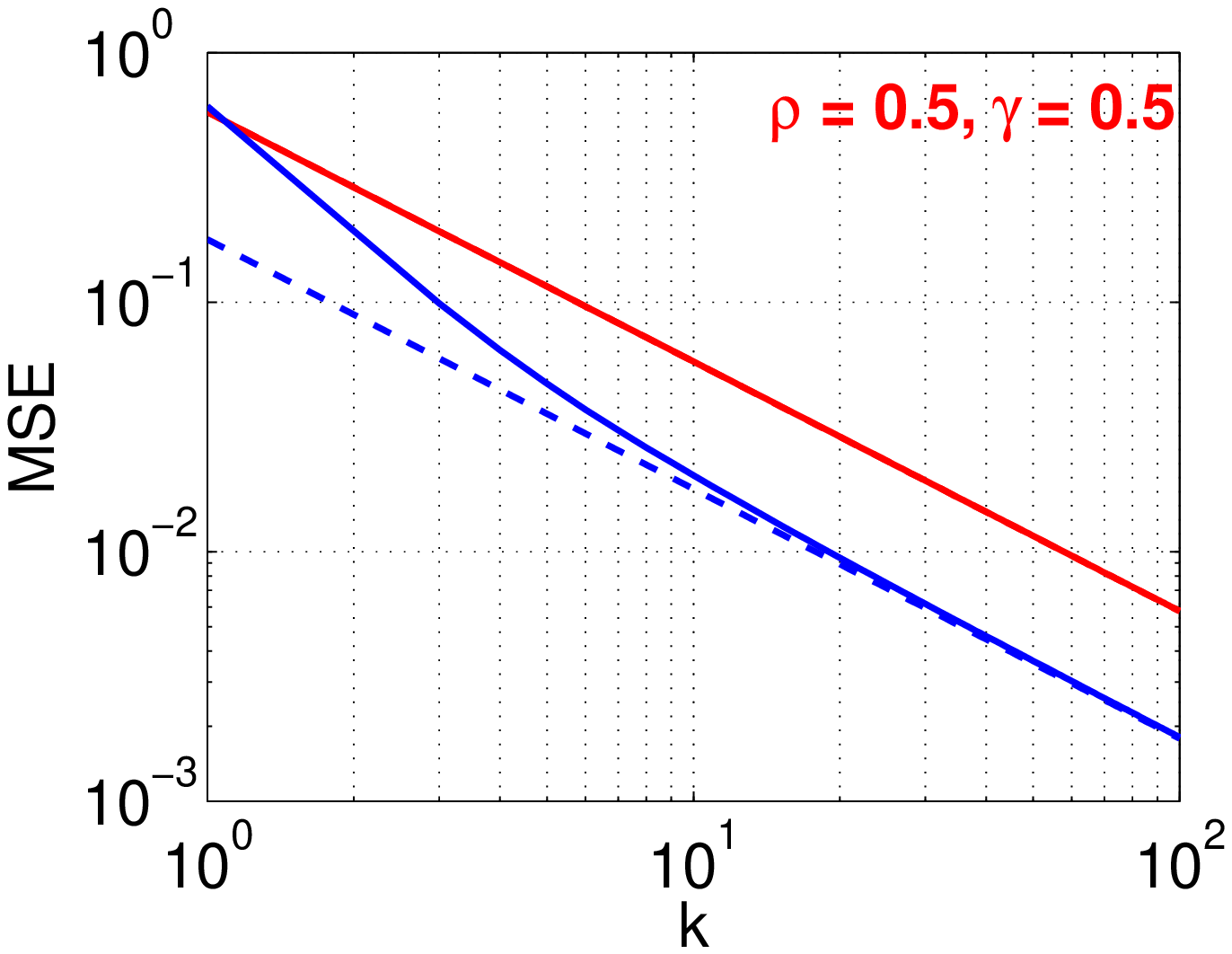}\hspace{-0.1in}
\includegraphics[width=2.1in]{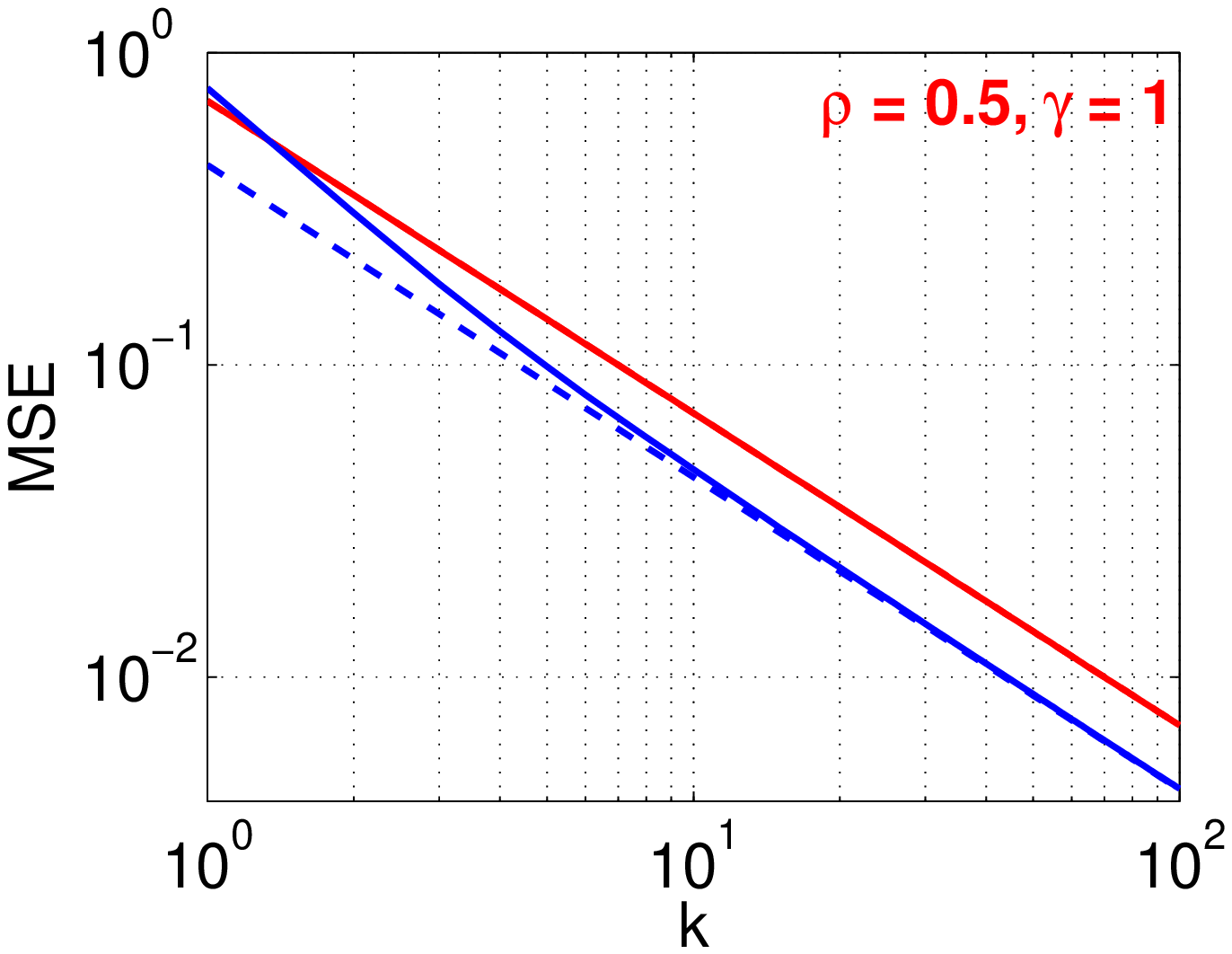}
}
\end{center}
\vspace{-0.3in}
\caption{A simulation study to verify the asymptotic theoretical results in Theorem~\ref{thm_NRFF}. With $k$ samples, we  estimate the RBF kernel $e^{-\gamma(1-\rho)}$, using both the original RFF and the normalized RFF (NRFF). With $10^5$ repetitions at each $k$, we can compute the empirical mean square error: MSE = Bias$^2$+Var. Each panel presents the MSEs (solid curves) for a particular choice of $(\rho,\gamma)$, along with the theoretical variances: $\frac{V_{\rho,\gamma}}{k}$  and $\frac{V_{n,\rho,\gamma}}{k}$ (dashed curves). The variance of the original RFF (curves above, or red if color is available) can be substantially larger than the MSE of the normalized RFF (curves below, or blue). When $k>10$, the normalized RFF provides an unbiased estimate of the RBF kernel and its empirical MSE matches the theoretical asymptotic variance.}\label{fig_NRFF}
\end{figure}

Note that the theoretical results in Theorem~\ref{thm_NRFF} are asymptotic (i.e., for larger $k$). With $k$ samples, the variance of the original RFF is exactly $\frac{V_{\rho,\gamma}}{k}$, however  the variance of the normalized RFF (NRFF) is written as $\frac{V_{n,\rho,\gamma}}{k} + O\left(\frac{1}{k^2}\right)$. It is important to understand the behavior when $k$ is not large.  For this purpose, Figure~\ref{fig_NRFF} presents the simulated mean square error (MSE) results for estimating the RBF kernel $e^{-\gamma(1-\rho)}$, confirming that a): the improvement due to normalization can be substantial, and b): the asymptotic variance formula (\ref{eqn_NRFF_Var})  becomes  accurate  for merely $k>10$.

\vspace{-0.in}

\newpage\clearpage

\begin{figure}[h!]
\begin{center}
\includegraphics[width=3in]{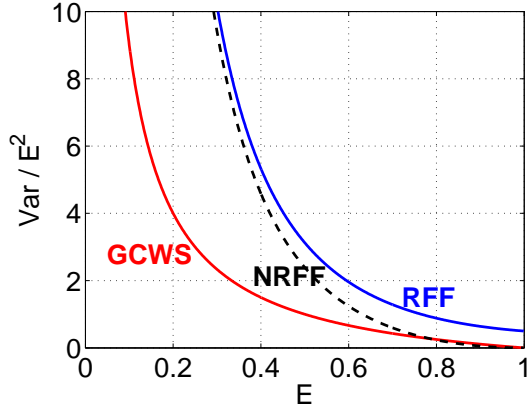}
\end{center}
\vspace{-0.3in}
\caption{Ratio of the variance  over the squared expectation, denoted as ${Var}/{E^2}$, for the convenience of comparing RFF/NRFF with GCWS. Smaller (lower) is  better.
}\label{fig_VarE}
\end{figure}

Next, we attempt to compare RFF with GCWS. While ultimately we can rely on classification accuracy as a metric for performance, here we compare their variances ($Var$) relative to their expectations ($E$) in terms of $Var/E^2$, as shown in Figure~\ref{fig_VarE}. For GCWS,  we know $Var/E^2 = E(1-E)/E^2=(1-E)/E$. For the original RFF, we have $Var/E^2 = \left[ \frac{1}{2}+\frac{1}{2}\left(1-E^2\right)^2\right]/E^2$, etc.

Figure~\ref{fig_VarE} shows that the relative variance of GCWS is substantially smaller than that of the original RFF and the normalized RFF (NRFF), especially when   $E$ is not large. For the very high similarity region (i.e., $E\rightarrow 1$), the variances of both GCWS and NRFF approach zero.

The results from Figure~\ref{fig_VarE} provide one explanation why later we will observe that, in the classification experiments, GCWS typically needs substantially fewer samples than the normalized RFF, in order to achieve similar classification accuracies. Note that for practical data, the similarities among most data points are usually small (i.e., small $E$) and  hence it is not surprising that GCWS may perform substantially better. Also see Section~\ref{sec_VarNRFFGMM} and Figure~\ref{fig_VarNRFFGMM} for a comparison from the perspective of  estimating RBF using GCWS based on a model assumption.

\vspace{0.08in}

In a sense, this drawback of RFF is expected, due to nature  of random projections. For example, as shown in~\cite{Proc:Li_Hastie_Church_COLT06,Proc:Li_Hastie_Church_KDD06}, the linear estimator of the correlation  $\rho$ using random projections has variance $\frac{1+\rho^2}{k}$, where $k$ is the number of projections. In order to make the variance small, one will have to use many projections (i.e., large $k$).

\vspace{0.2in}

\noindent\textbf{Proof of Theorem~\ref{thm_RFF}}: \  The following three  integrals  will be useful in our proof:
\begin{align}\notag
\int_{-\infty}^\infty \cos(cx)e^{-x^2/2}dx = \sqrt{2\pi} e^{-c^2/2}
\end{align}

\begin{align}\notag
\int_{-\infty}^\infty \cos(c_1x)\cos(c_2x)e^{-x^2/2}dx
=&\frac{1}{2}\int_{-\infty}^\infty \left[\cos((c_1+c_2)x) + \cos((c_1-c_2)x)\right]e^{-x^2/2}dx \\\notag
=& \frac{\sqrt{2\pi}}{{2}}\left[ e^{-(c_1+c_2)^2/2} + e^{-(c_1-c_2)^2/2} \right]
\end{align}
\begin{align}\notag
&\int_{-\infty}^\infty \sin(c_1x)\sin(c_2x)e^{-x^2/2}dx = \frac{\sqrt{2\pi}}{{2}}\left[ e^{-(c_1-c_2)^2/2} - e^{-(c_1+c_2)^2/2} \right]
\end{align}

Firstly, we consider integers $b_1, b_2 = 1, 2, 3, ..., $ and evaluate the  following general integral:
\begin{align}\notag
&E\left(\cos(c_1x+b_1w)\cos(c_2y+b_2w)\right)\\\notag
=&\frac{1}{2\pi}\int_0^{2\pi}E(\cos(c_1x+b_1t)\cos(c_2y+b_2t))dt\\\notag
=&\frac{1}{2\pi}\int_0^{2\pi}\int_{-\infty}^\infty\int_{-\infty}^\infty (\cos(c_1x+b_1t)\cos(c_2y+b_2t)) \frac{1}{2\pi}\frac{1}{\sqrt{1-\rho^2}}e^{-\frac{x^2+y^2-2\rho xy}{2(1-\rho^2)}} dxdydt\\\notag
=&\frac{1}{2\pi}\int_0^{2\pi}\int_{-\infty}^\infty\int_{-\infty}^\infty (\cos(c_1x+b_1t)\cos(c_2y+b_2t)) \frac{1}{2\pi}\frac{1}{\sqrt{1-\rho^2}}e^{-\frac{x^2+y^2-2\rho xy+\rho^2x^2-\rho^2x^2}{2(1-\rho^2)}} dxdydt\\\notag
=&\frac{1}{2\pi}\int_0^{2\pi}\int_{-\infty}^\infty\frac{1}{2\pi}\frac{1}{\sqrt{1-\rho^2}}e^{-\frac{x^2}{2}} \cos(c_1x+b_1t) dx\int_{-\infty}^\infty\cos(c_2y+b_2t)e^{-\frac{(y-\rho x)^2}{2(1-\rho^2)}} dydt\\\notag
=&\frac{1}{2\pi}\int_0^{2\pi}\int_{-\infty}^\infty\frac{1}{2\pi}e^{-\frac{x^2}{2}} \cos(c_1x+b_1t) dx\int_{-\infty}^\infty\cos(c_2y\sqrt{1-\rho^2}+c_2\rho x+b_2t)e^{-y^2/2} dydt\\\notag
=&\frac{1}{2\pi}\int_0^{2\pi}\int_{-\infty}^\infty\frac{1}{2\pi}e^{-\frac{x^2}{2}} \cos(c_1x+b_1t)\cos(c_2\rho x+b_2t) dx\int_{-\infty}^\infty\cos(c_2y\sqrt{1-\rho^2})e^{-y^2/2} dydt\\\notag
=&\frac{1}{2\pi}\int_0^{2\pi}\int_{-\infty}^\infty\frac{1}{2\pi}e^{-\frac{x^2}{2}} \cos(c_1x+b_1t)\cos(c_2\rho x+b_2t) \sqrt{2\pi}e^{-\frac{c_2^2(1-\rho^2)}{2}} dxdt\\\notag
=&\frac{1}{2\pi}\frac{1}{\sqrt{2\pi}}e^{-\frac{c_2^2(1-\rho^2)}{2}}\int_0^{{2\pi}} \int_{-\infty}^\infty e^{-\frac{x^2}{2}} \cos(c_1x+b_1t)\cos(c_2\rho x+b_2t) dxdt
\end{align}

Note that
\begin{align}\notag
&\int_0^{2\pi}\cos(c_1x+b_1t)\cos(c_2\rho x+b_2t)dt \\\notag
=&\int_0^{2\pi}\cos(c_1x)\cos(b_1t)\cos(c_2\rho x)\cos(b_2t)dt + \int_0^{2\pi}\sin(c_1x)\sin(b_1t)\sin(c_2\rho x)\sin(b_2t)dt\\\notag
-&\int_0^{2\pi}\cos(c_1x)\cos(b_1t)\sin(c_2\rho x)\sin(b_2t)dt - \int_0^{2\pi}\sin(c_1x)\sin(b_1t)\cos(c_2\rho x)\cos(b_2t)dt
\end{align}

When $b_1\neq b_2$, we have
\begin{align}\notag
&\int_0^{2\pi}\cos(b_1t)\cos(b_2t)dt  = \frac{1}{2}\int_0^{2\pi}\cos(b_1t-b_2t) + \cos(b_1t+ b_2t)dt = 0\\\notag
&\int_0^{2\pi}\sin(b_1t)\sin(b_2t)dt  = \frac{1}{2}\int_0^{2\pi}\cos(b_1t-b_2t) - \cos(b_1t+ b_2t)dt = 0
\end{align}

If $b_1 = b_2$, then
\begin{align}\notag
&\int_0^{2\pi}\cos(b_1t)\cos(b_2t)dt  = \int_0^{2\pi}\sin(b_1t)\sin(b_2t)dt = \pi
\end{align}

In addition, for any $b_1, b_2 = 1, 2, 3, ...$, we always have
\begin{align}\notag
&\int_0^{2\pi}\sin(b_1t)\cos(b_2t)dt  = \frac{1}{2}\int_0^{2\pi}\sin(b_1t-b_2t) + \sin(b_1t+ b_2t)dt = 0\\\notag
\end{align}

Thus, only when $b_1=b_2$ we have
\begin{align}\notag
&\int_0^{2\pi}\cos(c_1x+b_1t)\cos(c_2\rho x+b_2t)dt  = \pi \cos(c_1x) \cos(c_2\rho x) + \pi \sin(c_1x) \sin(c_2\rho x) = \pi \cos((c_1-c_2\rho)x)
\end{align}
Otherwise, $\int_0^{2\pi}\cos(c_1x+b_1t)\cos(c_2\rho x+b_2t)dt =0$. Therefore, when $b_1=b_2$,  we have
\begin{align}\notag
&E\left(\cos(c_1x+b_1w)\cos(c_2y+b_2w)\right)\\\notag
=&\frac{1}{2\pi}\frac{1}{\sqrt{2\pi}}e^{-\frac{c_2^2(1-\rho^2)}{2}}\int_0^{{2\pi}} \int_{-\infty}^\infty e^{-\frac{x^2}{2}} \cos(c_1x+b_1t)\cos(c_2\rho x+b_2t) dxdt\\\notag
=&\frac{1}{2\pi}\frac{1}{\sqrt{2\pi}}e^{-\frac{c_2^2(1-\rho^2)}{2}}  \int_{-\infty}^\infty e^{-\frac{x^2}{2}} \pi \cos((c_1-c_2\rho)x) dx \\\notag
=&\frac{1}{2\pi}\frac{1}{\sqrt{2\pi}}e^{-\frac{c_2^2(1-\rho^2)}{2}}  \pi\sqrt{2\pi} e^{-(c_1-c_2\rho)^2/2}\\\notag
=&\frac{1}{2}e^{-\frac{c_1^2+c_2^2-2c_1c_2\rho}{2}} \\\notag
=&\frac{1}{2}e^{-c^2(1-\rho)},\hspace{0.3in} \text{when } \ c_1 = c_2 = c
\end{align}

This completes the proof of the first moment.  Next, using the following fact
\begin{align}\notag
E \cos(2cx+2w)
 =&\frac{1}{2\pi}\int_0^{2\pi} \frac{1}{\sqrt{2\pi}}\int_{-\infty}^\infty \cos(2cx+2t) e^{-x^2/2} dx dt\\\notag
 =&\frac{1}{2\pi}\int_0^{2\pi} \frac{1}{\sqrt{2\pi}}\frac{1}{2}\sin 2t\int_{-\infty}^\infty \cos (2cx) e^{-x^2/2} dx dt\\\notag
 =&\frac{1}{4\pi}  e^{-2c^2}\int_0^{2\pi}\sin 2t dt = 0
\end{align}
we are ready to compute the second moment
\begin{align}\notag
&E\left[\cos(cx+w)\cos(cy+w)\right]^2\\\notag
=&\frac{1}{4}E\left[\cos(2cx+2w)\cos(2cy+2w) + \cos(2cx+2w) +\cos(2cy+2w)\right] + \frac{1}{4}\\\notag
=&\frac{1}{4}E\left[\cos(2cx+2w)\cos(2cy+2w)\right] + \frac{1}{4}\\\notag
=&\frac{1}{8}e^{-4c^2(1-\rho)} + \frac{1}{4}
\end{align}
and the variance
\begin{align}\notag
&Var\left[\cos(cx+w)\cos(cy+w)\right] = \frac{1}{8}e^{-4c^2(1-\rho)} + \frac{1}{4} - \frac{1}{4}e^{-2c^2(1-\rho)}
\end{align}

Finally, we prove the first moment without the ``$w$'' random variable:
\begin{align}\notag
E\left(\cos(c x)\cos(c y)\right)
=&\int_{-\infty}^\infty\int_{-\infty}^\infty \cos(c x)\cos(c y) \frac{1}{2\pi}\frac{1}{\sqrt{1-\rho^2}}e^{-\frac{x^2+y^2-2\rho xy+\rho^2x^2-\rho^2x^2}{2(1-\rho^2)}} dxdy\\\notag
=&\int_{-\infty}^\infty\frac{1}{2\pi}\frac{1}{\sqrt{1-\rho^2}}e^{-\frac{x^2}{2}} \cos(c x) dx\int_{-\infty}^\infty\cos(c y)e^{-\frac{(y-\rho x)^2}{2(1-\rho^2)}} dy\\\notag
=&\int_{-\infty}^\infty\frac{1}{2\pi}e^{-\frac{x^2}{2}} \cos(c x) dx\int_{-\infty}^\infty\cos(c y\sqrt{1-\rho^2}+c \rho x)e^{-y^2/2} dy\\\notag
=&\int_{-\infty}^\infty\frac{1}{2\pi}e^{-\frac{x^2}{2}} \cos(c x)\cos(c \rho x) dx\int_{-\infty}^\infty\cos(c y\sqrt{1-\rho^2})e^{-y^2/2} dy\\\notag
=&\int_{-\infty}^\infty\frac{1}{2\pi}e^{-\frac{x^2}{2}} \cos(c x)\cos(c \rho x) \sqrt{2\pi}e^{-c^2\frac{1-\rho^2}{2}} dx\\\notag
=&\frac{1}{\sqrt{2\pi}}e^{-c^2\frac{1-\rho^2}{2}}\int_{-\infty}^\infty e^{-\frac{x^2}{2}} \cos(c x)\cos(c\rho x) dx\\\notag
=&\frac{1}{\sqrt{2\pi}}e^{-c^2\frac{1-\rho^2}{2}}\frac{\sqrt{2\pi}}{2}  \left[e^{-c^2\frac{(1-\rho)^2}{2}} + e^{-c^2\frac{(1+\rho)^2}{2}}\right]\\\notag
=&\frac{1}{2}e^{-c^2(1-\rho)}+\frac{1}{2}e^{-c^2(1+\rho)}
\end{align}

This completes the proof of Theorem~\ref{thm_RFF}.$\hfill\Box$

\vspace{0.2in}

\noindent\textbf{Proof of Theorem~\ref{thm_NRFF}}:  \ \ We will use some of the results from the proof of Theorem~\ref{thm_RFF}. Define
\begin{align}\notag
X_j = \sqrt{2}\cos(\sqrt{\gamma}x_j+w_j),\hspace{0.2in}Y_j = \sqrt{2}\cos(\sqrt{\gamma}y_j+w_j),\hspace{0.2in}
Z_k = \frac{\sum_{j=1}^k X_j Y_j}{\sqrt{\sum_{j=1}^k X_j^2}\sqrt{\sum_{j=1}^k Y_j^2}}
\end{align}

From Theorem~2, it is easy to see that, as $k\rightarrow\infty$, we have
\begin{align}\notag
&\frac{1}{k}\sum_{j=1}^k X_j^2 \rightarrow E\left(X_j^2\right) = e^{-\gamma(1-1)} = 1, \ \ a.s. \hspace{0.5in} \frac{1}{k}\sum_{j=1}^k Y_j^2 \rightarrow 1, \ \ a.s.\\\notag
&Z_k = \frac{\frac{1}{k}\sum_{j=1}^k X_j Y_j}{\sqrt{\frac{1}{k}\sum_{j=1}^k X_j^2}\sqrt{\frac{1}{k}\sum_{j=1}^k Y_j^2}} \rightarrow e^{-\gamma(1-\rho)} = Z_\infty, \ \ a.s.
\end{align}

We express the deviation $Z_k - Z_\infty$ as
\begin{align}\notag
Z_k - Z_\infty =& \frac{\frac{1}{k}\sum_{j=1}^k X_j Y_j - Z_\infty + Z_\infty}{\sqrt{\frac{1}{k}\sum_{j=1}^k X_j^2}\sqrt{\frac{1}{k}\sum_{j=1}^k Y_j^2}} - Z_\infty\\\notag
=& \frac{\frac{1}{k}\sum_{j=1}^k X_j Y_j - Z_\infty }{\sqrt{\frac{1}{k}\sum_{j=1}^k X_j^2}\sqrt{\frac{1}{k}\sum_{j=1}^k Y_j^2}} + Z_\infty \frac{1-\sqrt{\frac{1}{k}\sum_{j=1}^k X_j^2}\sqrt{\frac{1}{k}\sum_{j=1}^k Y_j^2} }{\sqrt{\frac{1}{k}\sum_{j=1}^k X_j^2}\sqrt{\frac{1}{k}\sum_{j=1}^k Y_j^2}}\\\notag
=& \frac{1}{k}\sum_{j=1}^k X_j Y_j - Z_\infty + Z_\infty \frac{1-\frac{1}{k}\sum_{j=1}^k X_j^2\frac{1}{k}\sum_{j=1}^k Y_j^2}{2} + O_P(1/k)\\\notag
=& \frac{1}{k}\sum_{j=1}^k X_j Y_j - Z_\infty + Z_\infty \frac{1-\frac{1}{k}\sum_{j=1}^k X_j^2}{2}+ Z_\infty \frac{1-\frac{1}{k}\sum_{j=1}^k Y_j^2}{2} + O_P(1/k)
\end{align}

Note that if $a\approx 1$ and $b\approx1$, then
\begin{align}\notag
1-ab = 1-(1-(1-a))(1-(1-b)) =(1-a)+(1-b) -(1-a)(1-b)
\end{align}
and  we can ignore the higher-order term.\\

Therefore, to analyze the asymptotic variance, it suffices to study the following expectation
\begin{align}\notag
&E\left( XY - Z_\infty + Z_\infty \frac{1-X^2}{2}+ Z_\infty \frac{1-Y^2}{2}\right)^2\\\notag
=& E\left(XY-Z_\infty(X^2+Y^2)/2\right)^2\\\notag
=&E(X^2Y^2)+Z_\infty^2E(X^4+Y^4+2X^2Y^2)/4  - Z_\infty E(X^3Y) - Z_\infty E(XY^3)
\end{align}
which can be obtained from the results in the proof of Theorem 2. In particular, if $b_1=b_2$, then
\begin{align}\notag
&E\left(\cos(c_1x+b_1w)\cos(c_2y+b_2w)\right) =\frac{1}{2}e^{-\frac{c_1^2+c_2^2-2c_1c_2\rho}{2}}
\end{align}
Otherwise $E\left(\cos(c_1x+b_1w)\cos(c_2y+b_2w)\right) = 0$.  We can now compute
\begin{align}\notag
&E\left[\cos(cx+w)^3\cos(cy+w)\right]\\\notag
=&E\left[\frac{1}{4}\cos(3(cx+w))\cos(cy+w) +\frac{3}{4}\cos(cx+w)\cos(cy+w) \right]\\\notag
=&\frac{3}{8}e^{-c^2(1-\rho)}
\end{align}
\begin{align}\notag
&E\left[\cos(cx+w)\cos(cy+w)\right]^2 =\frac{1}{8}e^{-4c^2(1-\rho)} + \frac{1}{4}
\end{align}
\begin{align}\notag
&E\left[\cos(cx+w)\right]^4 = \frac{1}{8} + \frac{1}{4} = \frac{3}{8}
\end{align}

\begin{align}\notag
V_{n,\rho,\gamma} =& E\left( XY - Z_\infty + Z_\infty \frac{1-X^2}{2}+ Z_\infty \frac{1-Y^2}{2}\right)^2\\\notag
=&E(X^2Y^2)+Z_\infty^2E(X^4+Y^4+2X^2Y^2)/4  - Z_\infty E(X^3Y) - Z_\infty E(XY^3)\\\notag
=&\frac{1}{2}e^{-4c^2(1-\rho)} +1+e^{-2c^2(1-\rho)}\left(\frac{3}{8}+\frac{3}{8} + \frac{1}{4}e^{-4c^2(1-\rho)} + \frac{1}{2}\right) -e^{-c^2(1-\rho)}\left( \frac{3}{2}e^{-c^2(1-\rho)}+\frac{3}{2}e^{-c^2(1-\rho)} \right)\\\notag
=&\frac{1}{2}e^{-4c^2(1-\rho)} +1+e^{-2c^2(1-\rho)}\left(\frac{5}{4} + \frac{1}{4}e^{-4c^2(1-\rho)}\right) -3e^{-2c^2(1-\rho)} \\\notag
=&\frac{1}{2}e^{-4c^2(1-\rho)} +1+\frac{1}{4}e^{-6c^2(1-\rho)}-\frac{7}{4}e^{-2c^2(1-\rho)} \\\notag
=& V_{\rho,\gamma}- \frac{1}{4}e^{-2c^2(1-\rho)} \left[3-e^{-4c^2(1-\rho)}  \right]
\end{align}
where $V_{\rho,\gamma}$ is the corresponding variance factor without using normalization:
\begin{align}\notag
V_{\rho,\gamma} = \frac{1}{2}+\frac{1}{2}\left(1-e^{-2c^2(1-\rho)}\right)^2
\end{align}

This completes the proof of Theorem~\ref{thm_NRFF}.$\hfill\Box$


\section{Another Comparison Based on Asymptotic of GMM}\label{sec_VarNRFFGMM}

As proved in a technical report following this paper~\cite{Report:Li_GMM_Theory16}, under mild model assumption, as the dimension $D$ becomes large, the GMM kernel converges to a function of the true data correlation:
\begin{align}
GMM \rightarrow \frac{1-\sqrt{(1-\rho)/2}}{1+\sqrt{(1-\rho)/2}}  = g
\end{align}
The convergence holds almost surely for data with bounded first moment. Using the expression of $g$ we can express RBF $e^{-\gamma(1-\rho)}$ in terms of $g$:
\begin{align}
\rho = 1-2\left(\frac{1-g}{1+g}\right)^2,\hspace{0.3in}
e^{-\gamma(1-\rho)} = e^{-2\gamma\left(\frac{1-g}{1+g}\right)^2}
\end{align}
For the convenience of conducting theoretical analysis, we assume  $GMM =\frac{1-\sqrt{(1-\rho)/2}}{1+\sqrt{(1-\rho)/2}}  = g$, exactly instead of asymptotically. Then we have another estimator of the RBF kernel from GCWS. Note that with $k$ hashes, the estimate of GMM follows a binomial distribution $binomial(k,g)$.

\begin{theorem}\label{thm_g}
Assume  $g = \frac{1-\sqrt{(1-\rho)/2}}{1+\sqrt{(1-\rho)/2}}$ and $X\sim binomial(k,g)$. Then, denoting $\bar{X} = \frac{1}{k}\sum_{i=1}^k X_i$, we have
\begin{align}
&E\left( e^{-2\gamma\left(\frac{1-\bar{X}}{1+\bar{X}}\right)^2}\right) = e^{-\gamma(1-\rho)} + O\left(\frac{1}{k}\right)\\
&Var\left( e^{-2\gamma\left(\frac{1-\bar{X}}{1+\bar{X}}\right)^2}\right)  = \frac{V_{g,\gamma}}{k} + O\left(\frac{1}{k^2}\right)\\
\text{where } \ &V_{g,\gamma} =e^{-2\gamma(1-\rho)}\frac{g(1-g)^3}{(1+g)^6}64\gamma^2
\end{align}
\end{theorem}

\noindent\textbf{Proof of Theorem~\ref{thm_g}:}\hspace{0.2in} For an asymptotic analysis with large $k$, it suffices to consider $Z = \frac{1-\hat{X}}{1+\hat{X}}$ as a normal random variable, whose mean and variance can be calculated to be
$\mu = \frac{1-g}{1+g},\ \  \sigma^2 = \frac{1}{k}\frac{4g(1-g)}{(1+g)^4}$. \ Thus, it suffices to compute
\begin{align}\notag
E\left(e^{X^2t}\right) =& \int_{-\infty}^\infty e^{x^2t}\frac{1}{\sqrt{2\pi}\sigma} e^{-\frac{(x-\mu)^2}{2\sigma^2}}dx
= \int_{-\infty}^\infty \frac{1}{\sqrt{2\pi}\sigma} e^{-\frac{(x-\mu)^2-2\sigma^2x^2t}{2\sigma^2}}dx\\\notag
=&\int_{-\infty}^\infty \frac{1}{\sqrt{2\pi}\sigma} e^{-\frac{(x-\mu)^2-2\sigma^2x^2t}{2\sigma^2}}dx
=\int_{-\infty}^\infty \frac{1}{\sqrt{2\pi}\sigma} e^{-\frac{(1-2\sigma^2t)x^2 - 2\mu x + \mu^2}{2\sigma^2}}dx\\\notag
=&\int_{-\infty}^\infty \frac{1}{\sqrt{2\pi}\sigma} e^{-\frac{x^2 - 2\mu/c^2 x + \mu^2/c^2}{2\sigma^2/c^2}}dx,\hspace{0.3in}\text{where } c^2 = 1-2\sigma^2t\\\notag
=&\frac{1}{c}\int_{-\infty}^\infty \frac{1}{\sqrt{2\pi}\sigma/c} e^{-\frac{(x - \mu/c^2)^2 - \mu^2/c^4 + \mu^2/c^2}{2\sigma^2/c^2}}dx\\\notag
=&\frac{1}{c}e^{\frac{\mu^2(1-c^2)}{2\sigma^2c^2}} = \frac{1}{c}e^{\frac{\mu^2}{c^2}t} = \frac{1}{\sqrt{1-2\sigma^2t}}e^{\frac{\mu^2t}{1-2\sigma^2t}}
\end{align}
from which we can compute the variance (letting $\sigma^2 = \frac{1}{k}\lambda^2$)
\begin{align}\notag
Var\left(e^{X^2t}\right)
 =& E\left(e^{X^22t}\right) - E^2\left(e^{X^2t}\right)
= \frac{1}{\sqrt{1-2\sigma^22t}}e^{\frac{\mu^22t}{1-2\sigma^22t}} - \frac{1}{{1-2\sigma^2t}}e^{\frac{\mu^22t}{1-2\sigma^2t}}\\\notag
=&\left(1+\frac{2\lambda^2t}{k}+O\left(\frac{1}{k^2}\right)\right)e^{2\mu^2t\left(1+\frac{4\lambda^2t}{k} +O\left(\frac{1}{k^2}\right) \right)} -
\left(1+\frac{2\lambda^2t}{k}+O\left(\frac{1}{k^2}\right)\right)e^{2\mu^2t\left(1+\frac{2\lambda^2t}{k} +O\left(\frac{1}{k^2}\right) \right)}\\\notag
=&\left(1+O\left(\frac{1}{k}\right)\right)e^{2\mu^2t}\left(1+\frac{8\mu^2\lambda^2t^2}{k} +O\left(\frac{1}{k^2}\right)\right) -
\left(1+O\left(\frac{1}{k}\right)\right)e^{2\mu^2t}\left(1+\frac{4\mu^2\lambda^2t^2}{k} +O\left(\frac{1}{k^2}\right) \right)\\\notag
=&\frac{4\mu^2\lambda^2t^2}{k} e^{2\mu^2t}+O\left(\frac{1}{k^2}\right)
\end{align}
Plugging in $t = -2\gamma$, $\mu = \frac{1-g}{1+g}$, and $\lambda^2 = \frac{4g(1-g)}{(1+g)^4}$, yields
\begin{align}\notag
Var\left(e^{-2\gamma \left(\frac{1-\bar{X}}{1+\bar{X}}\right)^2}\right)
=\frac{64\gamma^2}{k}\frac{g(1-g)^3}{(1+g)^6}e^{-4\gamma\left(\frac{1-g}{1+g}\right)^2}+O\left(\frac{1}{k^2}\right)
=\frac{64\gamma^2}{k}\frac{g(1-g)^3}{(1+g)^6}e^{-2\gamma(1-\rho)}+O\left(\frac{1}{k^2}\right)
\end{align}$\hfill\Box$

\begin{figure}[h!]
\begin{center}
\mbox{
\includegraphics[width=3in]{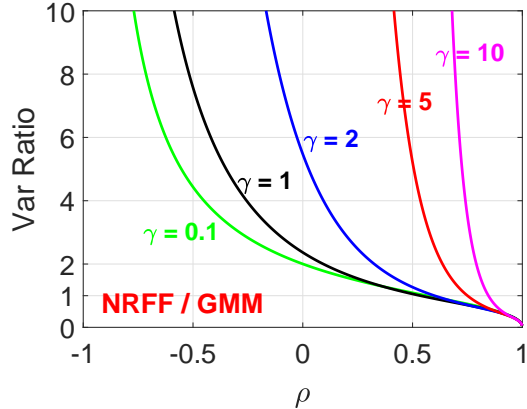}
}
\end{center}
\vspace{-0.3in}
\caption{The variance ratio:  $\frac{V_{n,\rho,\gamma}}{V_{g,\gamma}}$ provides another comparison of GCWS with   NRFF. $V_{g,\gamma}$ is derived in Theorem~\ref{thm_g} and $V_{n,\rho,\gamma}$ is derived in Theorem~\ref{thm_NRFF}. The ratios are significantly larger than 1 except for the very high similarity region (where the variances of both methods are close to zero). }\label{fig_VarNRFFGMM}
\end{figure}

This theoretical result provides a  direct comparison of GCWS with  NRFF for estimating the same object, by visualizing the variance ratio: $\frac{V_{n,\rho,\gamma}}{V_{g,\gamma}}$, using results from Theorem~\ref{thm_NRFF}. As shown in Figure~\ref{fig_VarNRFFGMM}, for estimating the RBF kernel, the variance of GCWS is substantially smaller than the variance of NRFF, except for the very high similarity region (depending on $\gamma$). At high similarity, the variances of both methods approach zero.  This provides another explanation for the superb empirical performance of GCWS compared to NRFF, as will be reported later in the paper.

\vspace{-0.25in}
\section{An Experimental Study on Kernel SVMs}\label{sec_kernel}
\vspace{-0.055in}

\begin{figure}[h!]
\begin{center}
\mbox{
\includegraphics[width=2.2in]{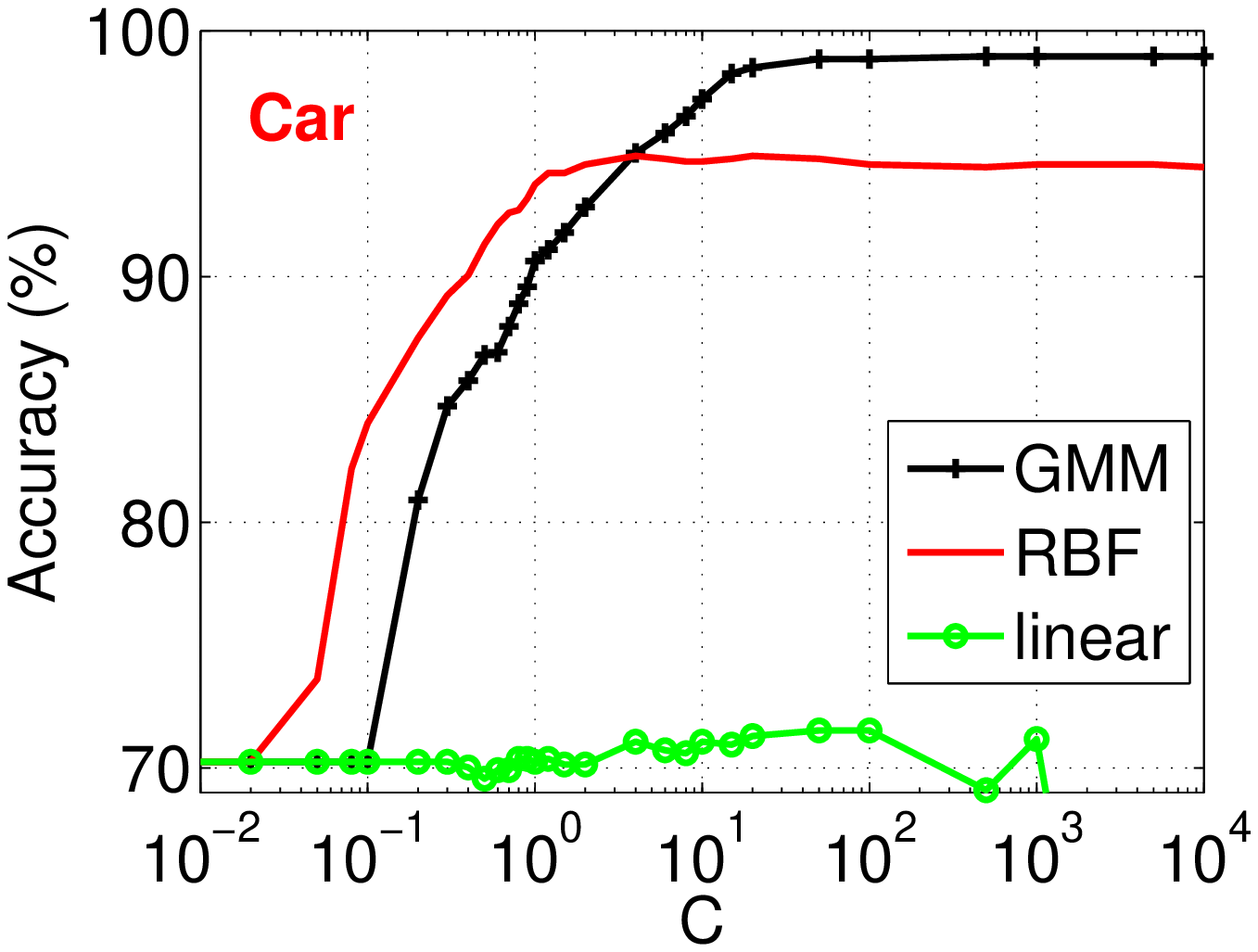}
\includegraphics[width=2.2in]{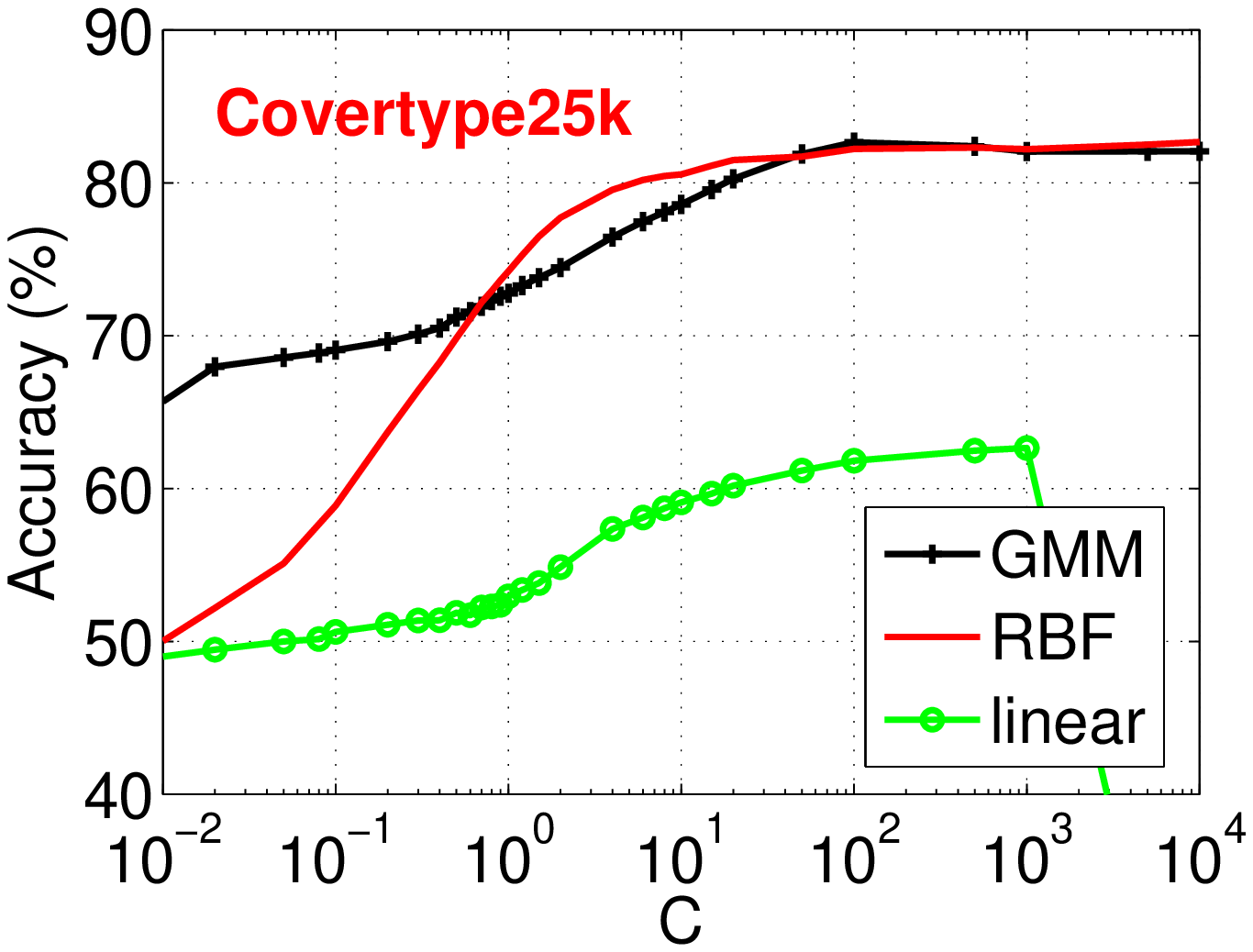}
\includegraphics[width=2.2in]{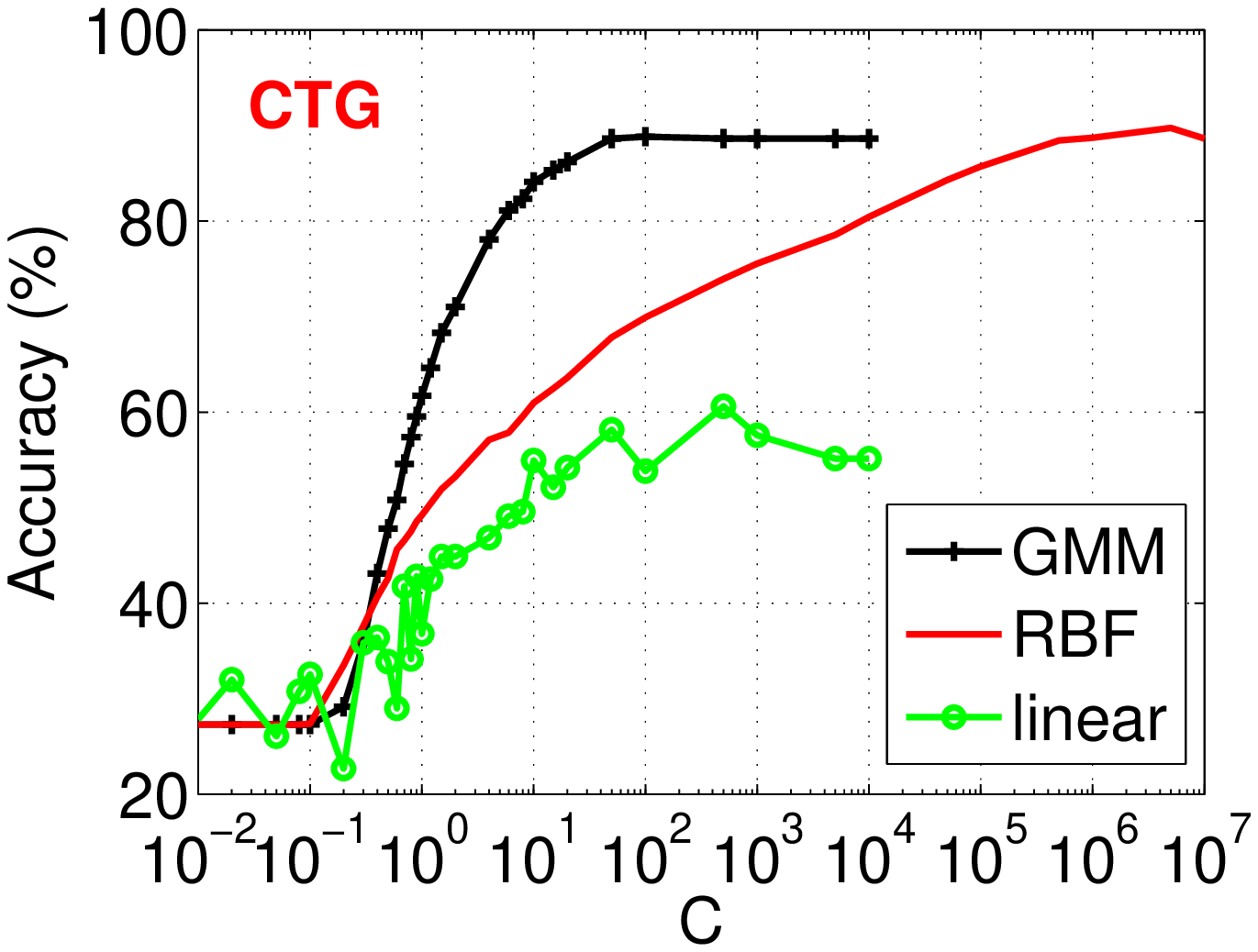}

}

\mbox{
\includegraphics[width=2.2in]{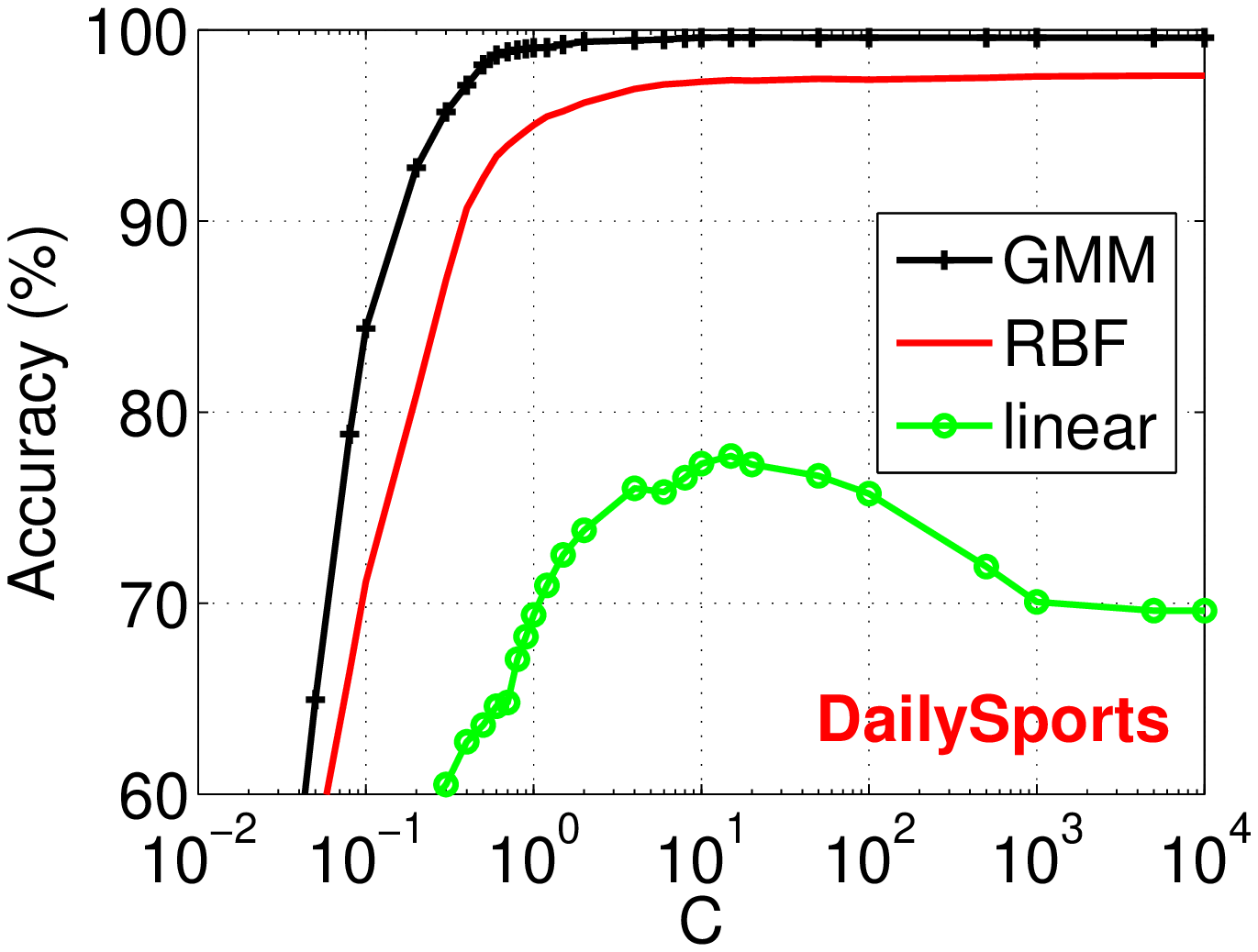}
\includegraphics[width=2.2in]{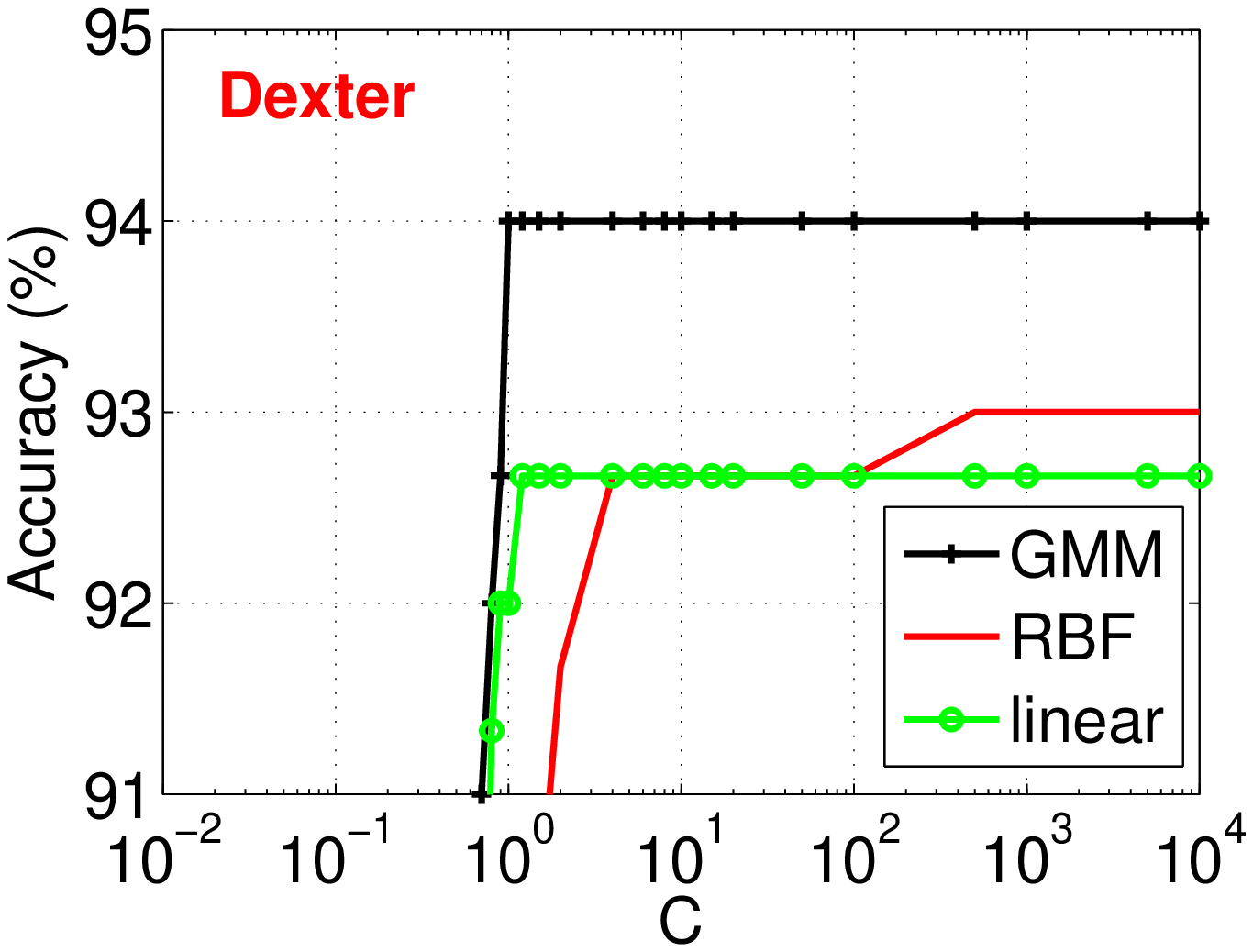}
\includegraphics[width=2.2in]{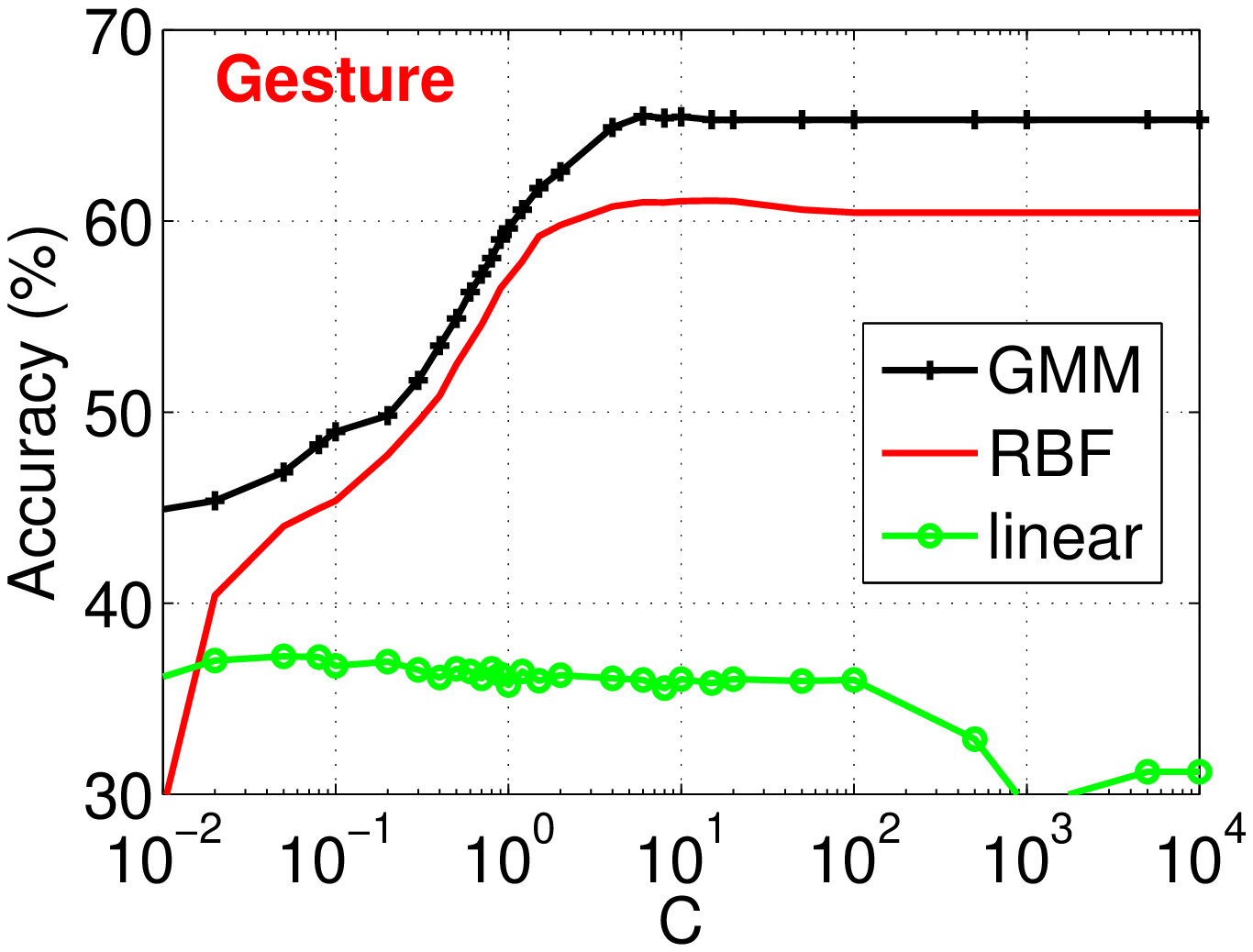}
}

\mbox{
\includegraphics[width=2.2in]{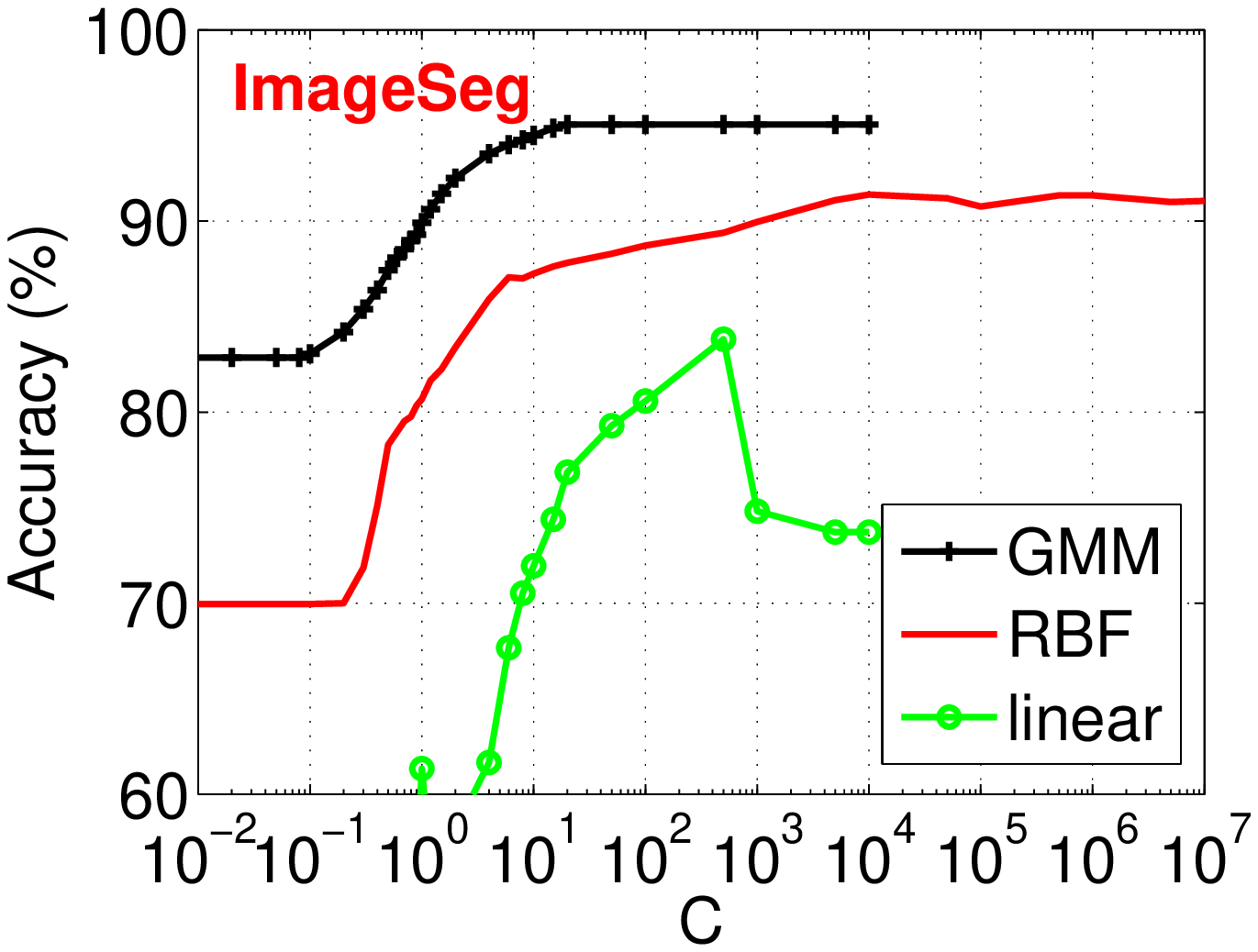}
\includegraphics[width=2.2in]{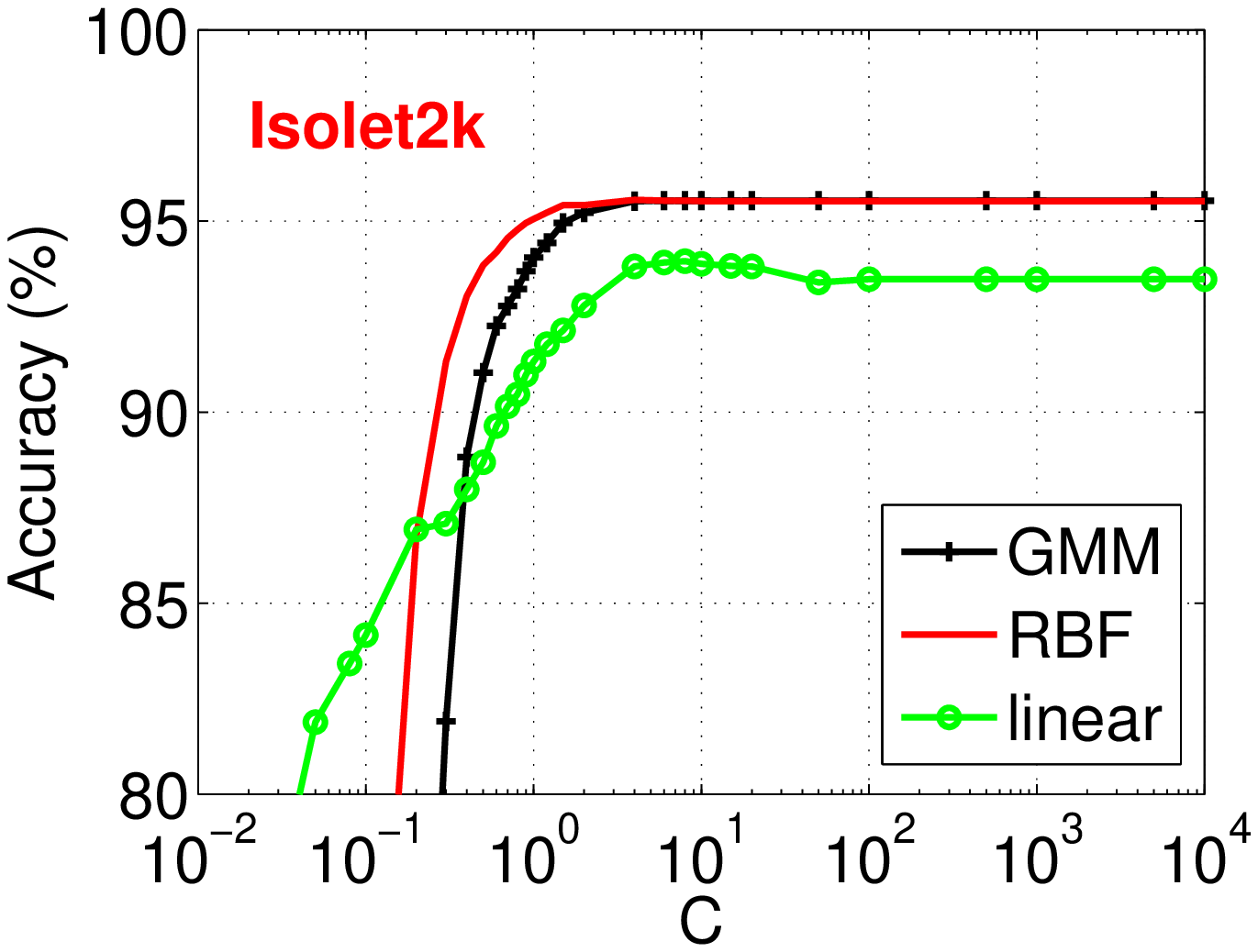}
\includegraphics[width=2.2in]{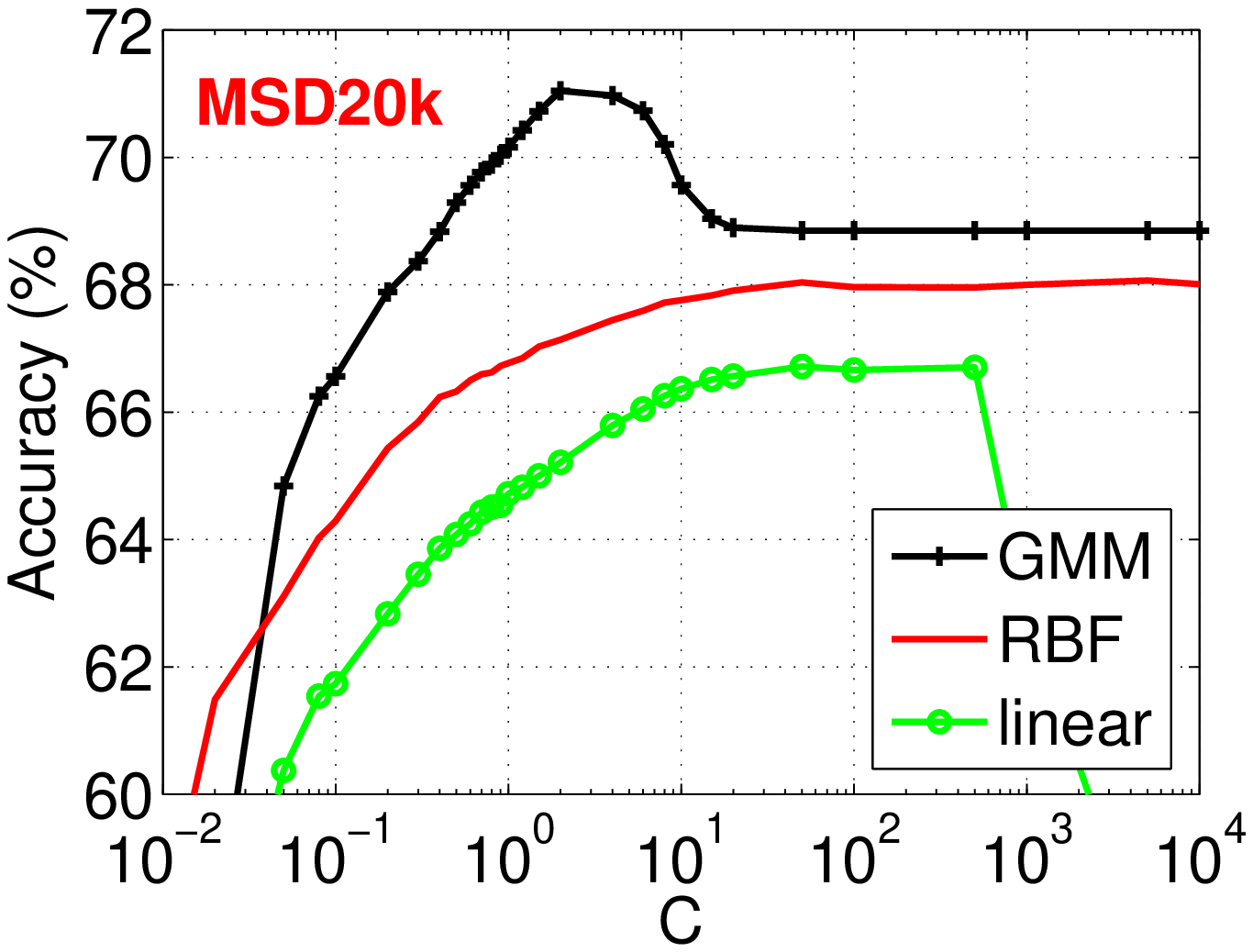}
}

\mbox{
\includegraphics[width=2.2in]{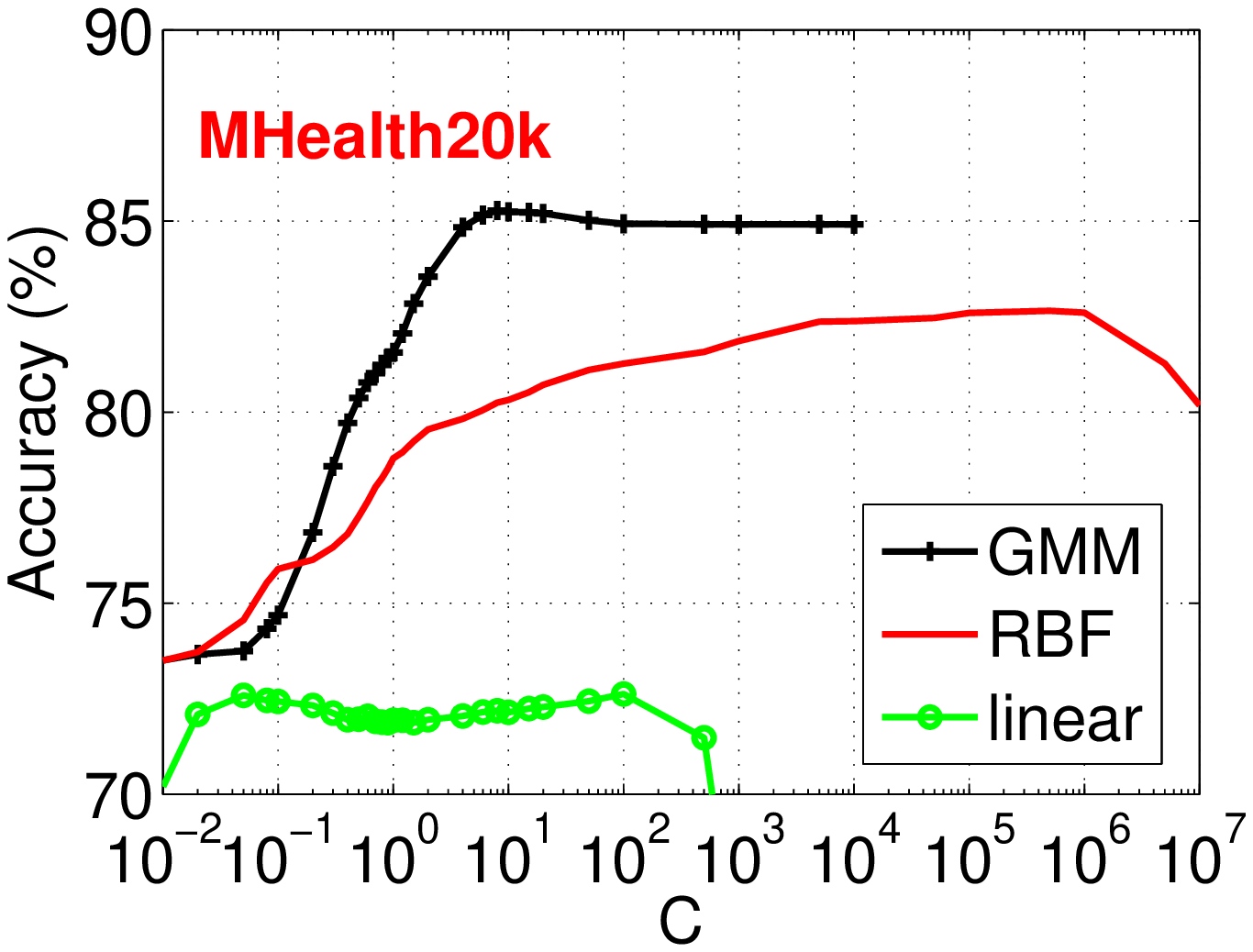}
\includegraphics[width=2.2in]{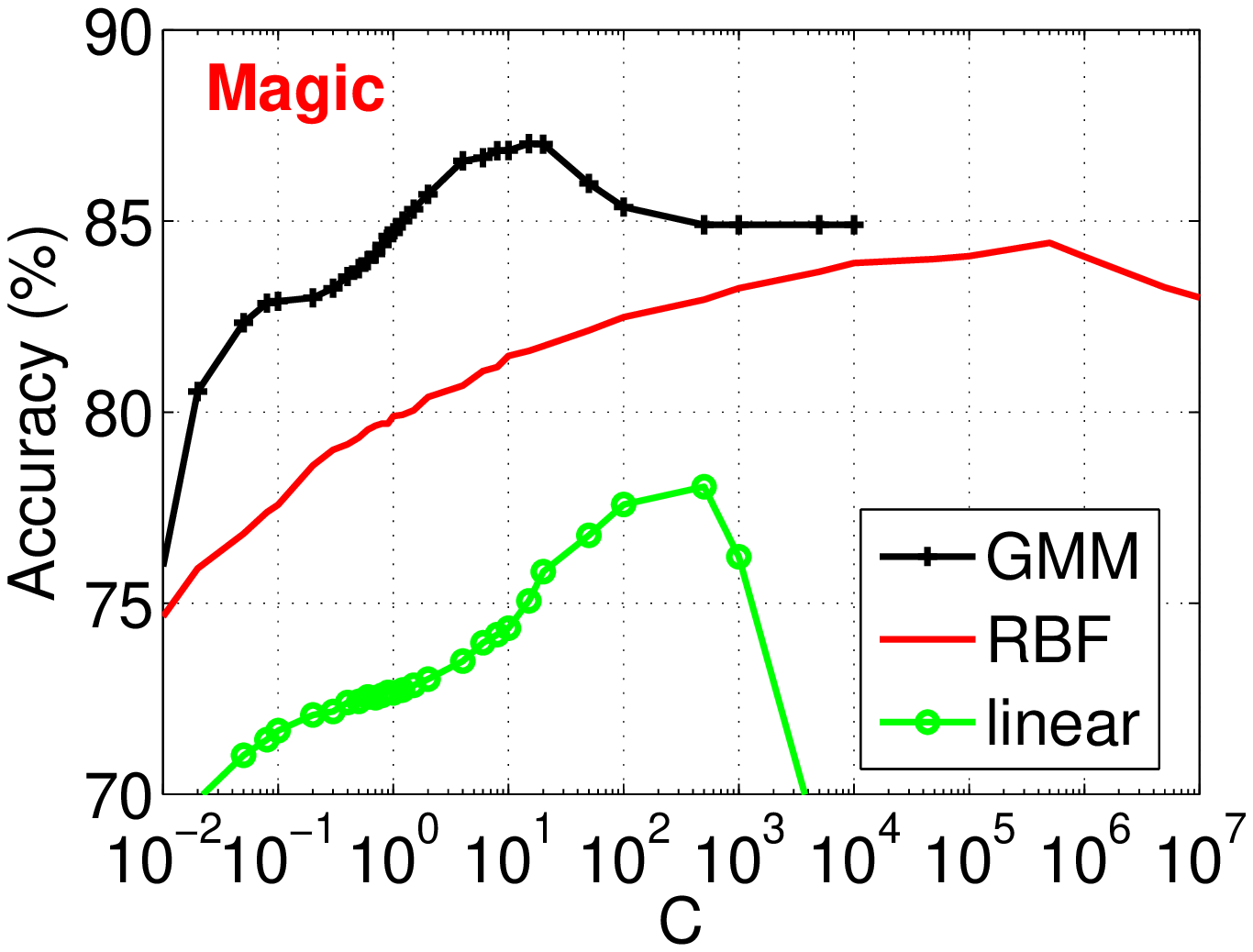}
\includegraphics[width=2.2in]{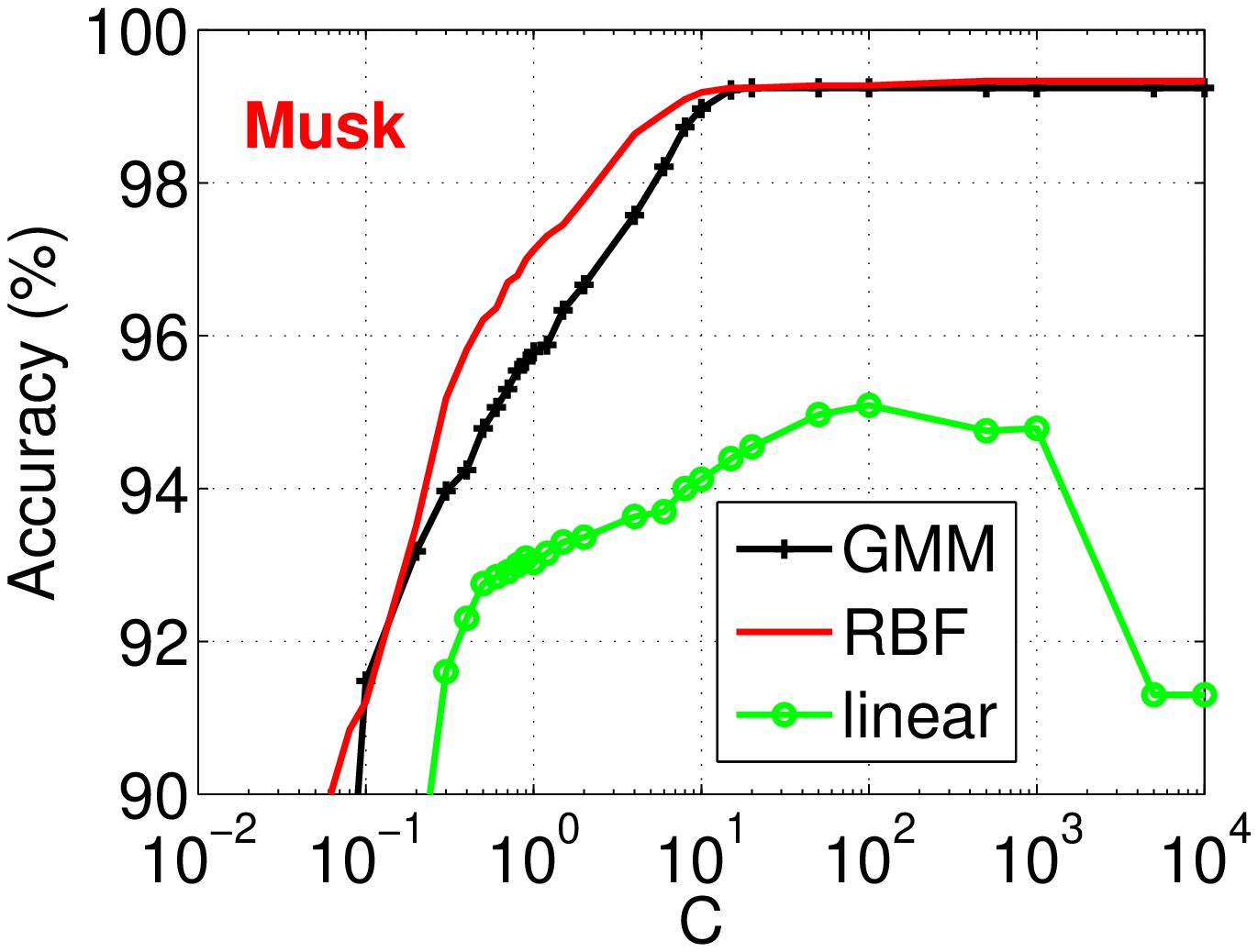}
}

\end{center}
\vspace{-0.3in}
\caption{\textbf{Test classification accuracies using kernel SVMs}. Both the GMM kernel and RBF kernel substantially improve linear SVM. $C$ is the $l_2$-regularization parameter of SVM. For the RBF kernel, we report the result at the best $\gamma$ value for every $C$ value.  }\label{fig_KernelSVM1}
\end{figure}

\begin{figure}[h!]
\begin{center}

\mbox{
\includegraphics[width=2.2in]{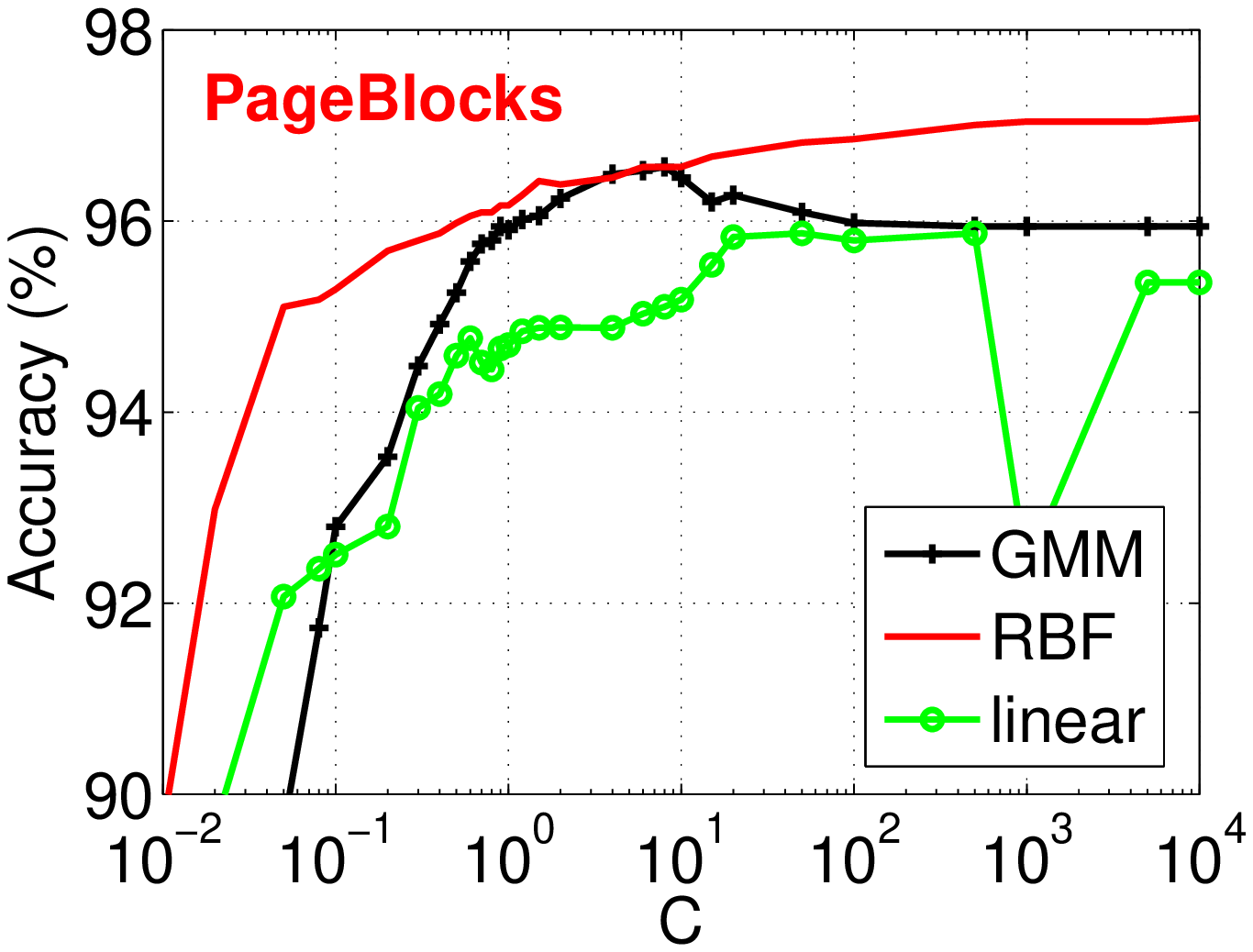}
\includegraphics[width=2.2in]{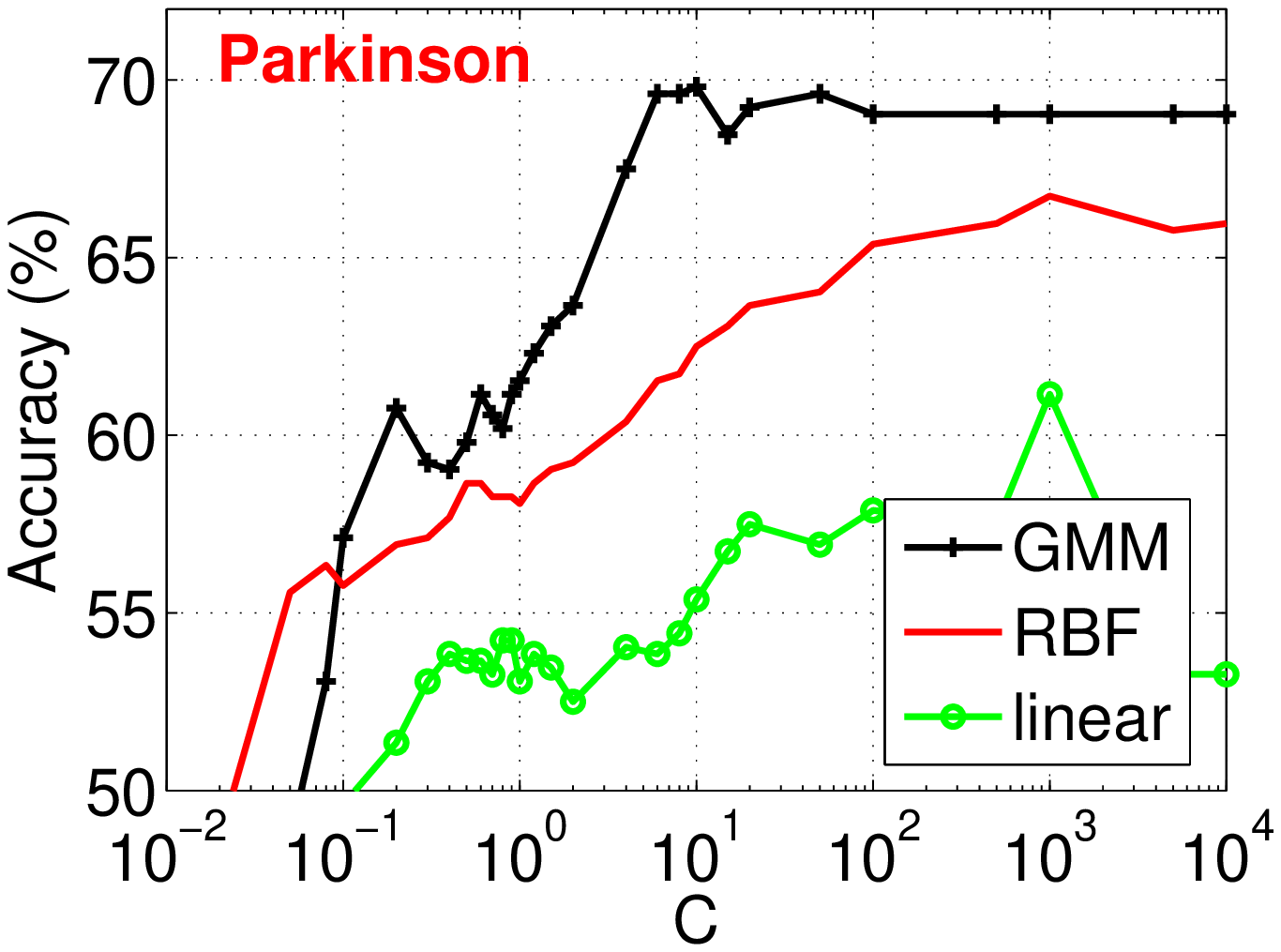}
\includegraphics[width=2.2in]{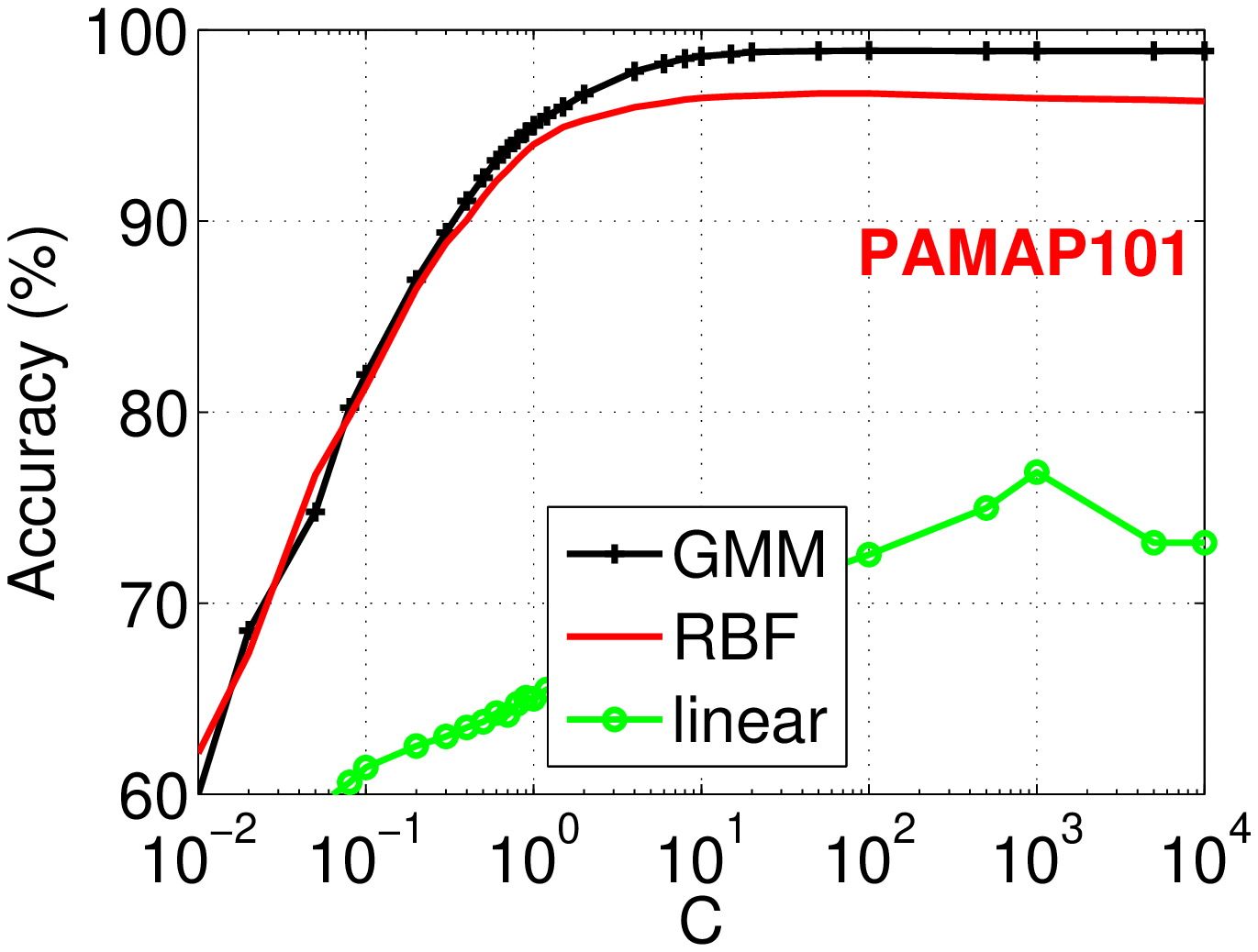}
}

\mbox{
\includegraphics[width=2.2in]{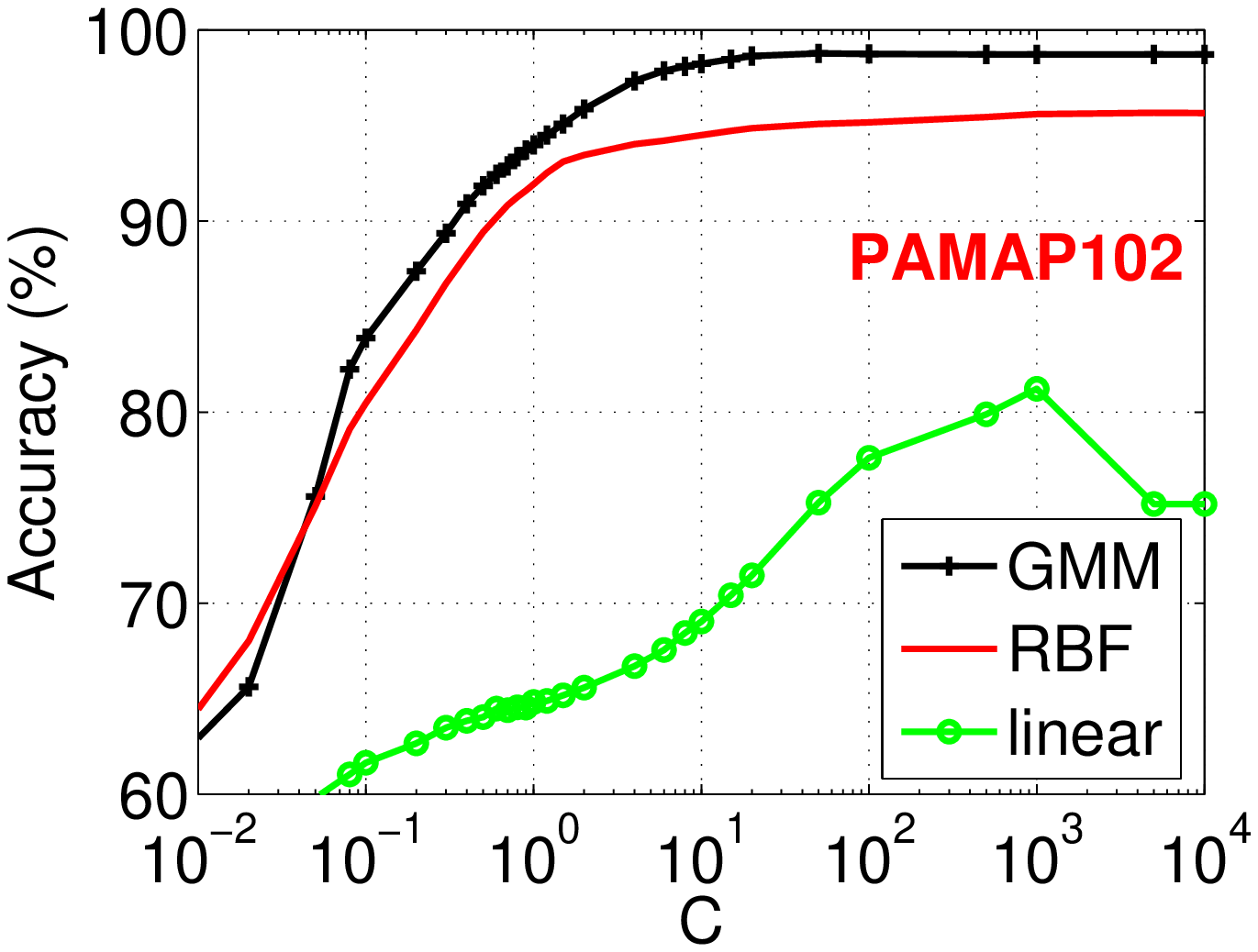}
\includegraphics[width=2.2in]{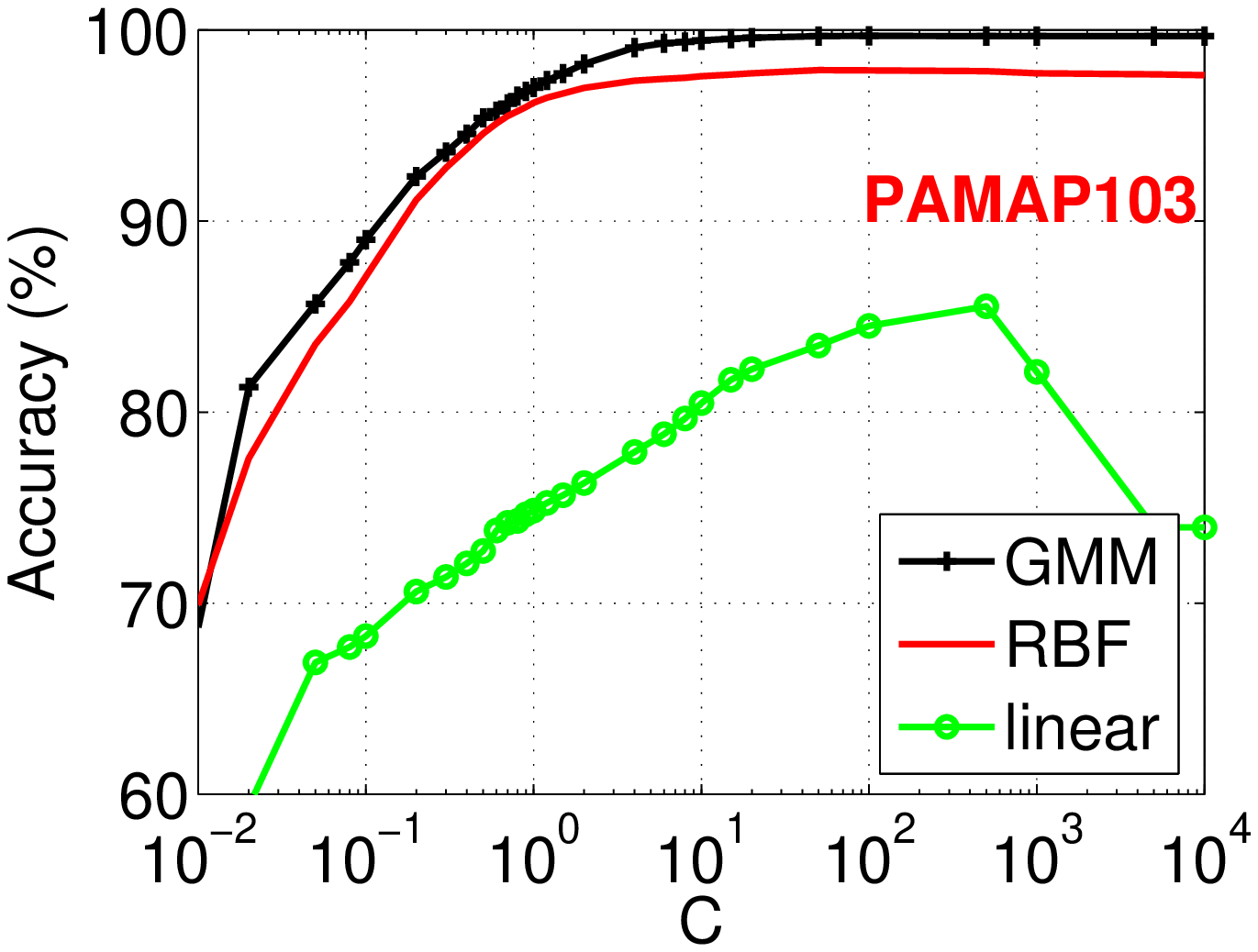}
\includegraphics[width=2.2in]{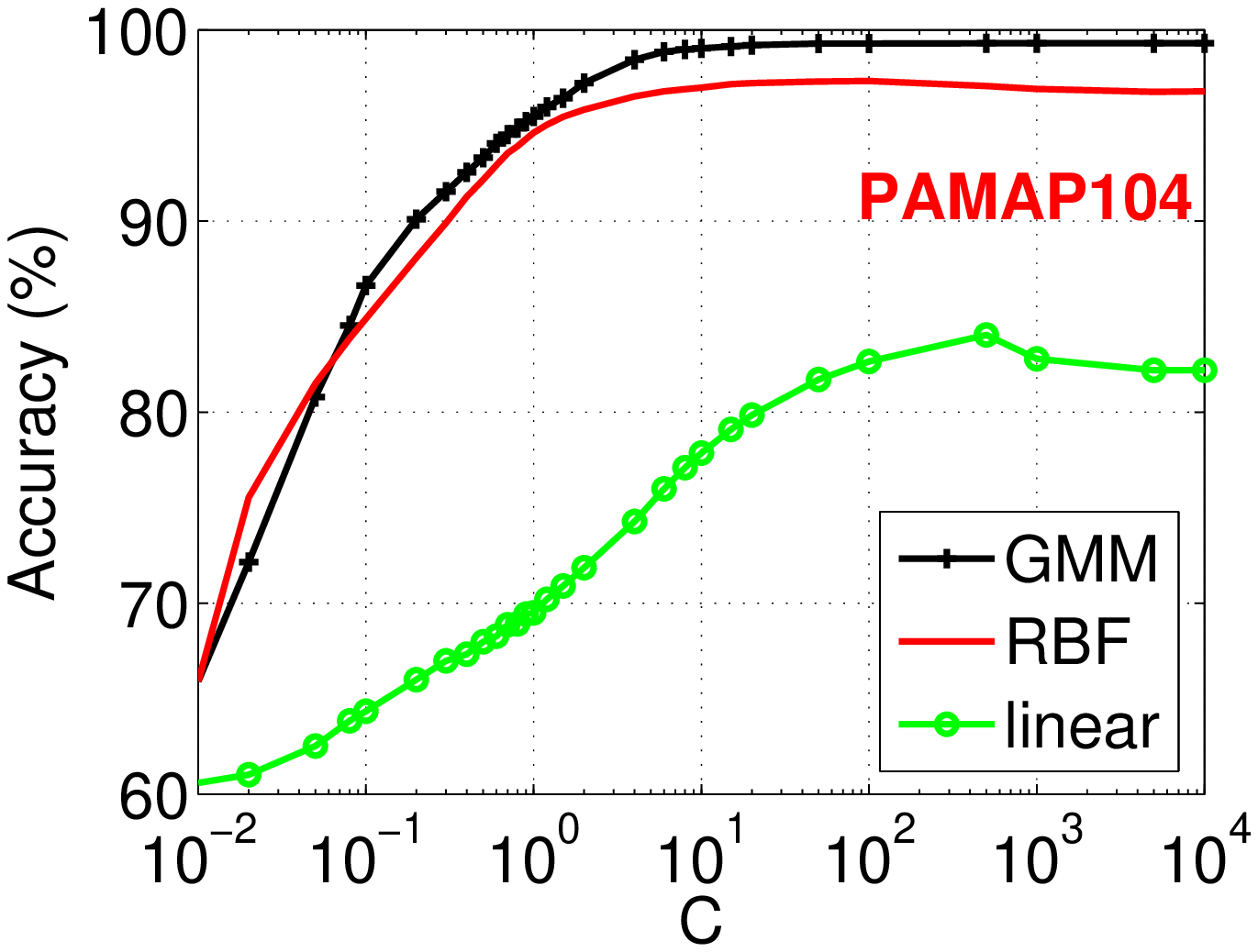}
}

\mbox{
\includegraphics[width=2.2in]{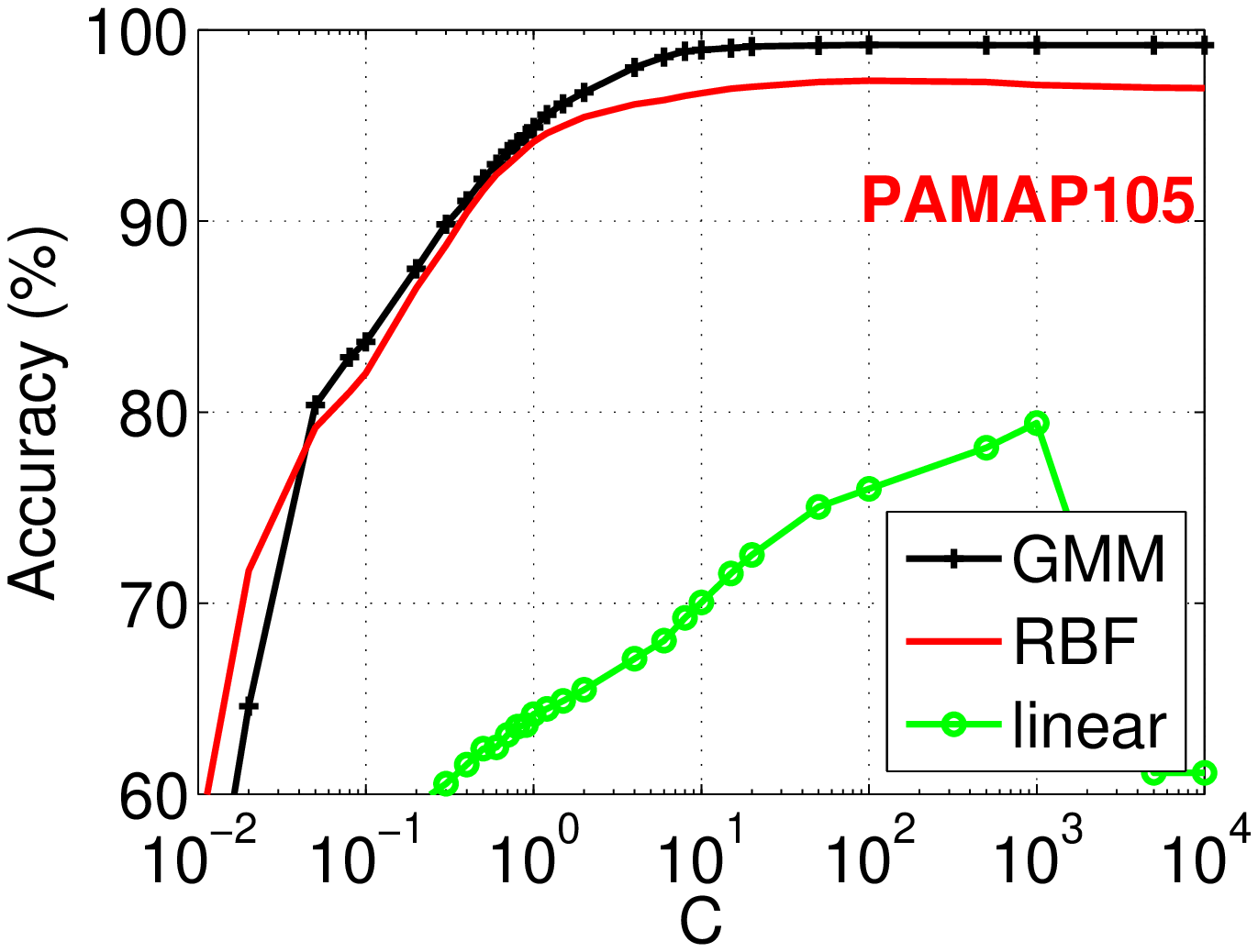}
\includegraphics[width=2.2in]{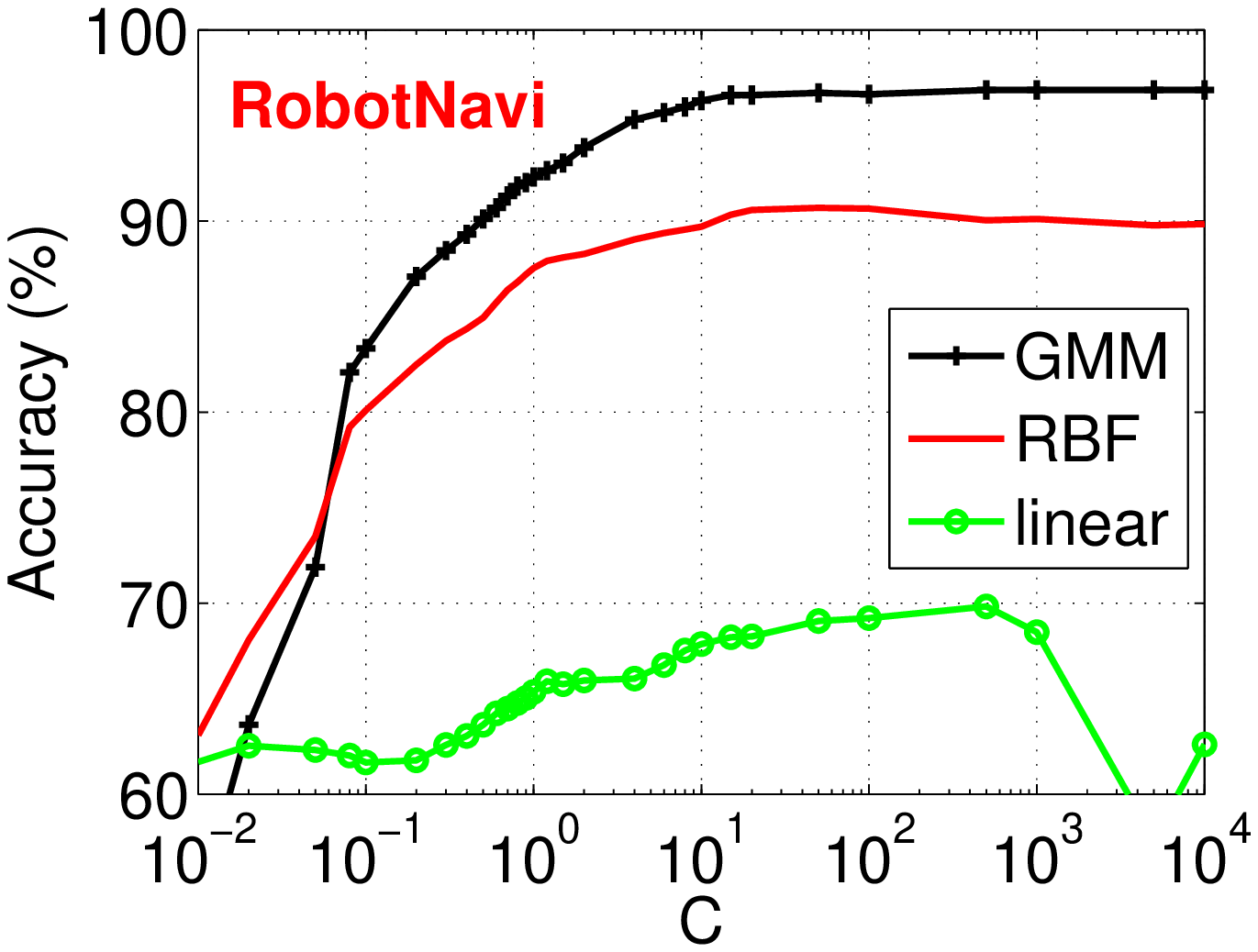}
\includegraphics[width=2.2in]{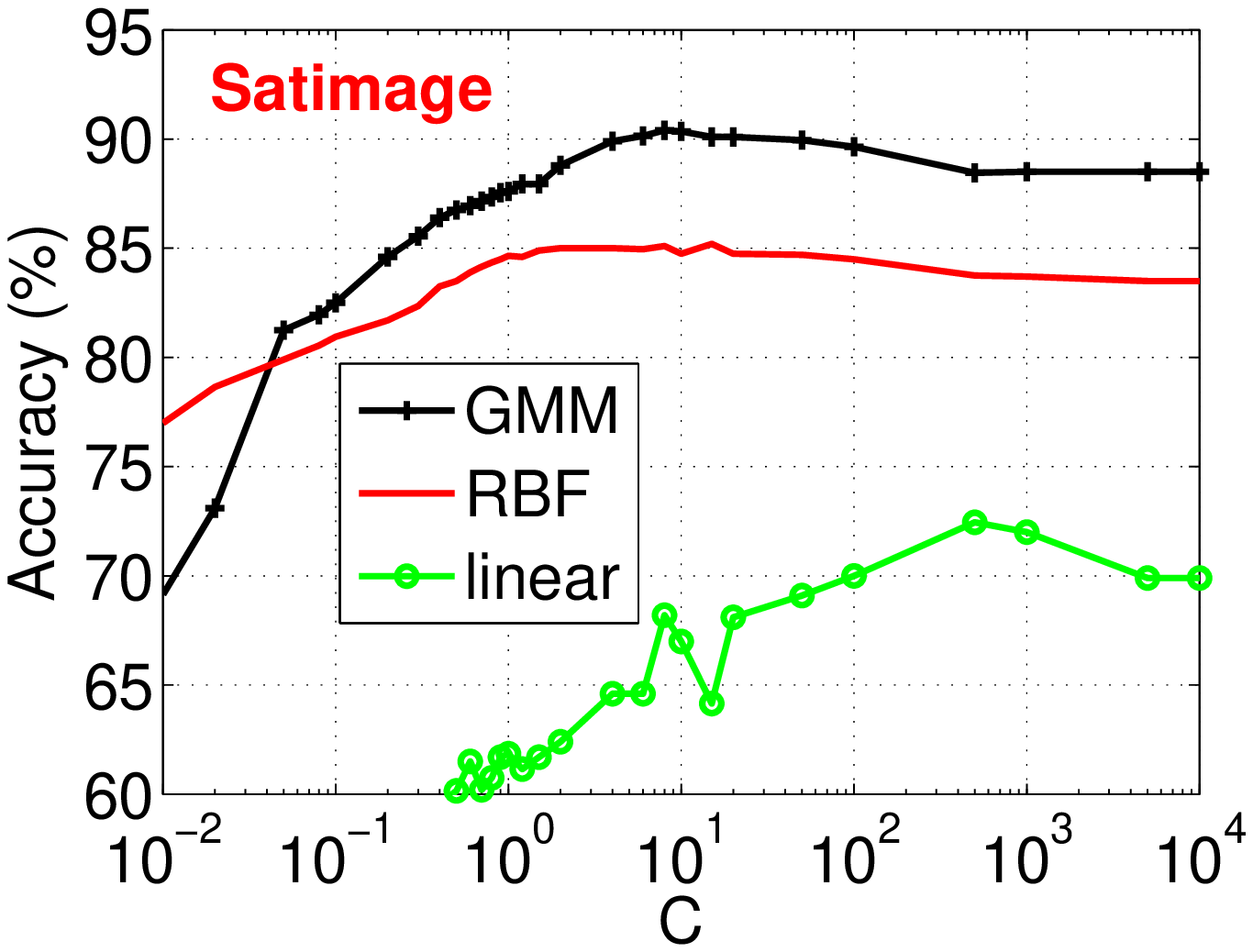}
}

\mbox{
\includegraphics[width=2.2in]{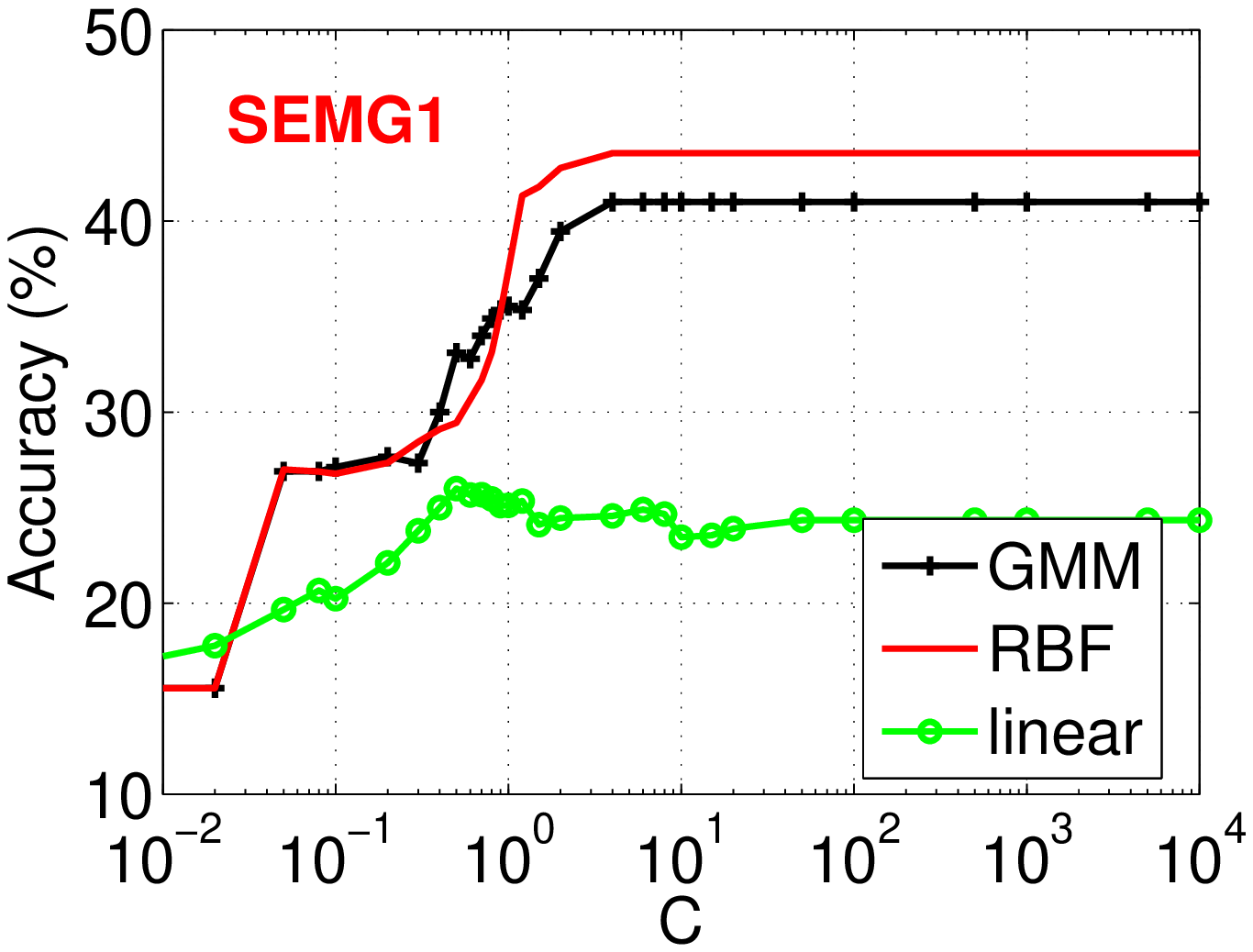}
\includegraphics[width=2.2in]{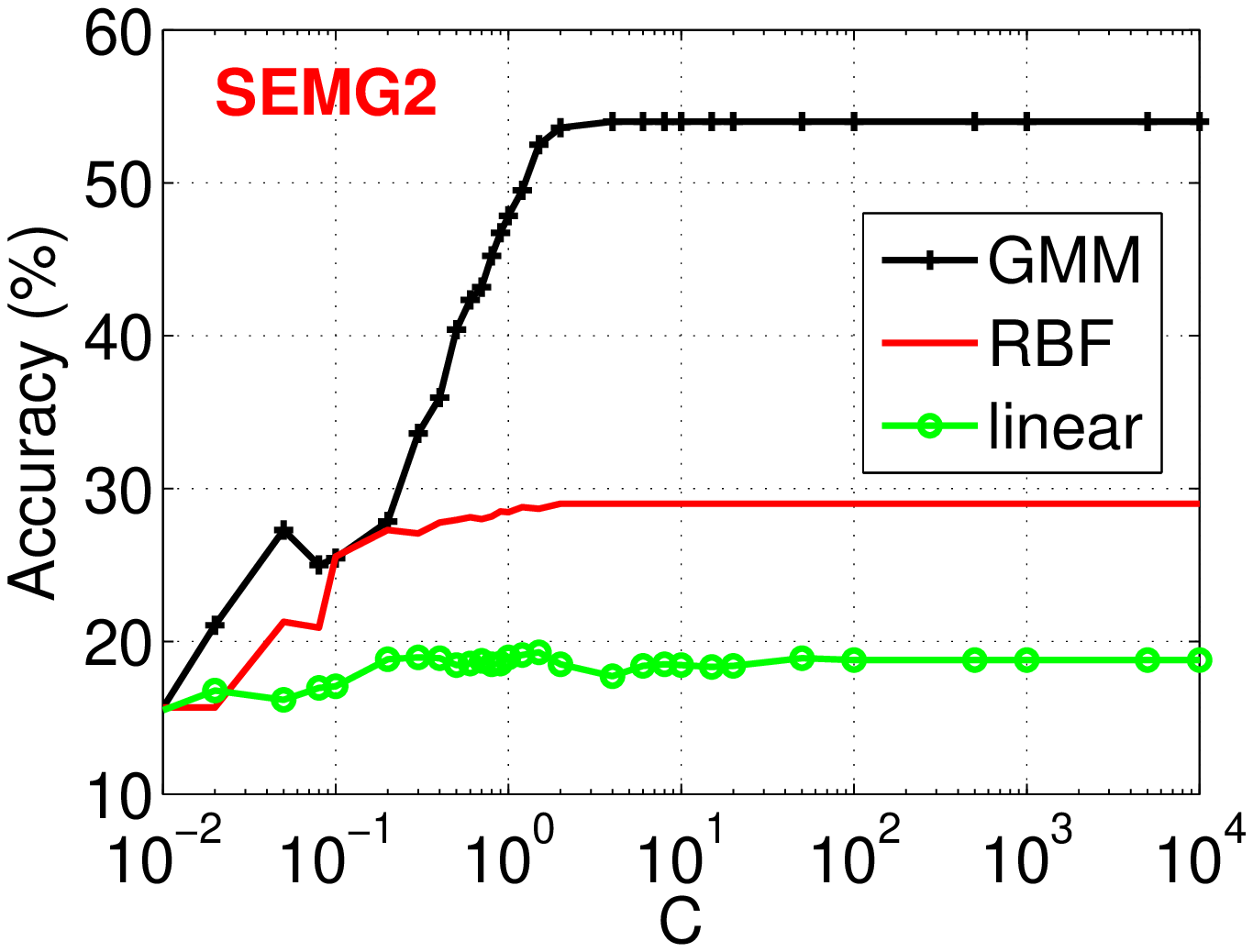}
\includegraphics[width=2.2in]{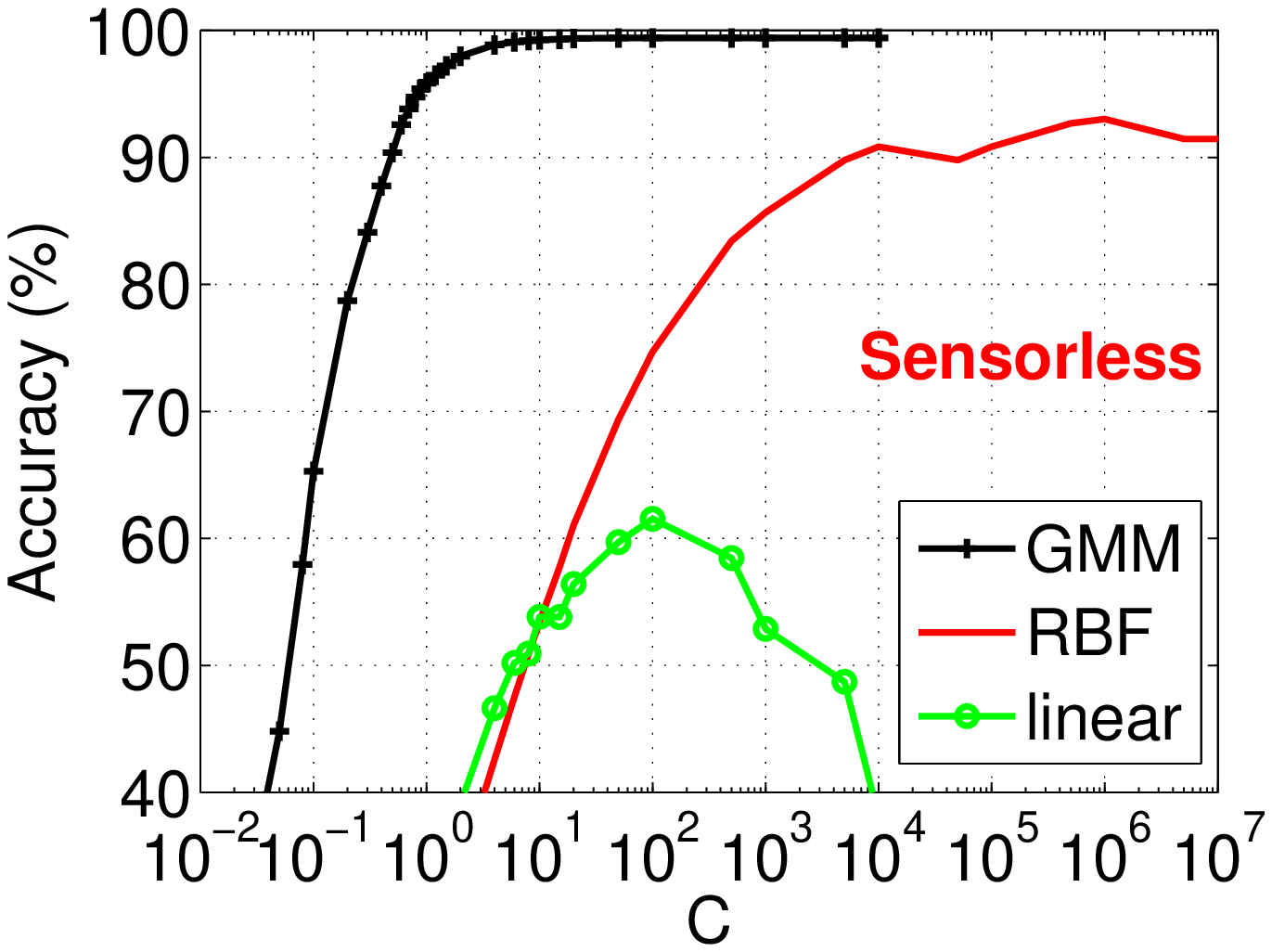}
}

\end{center}
\vspace{-0.3in}
\caption{\textbf{Test classification accuracies using kernel SVMs}. Both the GMM kernel and RBF kernel substantially improve linear SVM. $C$ is the $l_2$-regularization parameter of SVM. For the RBF kernel, we report the result at the best $\gamma$ value for every $C$ value.  }\label{fig_KernelSVM2}
\end{figure}

\begin{figure}[h!]
\begin{center}

\mbox{
\includegraphics[width=2.2in]{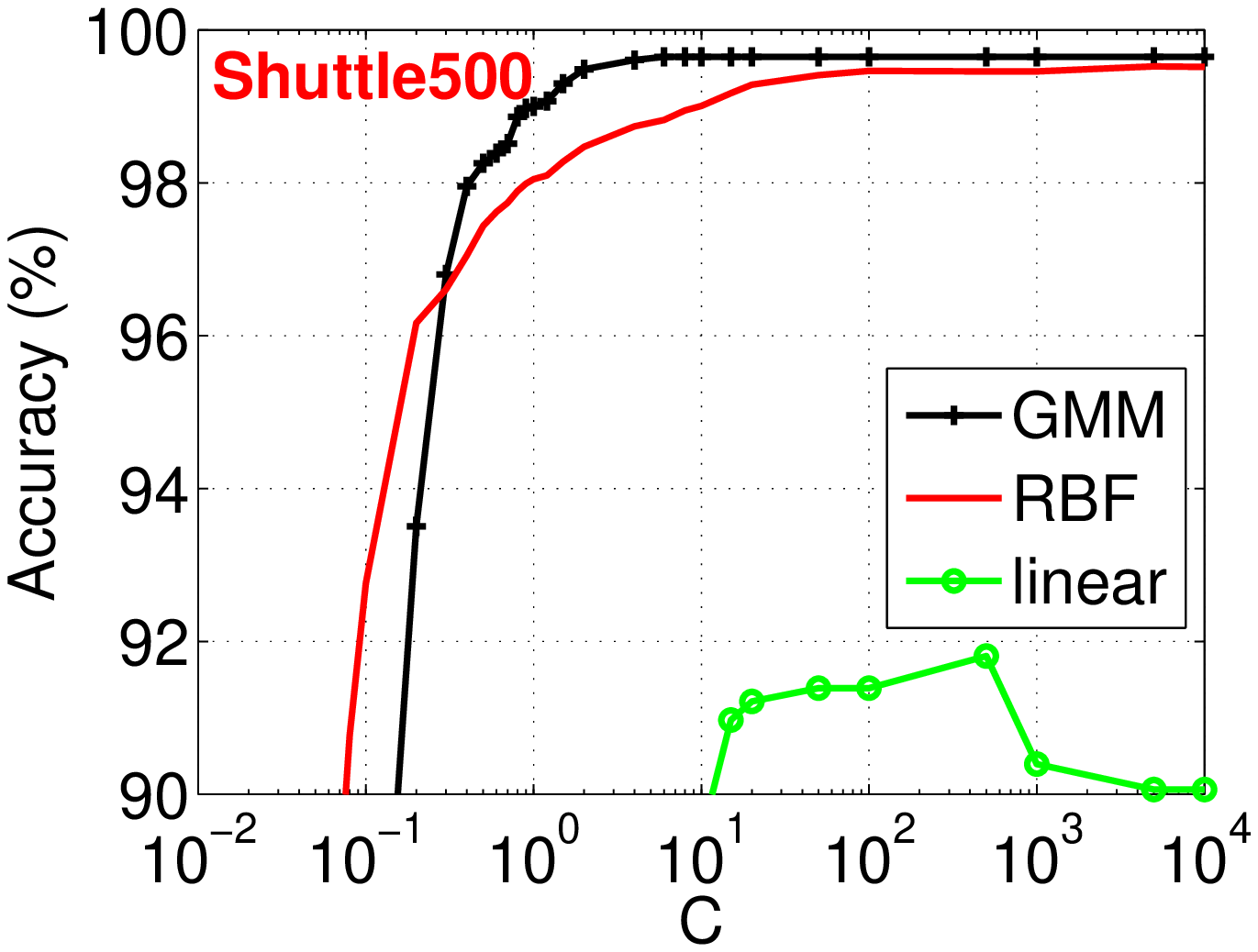}
\includegraphics[width=2.2in]{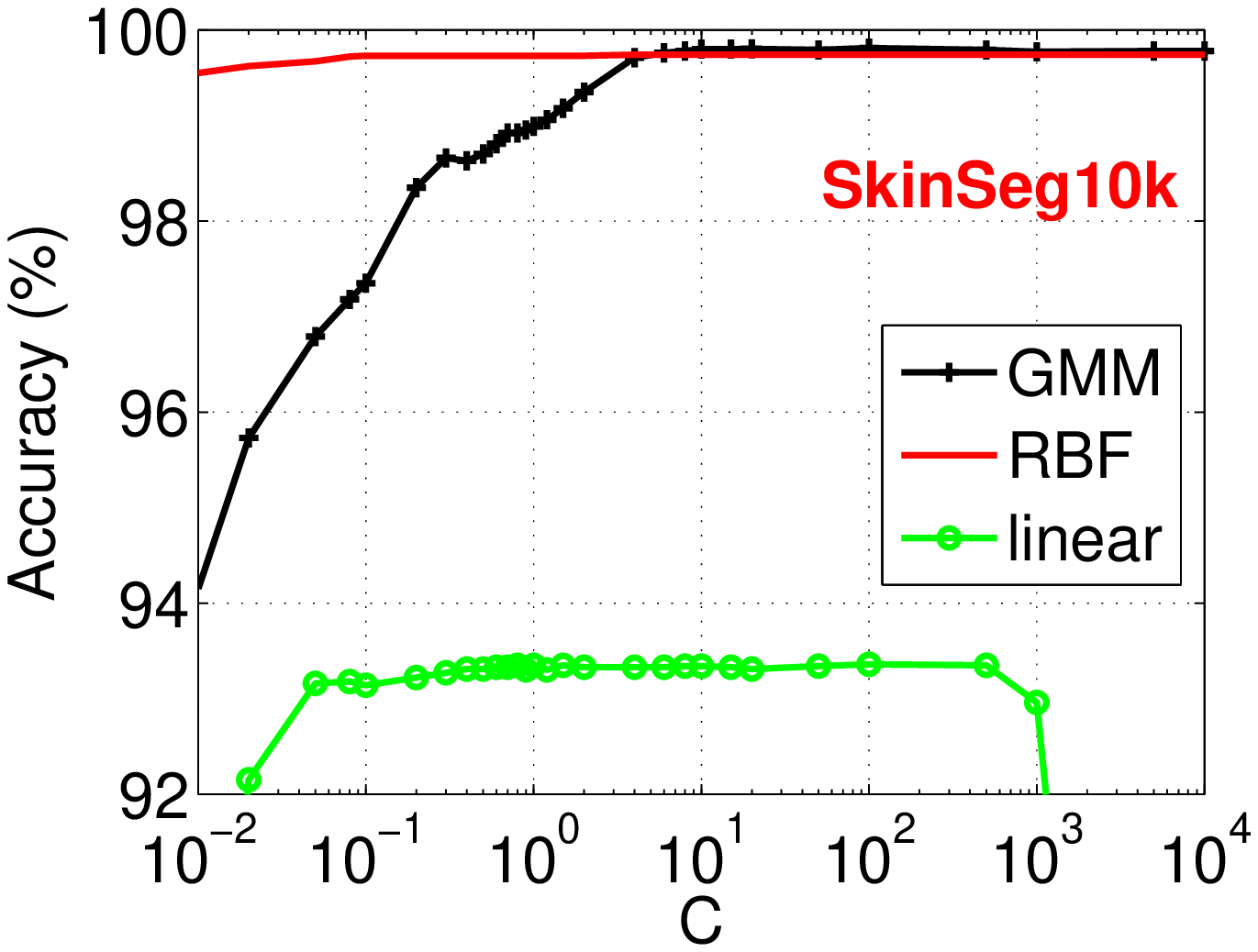}
\includegraphics[width=2.2in]{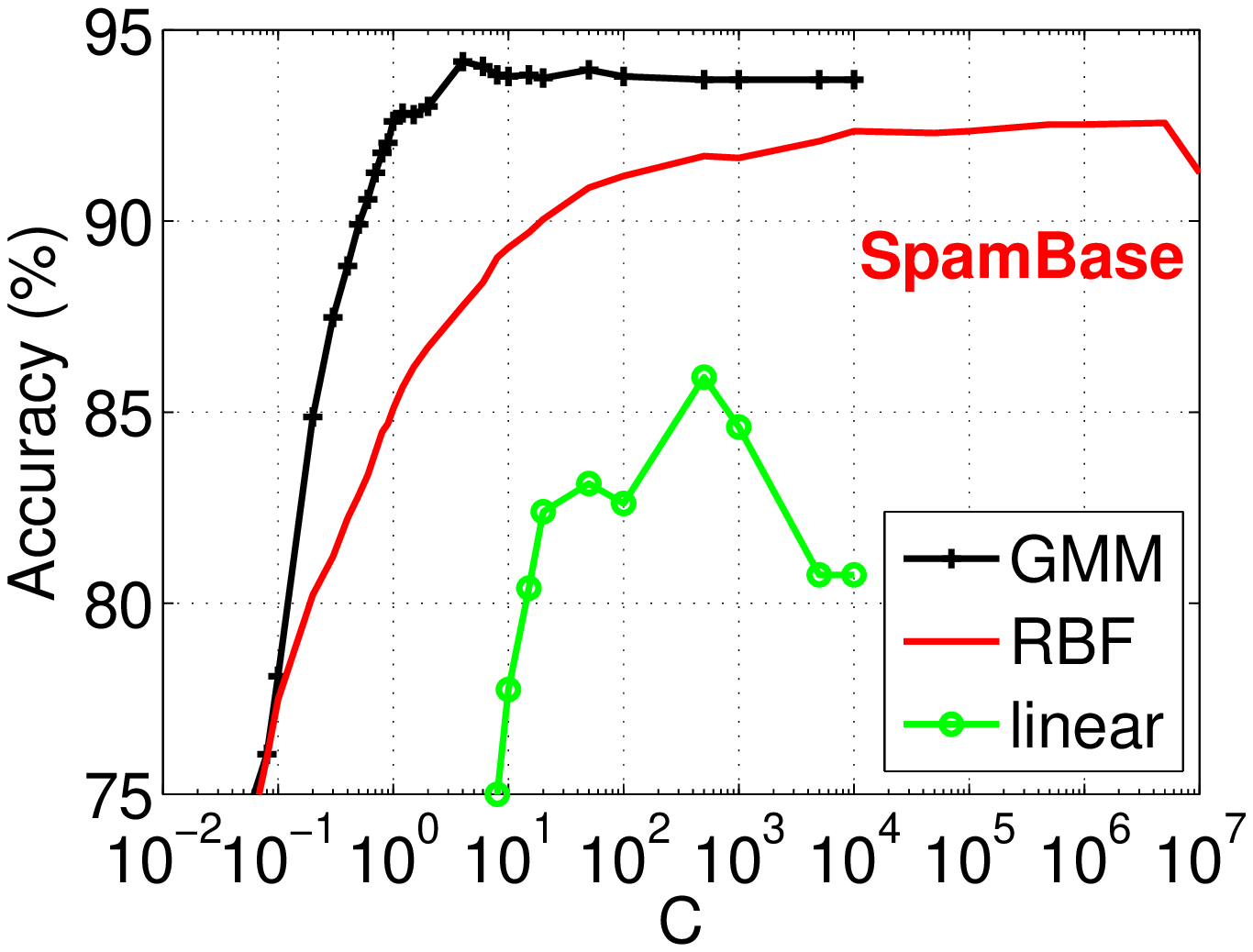}
}

\mbox{
\includegraphics[width=2.2in]{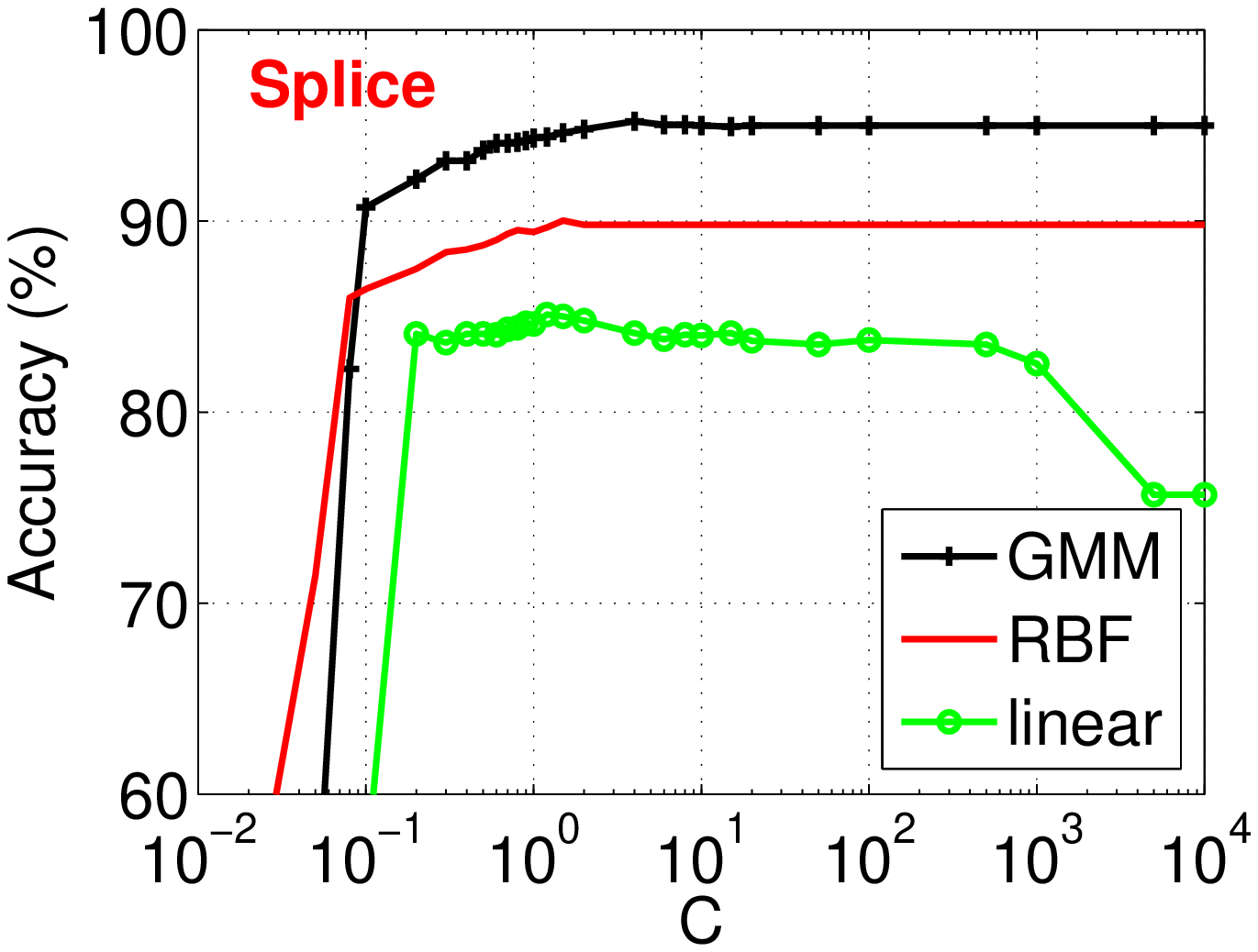}
\includegraphics[width=2.2in]{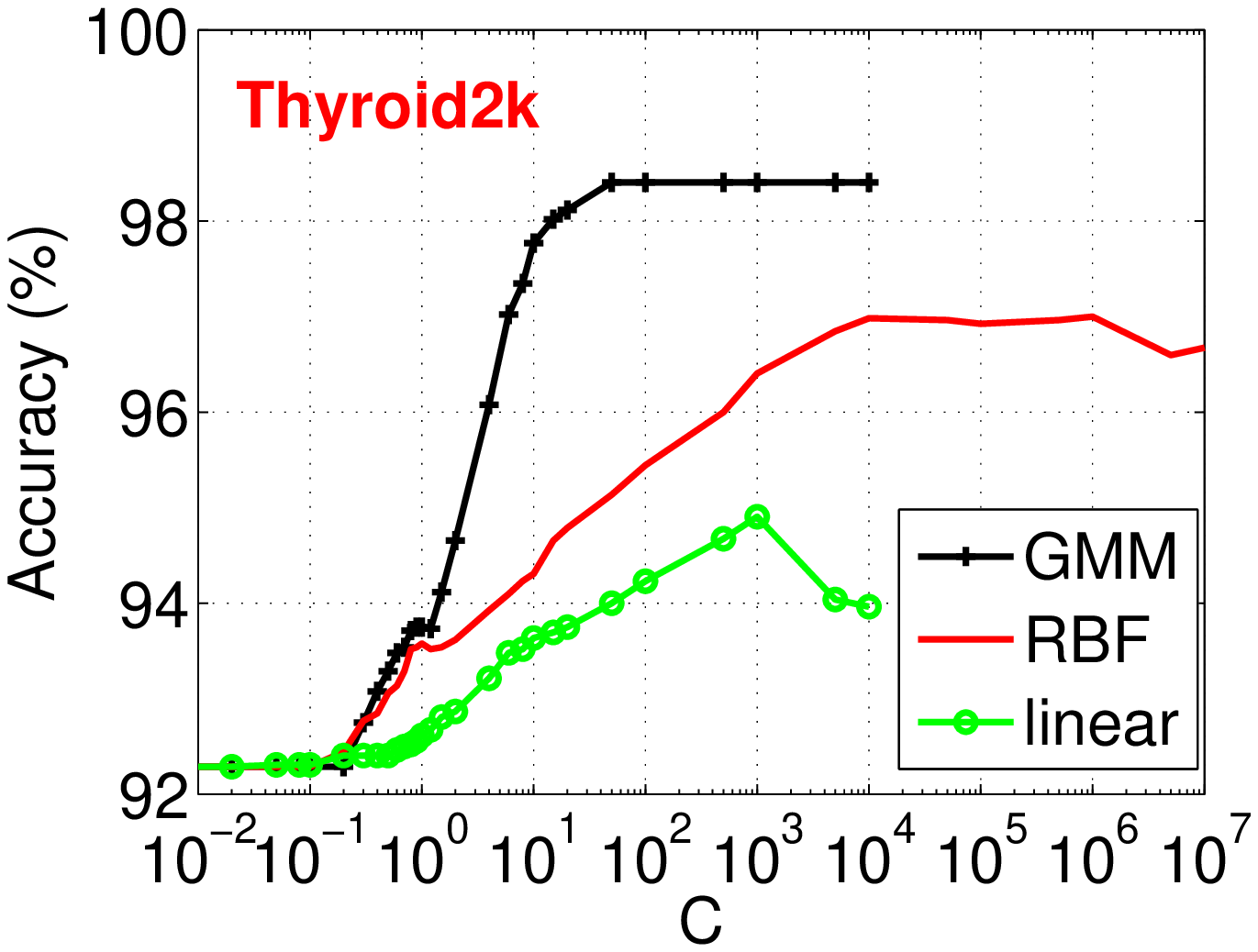}
\includegraphics[width=2.2in]{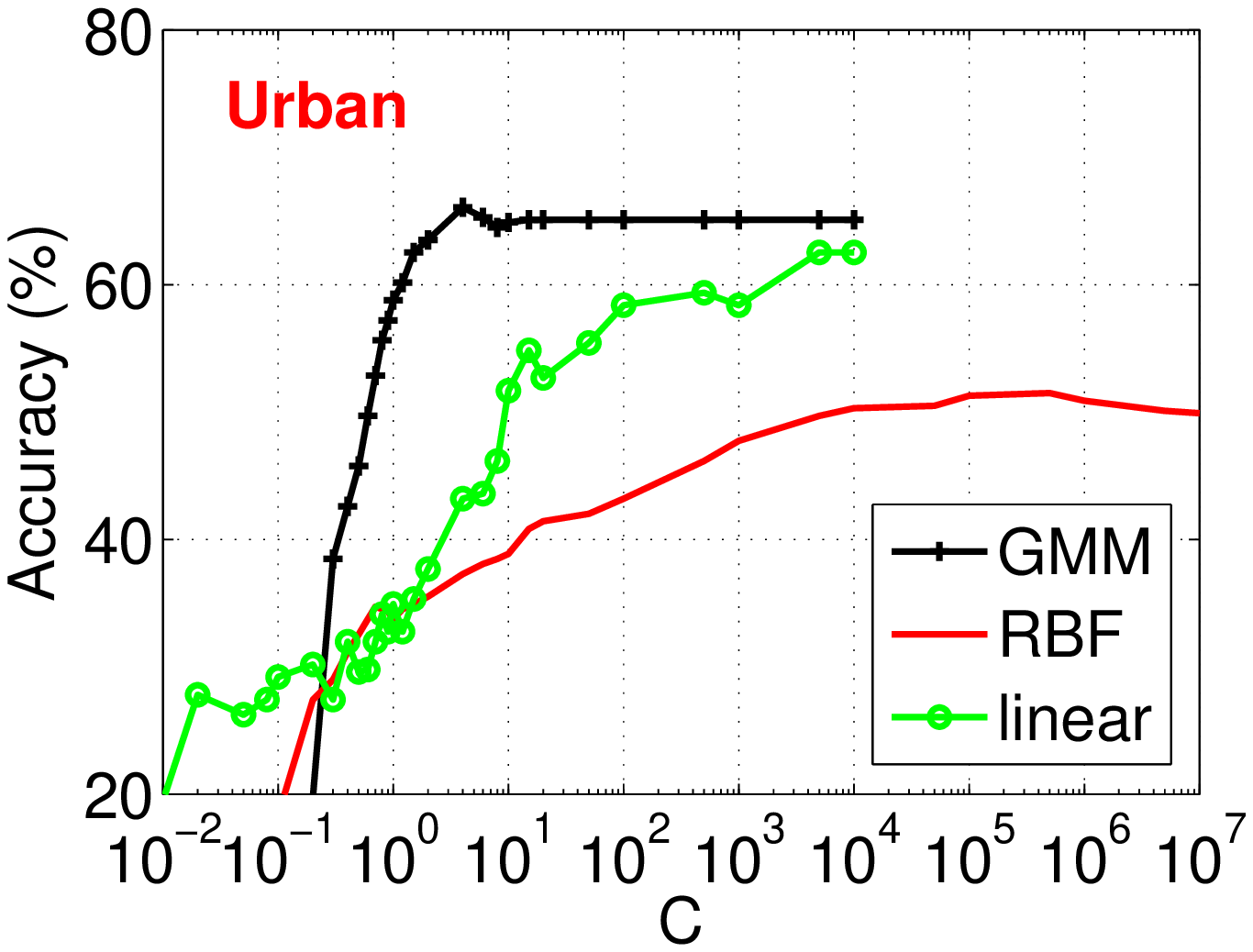}
}

\mbox{
\includegraphics[width=2.2in]{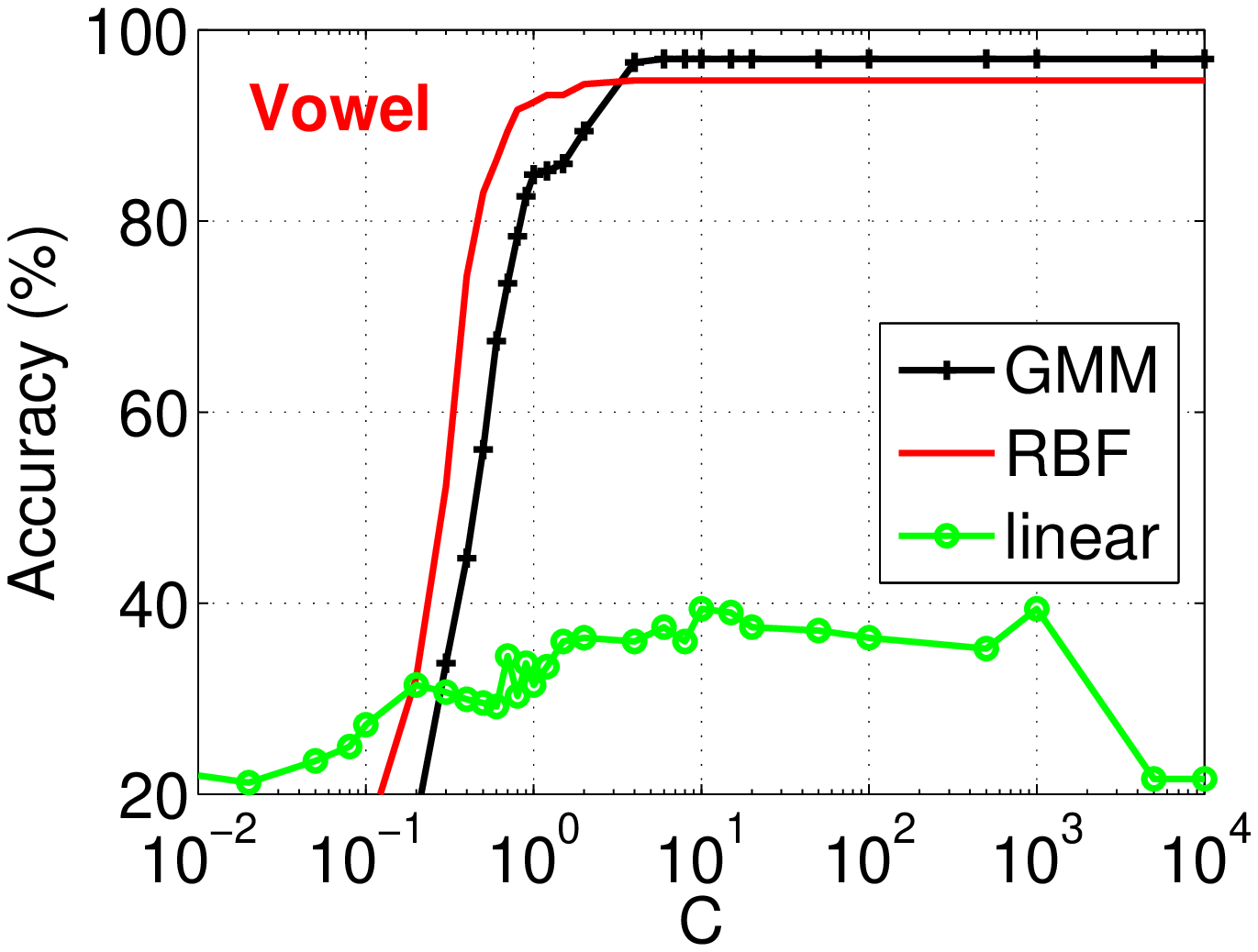}
\includegraphics[width=2.2in]{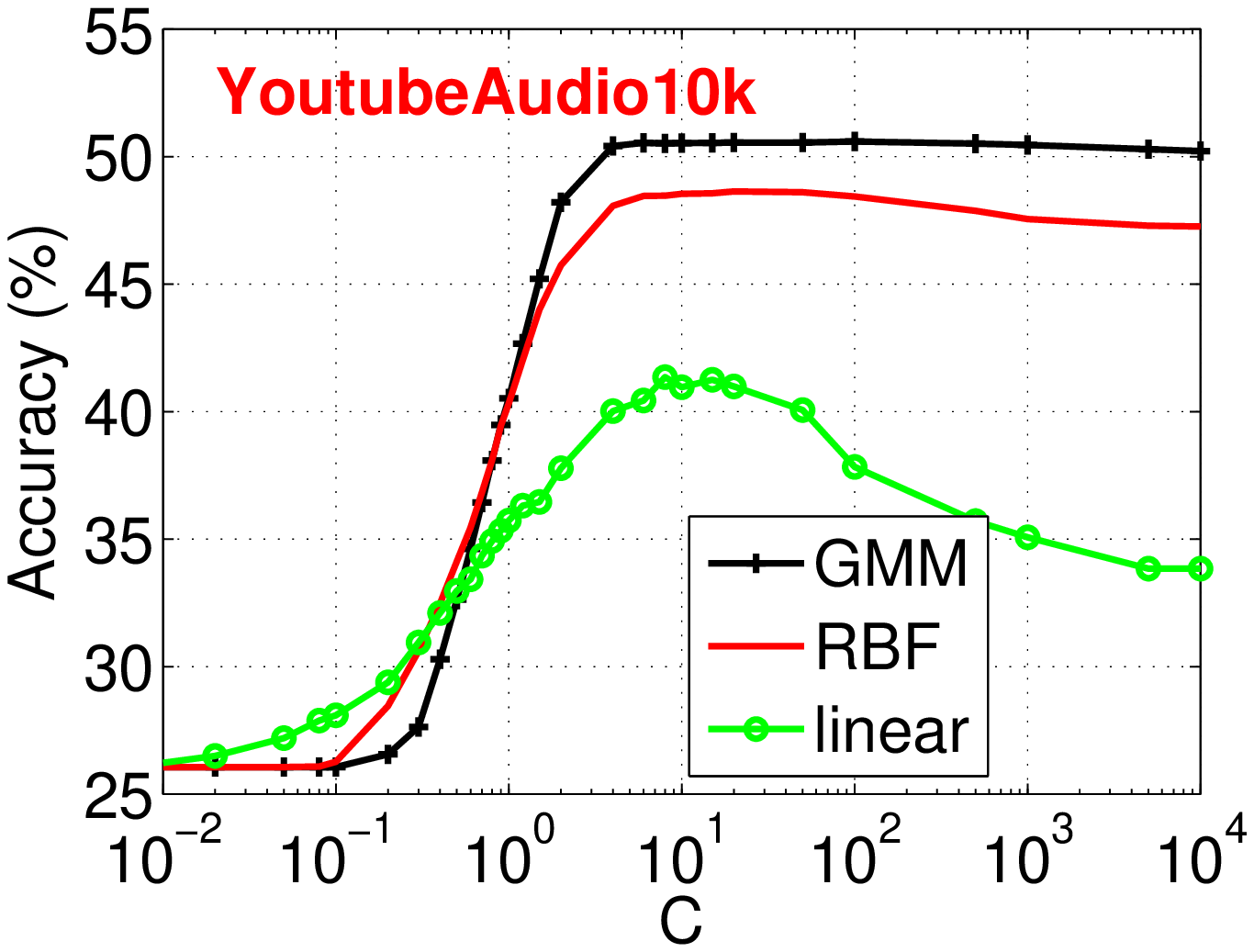}
\includegraphics[width=2.2in]{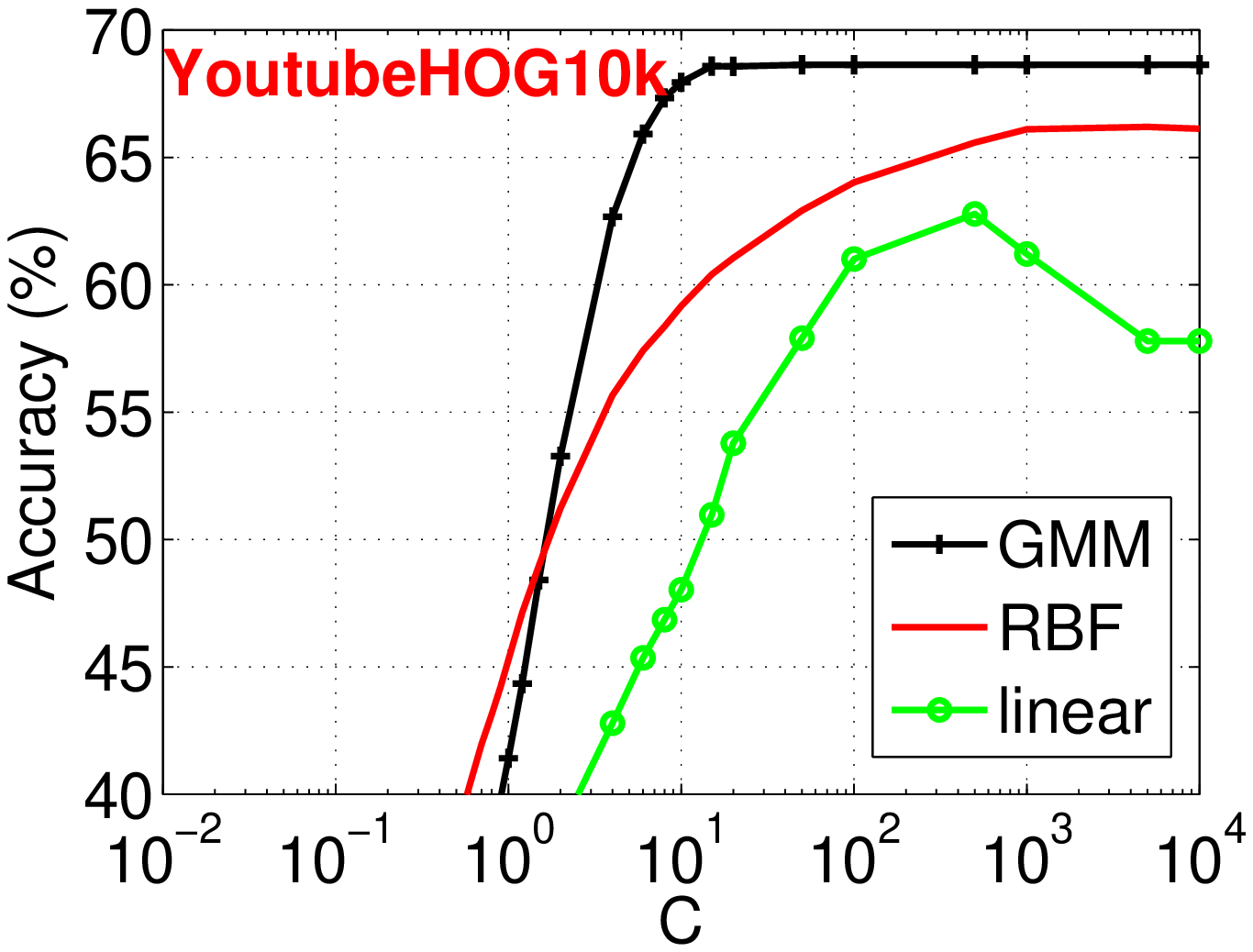}
}

\mbox{
\includegraphics[width=2.2in]{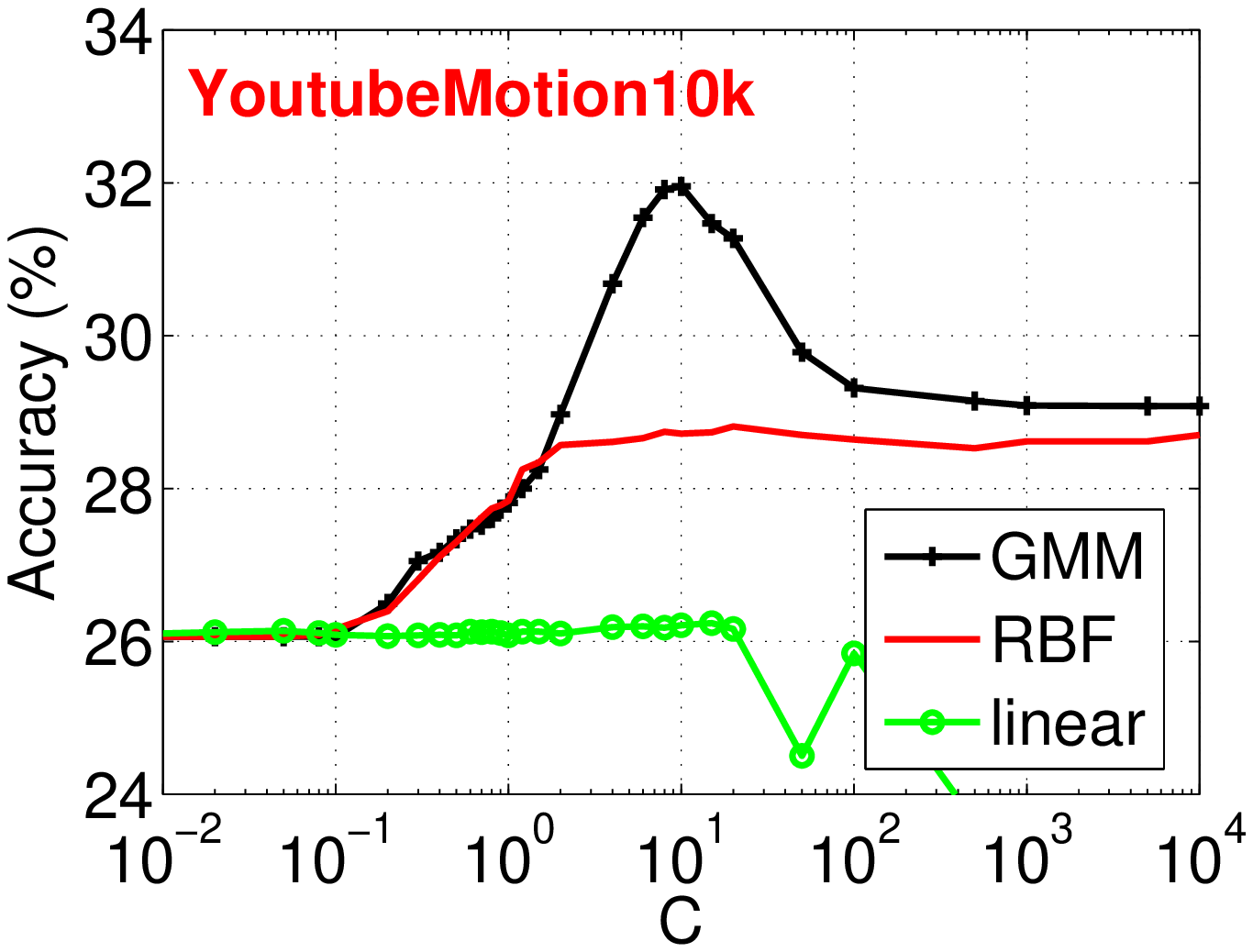}
\includegraphics[width=2.2in]{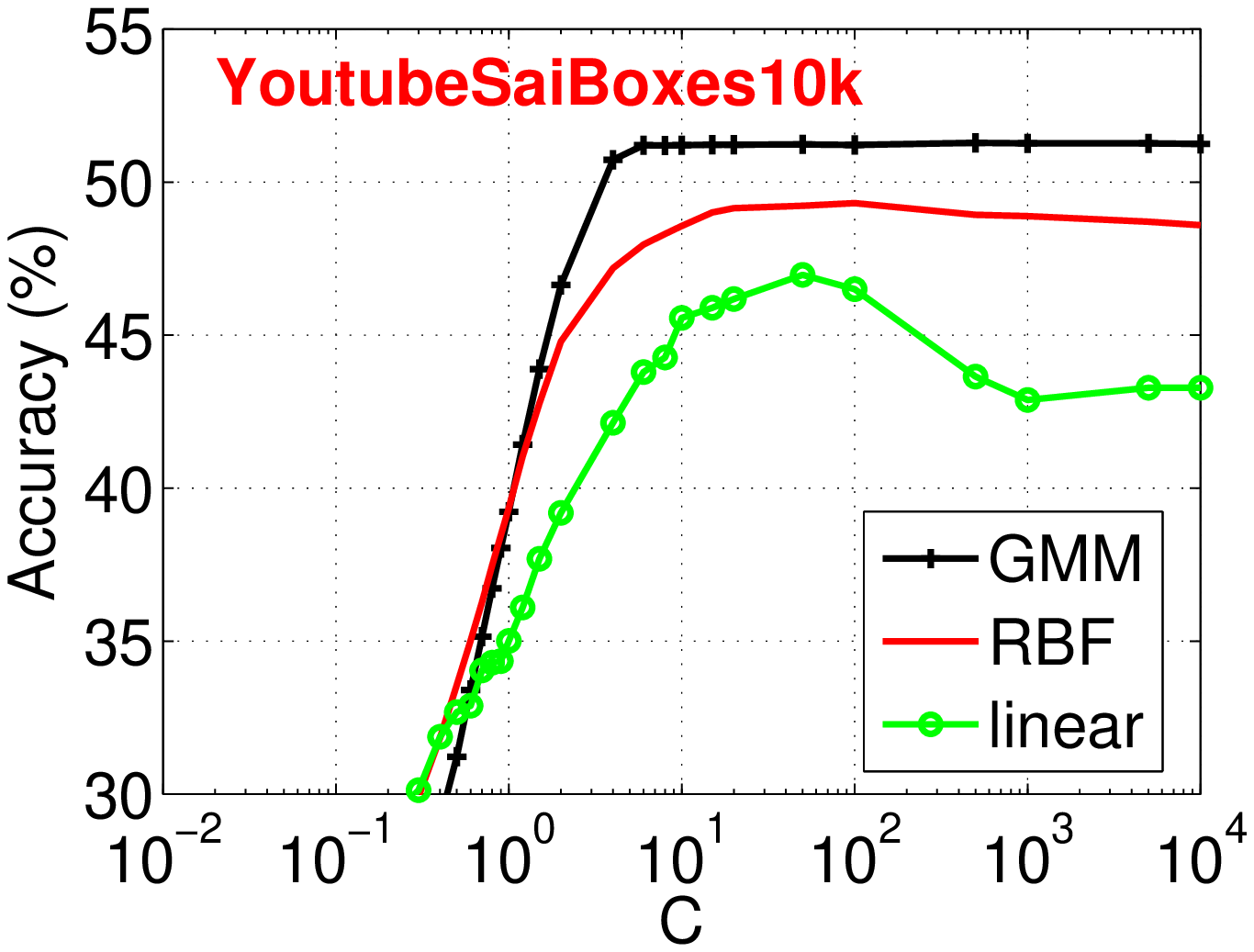}
\includegraphics[width=2.2in]{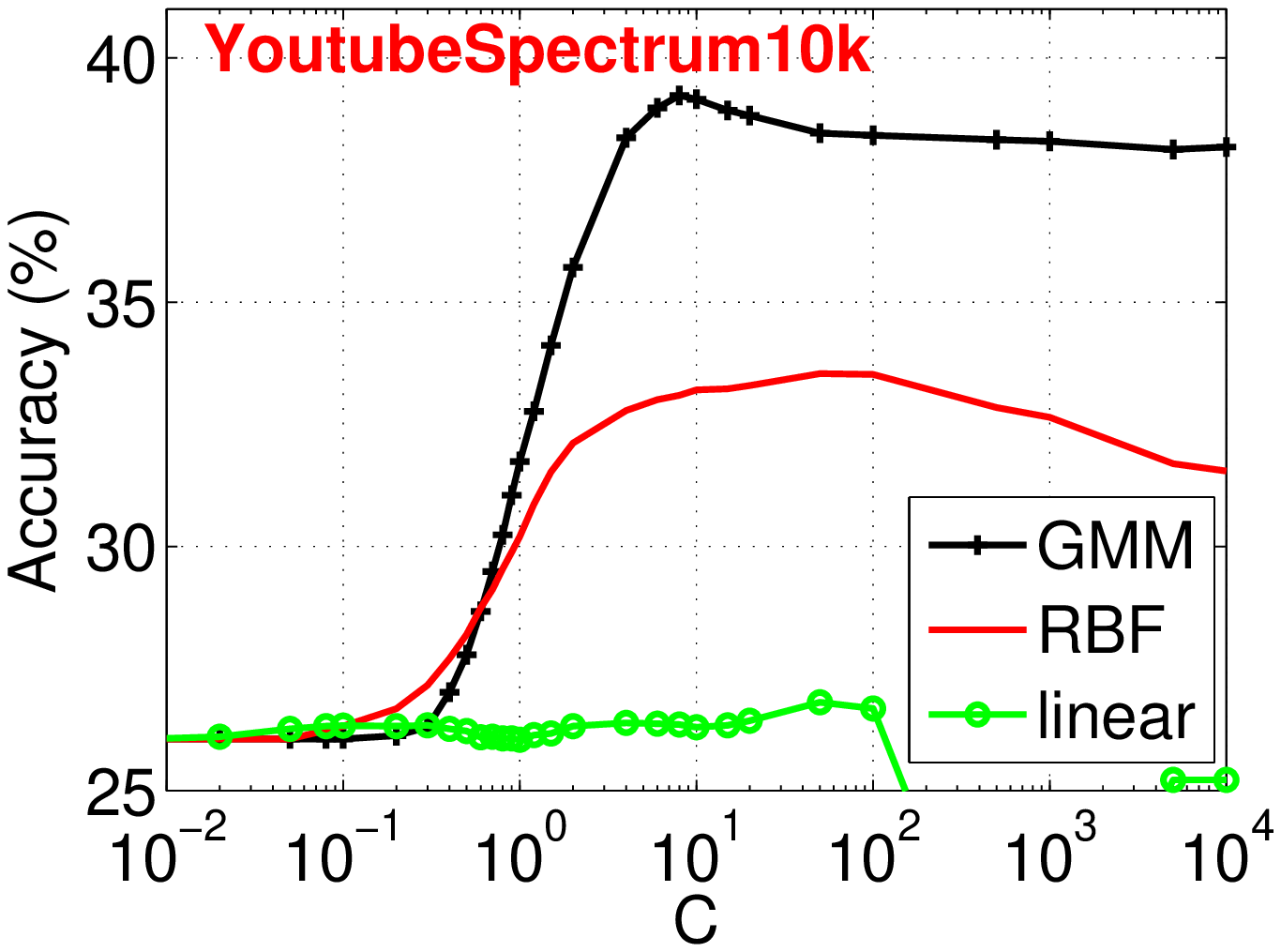}
}

\end{center}
\vspace{-0.3in}
\caption{\textbf{Test classification accuracies using kernel SVMs}. Both the GMM kernel and RBF kernel substantially improve linear SVM. $C$ is the $l_2$-regularization parameter of SVM. For the RBF kernel, we report the result at the best $\gamma$ value for every $C$ value.  }\label{fig_KernelSVM3}
\end{figure}

\begin{figure}[h!]

\begin{center}

\mbox{
\includegraphics[width=2.2in]{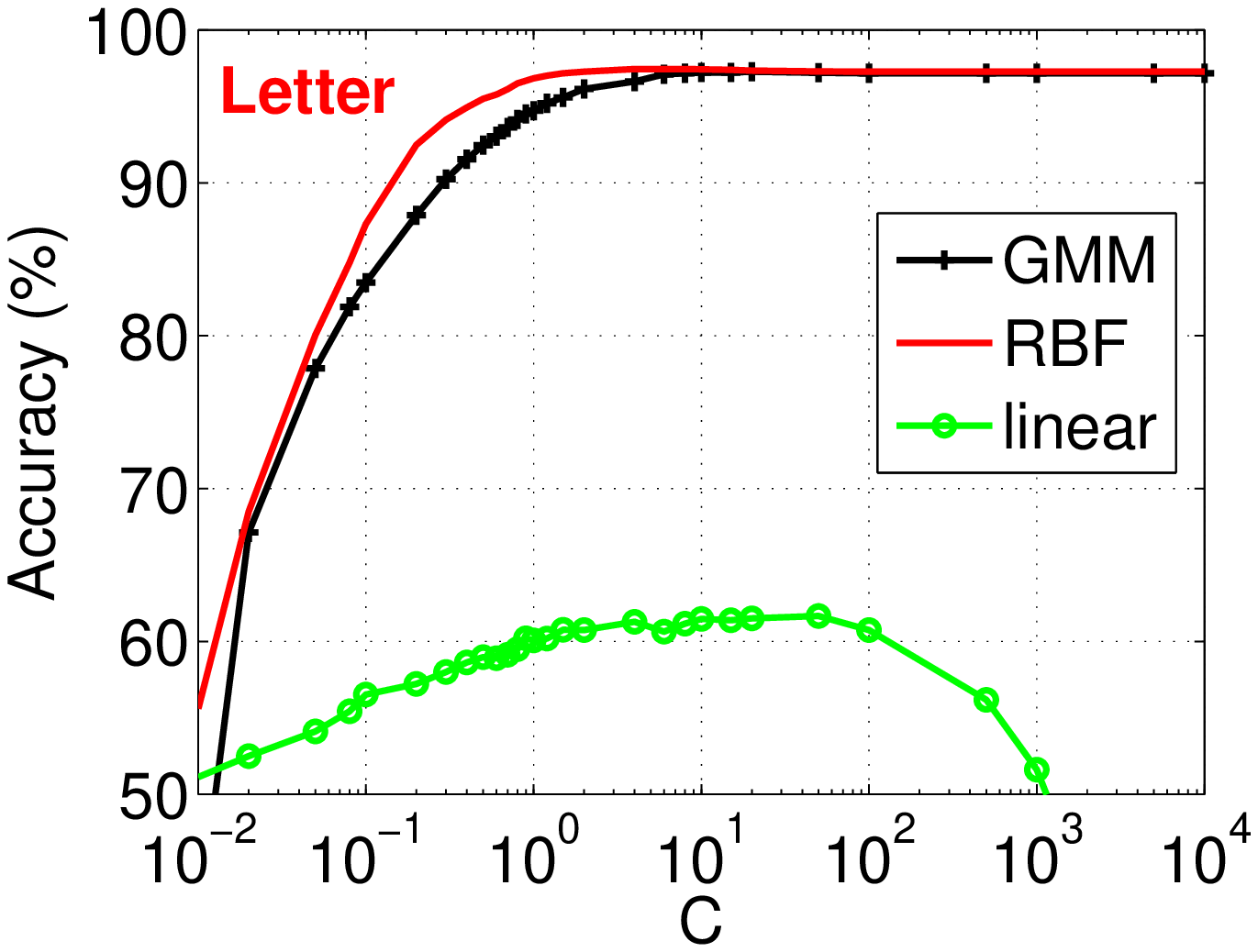}
\includegraphics[width=2.2in]{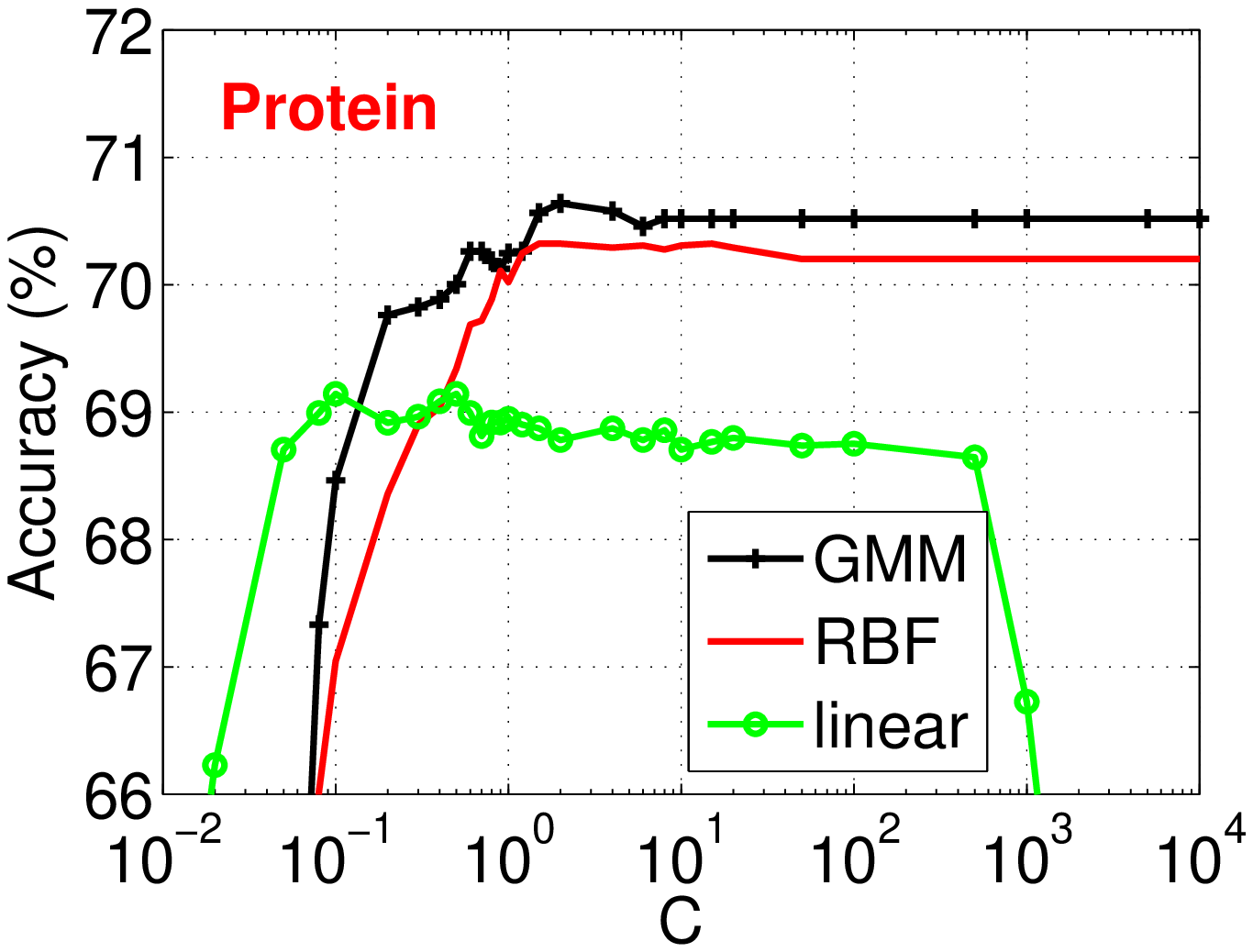}
\includegraphics[width=2.2in]{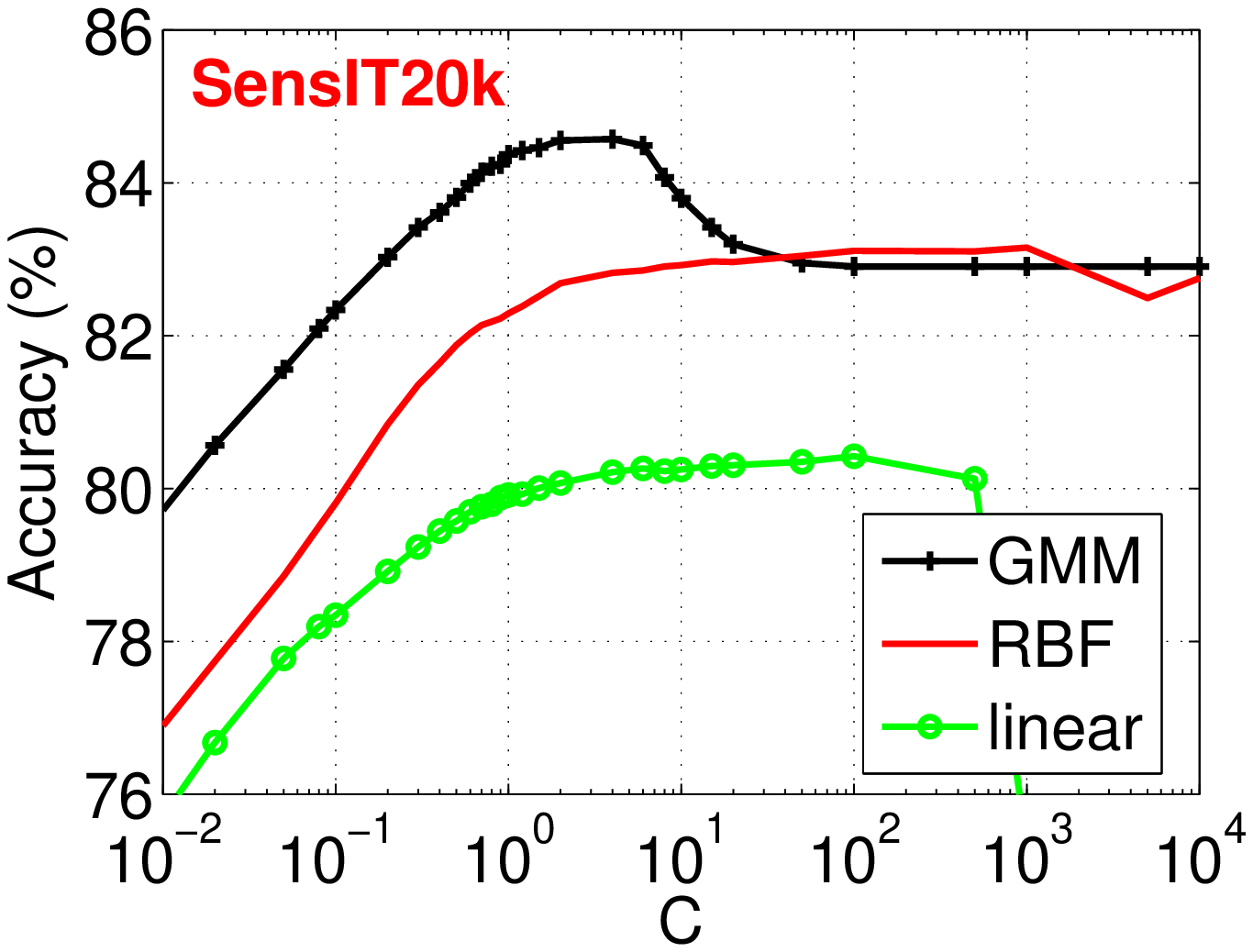}
}

\mbox{
\includegraphics[width=2.2in]{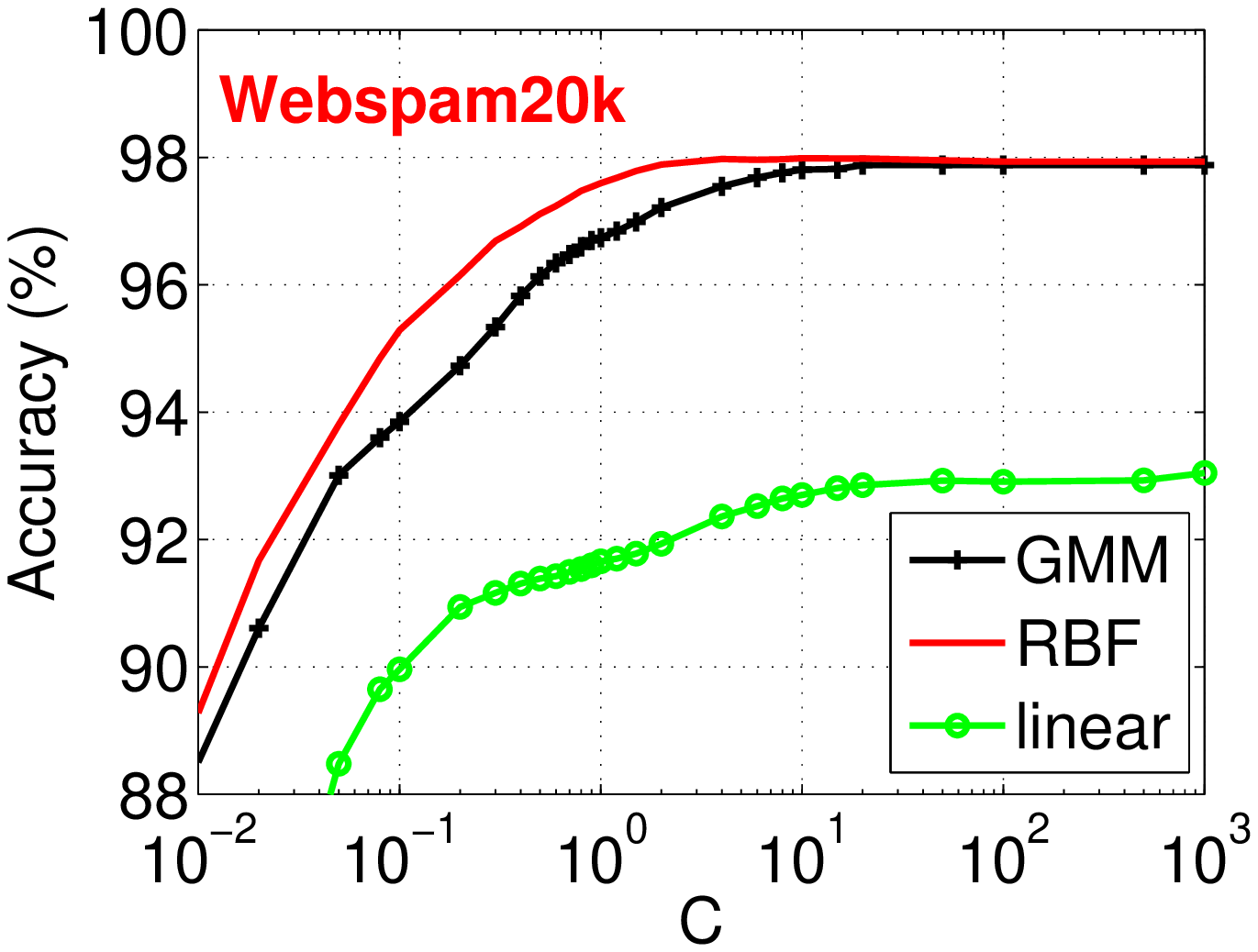}
\includegraphics[width=2.2in]{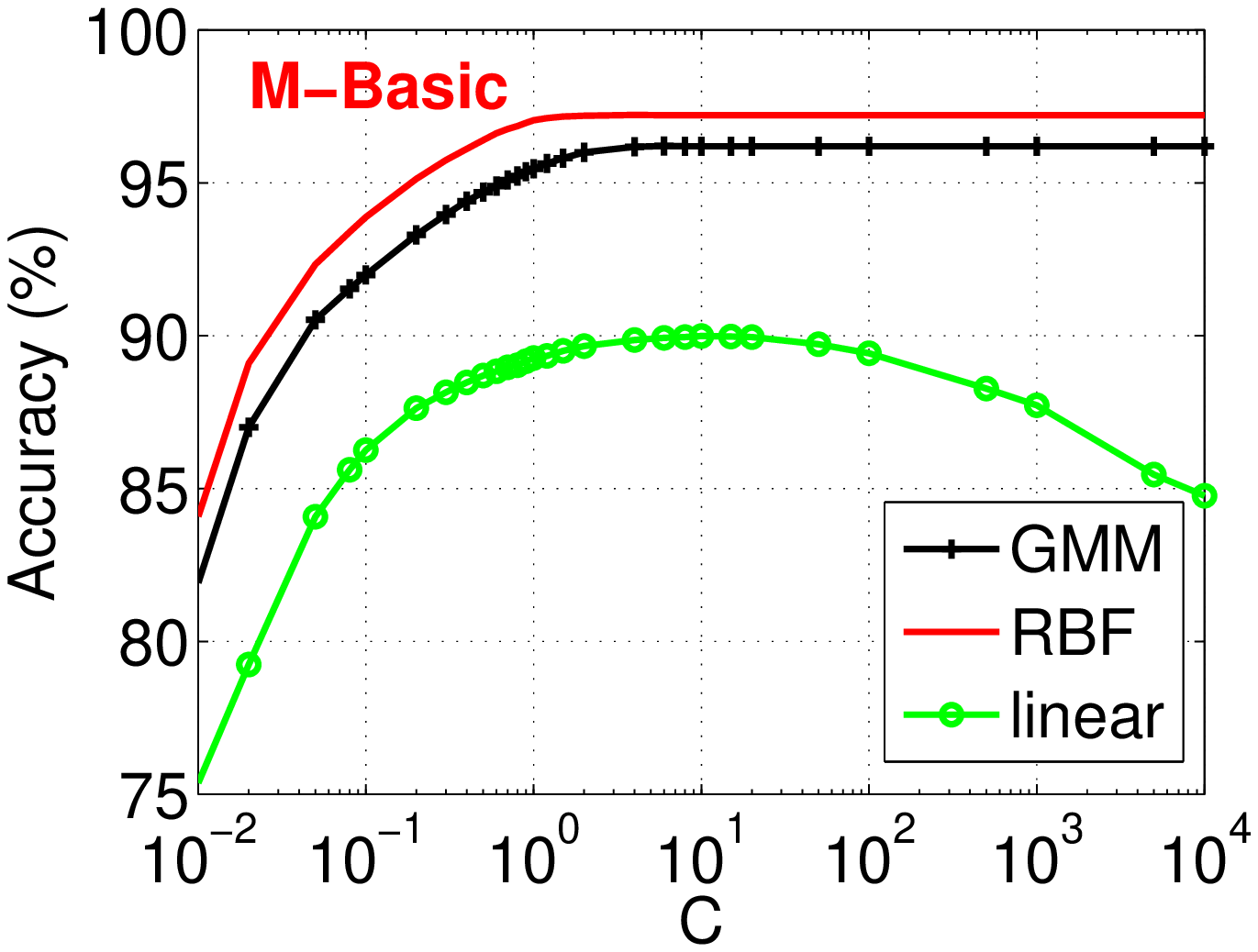}
\includegraphics[width=2.2in]{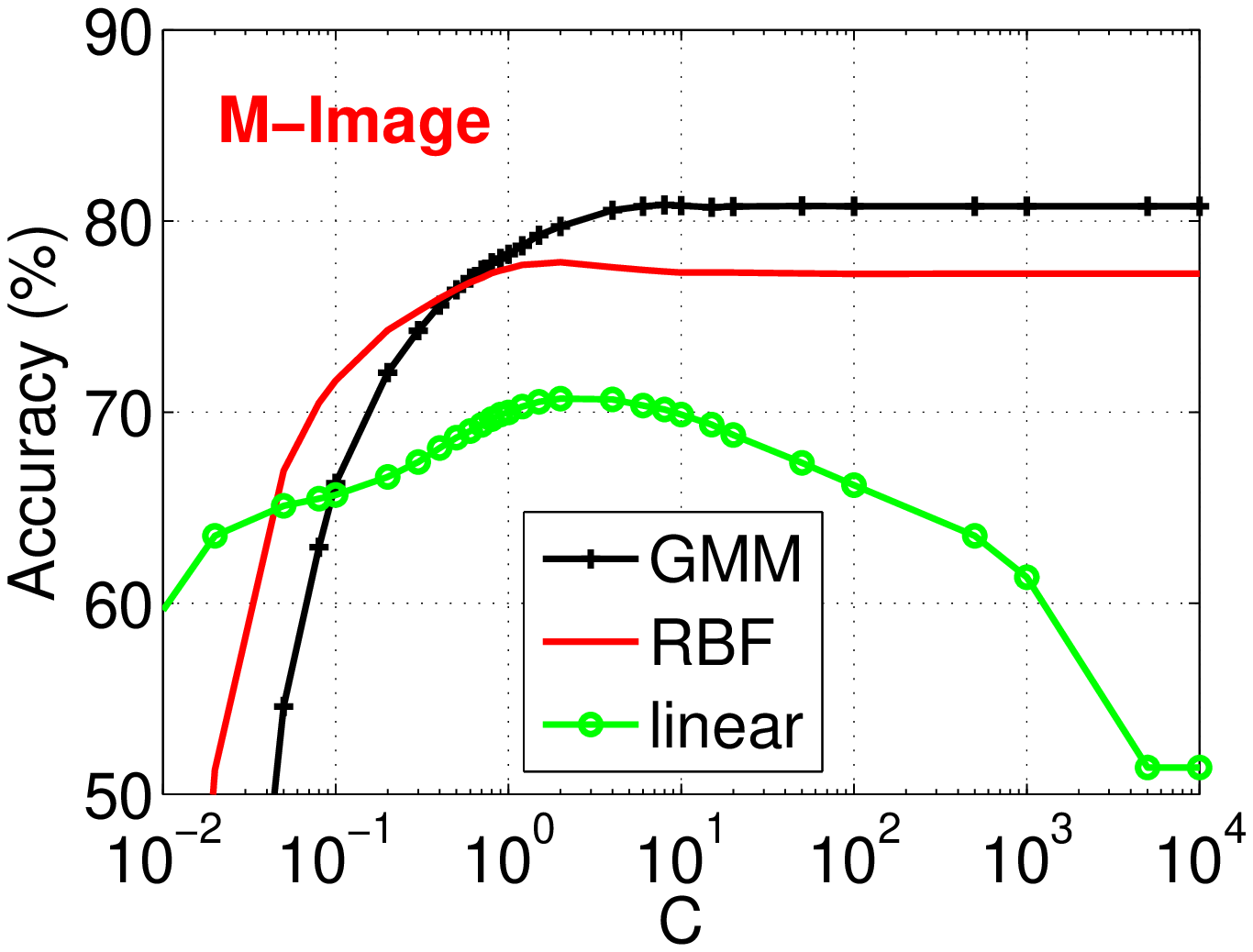}
}

\mbox{
\includegraphics[width=2.2in]{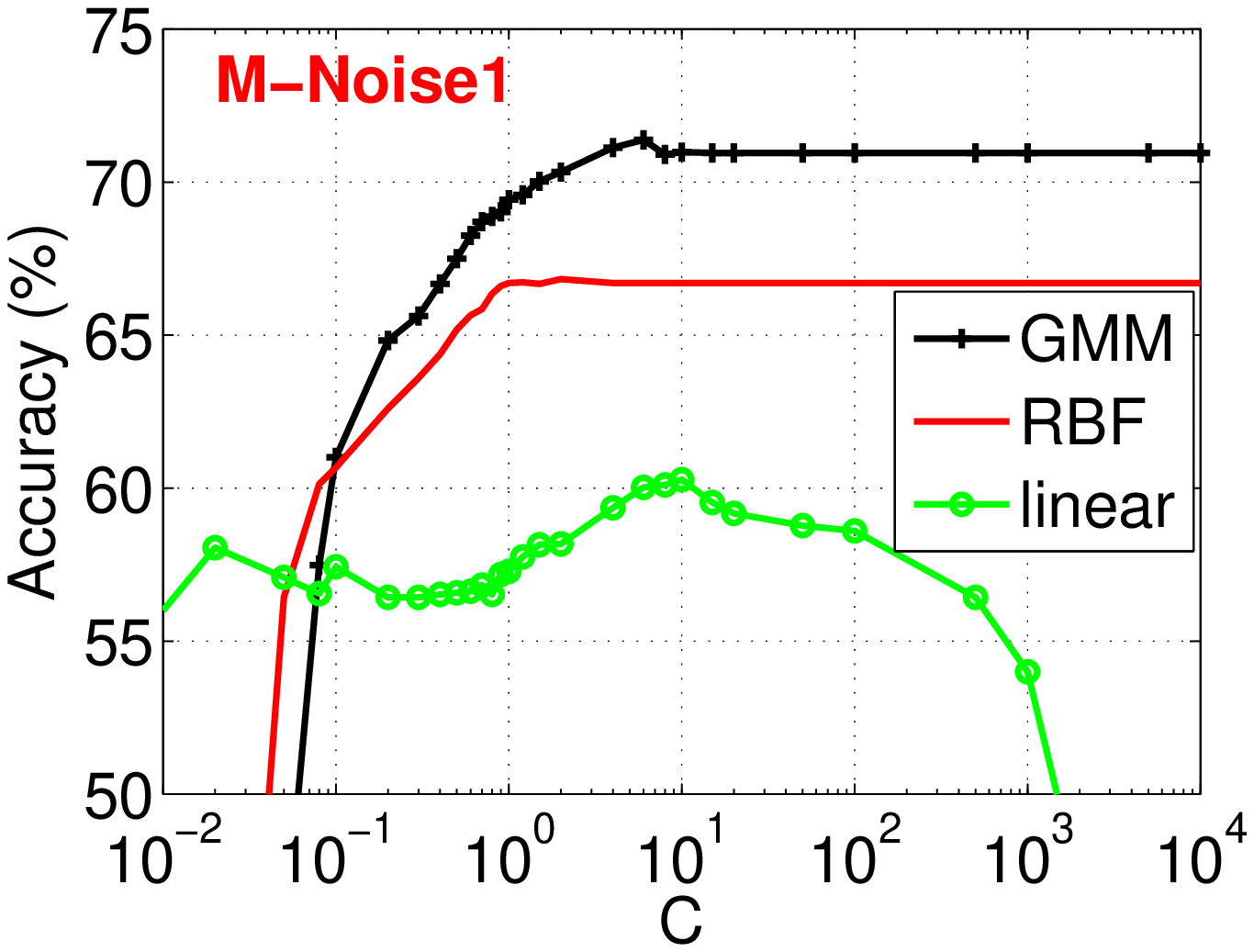}
\includegraphics[width=2.2in]{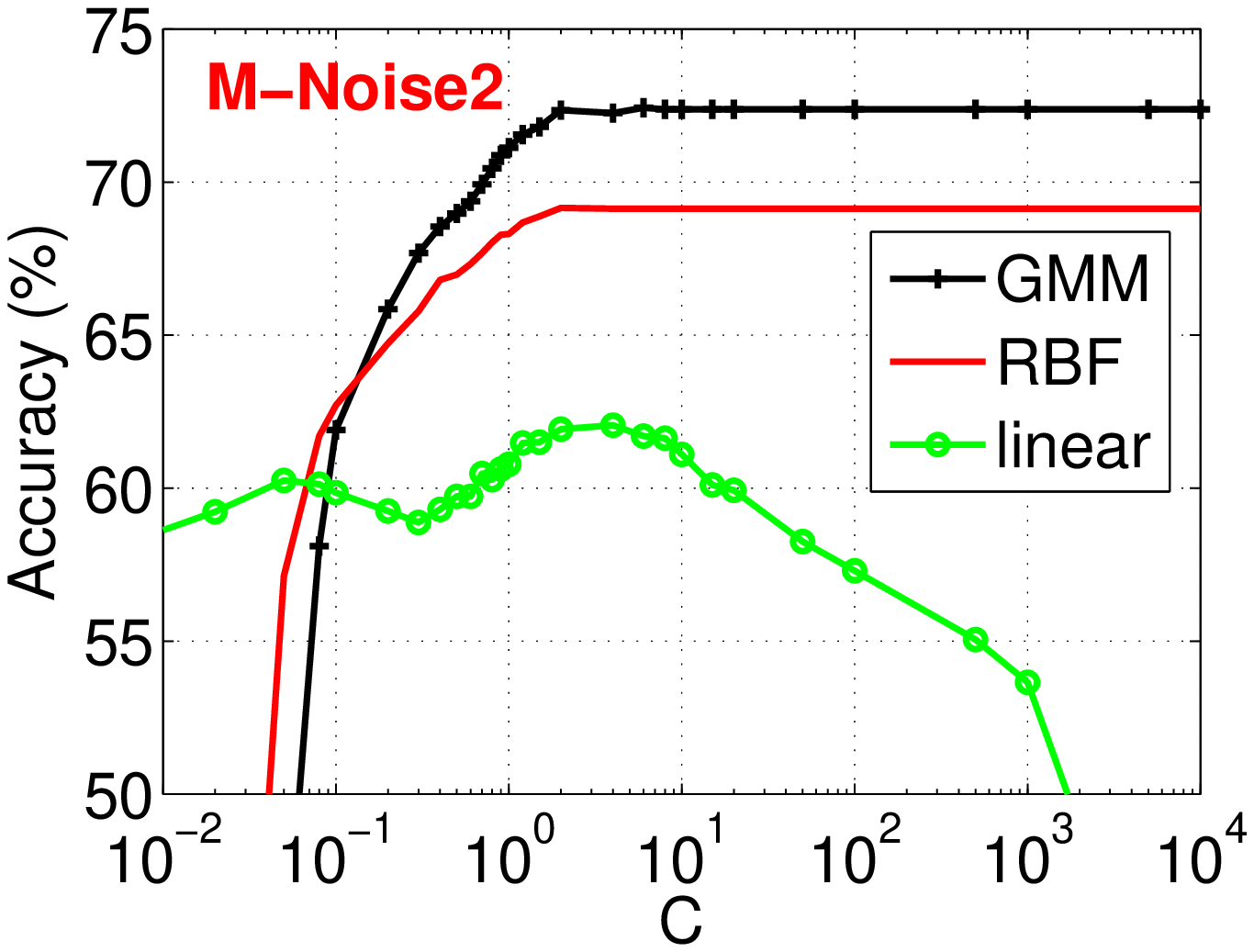}
\includegraphics[width=2.2in]{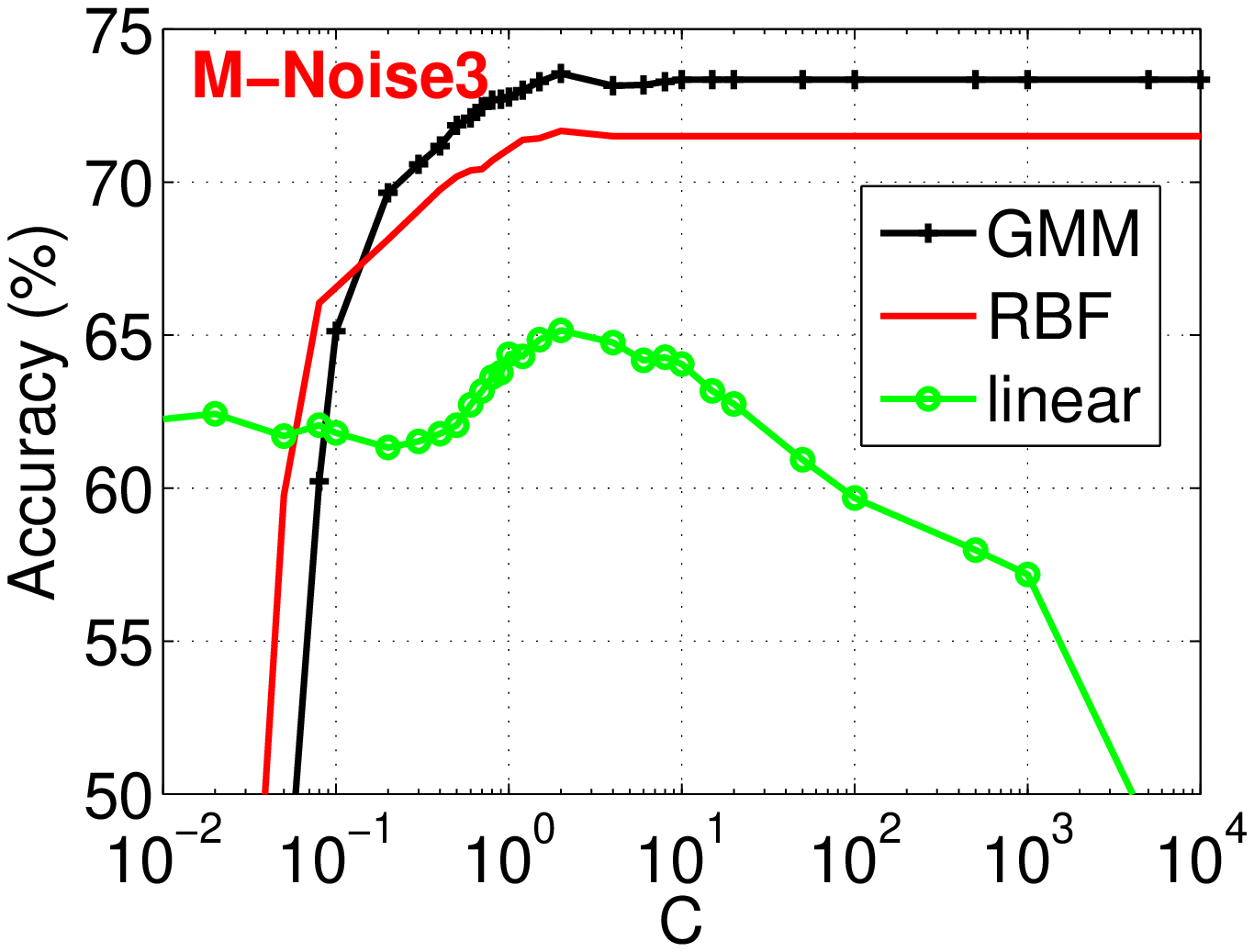}
}

\mbox{
\includegraphics[width=2.2in]{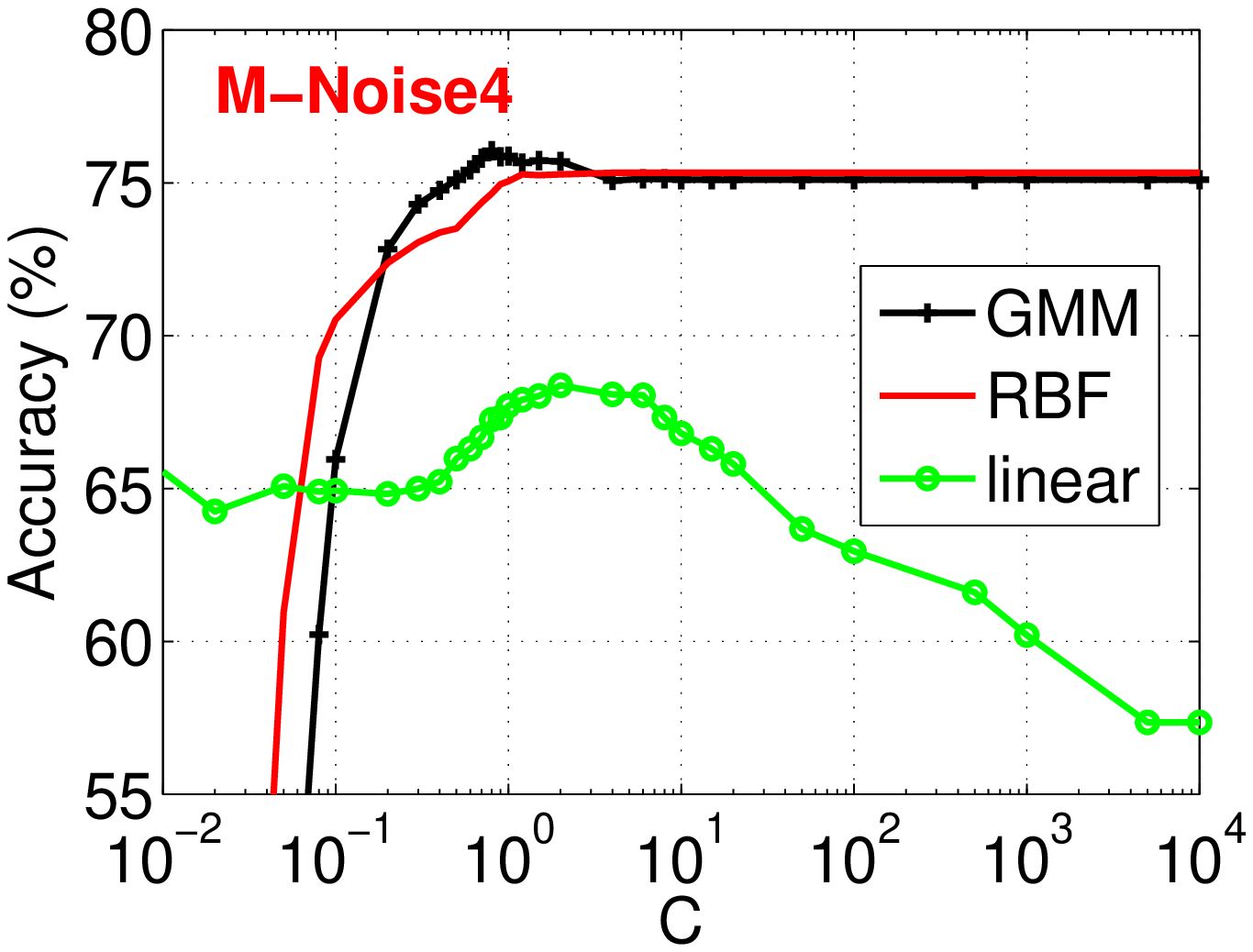}
\includegraphics[width=2.2in]{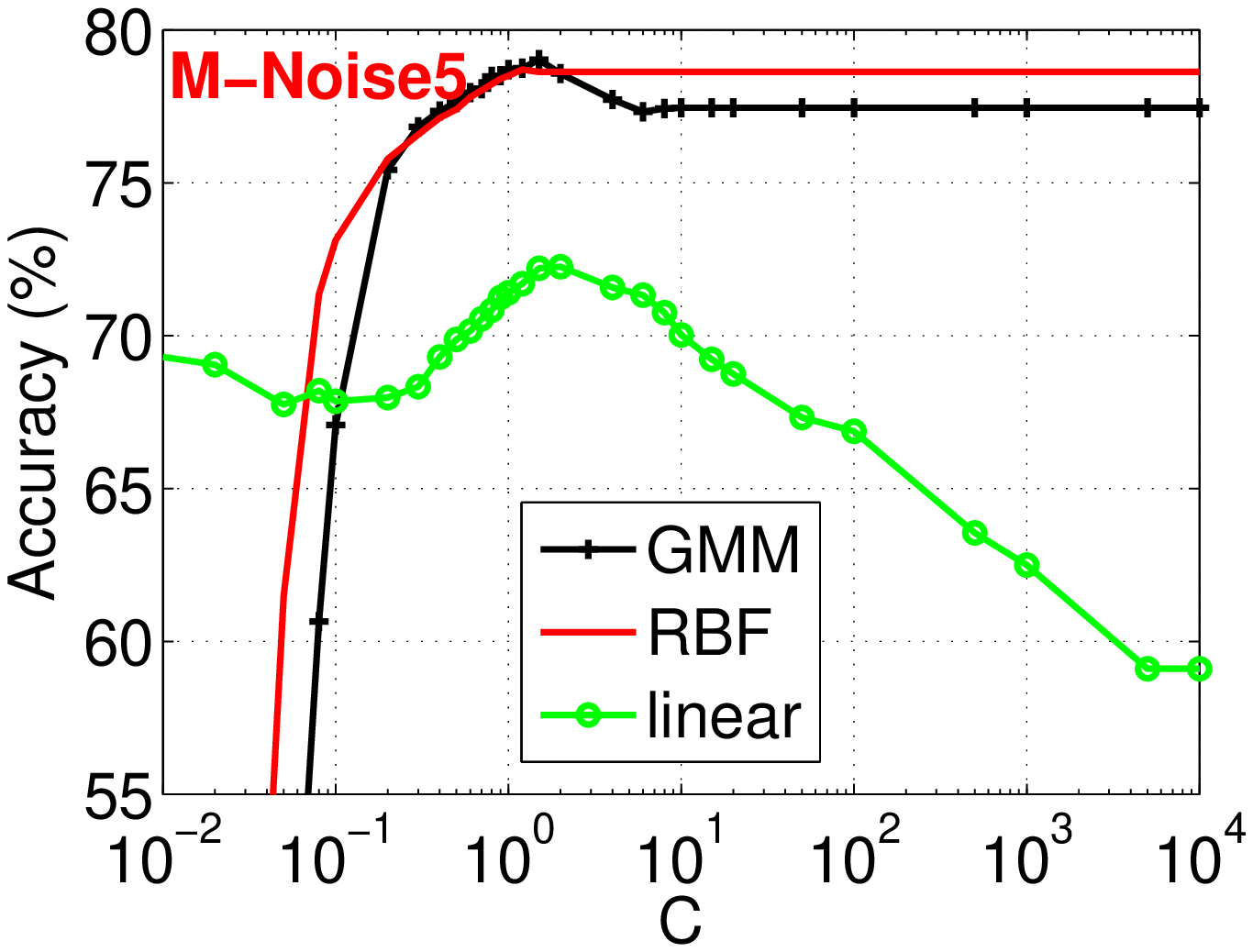}
\includegraphics[width=2.2in]{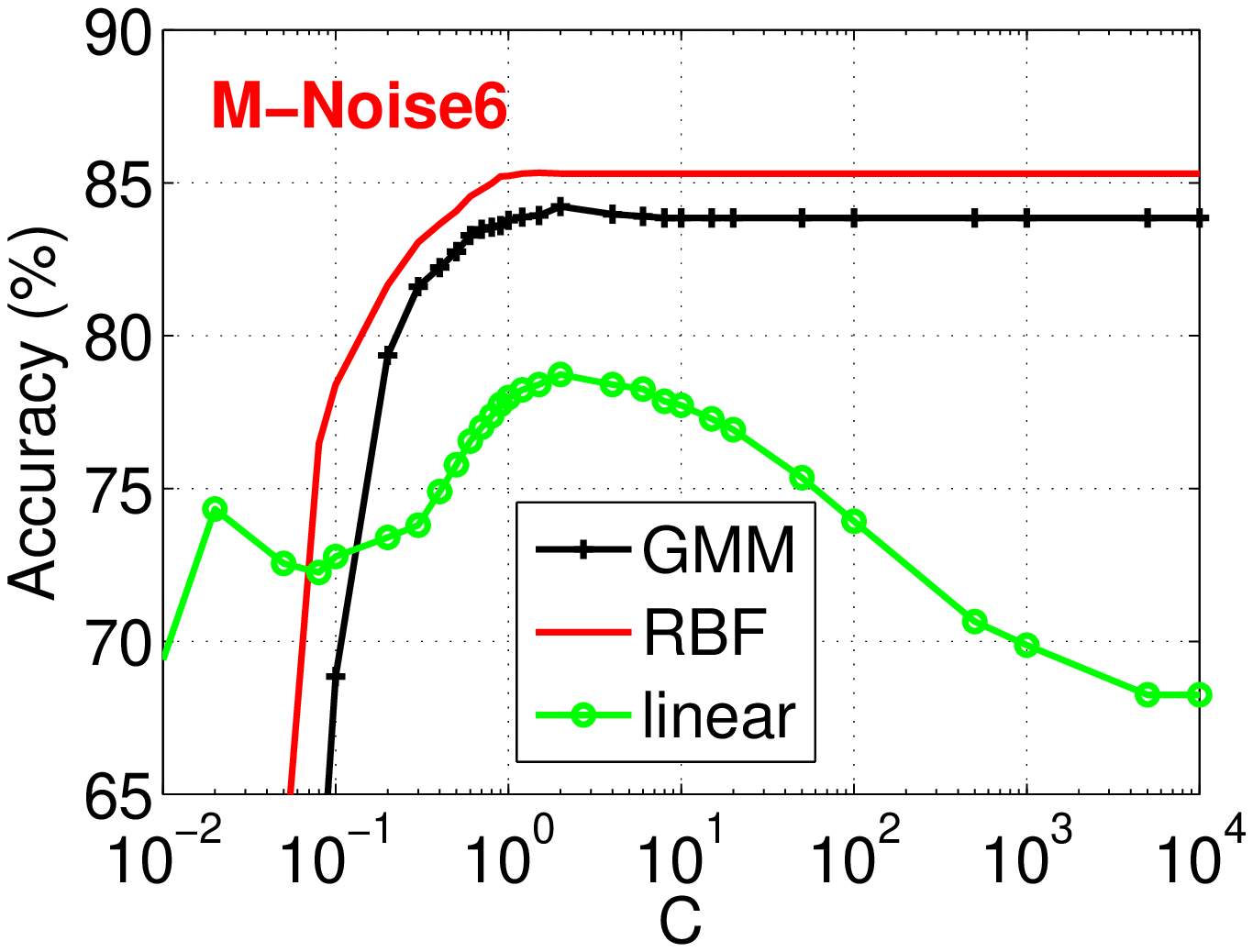}
}

\mbox{
\includegraphics[width=2.2in]{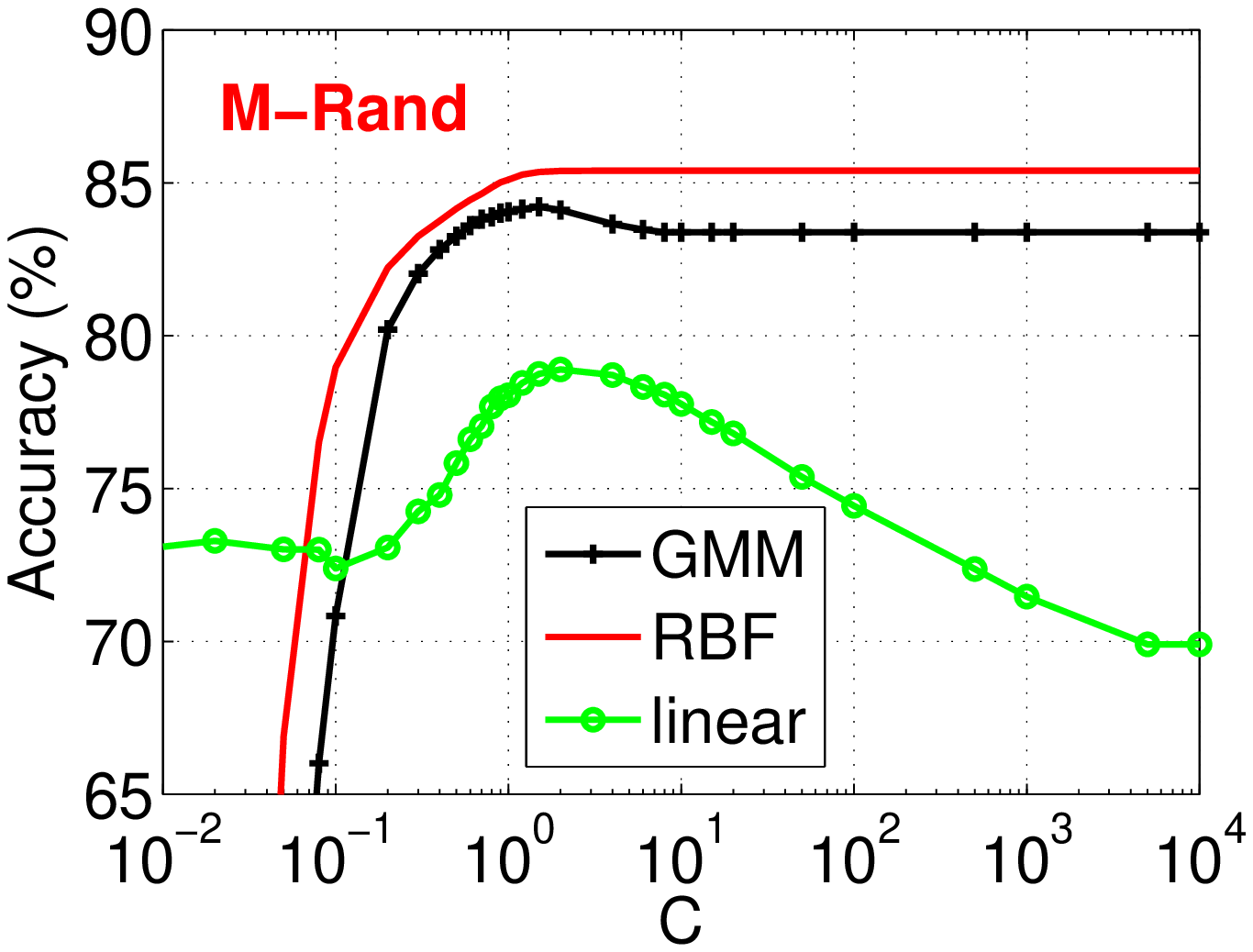}
\includegraphics[width=2.2in]{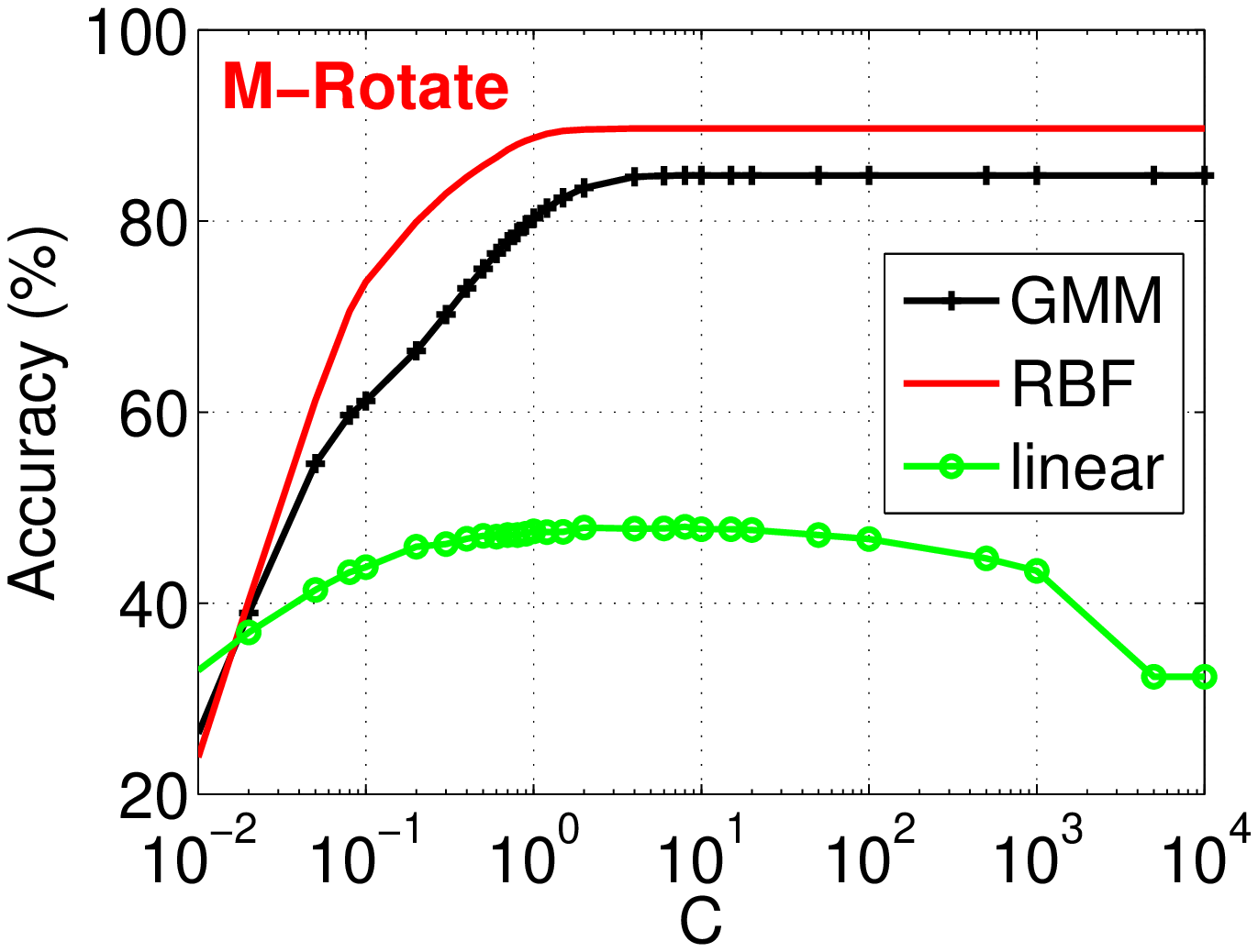}
\includegraphics[width=2.2in]{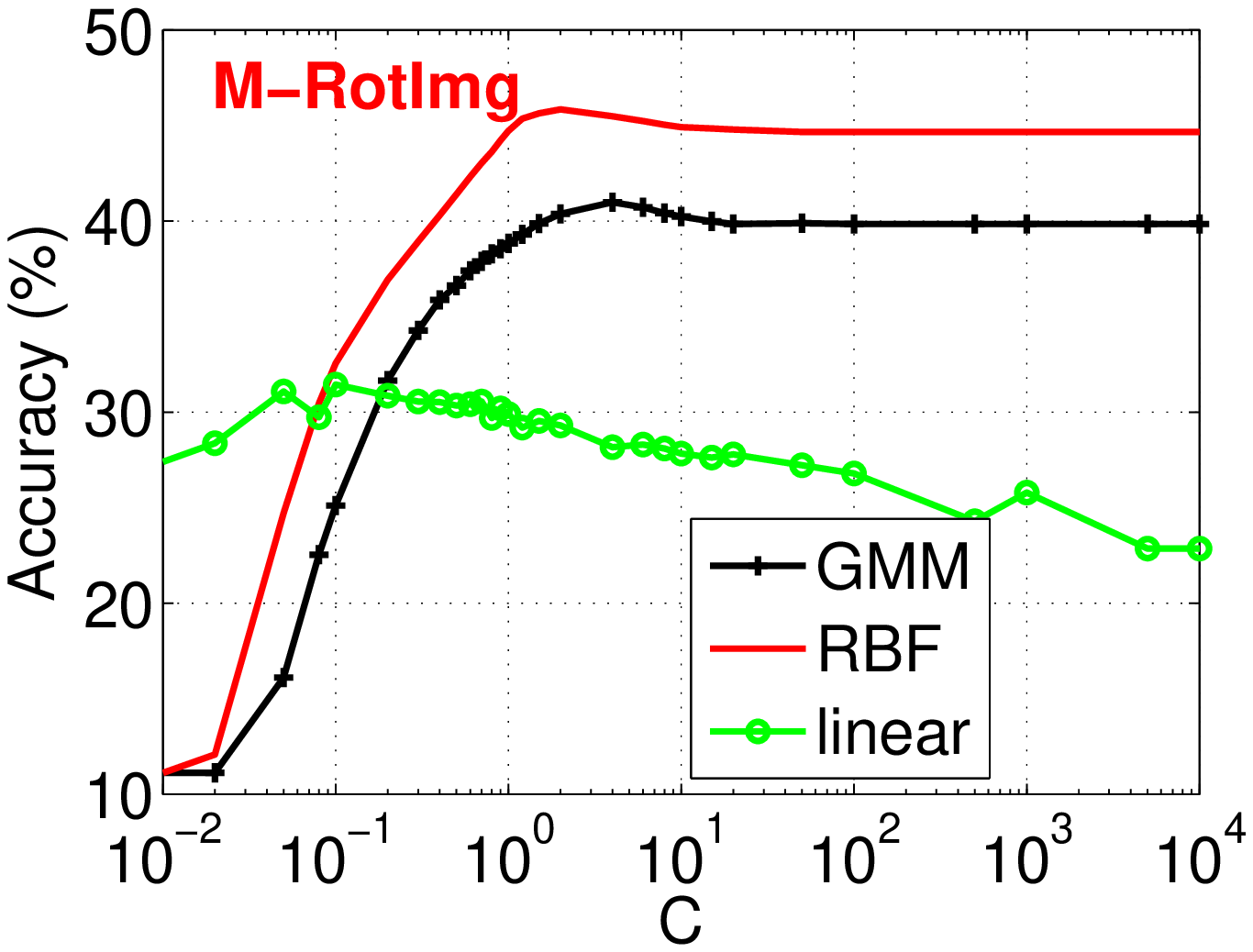}
}

\end{center}
\vspace{-0.3in}
\caption{ \textbf{Test classification accuracies using kernel SVMs}. Both the GMM kernel and RBF kernel substantially improve linear SVM. $C$ is the $l_2$-regularization parameter of SVM. For the RBF kernel, we report the result at the best $\gamma$ value for every $C$ value. }\label{fig_KernelSVM4}
\end{figure}

Table~\ref{tab_UCI} lists datasets from the UCI repository. Table~\ref{tab_Large} presents datasts from the LIBSVM website as well as datasets which are fairly large. Table~\ref{tab_MNIST} contains datasets used for evaluating   deep learning and trees~\cite{Proc:Larochelle_ICML07,Proc:ABC_UAI10}. Except for the relatively large datasets in Table~\ref{tab_Large}, we also report the classification accuracies for the linear SVM, kernel SVM with RBF, and kernel SVM with GMM, at the best $l_2$-regularization $C$ values. More detailed results (for all regularization $C$ values) are available in Figures~\ref{fig_KernelSVM1}, ~\ref{fig_KernelSVM2},  ~\ref{fig_KernelSVM3}, and~\ref{fig_KernelSVM4}. To ensure repeatability, we use the LIBSVM pre-computed kernel functionality. This also means we can not (easily) test nonlinear kernels on  larger datasets.

For the RBF kernel, we exhaustively experimented with 58 different values of $\gamma\in\{$0.001, 0.01, 0.1:0.1:2, 2.5, 3:1:20 25:5:50, 60:10:100, 120, 150, 200, 300, 500, 1000$\}$. Basically, Tables~\ref{tab_UCI}, ~\ref{tab_Large}, and ~\ref{tab_MNIST} reports the best RBF results among all $\gamma$ and $C$ values in our experiments.

\vspace{0.08in}

The classification results indicate that, on these datasets, kernel (GMM and RBF) SVM classifiers   improve over linear classifiers substantially. For more than half of the datasets, the GMM kernel (which has no tuning parameter) outperforms the best-tuned RBF kernel.  For a small number of  datasets (e.g., ``SEMG1''), even though the RBF kernel performs better, we will show in Section~\ref{sec_hashing} that the GCWS hashing can still be substantially better than the NRFF hashing.

\begin{table}[h!]
\caption{\textbf{Public (UCI) classification datasets  and $l_2$-regularized kernel SVM results}. We report the test classification accuracies for the linear kernel, the best-tuned RBF kernel (and the best $\gamma$), and the GMM kernel, at their individually- best SVM regularization  $C$ values.\vspace{-0.1in}
}
\begin{center}{\small
{\begin{tabular}{l r r r c l c c c}
\hline \hline
Dataset     &\# train  &\# test  &\# dim &linear  &RBF ($\gamma$)  &GMM \\
\hline
Car &864 &864 &6 &71.53   &94.91 (100)  &{\bf98.96} \\
Covertype25k &25000 &25000 &54 &62.64 &{\bf82.66} (90) &82.65 \\
CTG &1063 &1063 &35 &60.59   &{\bf89.75} (0.1)  &88.81\\
DailySports &4560 &4560 &5625  & 77.70 &97.61 (4) &\textbf{99.61}\\
Dexter &300 &300 &19999 &92.67   &93.00 (0.01) &{\bf94.00}\\
Gesture &4937 &4936 &32 & 37.22   &61.06 (9)  &{\bf65.50}\\
ImageSeg &210 &2100 &19 &83.81   &91.38 (0.4)  &{\bf95.05} \\
Isolet2k &2000 &5797 &617 & 93.95   &{\bf95.55} (3) &95.53\\
MSD20k &20000 &20000 &90 &66.72 &68.07 (0.1) &{\bf71.05} \\
MHealth20k&20000&20000&23&72.62   &82.65 (0.1)   &{\bf85.28}   \\
Magic &9150 &9150 &10 &78.04   &84.43 (0.8) &{\bf87.02}  \\
Musk &3299 &3299 &166  & 95.09 &\textbf{99.33} (1.2) &99.24 \\
PageBlocks &2737  &2726 &10 &95.87   &{\bf97.08} (1.2)  &96.56\\
Parkinson &520&520&26&61.15   &66.73(1.9)   &{\bf69.81}\\
PAMAP101 &20000 &20000 &51 &76.86   &96.68 (15) &{\bf98.91}  \\
PAMAP102 &20000 &20000 &51 &81.22   &95.67  (1.1) &{\bf98.78} \\
PAMAP103 &20000 &20000 &51 & 85.54  &97.89 (19)  &{\bf99.69}  \\
PAMAP104 &20000 &20000 &51 &84.03  &97.32  (19) &{\bf99.30} \\
PAMAP105 &20000 &20000 &51 &79.43  &97.34  (18) &{\bf99.22} \\
RobotNavi &2728 &2728 &24 &69.83   &90.69 (10)  &{\bf96.85}  \\
Satimage &4435 &2000 &36 &72.45   &85.20  (200) &{\bf90.40}  \\
SEMG1 &900 &900 &3000  &26.00   &\textbf{43.56} (4)   &41.00 \\
SEMG2 &1800 &1800 &2500  &19.28  &29.00 (6)   &{\bf54.00} \\
Sensorless &29255 &29254 &48 &61.53 &93.01 (0.4)  &{\bf99.39}   \\
Shuttle500 &500 &14500 &9 &91.81   &99.52 (1.6)  &{\bf99.65} \\
SkinSeg10k&10000&10000&3& 93.36   &99.74 (120) &{\bf99.81}\\
SpamBase &2301&2300&57&   85.91&   92.57 (0.2) & {\bf94.17} \\
Splice &1000&2175&60&85.10   &90.02 (15)  &{\bf95.22}  \\
Thyroid2k &2000&5200&21&   94.90   &97.00 (2.5) &{\bf98.40} \\
Urban &168&507&147&   62.52   &51.48 (0.01)  &{\bf66.08}  \\
Vowel &264 &264 &10 &39.39 &94.70 (45) &{\bf96.97} \\
YoutubeAudio10k&10000&11930&2000&   41.35   &48.63 (2)  &{\bf50.59}  \\
YoutubeHOG10k&10000&11930&647&   62.77   &66.20 (0.5)  &{\bf68.63} \\
YoutubeMotion10k&10000&11930&64& 26.24  &28.81 (19)  &{\bf31.95}  \\
YoutubeSaiBoxes10k&10000&11930&7168&46.97   &49.31 (1.1)  &{\bf51.28}   \\
YoutubeSpectrum10k&10000&11930&1024&26.81   &33.54 (4)   &{\bf39.23} \\
\hline\hline
\end{tabular}}
}
\end{center}\label{tab_UCI}\vspace{-0.15in}
\end{table}

\begin{table}[h!]
\caption{Datasets in group 1 and group 3 are from the LIBSVM website. Datasets in group 2 are from the UCI repository. Datasets in group 2 and 3 are too large for LIBSVM pre-computed kernel functionality and are thus only used for testing hashing methods.
}
\begin{center}{\small
{\begin{tabular}{cl r r r c l c c c}
\hline \hline
Group &Dataset     &\# train  &\# test  &\# dim &linear  &RBF ($\gamma$)  &GMM \\
\hline
&Letter&15000 &5000 &16 &61.66 &\textbf{97.44} (11) &97.26\\
1&Protein&17766    & 6621      & 357  &69.14   &70.32 (4)   &\textbf{70.64}\\
&SensIT20k&20000 &19705&100& 80.42   &83.15 (0.1) &\textbf{84.57}\\
&Webspam20k&20000 &60000 &254&93.00  &\textbf{97.99} (35) &97.88 \\\hline
2&PAMAP101Large &186,581 &186,580 &51 &79.19   \\
&PAMAP105Large &185,548 &185,548 &51 &83.35 \\\hline
&IJCNN &49990 &91701 &22 &92.56 \\
3&RCV1 &338,699 &338,700 &47,236 &97.66 \\
&SensIT &78,823 &19,705 &100 &80.55 \\
&Webspam &175,000 &175,000 &254 &93.31\\
\hline\hline
\end{tabular}}
}
\end{center}\label{tab_Large}
\end{table}

\begin{table}[h!]
\caption{Datasets from \cite{Proc:Larochelle_ICML07,Proc:ABC_UAI10}. See the technical report~\cite{Report:Li_epGMM17} on ``tunable GMM kernels'' for substantially improved results, by introducing tuning parameters in the GMM kernel.
}
\begin{center}{
{\begin{tabular}{l r r c c l l c }
\hline \hline
Dataset     &\# train  &\# test  &\# dim &linear &RBF ($\gamma$)  &GMM \\
\hline
M-Basic   &12000 &50000 &784 & 89.98   &{\bf97.21} (5)  &96.20  \\
M-Image &12000 &50000 &784 & 70.71 &77.84 (16)  &{\bf80.85} \\
M-Noise1 &10000 &4000 &784 &60.28   &66.83 (10)   &{\bf71.38} \\
M-Noise2 &10000 &4000 &784 & 62.05  & 69.15 (11)   &{\bf72.43}\\
M-Noise3 &10000 &4000 &784 &65.15   &71.68 (11)   &{\bf73.55} \\
M-Noise4 &10000 &4000 &784 & 68.38  &75.33 (14)  &{\bf76.05} \\
M-Noise5 &10000 &4000 &784 &72.25   &78.70 (12)  &{\bf79.03} \\
M-Noise6 &10000 &4000 &784 &78.73   &{\bf85.33} (15)  &84.23  \\
M-Rand &12000 &50000 &784 & 78.90   &{\bf85.39} (12)  &84.22 \\
M-Rotate &12000 &50000   &784 &47.99  &{\bf89.68} (5)  & 84.76\\
M-RotImg &12000 &50000 &784 &31.44  &{\bf45.84} (18)  & 40.98\\
\hline\hline
\end{tabular}}
}
\end{center}\label{tab_MNIST}

\end{table}

\newpage\clearpage

\begin{figure}[h!]
\begin{center}
\includegraphics[width=3in]{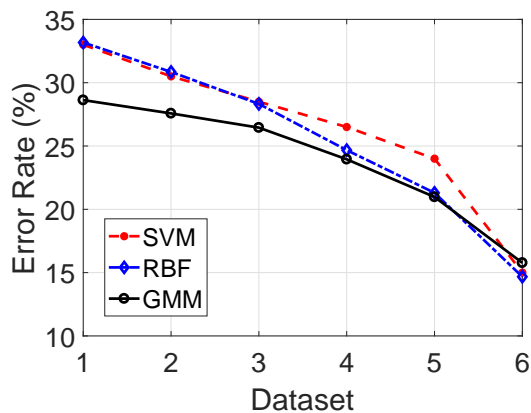}
\end{center}
\vspace{-0.3in}
\caption{Error rates on 6 datasets: M-Noise1 to M-Noise6 as in Table~\ref{tab_MNIST}. In this figure, the curve labeled as ``SVM'' represents the results on RBF kernel SVM conducted by~\cite{Proc:Larochelle_ICML07}, while the curve labeled as ``RBF'' presents our own experiments. The small discrepancies might be caused by the fact that we always use normalized data (i.e., $\rho$). }\label{fig_Noise6}
\end{figure}

For the datasets in Table~\ref{tab_MNIST}, since~\cite{Proc:Larochelle_ICML07} also conducted experiments on the RBF kernel, the polynomial kernel, and neural nets, we assembly the (error rate) results in Figure~\ref{fig_Noise6} and Table~\ref{tab_MNIST_result}.

\begin{table}[h]
\caption{Summary of test error rates of various algorithms  on other datasets used in ~\cite{Proc:Larochelle_ICML07,Proc:ABC_UAI10}.  Results in group 1 are reported by \cite{Proc:Larochelle_ICML07} for using RBF kernel, polynomial kernel, and neural nets.  Results in group 2 are from our own experiments. Also, see the technical report~\cite{Report:Li_epGMM17} on ``tunable GMM kernels'' for substantially improved results, by introducing tuning parameters in the GMM kernel. }
\begin{center}{
\begin{tabular}{c l r r r r r}
\hline \hline
Group &Method &M-Basic & M-Rotate &M-Image &M-Rand & M-RotImg\\\hline
&SVM-RBF &3.05\% &11.11\% &22.61\% &14.58\% &55.18\%\\
1&SVM-POLY &3.69\% &15.42\% &24.01\% &16.62\% &56.41\%\\
&NNET &4.69\% &18.11\% &27.41\% &20.04\% &62.16\%
\\\hline
&Linear &10.02\% &52.01\% &29.29\% &21.10\% &68.56\% \\
2&RBF &\textbf{2.79}\% &\textbf{10.30}\% &22.16\% &\textbf{14.61}\% &\textbf{54.16}\%\\
&GMM &3.80\% &15.24\% &\textbf{19.15}\% &15.78\% &59.02\%\\
\hline\hline
\end{tabular}

}
\end{center}
\label{tab_MNIST_result}
\end{table}

\newpage\clearpage

\section{Hashing for Linearizing Nonlinear Kernels}\label{sec_hashing}

It is known that a straightforward  implementation of nonlinear kernels can be difficult for large datasets~\cite{Book:Bottou_07}.  For example, for  a small dataset with merely $100,000$ data points, the $100,000 \times 100,000$ kernel matrix  has $10^{10}$ entries.  In practice, being able to linearize nonlinear kernels becomes very beneficial, as that would allow us to easily apply efficient linear algorithms especially online learning~\cite{URL:Bottou_SGD}. Randomization (hashing) is a popular tool for kernel linearization.

\vspace{0.08in}

In the introduction, we have explained how to linearize both the RBF kernel and the GMM kernel. From practitioner's perspective, while the kernel classification results in Tables~\ref{tab_UCI}, ~\ref{tab_Large}, and ~\ref{tab_MNIST} are informative, they are not sufficient for guiding  the choice of kernels. For example, as we will show, for some datasets, even though the RBF kernel outperform the GMM kernel, the linearization algorithm (i.e., the normalized RFF) requires substantially more samples (i.e., larger $k$). Note that in our SVM experiments, we always normalize the input features to the unit $l_2$ norm (i.e., we will always use NRFF instead of RFF).

\vspace{0.08in}

We will report detailed experimental results on 6 datasets. As shown in Table~\ref{tab_data}, on the first two datasets, the original RBF and GMM kernels perform similarly; in the second group, the GMM kernel noticeably outperforms the RBF kernel; in the last group, the RBF kernel noticeably outperforms the GMM kernel. We will show on all these 6 datasets, the GCWS hashing is substantially more accurate than the NRFF hashing at the same number of sample size ($k$). We will then present less detailed results on other datasets.

\begin{table}[h!]
\caption{6 datasets used for presenting detailed experimental results on GCWS and NRFF.}
\begin{center}{\small
{\begin{tabular}{c l r r r c l c c c}
\hline \hline
Group &Dataset     &\# train  &\# test  &\# dim &linear  &RBF ($\gamma$)  &GMM \\
\hline

1&Letter&15000 &5000 &16 &61.66 &\textbf{97.44} (11) &97.26\\
&Webspam20k&20000 &60000 &254&93.00  &\textbf{97.99} (35) &97.88 \\\hline
2&DailySports &4560 &4560 &5625  & 77.70 &97.61 (4) &\textbf{99.61}\\
&RobotNavi &2728 &2728 &24 &69.83   &90.69 (10)  &{\bf96.85}  \\\hline
3&SEMG1 &900 &900 &3000  &26.00   &\textbf{43.56} (4)   &41.00 \\
&M-Rotate &12000 &50000   &784 &47.99  &{\bf89.68} (5)  & 84.76\\
\hline\hline
\end{tabular}}
}
\end{center}\label{tab_data}

\end{table}

\begin{figure}[h!]
\begin{center}

\mbox{
\includegraphics[width=2.7in]{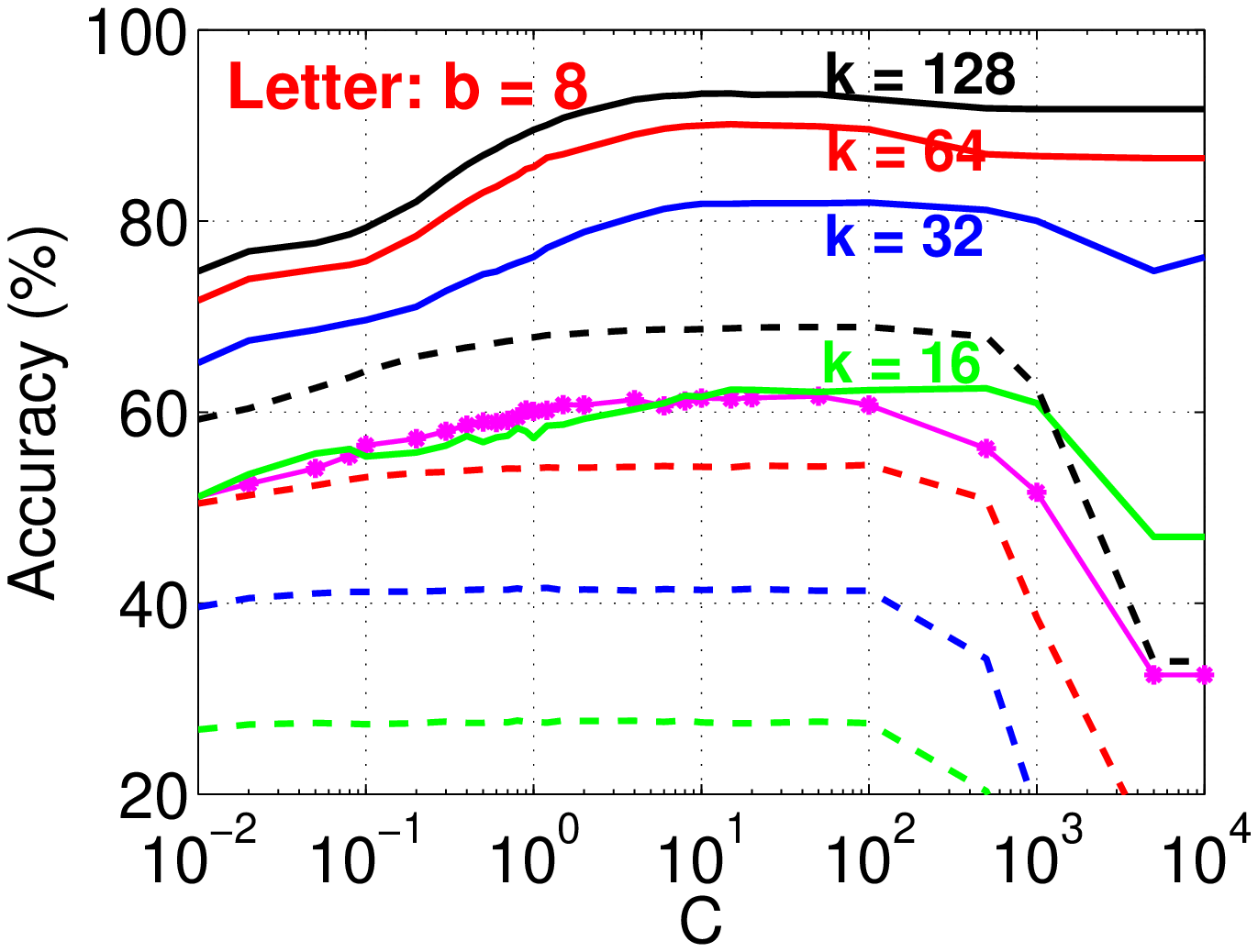}
\includegraphics[width=2.7in]{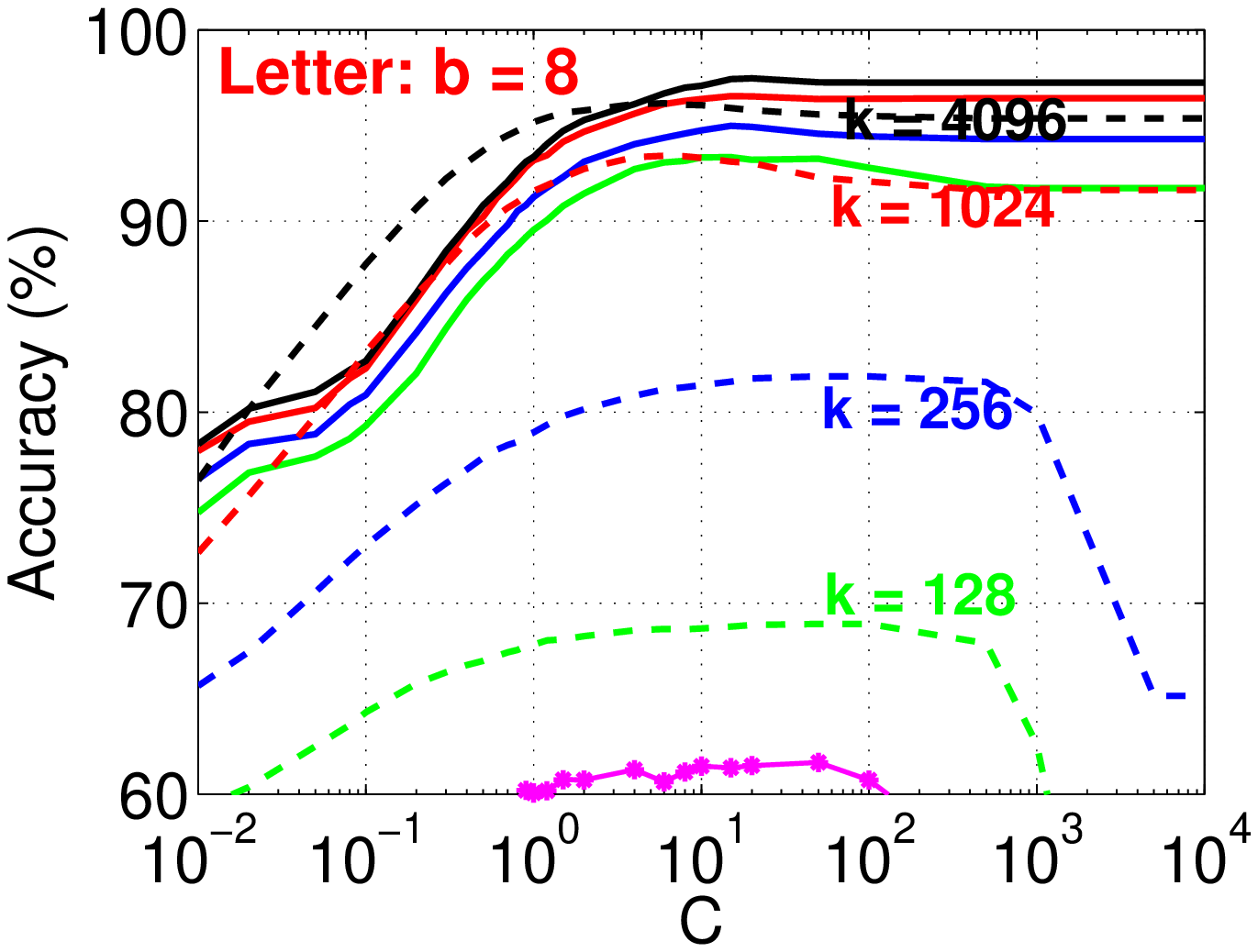}
}

\mbox{
\includegraphics[width=2.7in]{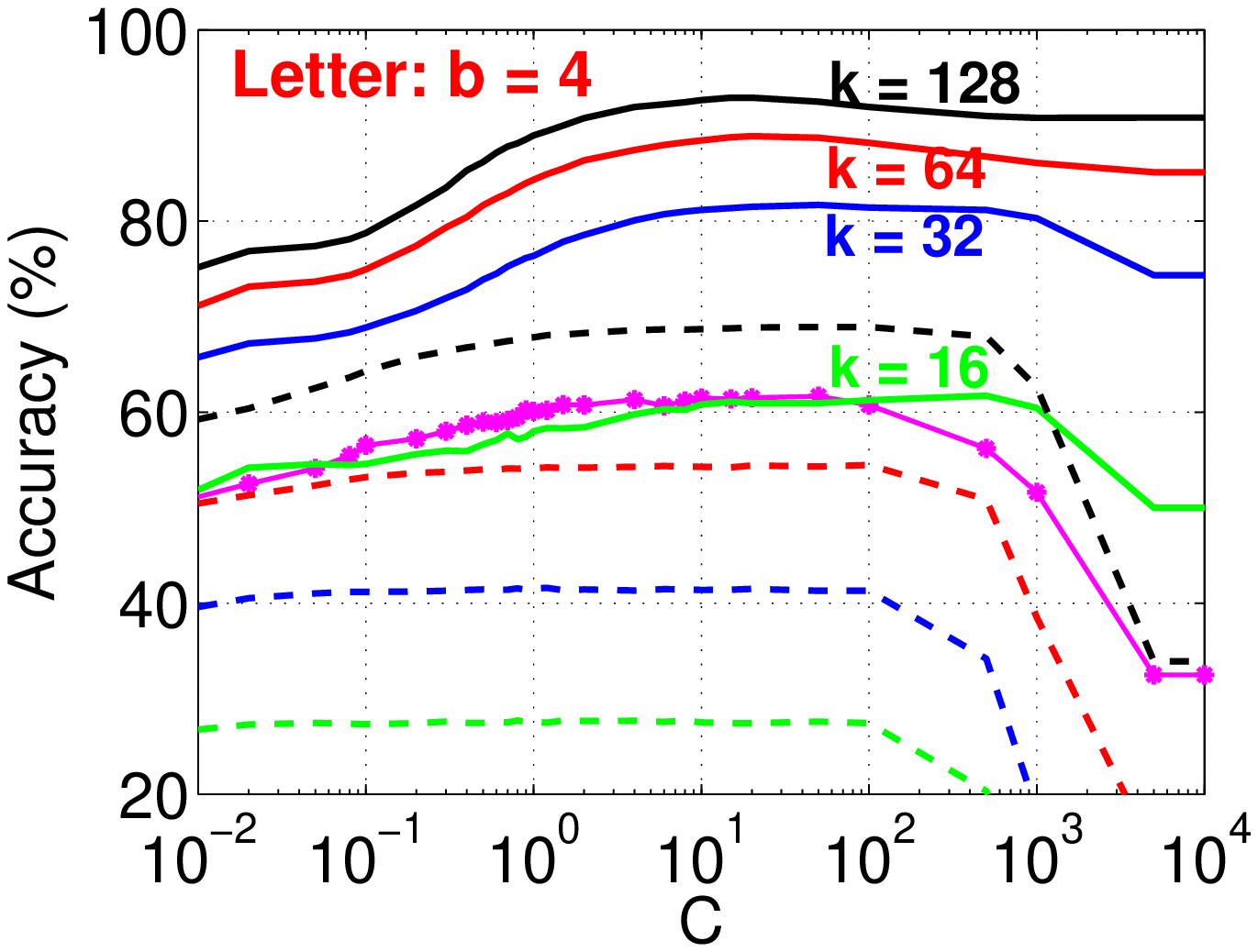}
\includegraphics[width=2.7in]{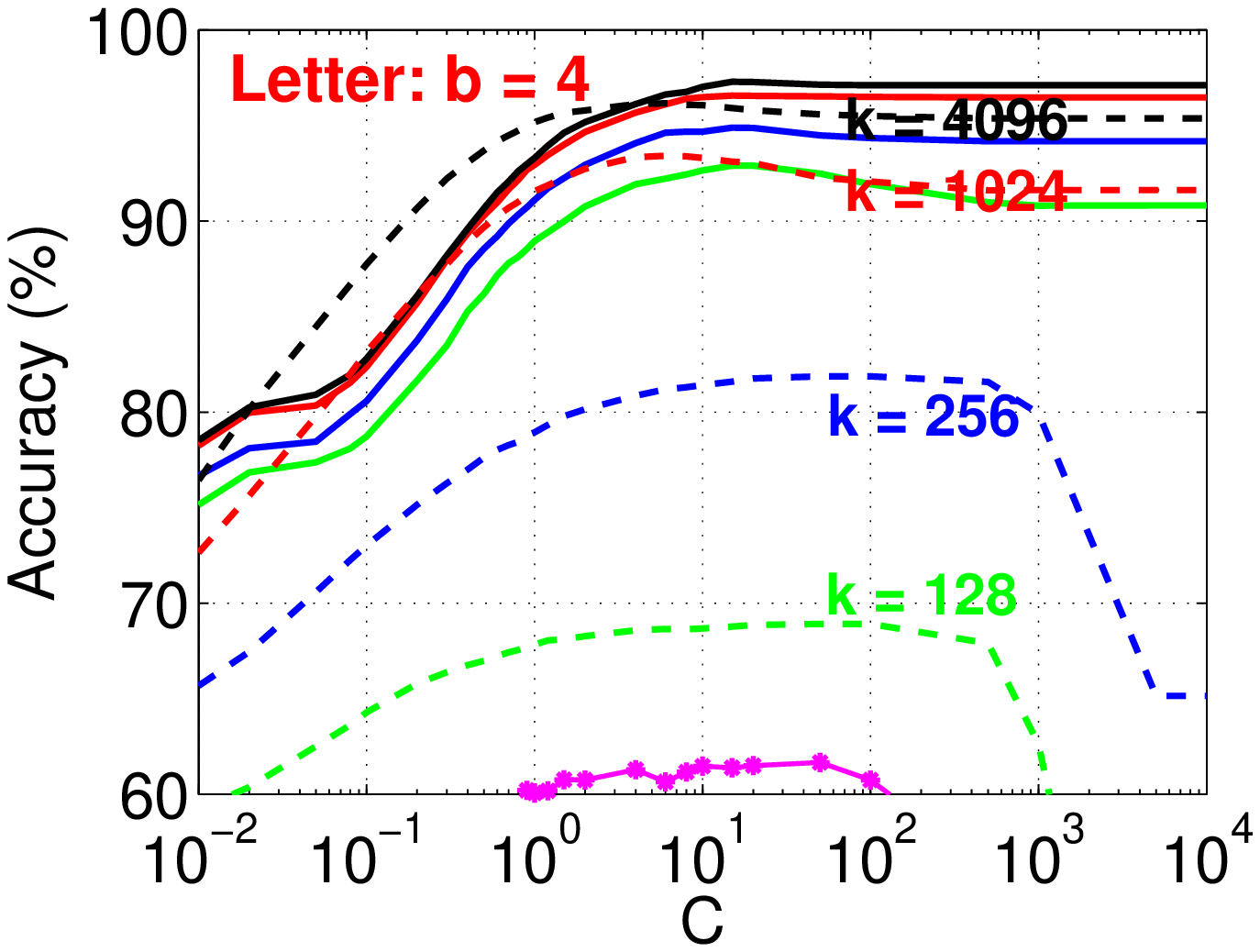}
}

\mbox{
\includegraphics[width=2.7in]{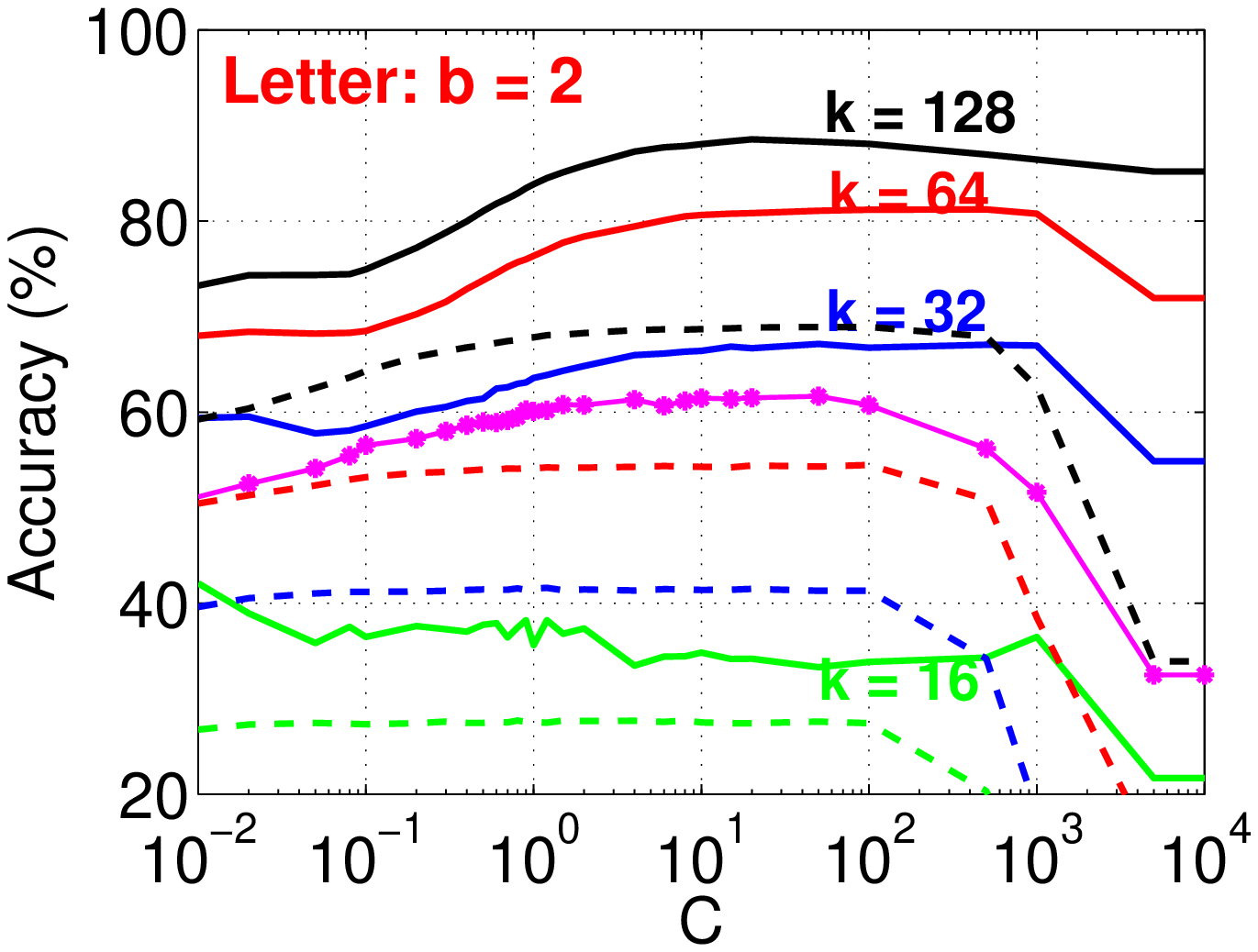}
\includegraphics[width=2.7in]{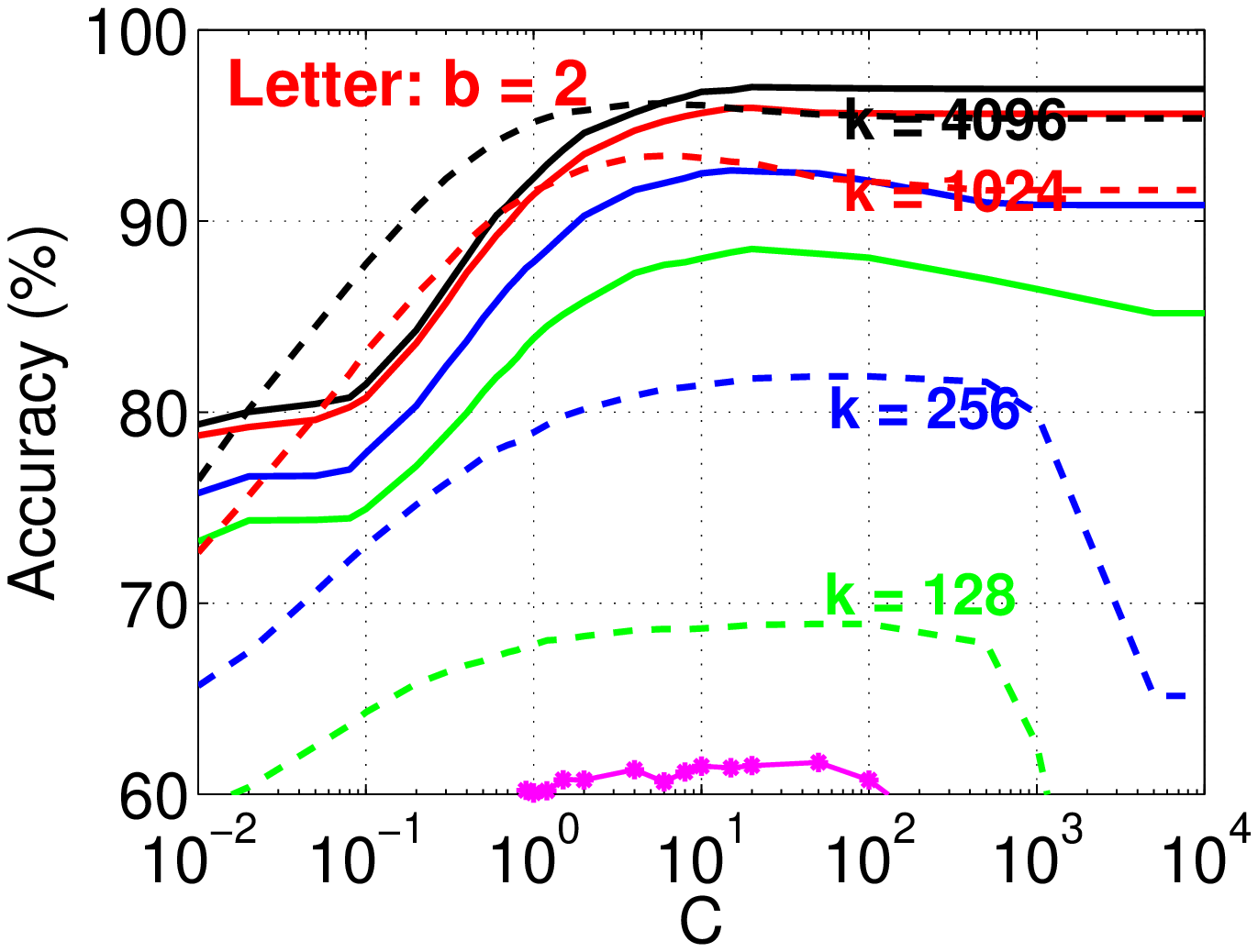}
}

\end{center}
\vspace{-0.3in}
\caption{\textbf{Letter}: Test classification accuracies of the linearized GMM kernel  (solid,  GCWS) and  linearized RBF kernel (dashed,  NRFF), using LIBLINEAR, averaged over 10 repetitions. In each panel, we report the results on 4 different $k$ (sample size) values: 128, 256, 1024, 4096 (right panels), and  16, 32, 64, 128 (left panels).  We can see that the linearized q
RBF  (using NRFF) would require substantially more samples in order to reach the same accuracies as the linearized GMM kernel (using GCWS).  Two interesting points: (i) Although the original (best-tuned) RBF kernel slightly outperforms the original GMM kernel, the results of GCWS are still more accurate than the results of RFF even at $k = 4096$, which is very large,  considering the original data dimension is merely 16. (ii) With merely $k=16$ samples ($b\geq4$), GCWS already produces better results than linear SVM based on the original dataset (the solid curve marked by *).
}\label{fig_Letter}
\end{figure}

Figure~\ref{fig_Letter} reports the test classification accuracies on the \textbf{Letter} dataset, for both linearized GMM kernel with GCWS and linearized RBF kernel (at the best $\gamma$) with NRFF, using LIBLINEAR. From Table~\ref{tab_data}, we can see that the original RBF kernel slightly outperforms the GMM kernel. Obviously, the results  obtained by GCWS hashing  are noticeably better than the results of NRFF hashing, especially when the number of samples ($k$) is not too large (i.e., the left panels).

\vspace{0.1in}

For the ``Letter'' dataset, the original dimension is merely 16.  It is known that, for modern linear algorithms, the computational cost is largely determined by the number of nonzeros. Hence the number of samples (i.e., $k$) is a crucial parameter which directly controls the training complexity. From the left panels of Figure~\ref{fig_Letter}, we can see that with merely $k=16$ samples, GCWS already produces better results than the original linear method. This phenomenon is exciting, because in industrial practice, the goal is often to produce better results than linear methods without consuming much more resources.

\begin{figure}[h!]
\begin{center}

\mbox{
\includegraphics[width=2.7in]{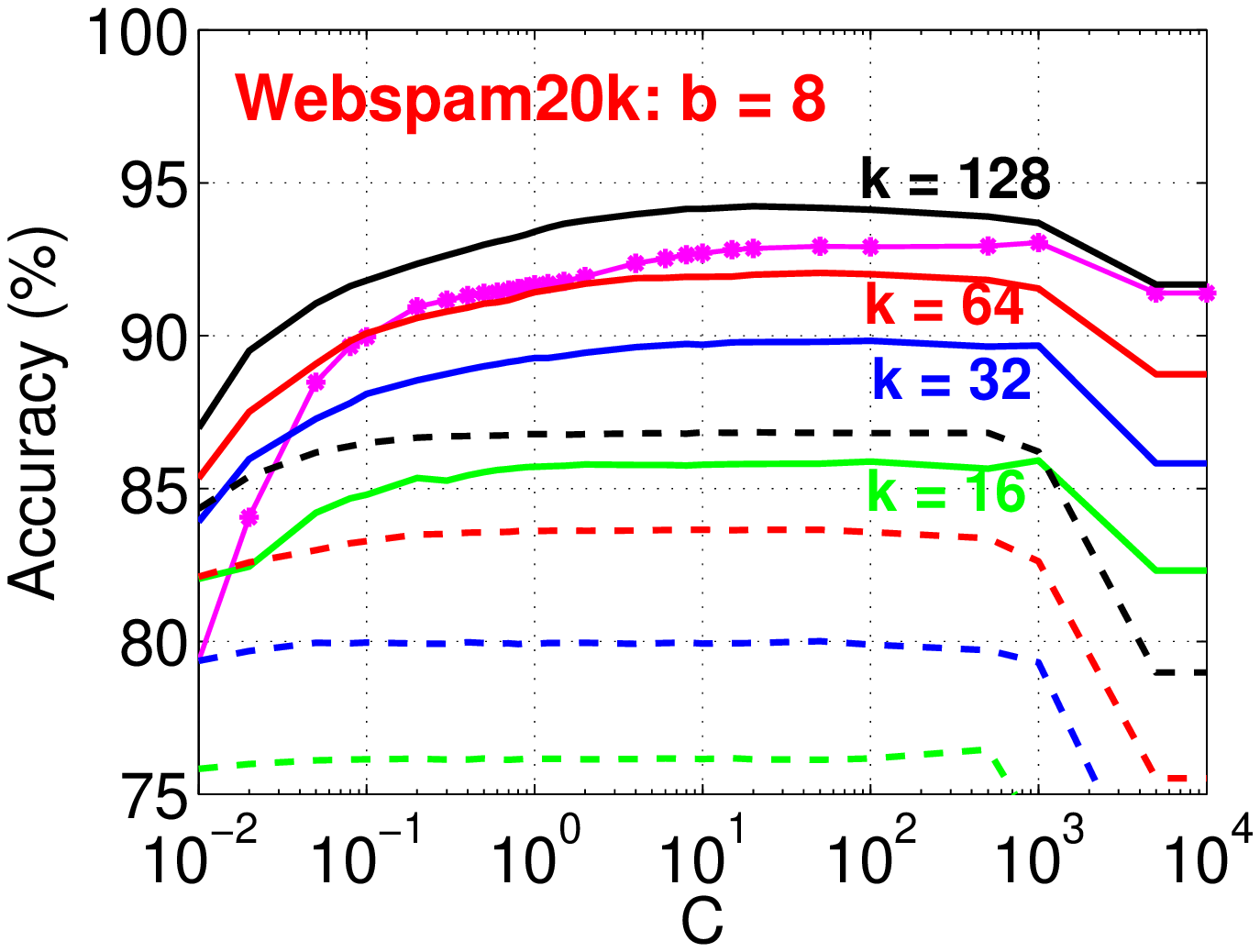}
\includegraphics[width=2.7in]{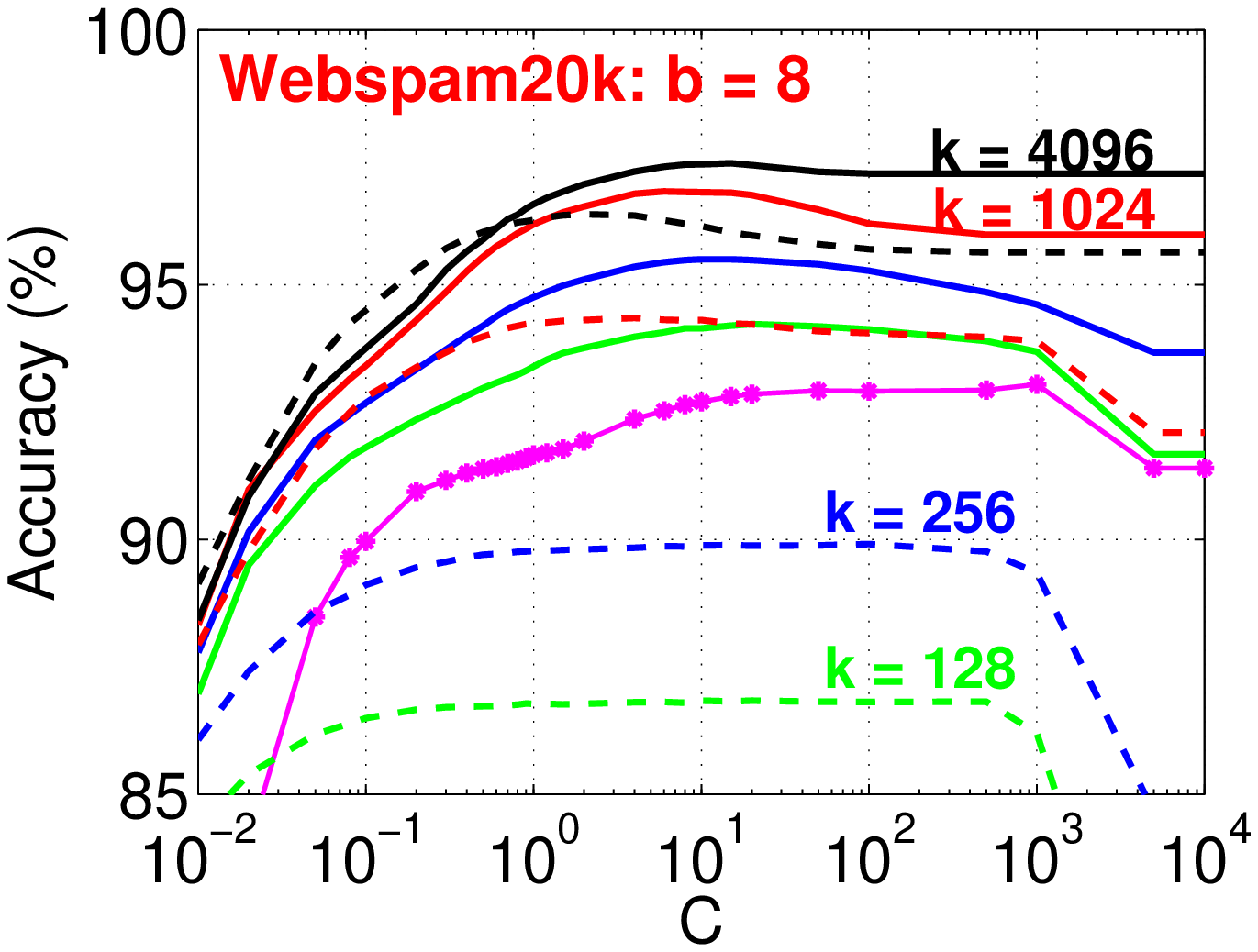}
}

\mbox{
\includegraphics[width=2.7in]{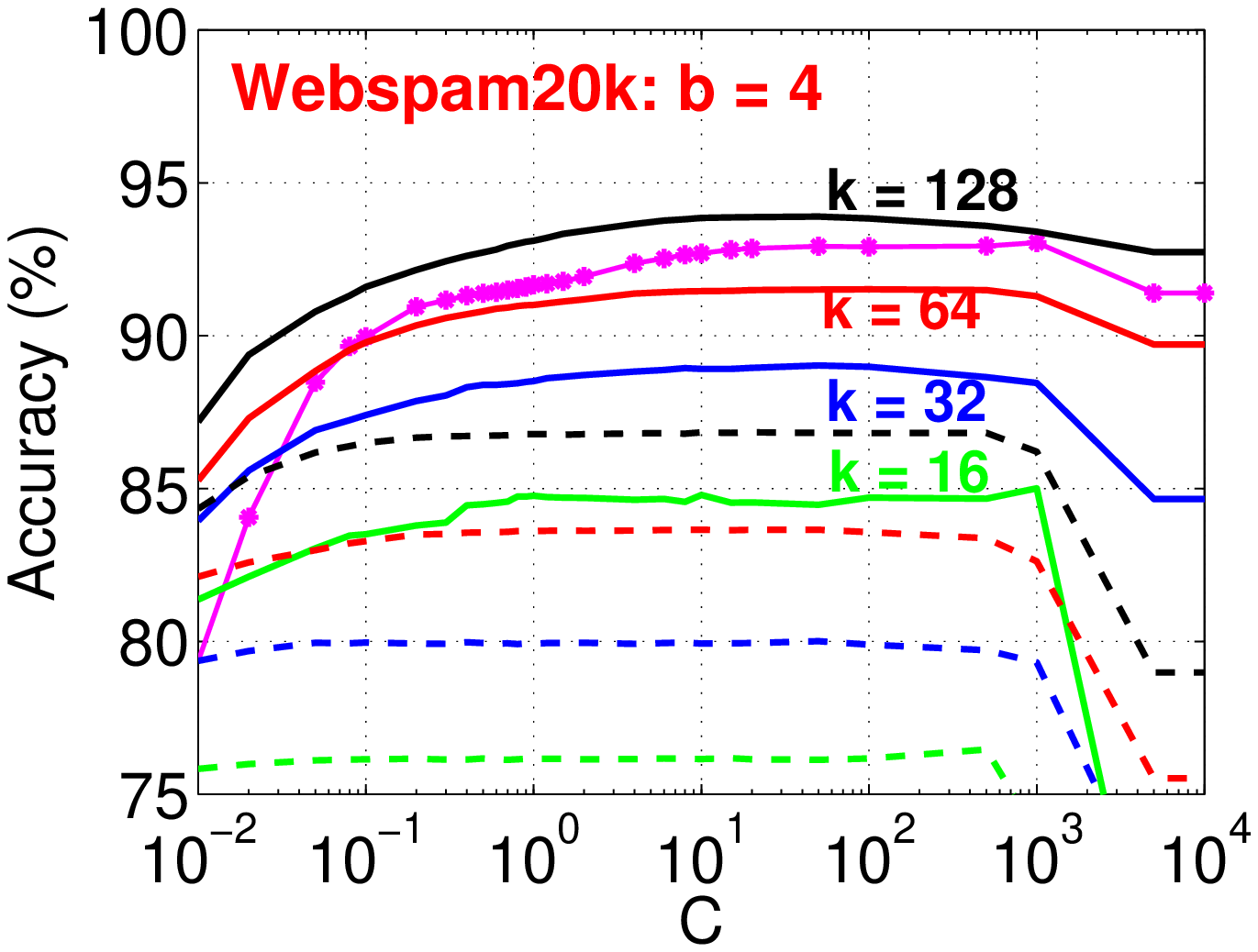}
\includegraphics[width=2.7in]{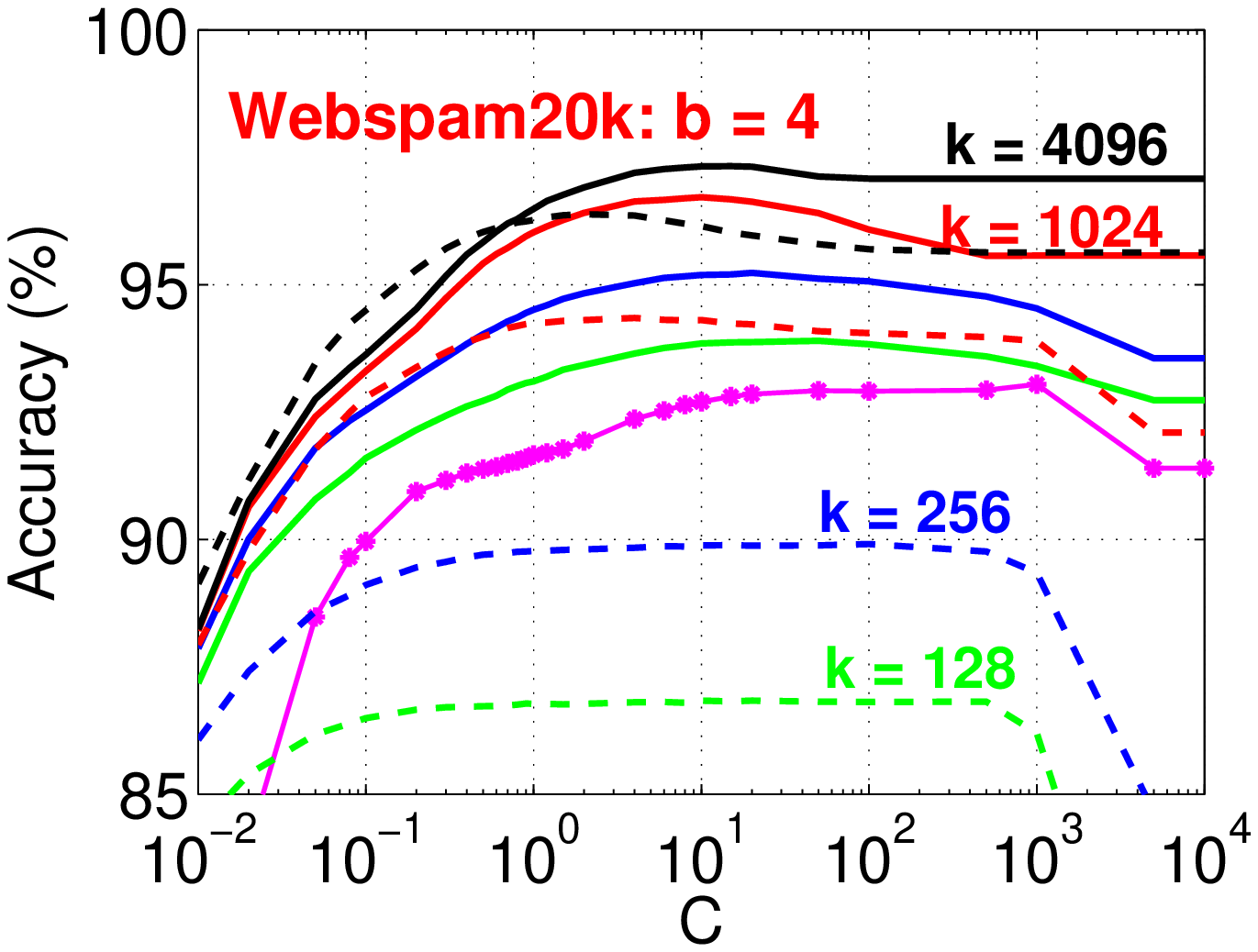}
}

\mbox{
\includegraphics[width=2.7in]{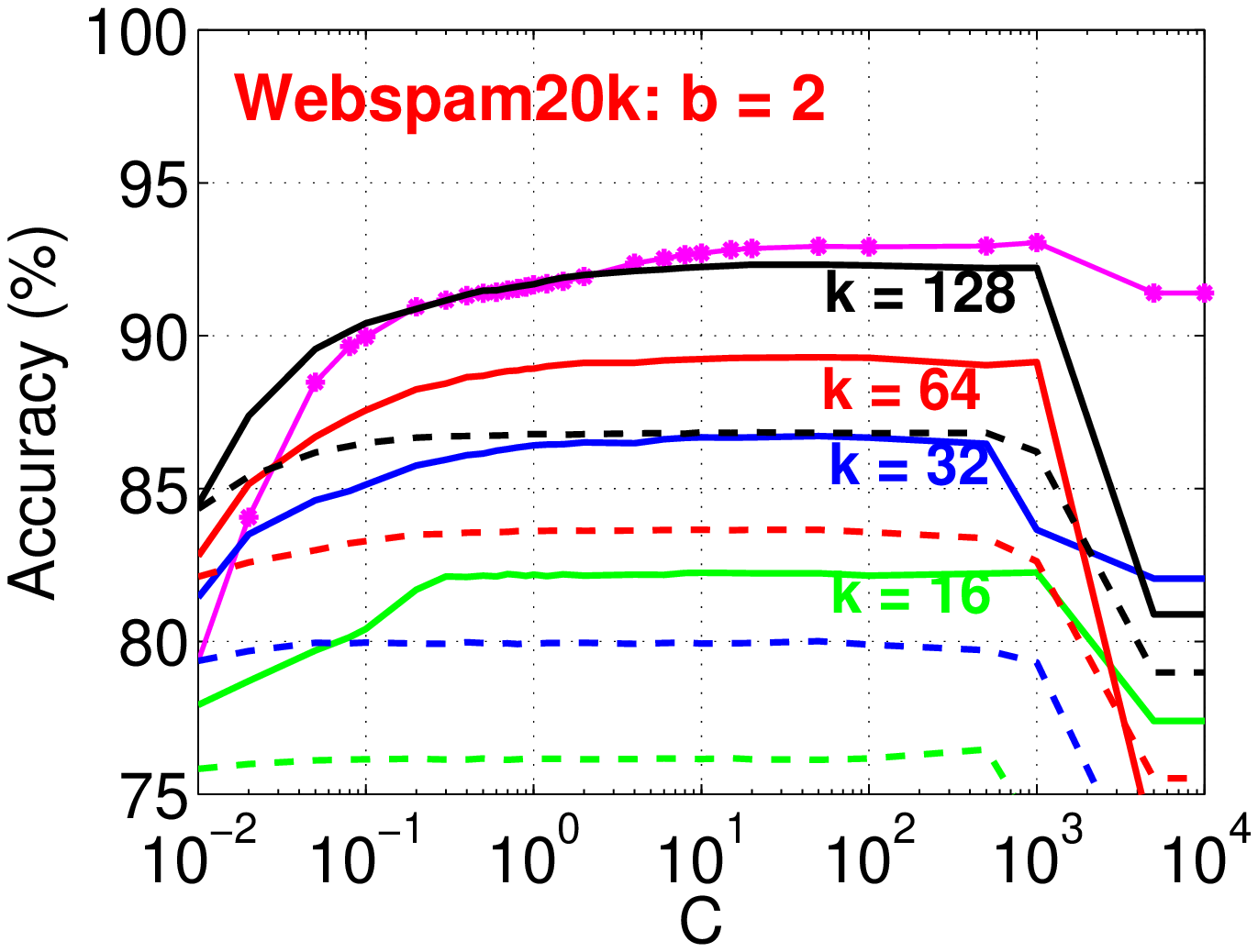}
\includegraphics[width=2.7in]{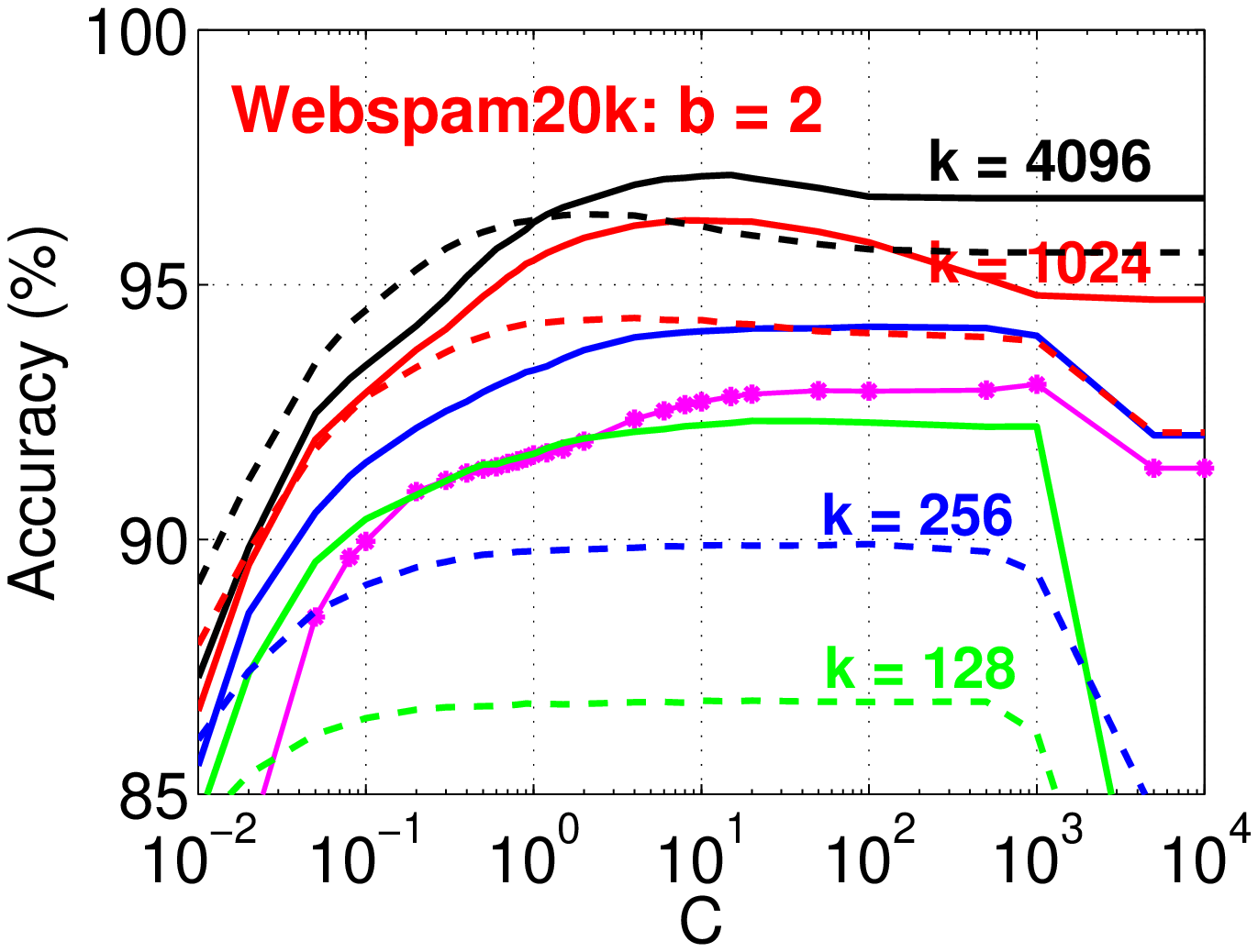}
}

\end{center}
\vspace{-0.3in}
\caption{\textbf{Webspam20k}: Test classification accuracies of the linearized GMM kernel  (solid, GCWS) and  linearized RBF kernel (dashed, NRFF), using LIBLINEAR, averaged over 10 repetitions. In each panel, we report the results on 4 different $k$ (sample size) values: 128, 256, 1024, 4096 (right panels), and  16, 32, 64, 128 (left panels).  We can see that the linearized RBF  (using NRFF) would require substantially more samples in order to reach the same accuracies as the linearized GMM kernel (using GCWS).  The linear SVM results are represented by  solid curves marked by *.
}\label{fig_Webspam20k}
\end{figure}

\newpage\clearpage

Figure~\ref{fig_Webspam20k} reports the test classification accuracies on the \textbf{Webspam20k} dataset. Again, the results  obtained by GCWS hashing and linear classification are noticeably better than the results of NRFF hashing and linear classification, especially when the number of samples ($k$) is not too large (i.e., the left panels). For this dataset, the original dimension is 254. With GCWS hashing and merely $k=128$, we can achieve higher accuracy than using linear classifier on the original data. However, with NRFF hashing, we need almost $k=1024$ in order to outperform linear classifier on the original data. Also, note that it is sufficient to use $b=4$ for GCWS hashing on this dataset.

\vspace{0.08in}

\begin{figure}[h!]
\begin{center}

\mbox{
\includegraphics[width=2.7in]{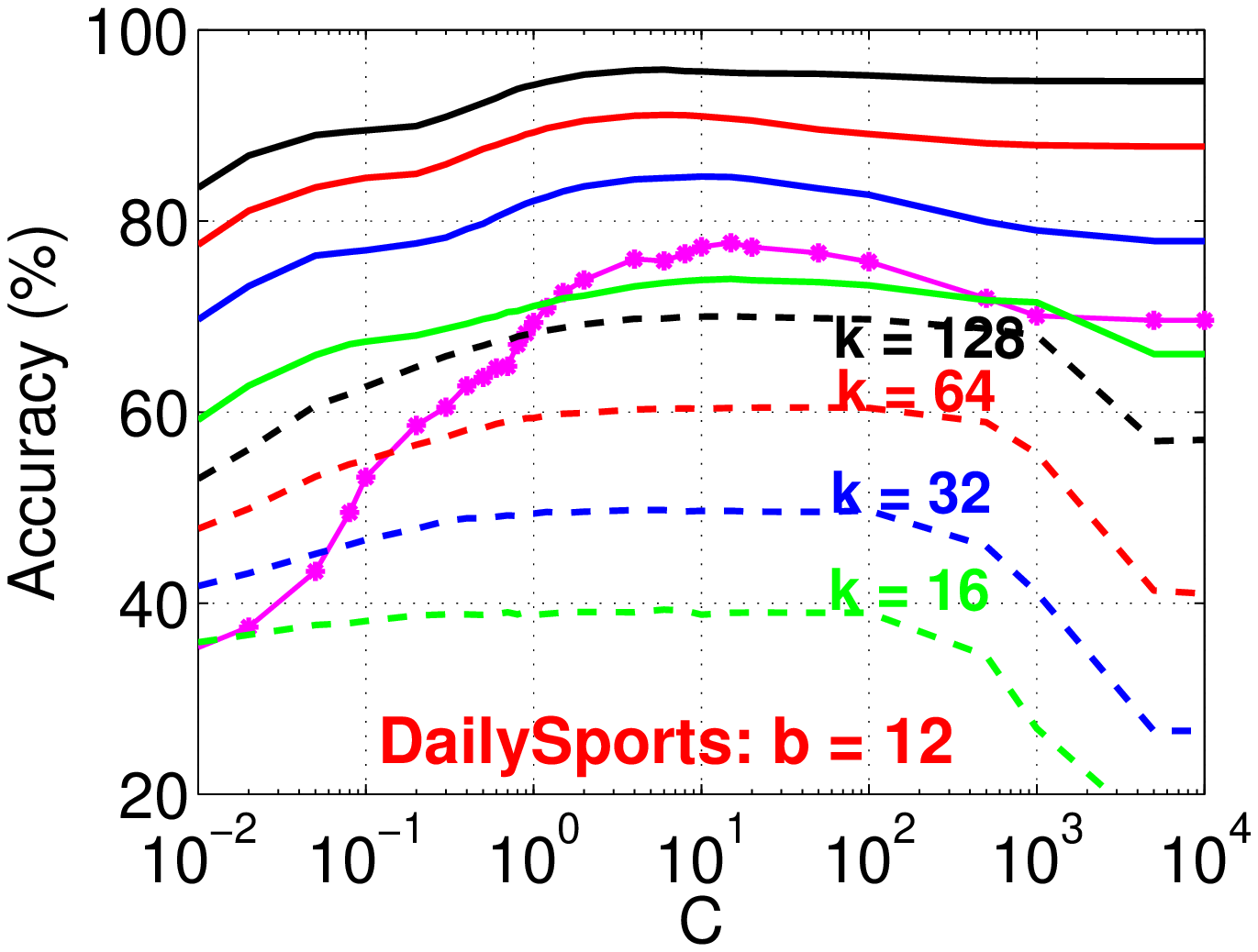}
\includegraphics[width=2.7in]{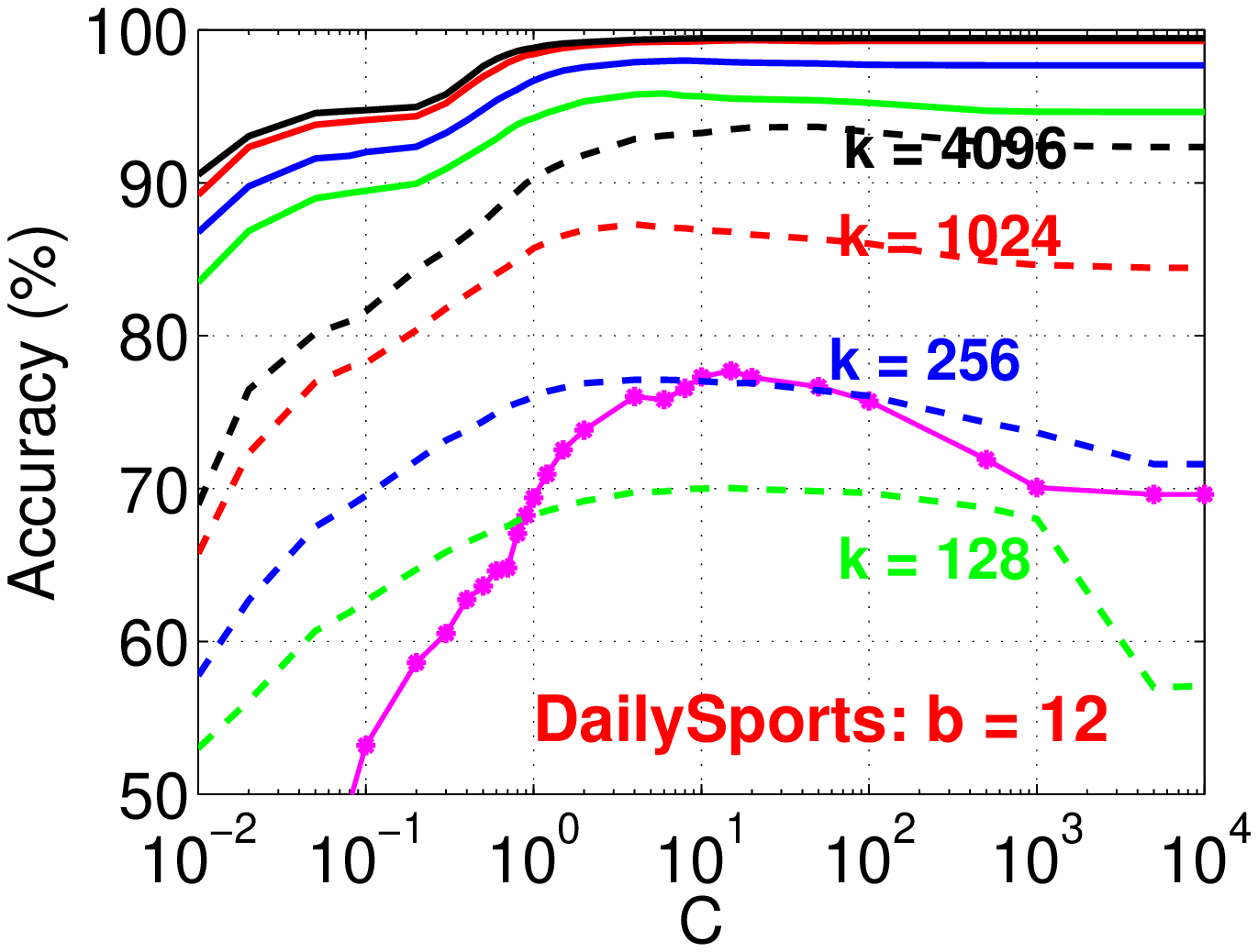}
}

\vspace{-0.04in}

\mbox{
\includegraphics[width=2.7in]{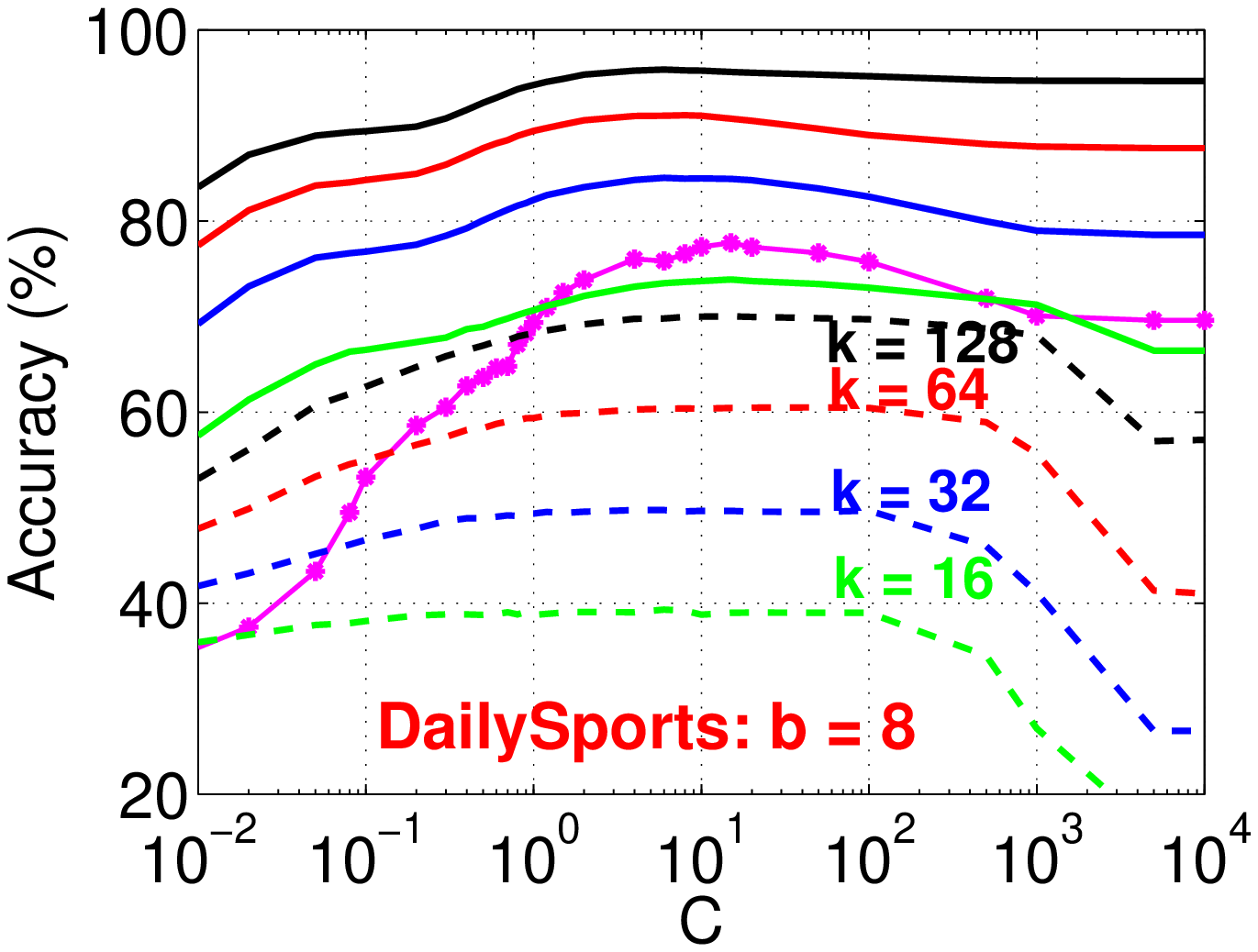}
\includegraphics[width=2.7in]{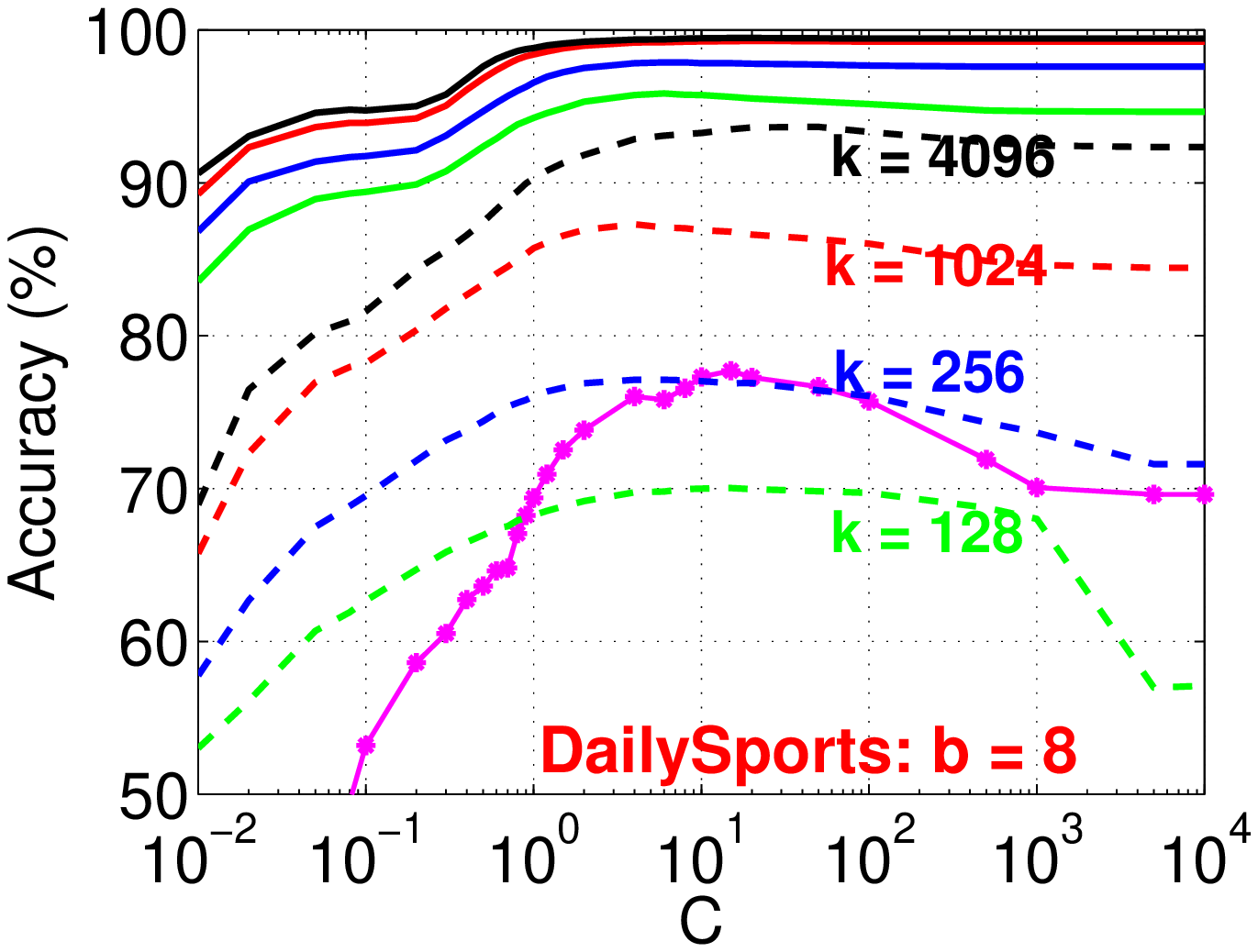}
}

\vspace{-0.04in}

\mbox{
\includegraphics[width=2.7in]{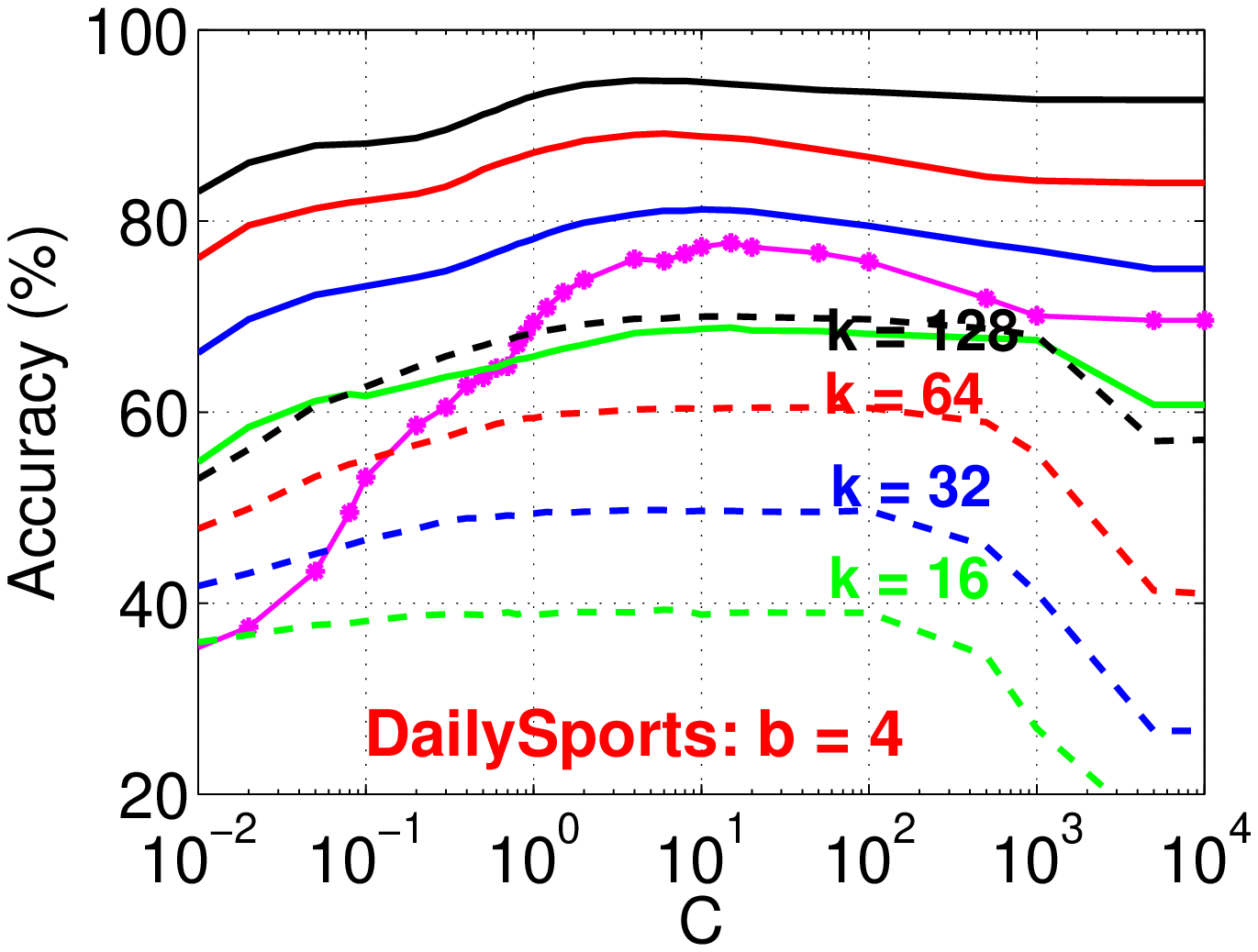}
\includegraphics[width=2.7in]{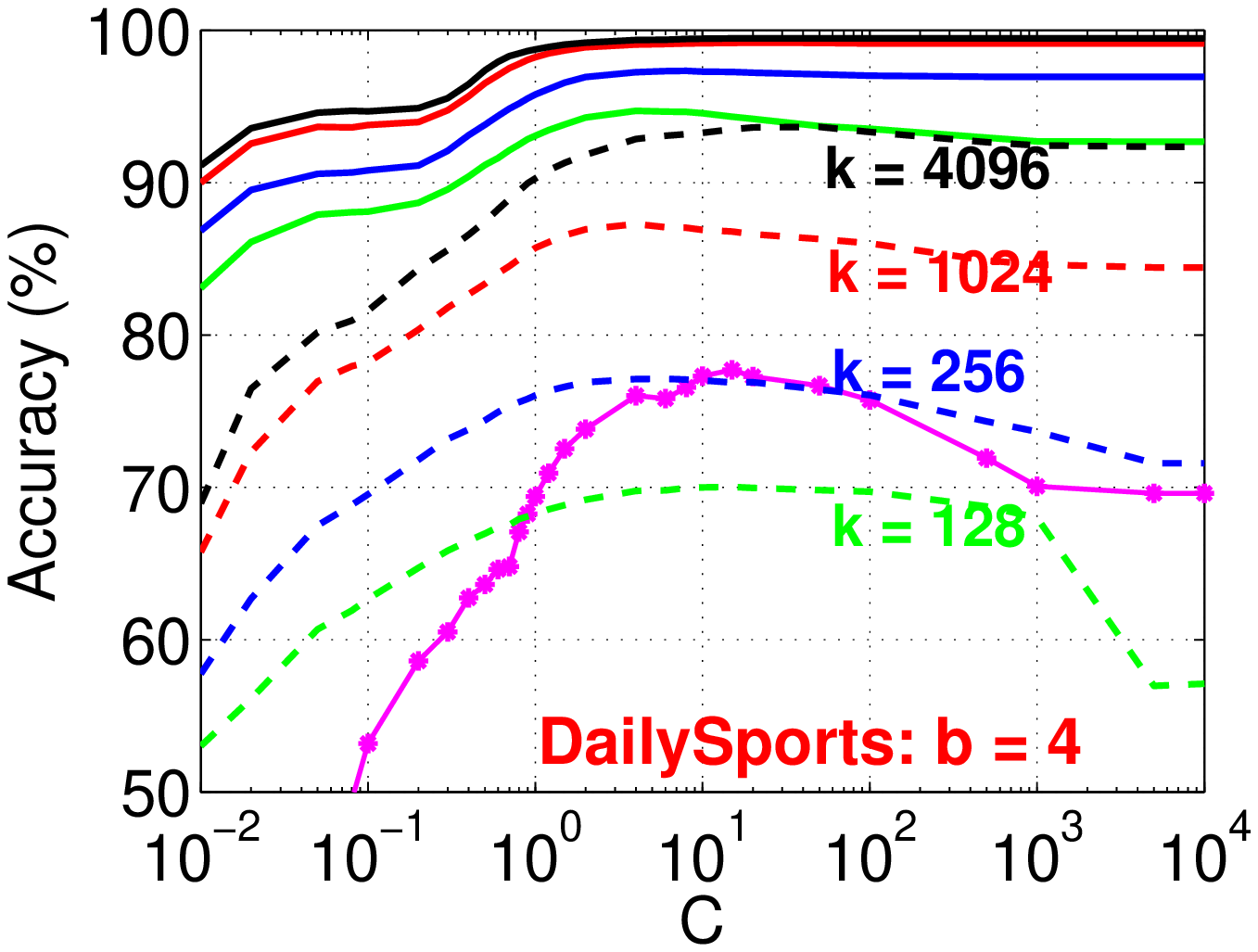}
}

\vspace{-0.04in}

\mbox{
\includegraphics[width=2.7in]{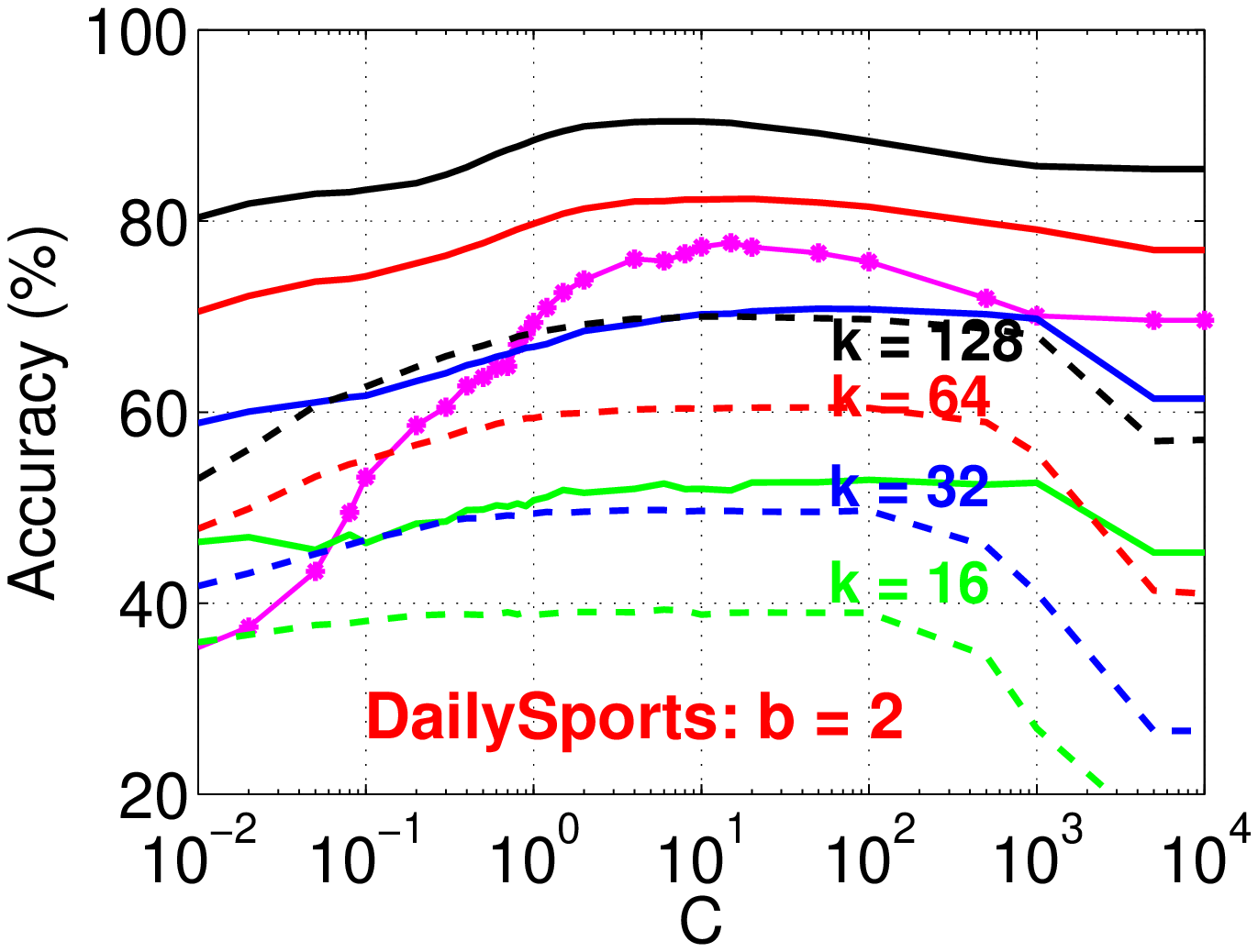}
\includegraphics[width=2.7in]{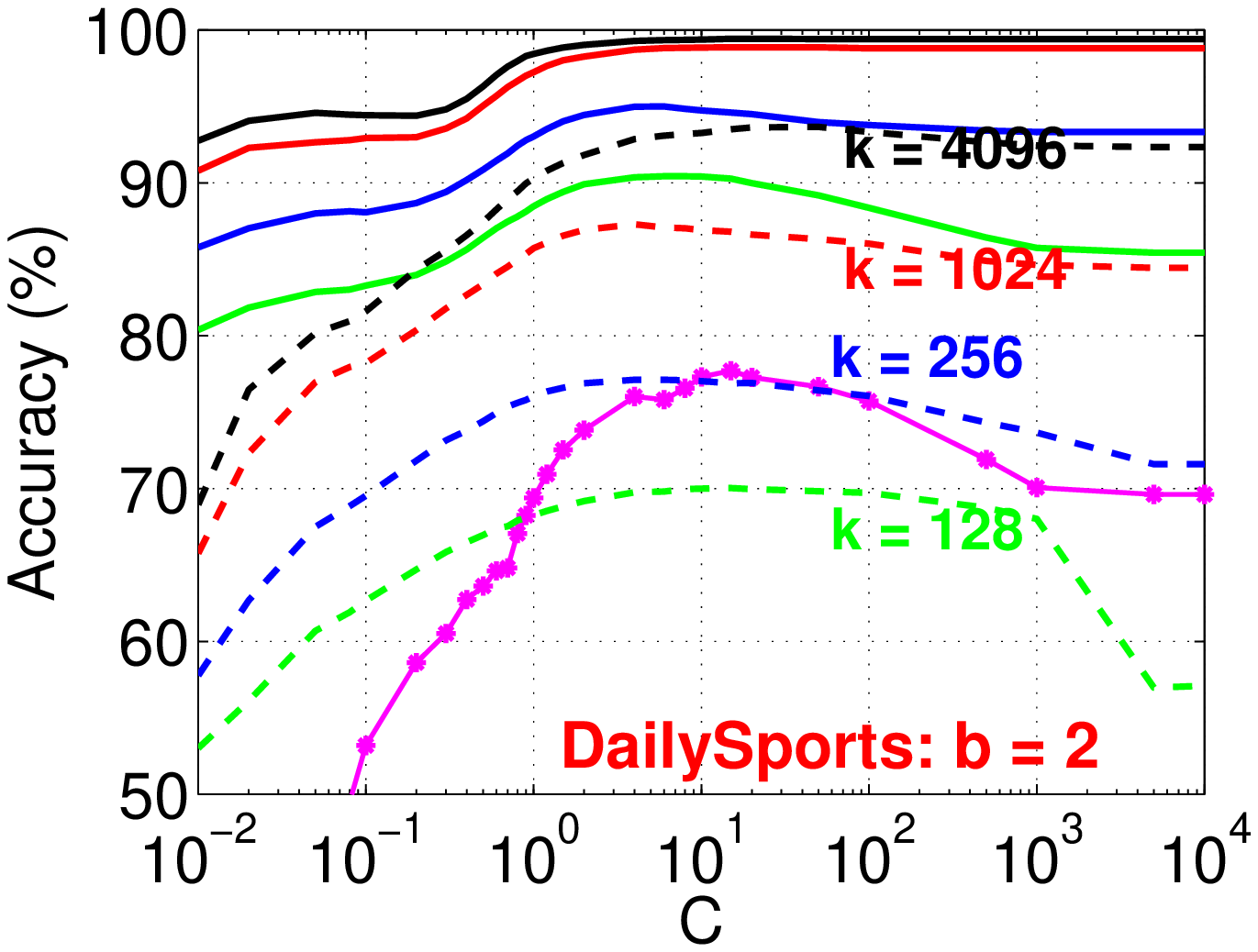}
}

\end{center}
\vspace{-0.3in}
\caption{\textbf{DailySports}: Test classification accuracies of the linearized GMM kernel  (solid) and  linearized RBF kernel (dashed) , using LIBLINEAR. In each panel, we report the results on 4 different $k$ (sample size) values: 128, 256, 1024, 4096 (right panels), and 16, 32, 64, 128 (left panels).  We can see that the linearized RBF  (using NRFF) would require substantially more samples in order to reach the same accuracies as the linearized GMM kernel (using GCWS).
}\label{fig_DailySports}
\end{figure}

\begin{figure}[h!]
\begin{center}

\mbox{
\includegraphics[width=2.7in]{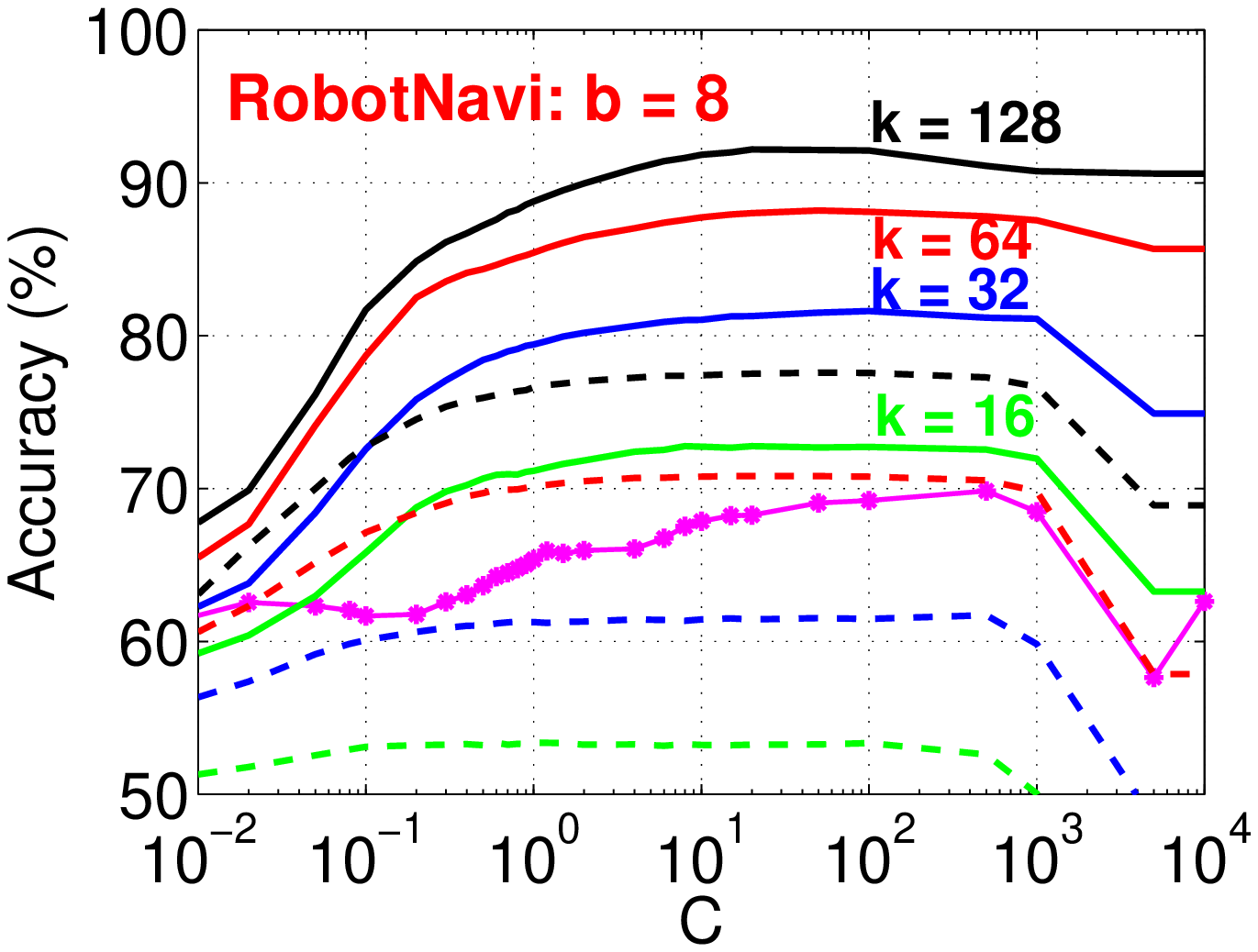}
\includegraphics[width=2.7in]{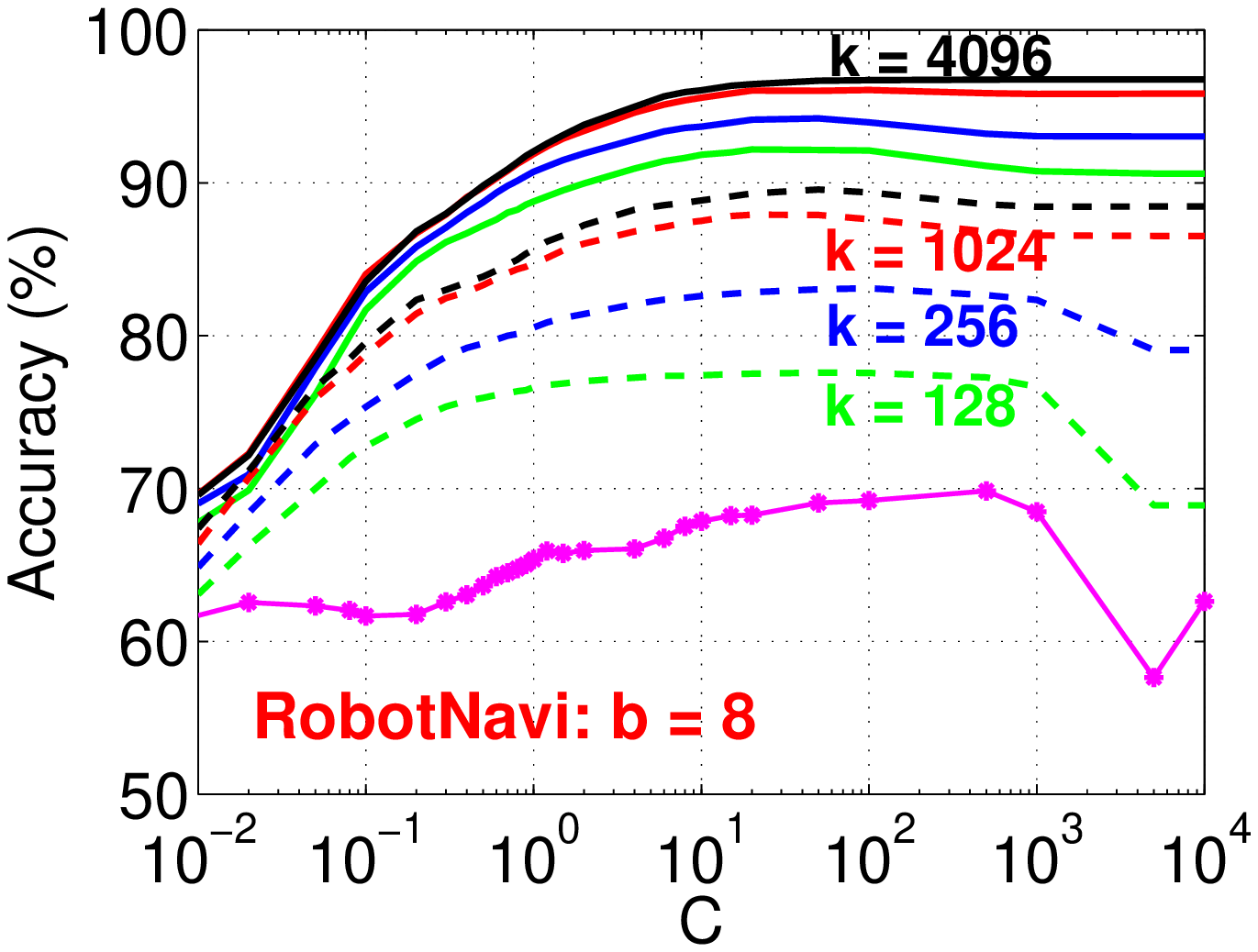}
}

\mbox{
\includegraphics[width=2.7in]{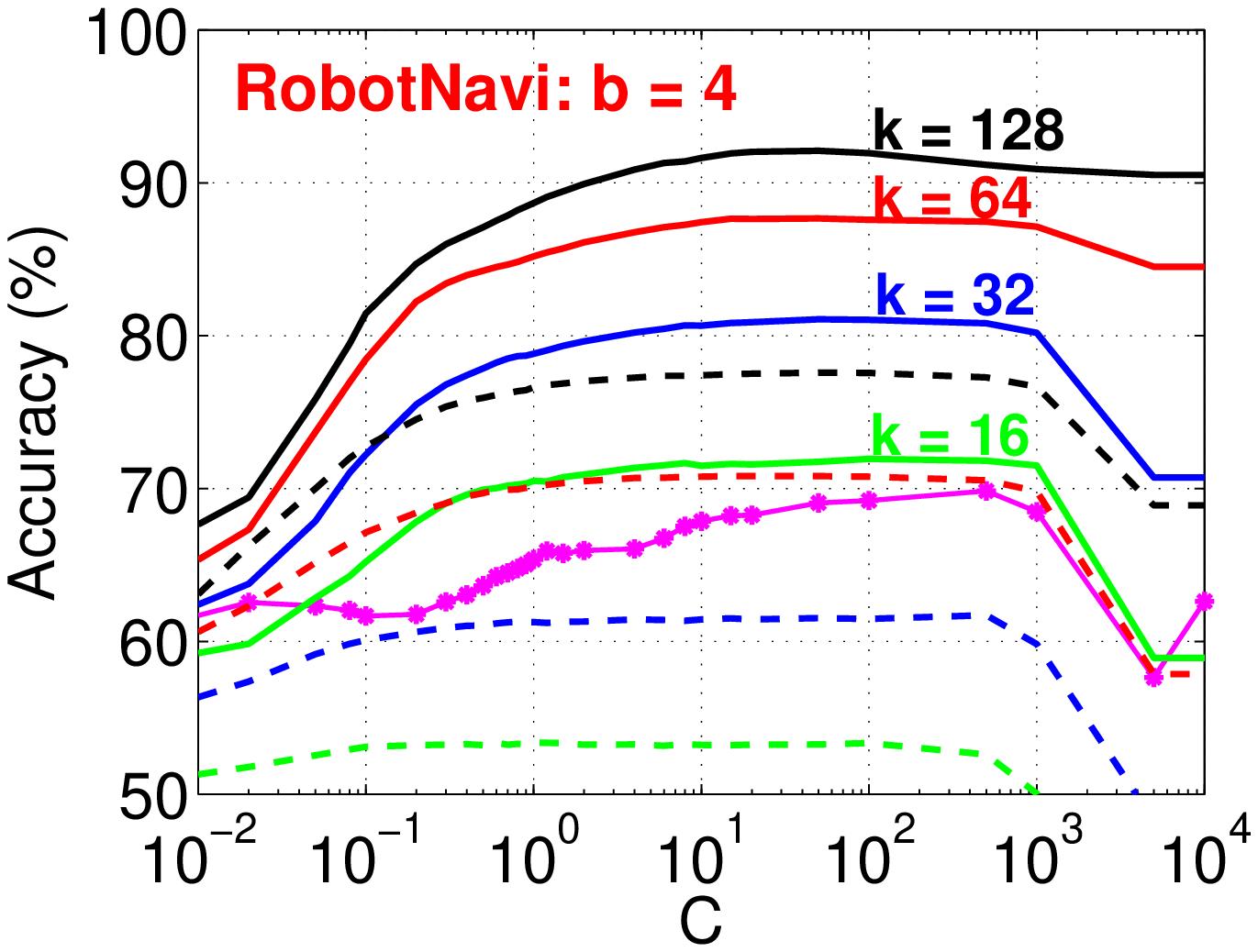}
\includegraphics[width=2.7in]{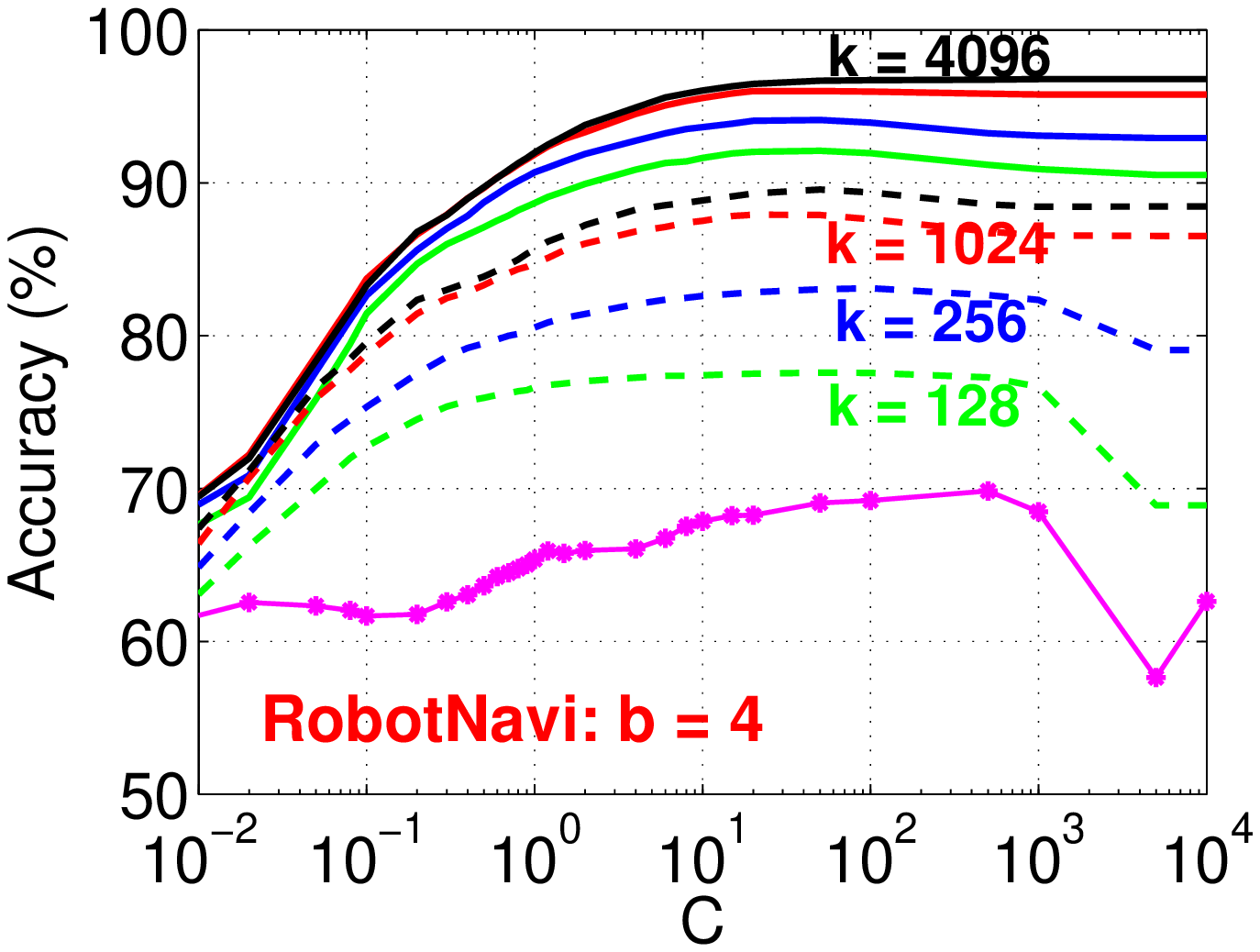}
}

\mbox{
\includegraphics[width=2.7in]{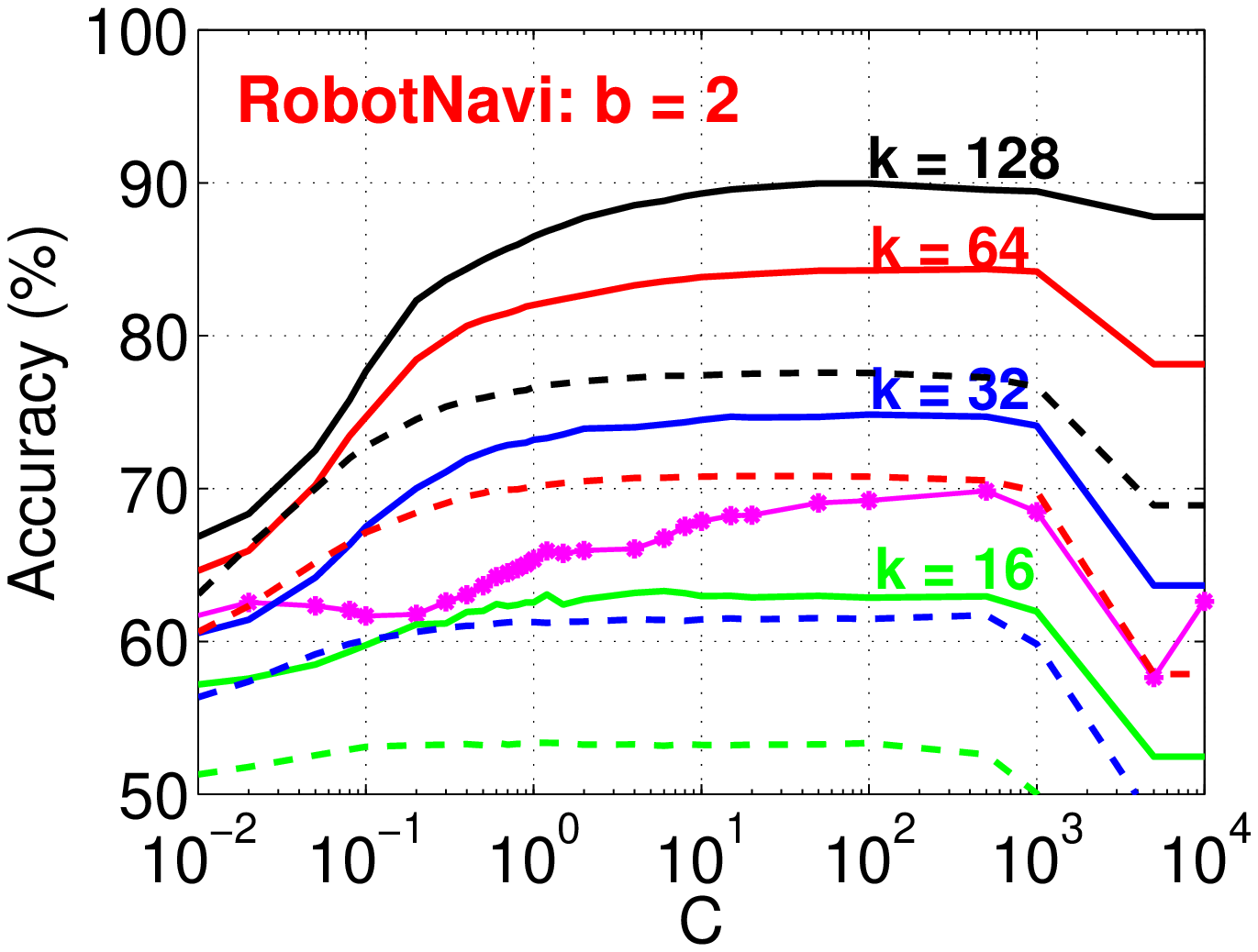}
\includegraphics[width=2.7in]{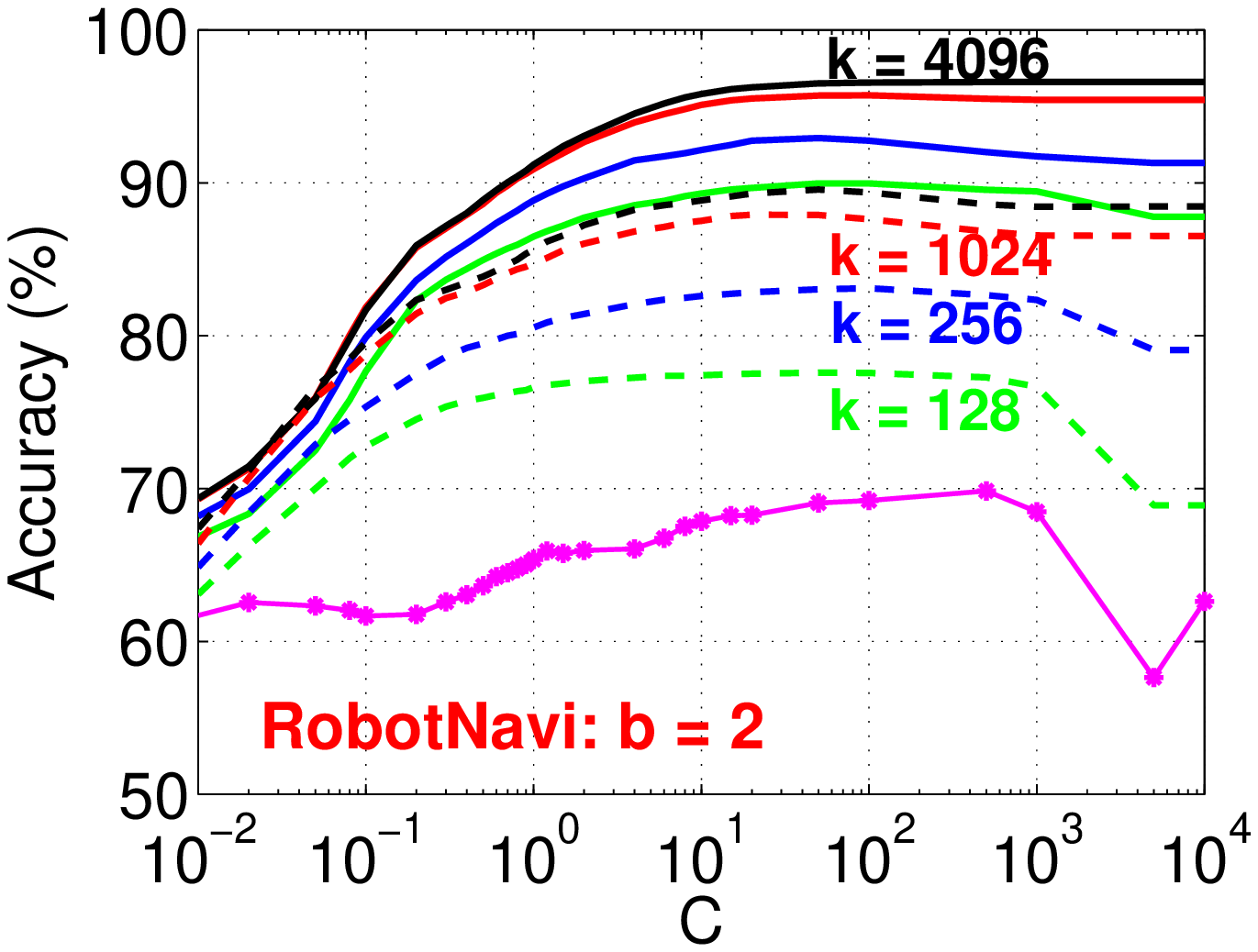}
}

\end{center}
\vspace{-0.3in}
\caption{\textbf{RobotNavi}: Test classification accuracies of the linearized GMM kernel  (solid) and  linearized RBF kernel (dashed), using LIBLINEAR. In each panel, we report the results on 4 different $k$ (sample size) values: 128, 256, 1024, 4096 (right panels), and 16, 32, 64, 128 (left panels).  We can see that the linearized RBF  (using NRFF) would require substantially more samples in order to reach the same accuracies as the linearized GMM kernel (using GCWS).
}\label{fig_RobotNavi}
\end{figure}

Figure~\ref{fig_DailySports} and Figure~\ref{fig_RobotNavi} report the test classification accuracies on the \textbf{DailySports} dataset and the \textbf{RobotNavi} dataset, respectively. For both datasets, the original GMM kernel noticeably outperforms the original RBF kernel. Not surprisingly, NRFF hashing requires substantially more samples in order to reach similar accuracy as GCWS hashing, on both datasets. The results also illustrate that the parameter $b$ (i.e., the number of bits we store for each GCWS hashed value $i^*$) does matter, but nevertheless, as long as $b\geq 4$, the results do not differ much.

\vspace{0.08in}

\begin{figure}[h!]
\begin{center}
\mbox{
\includegraphics[width=2.7in]{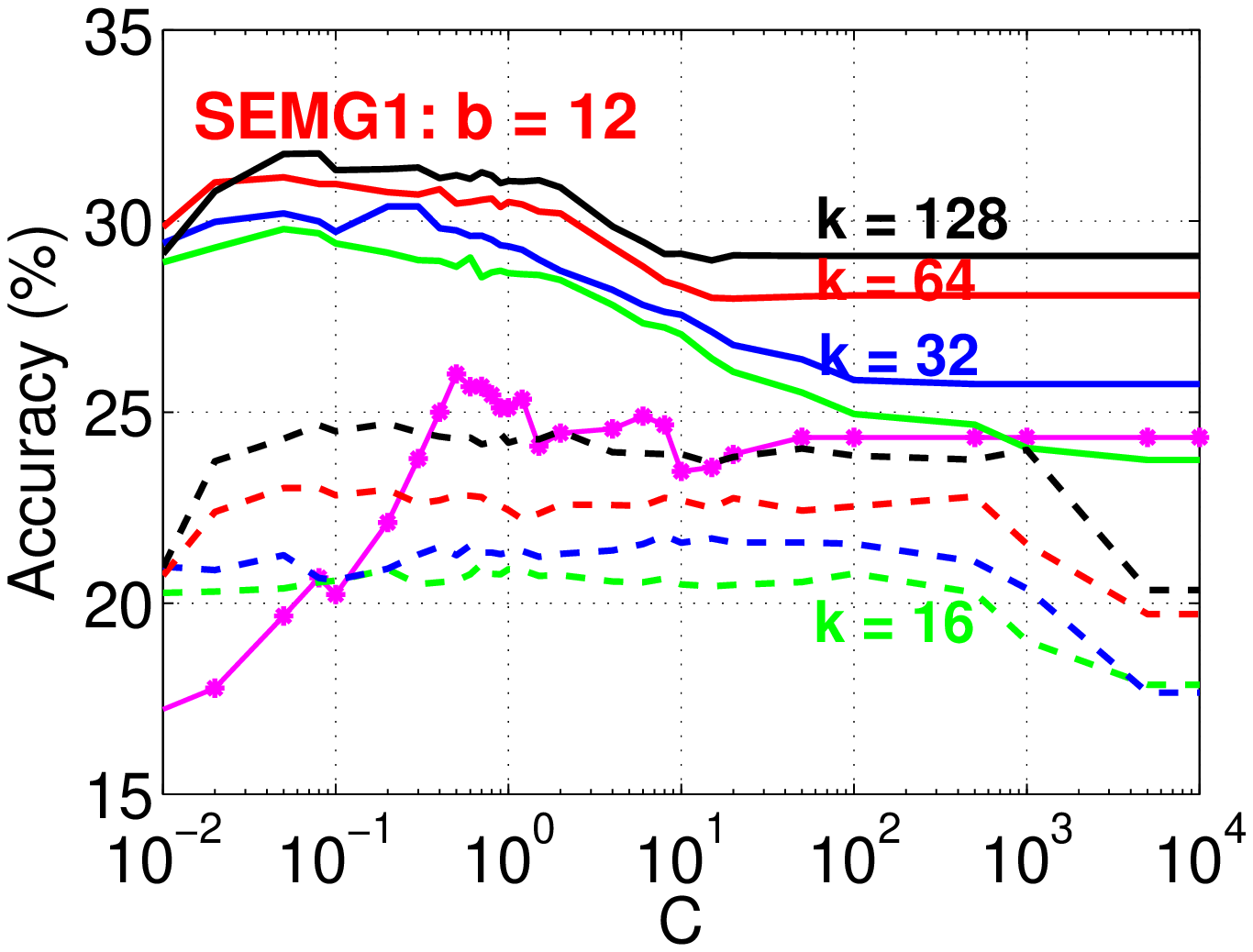}
\includegraphics[width=2.7in]{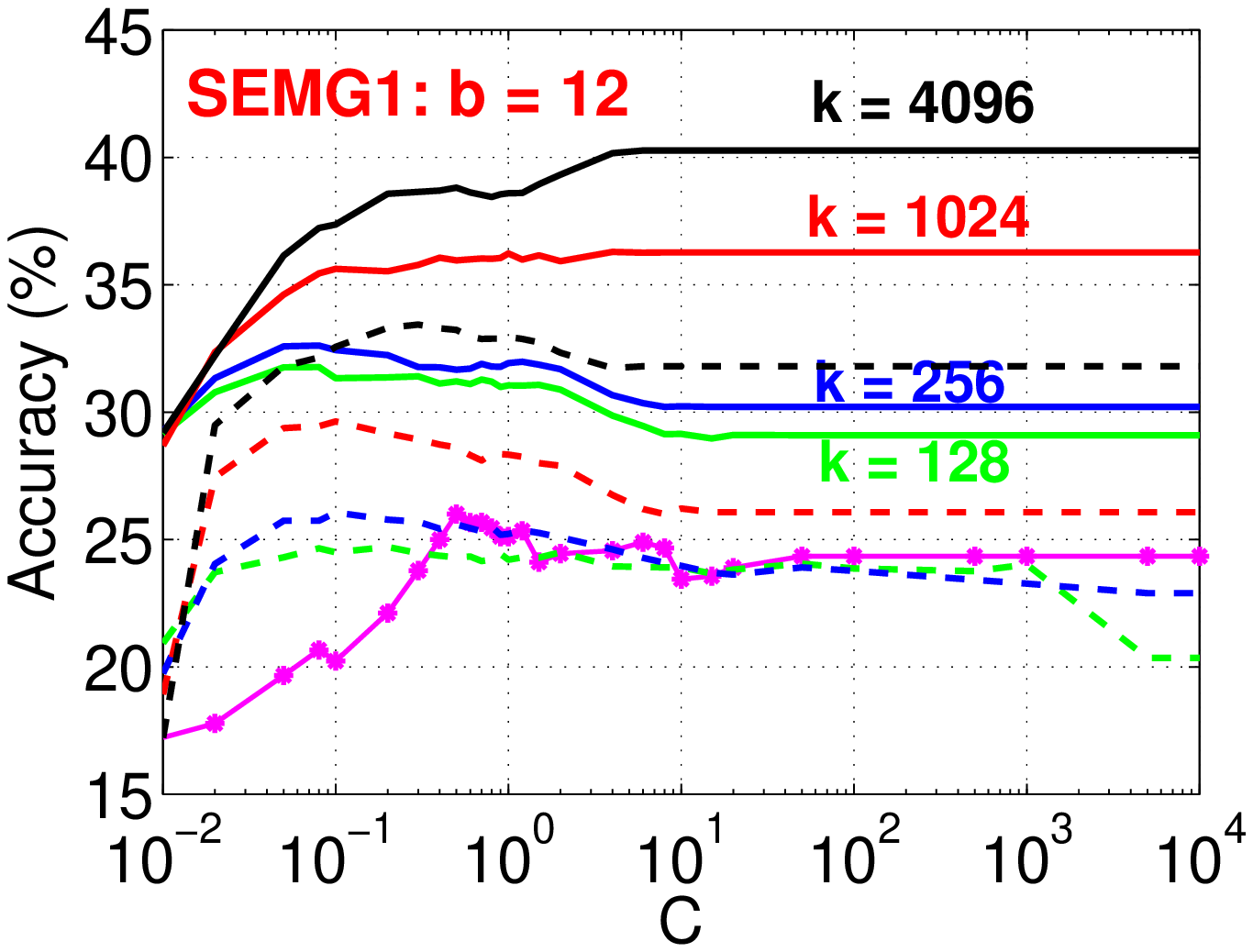}
}

\vspace{-0.04in}

\mbox{
\includegraphics[width=2.7in]{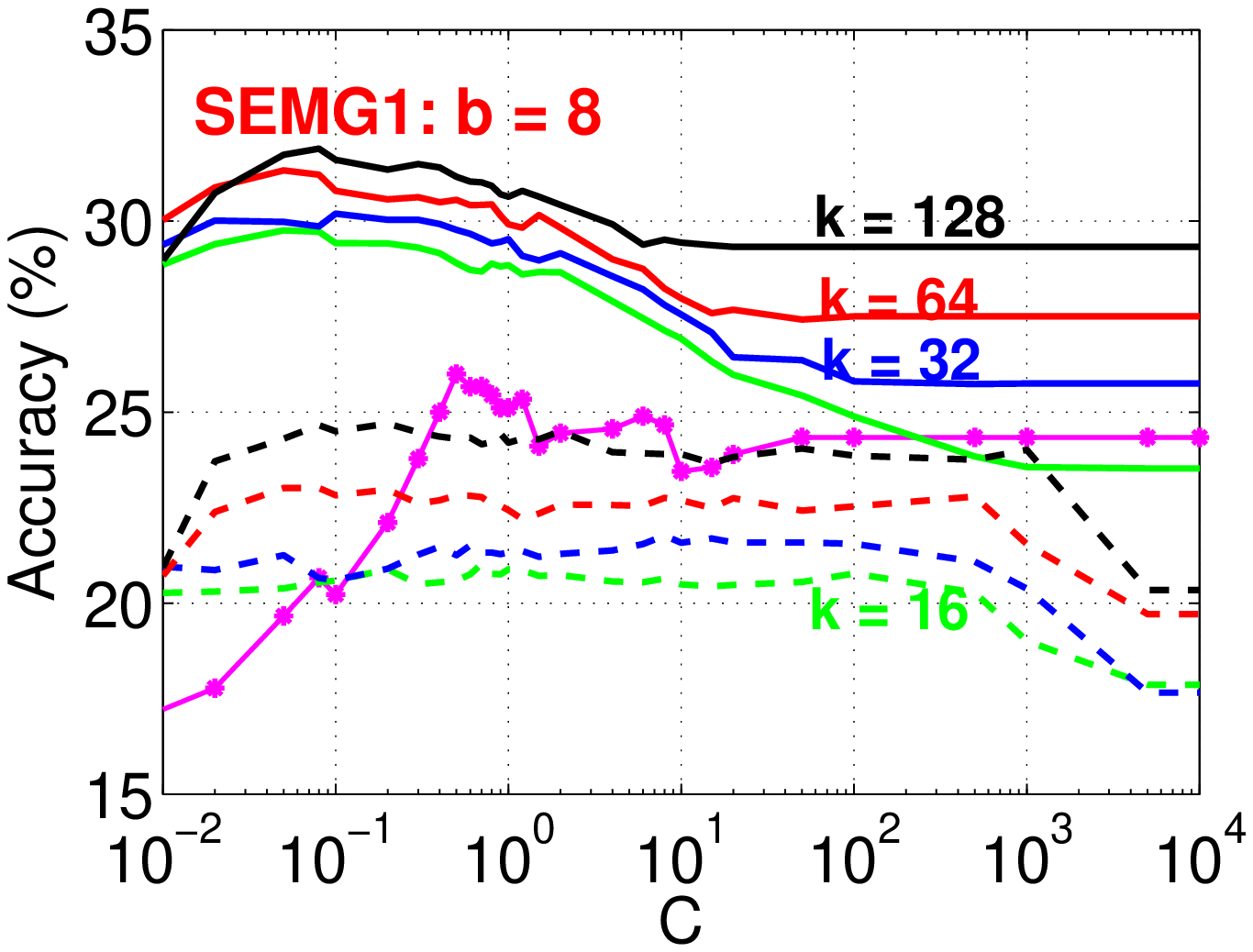}
\includegraphics[width=2.7in]{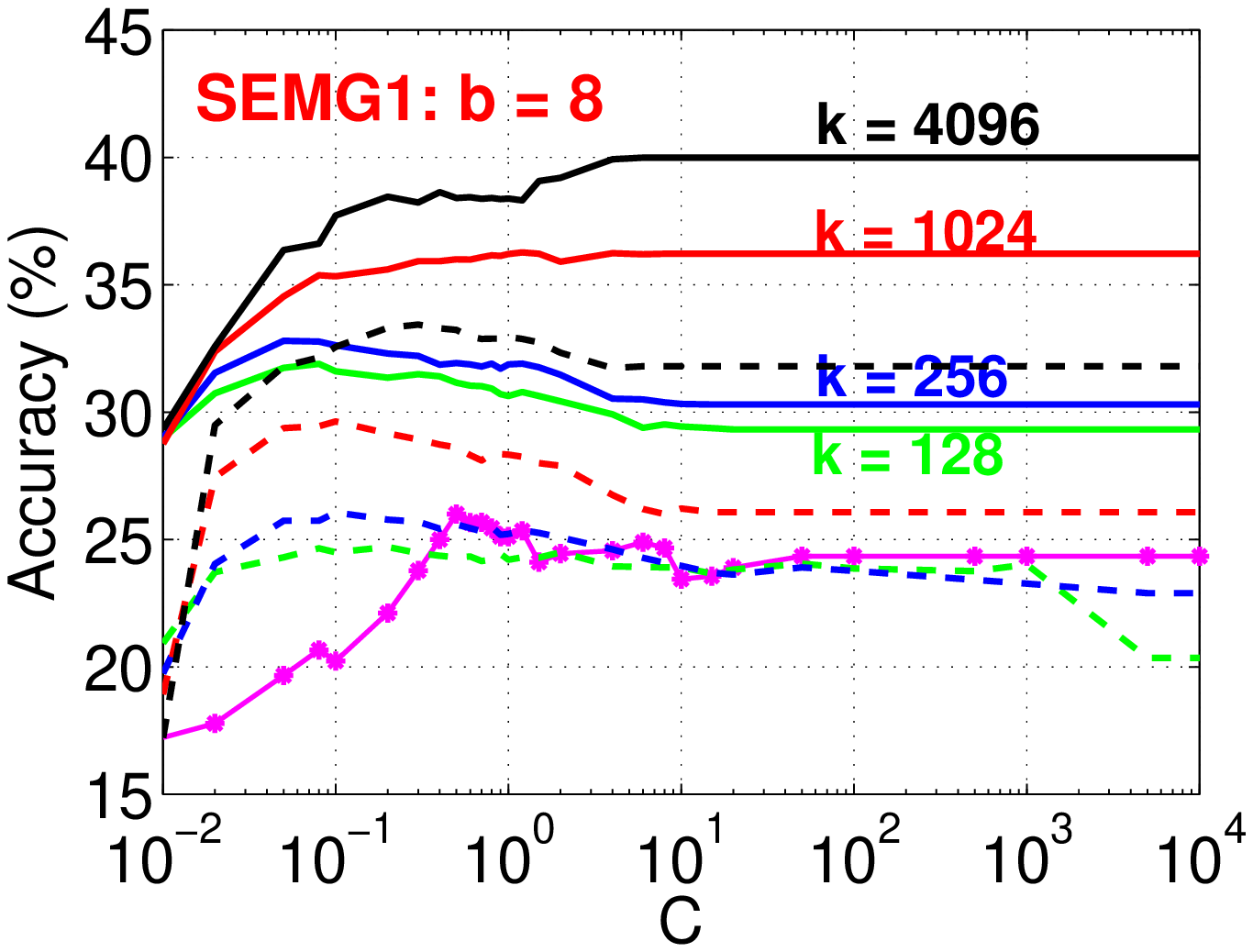}
}

\vspace{-0.04in}

\mbox{
\includegraphics[width=2.7in]{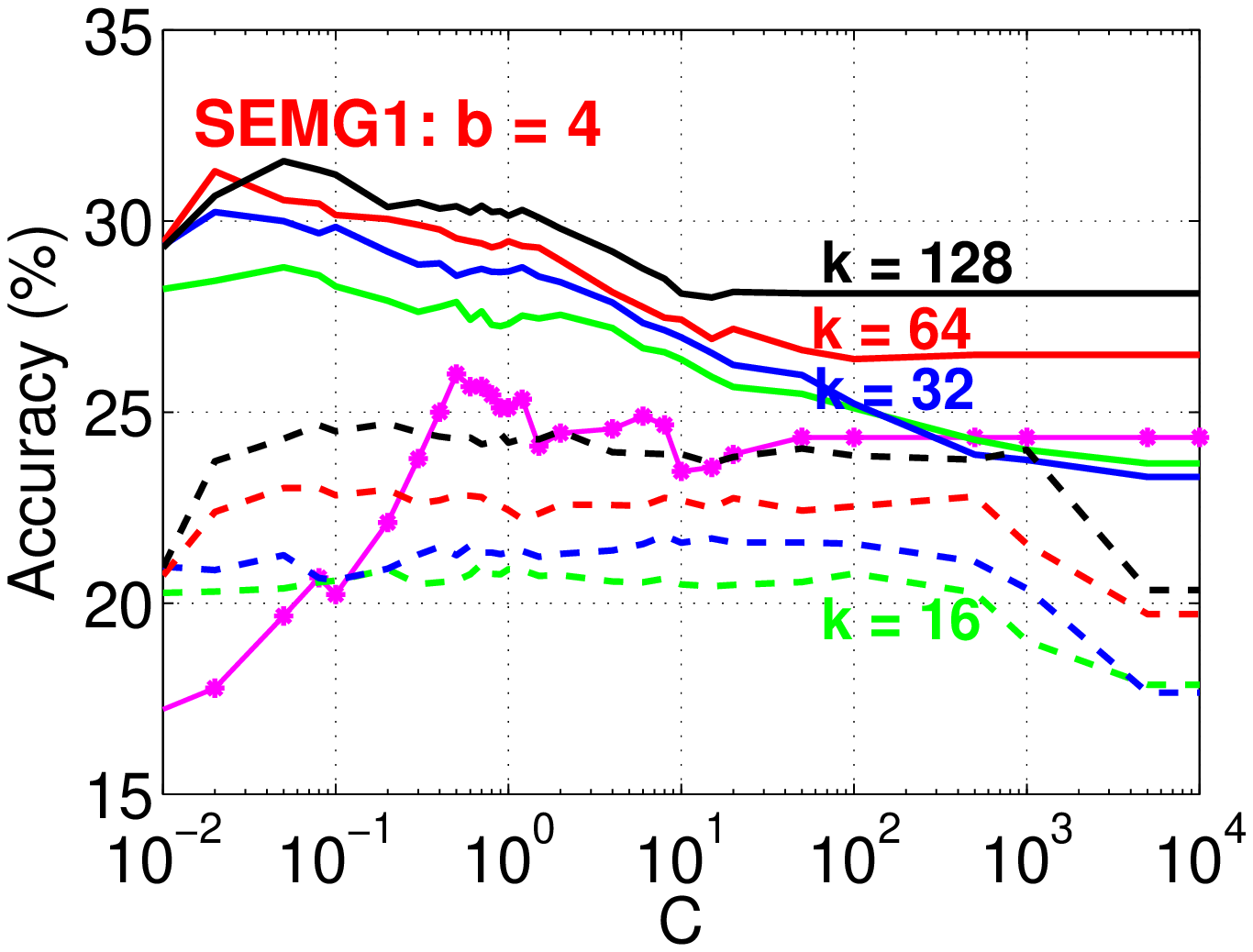}
\includegraphics[width=2.7in]{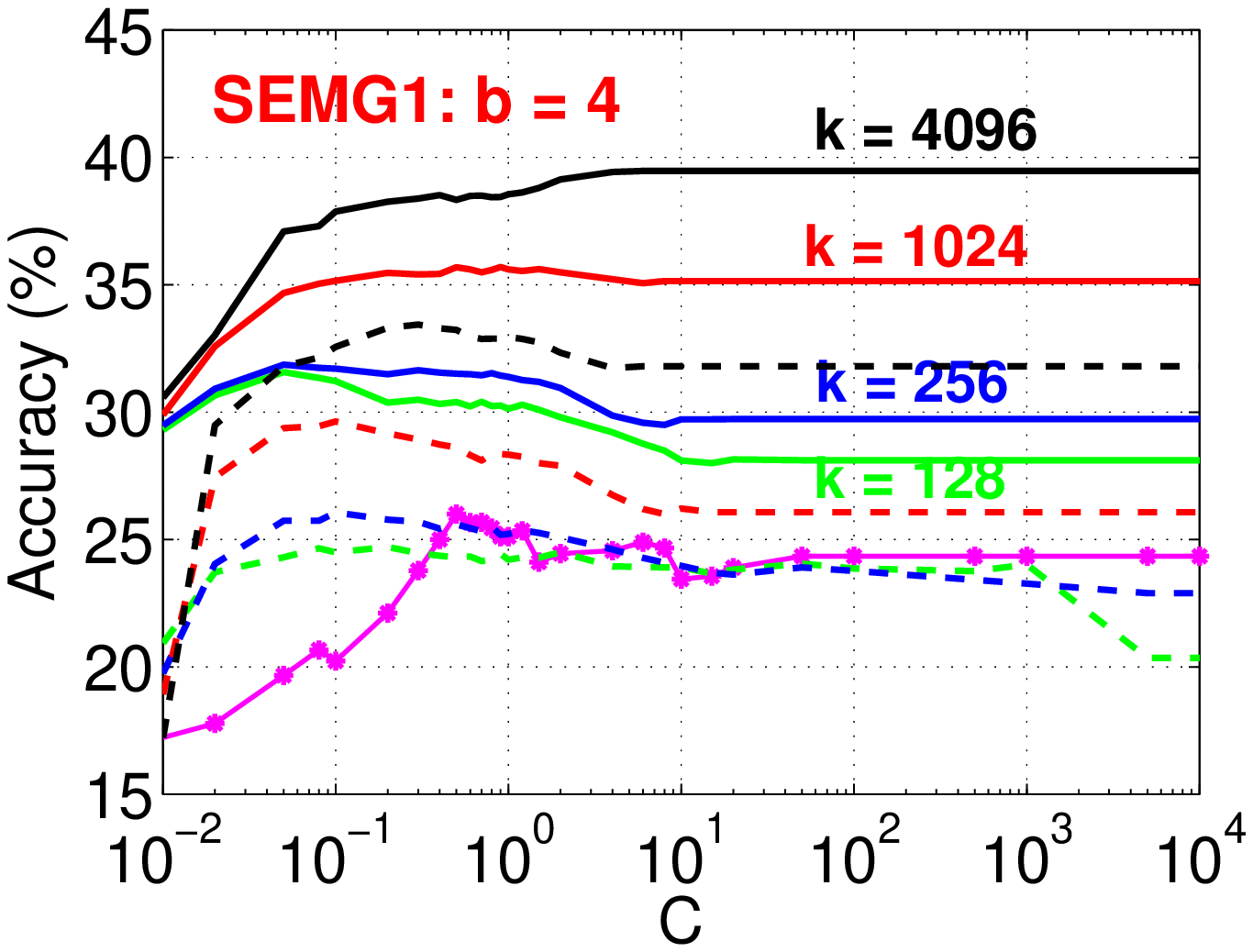}
}

\vspace{-0.04in}

\mbox{
\includegraphics[width=2.7in]{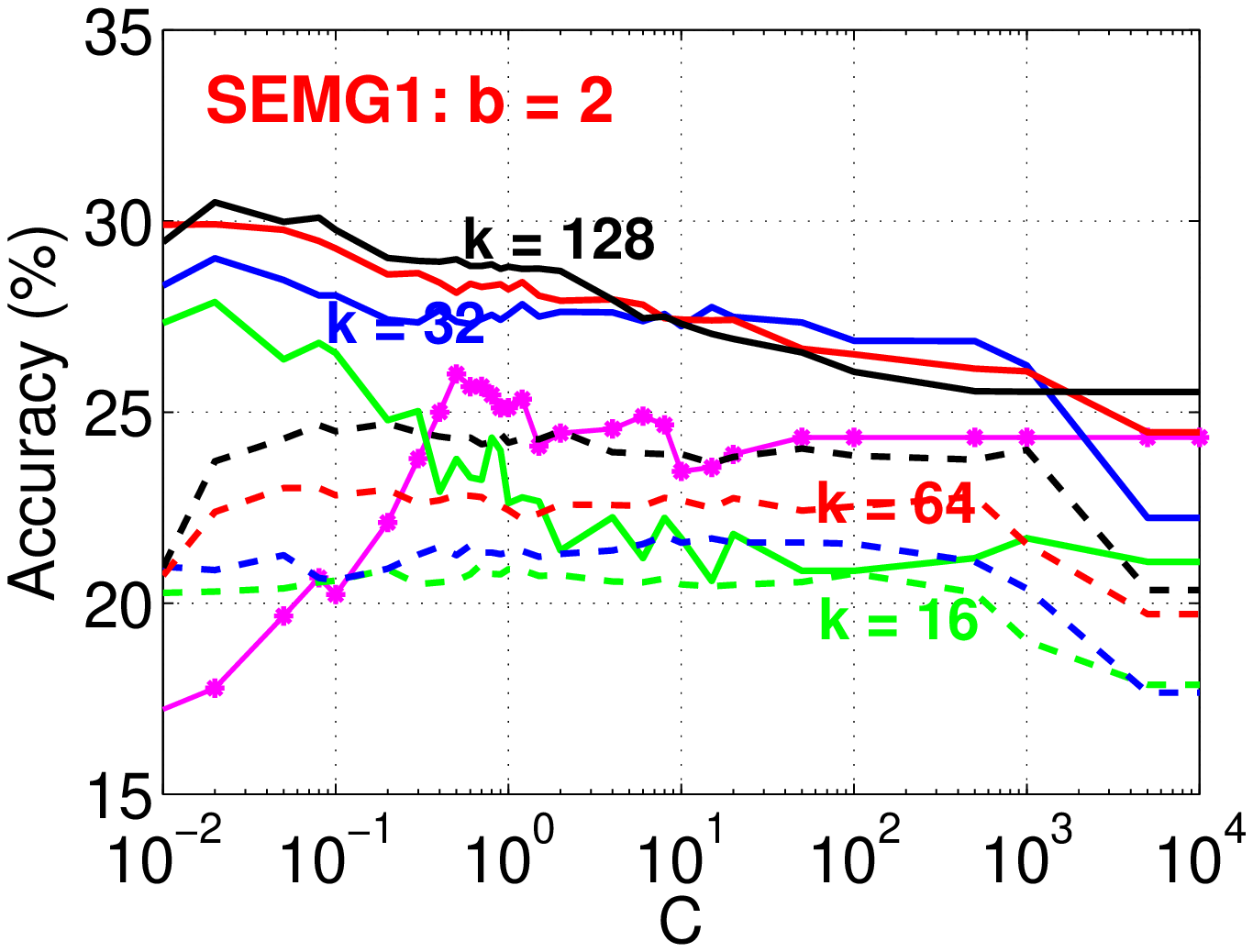}
\includegraphics[width=2.7in]{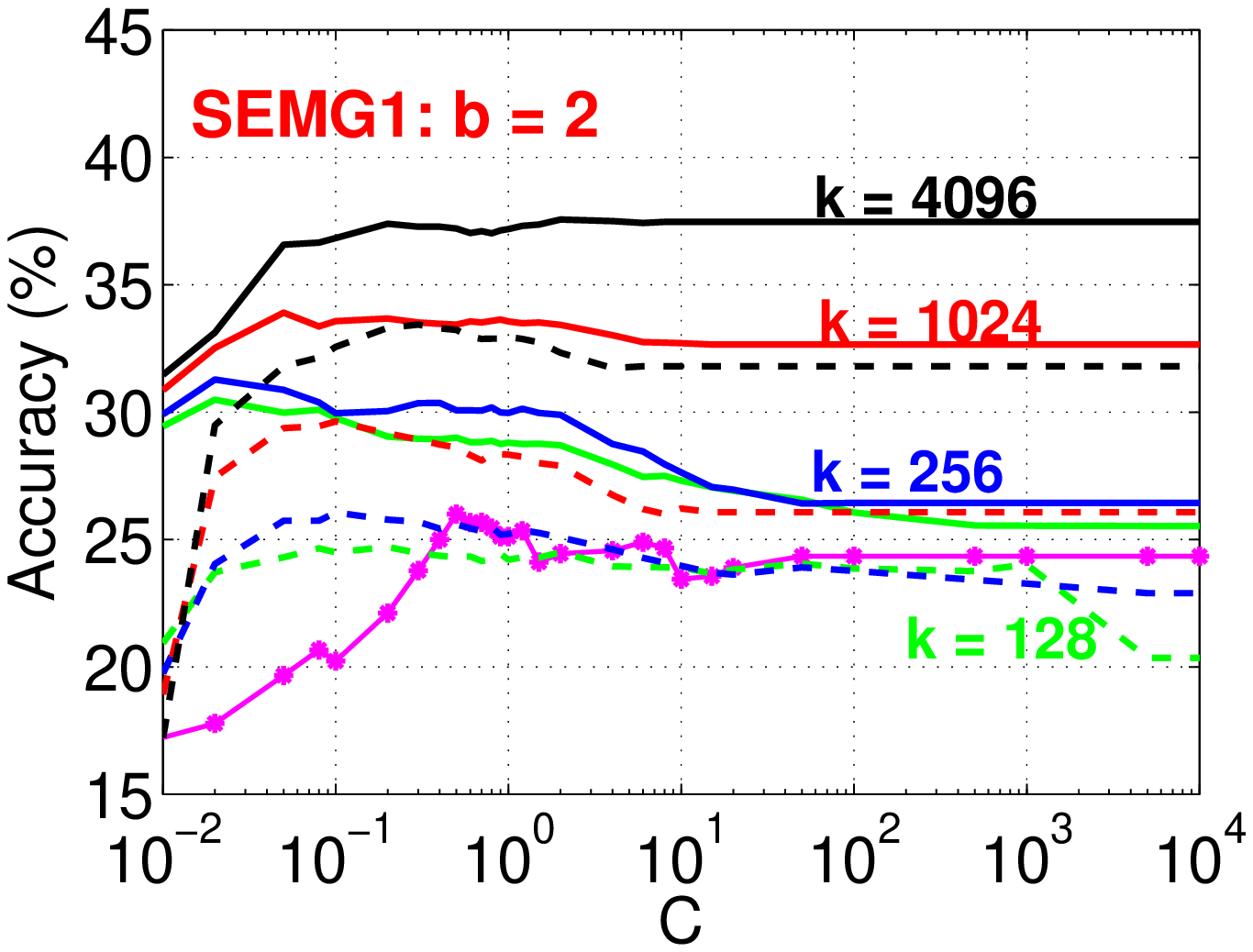}
}

\end{center}
\vspace{-0.3in}
\caption{
\textbf{SEMG1}: Test classification accuracies of the linearized GMM kernel  (solid) and  linearized RBF kernel (dashed), using LIBLINEAR. Again, we can see that the linearized RBF would require substantially more samples in order to reach the same accuracies as the linearized GMM kernel. Note that, for this dataset, the original RBF kernel actually outperforms the original GMM kernel as shown in Table~\ref{tab_UCI}.
}\label{fig_SEMG1}
\end{figure}

\begin{figure}[h!]
\begin{center}
\mbox{
\includegraphics[width=2.7in]{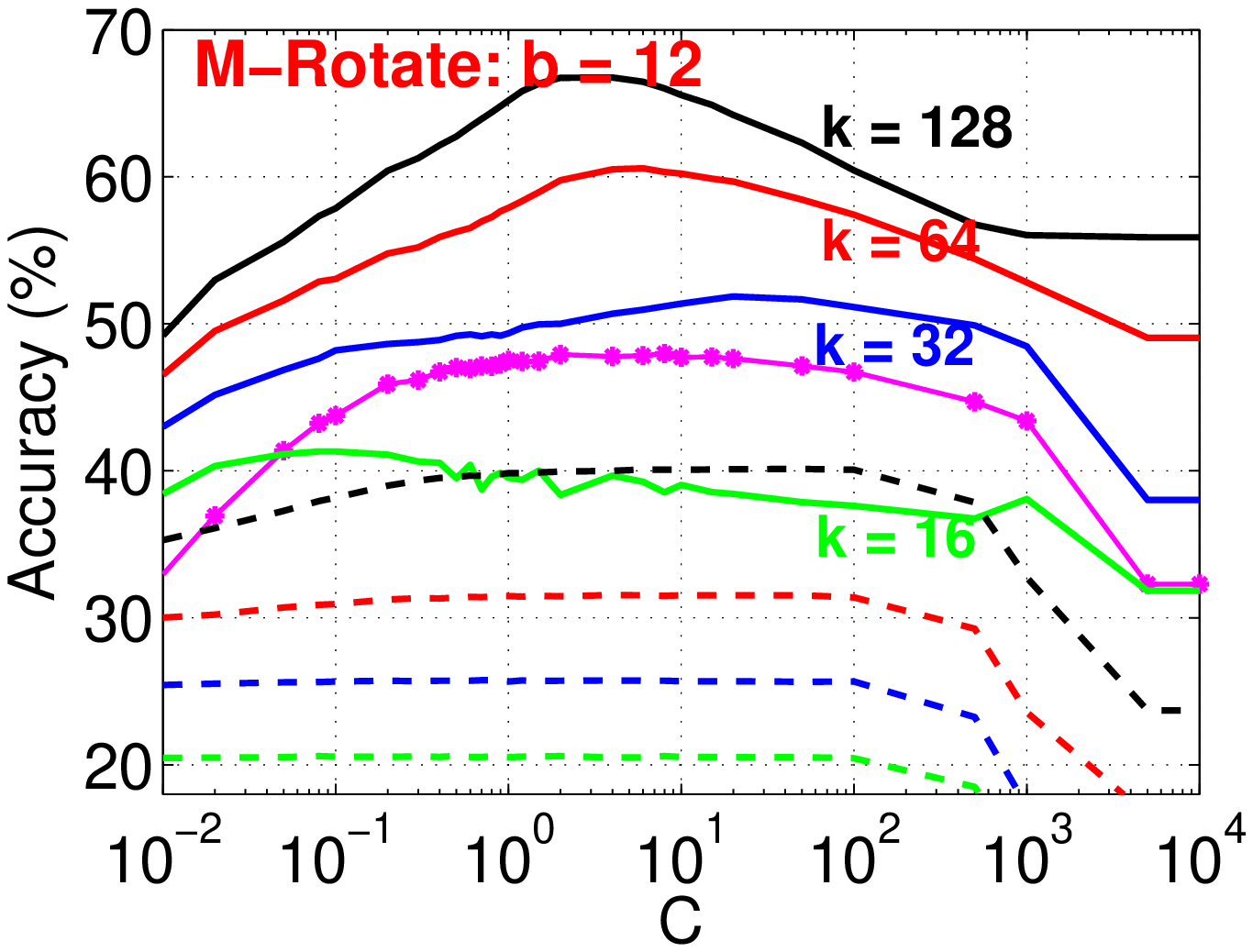}
\includegraphics[width=2.7in]{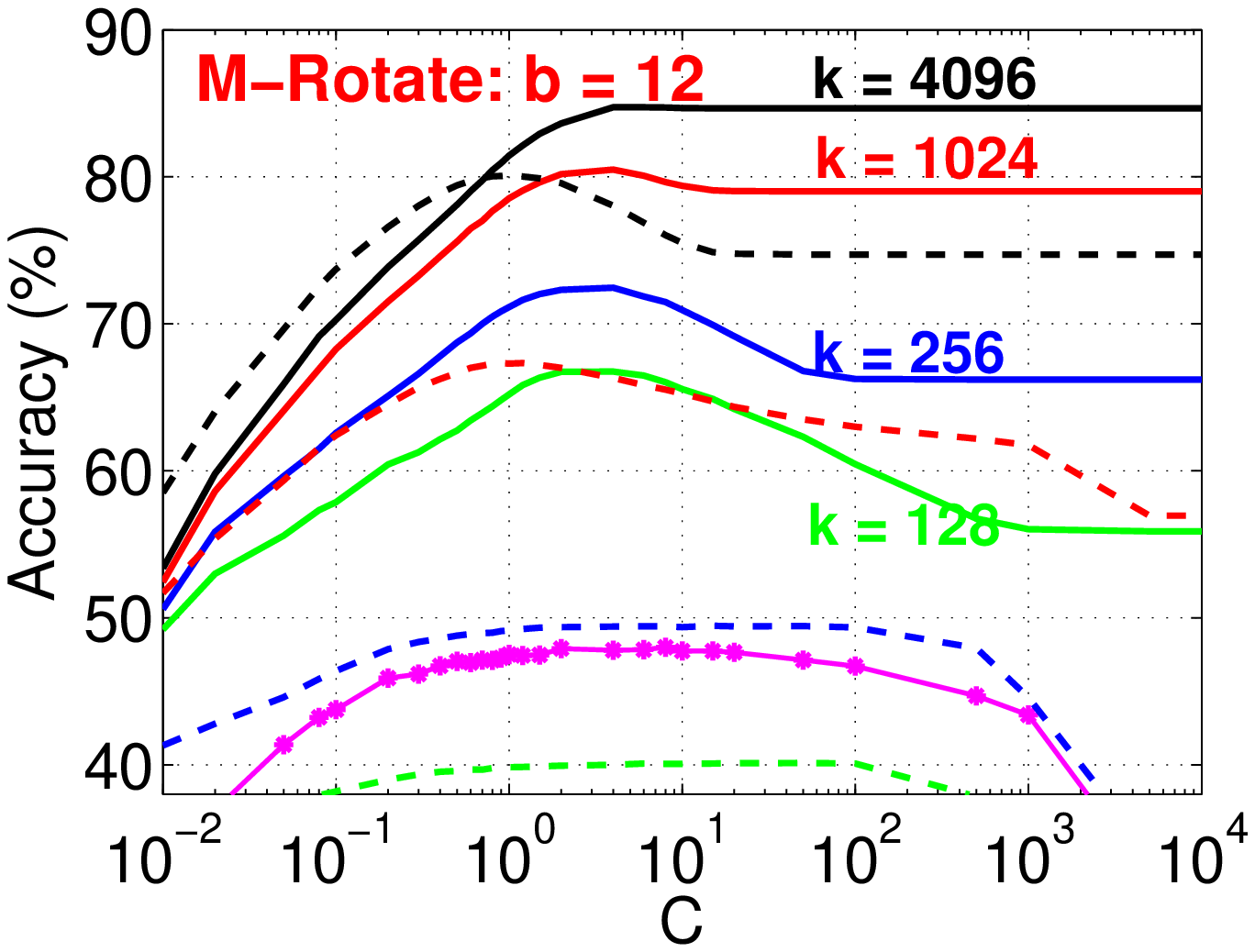}
}

\mbox{
\includegraphics[width=2.7in]{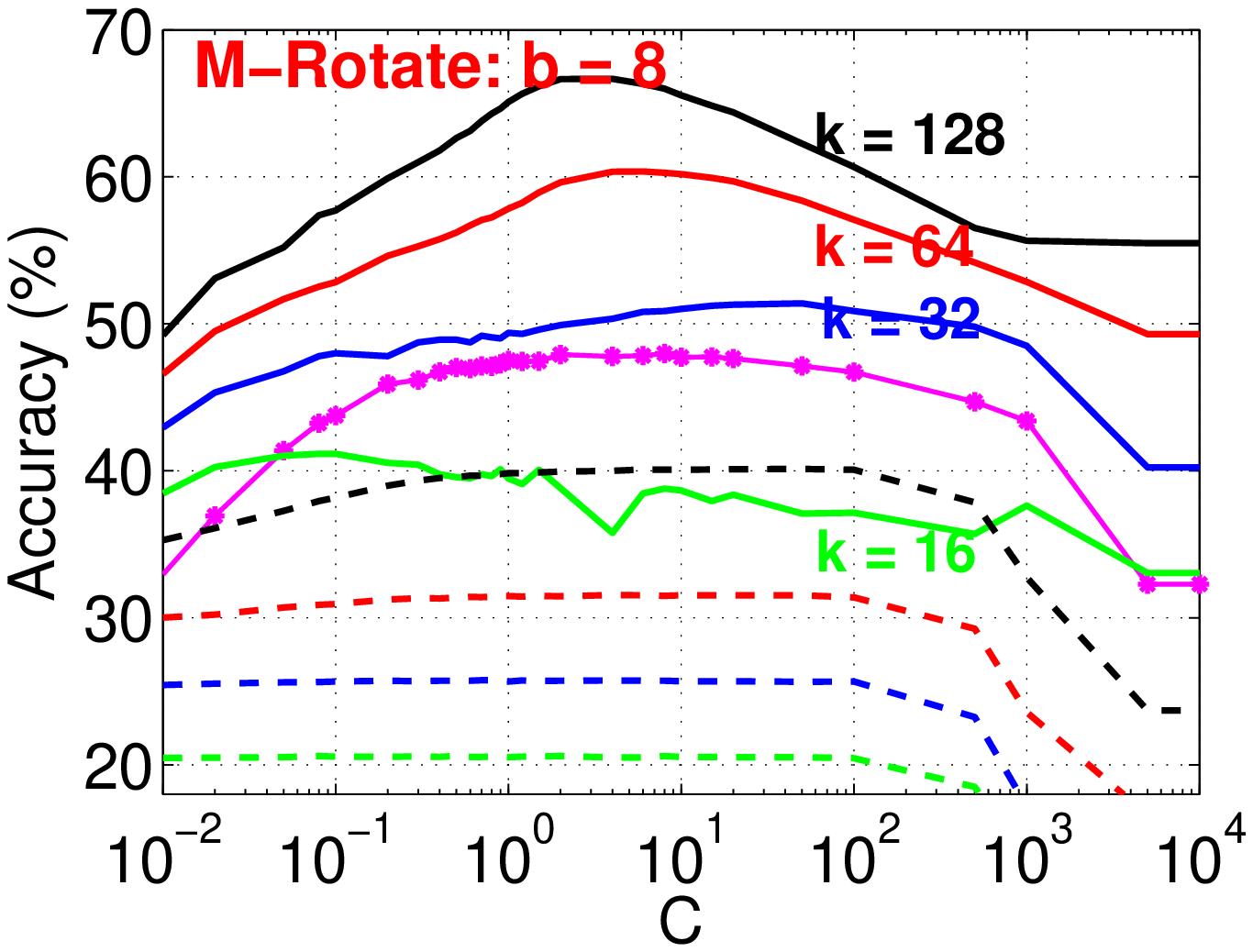}
\includegraphics[width=2.7in]{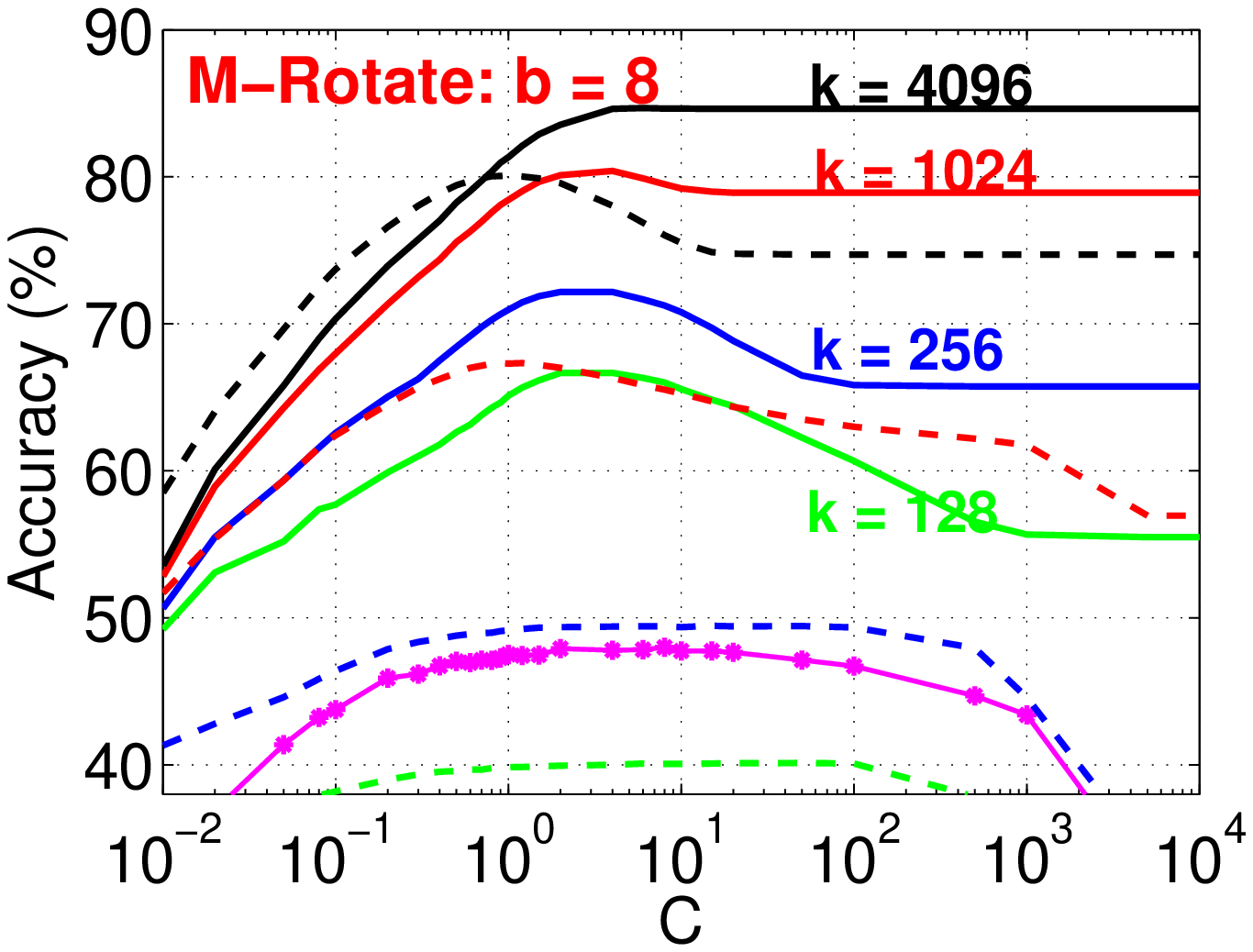}
}

\mbox{
\includegraphics[width=2.7in]{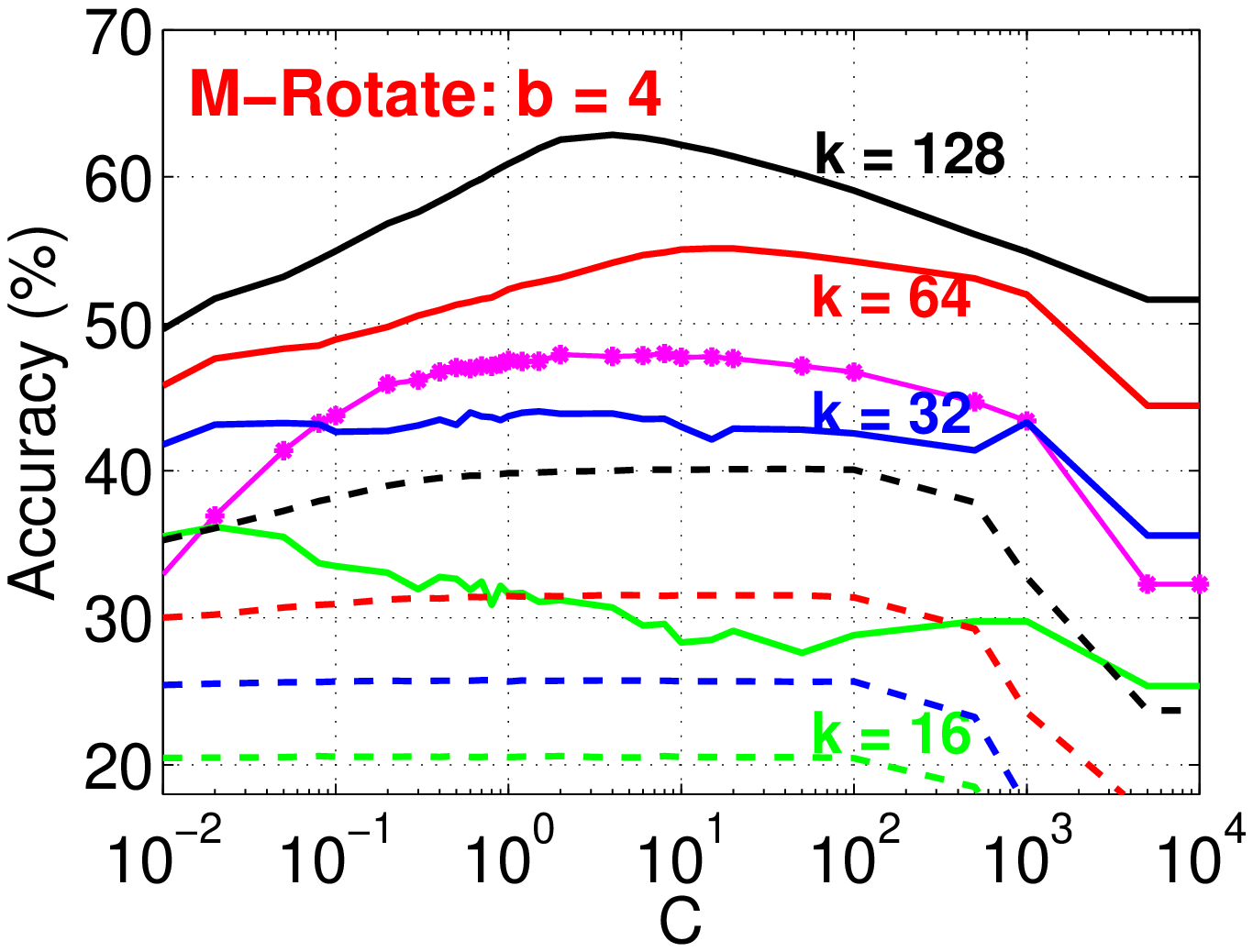}
\includegraphics[width=2.7in]{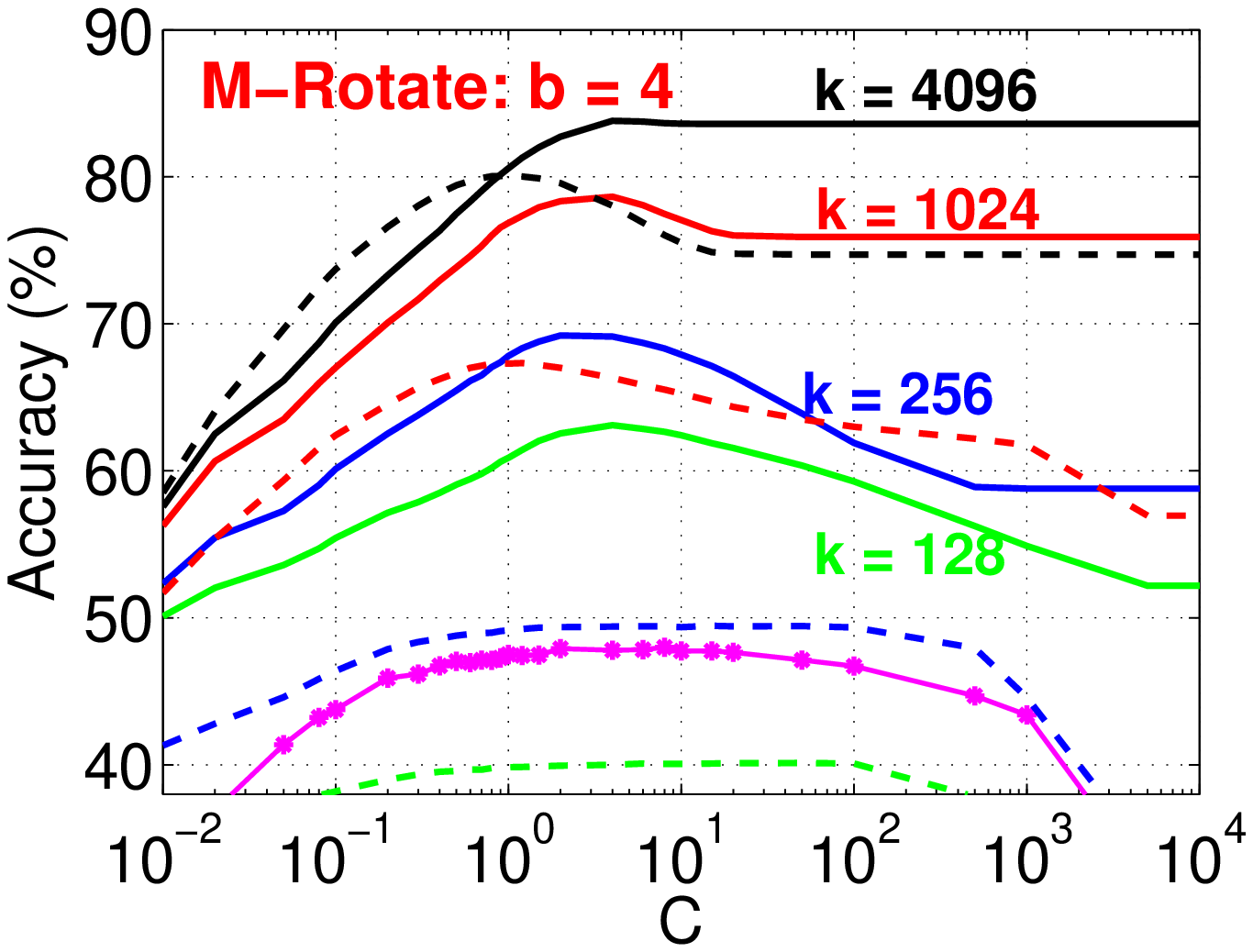}
}

\mbox{
\includegraphics[width=2.7in]{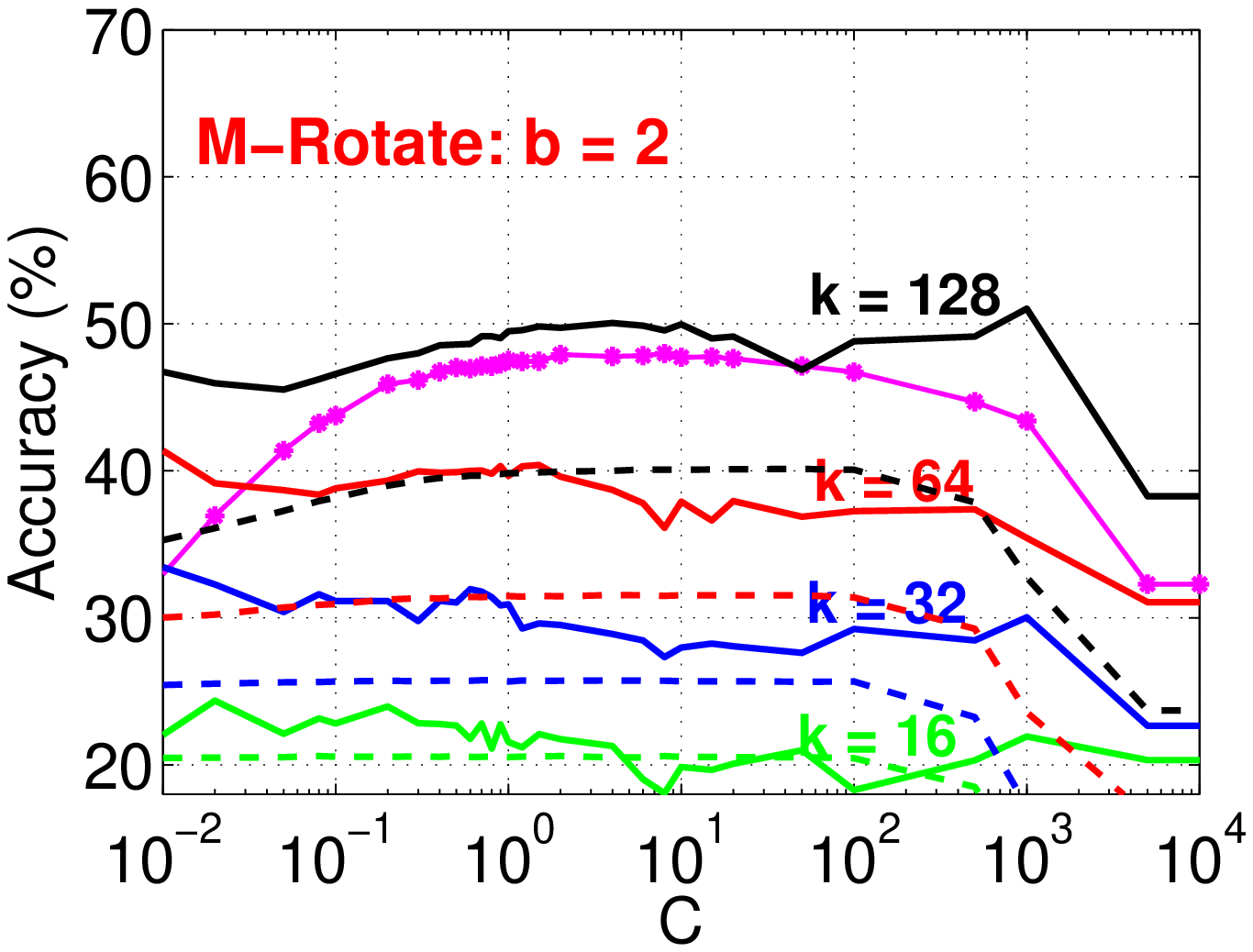}
\includegraphics[width=2.7in]{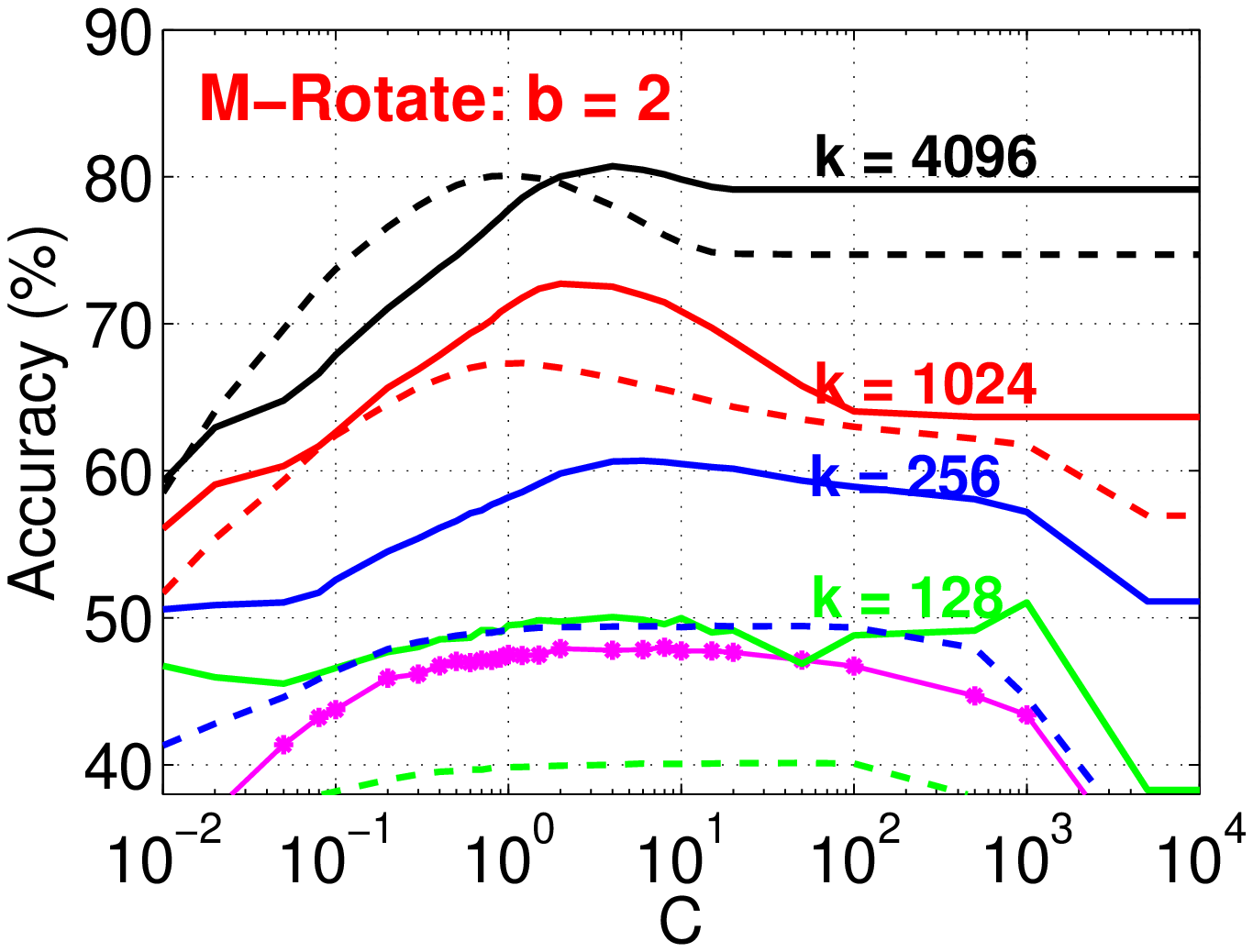}
}

\end{center}
\vspace{-0.3in}
\caption{
\textbf{M-Rotate}: Test classification accuracies of the linearized GMM kernel  (solid) and  linearized RBF kernel (dashed) , using LIBLINEAR. Again, we can see that the linearized RBF would require substantially more samples in order to reach the same accuracies as the linearized GMM kernel. For M-Rotate, the original RBF kernel actually outperforms the original GMM kernel as shown in Table~\ref{tab_MNIST}.
}\label{fig_M-Rotate}
\end{figure}

Figure~\ref{fig_SEMG1} and Figure~\ref{fig_M-Rotate} report the test classification accuracies on the \textbf{SEMG1} dataset and \textbf{M-Rotate} dataset, respectively. For both datasets, the original RBF  kernel considerably outperforms the original GMM kernel. Nevertheless, NRFF hashing still needs substantially more samples than GCWS hashing on both datasets. Again, for GCWS, the results do not differ much once we use $b\geq4$. These results again confirm the advantage of GCWS hashing.

\vspace{0.08in}

\begin{figure}
\begin{center}

\mbox{
\includegraphics[width=2.7in]{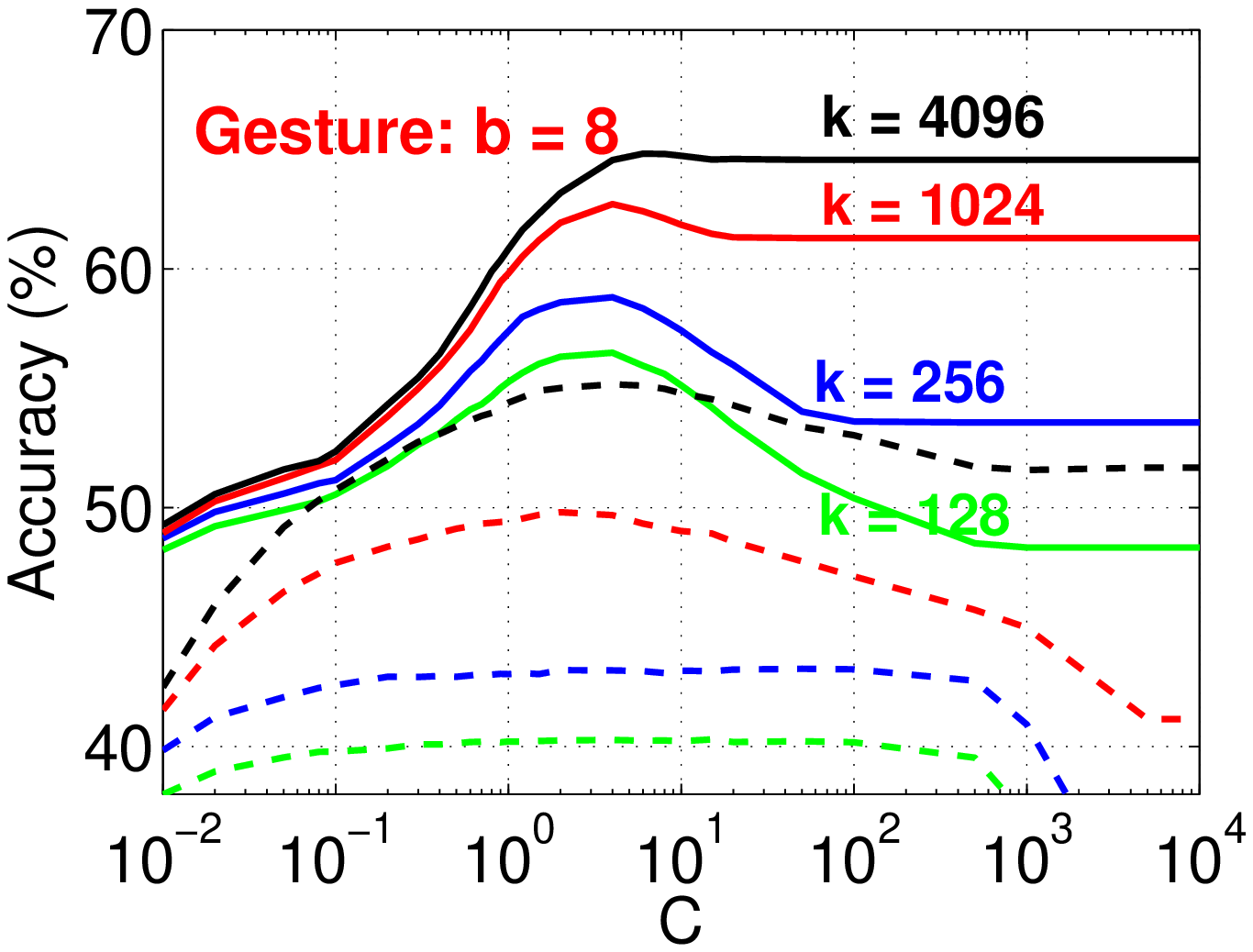}
\includegraphics[width=2.7in]{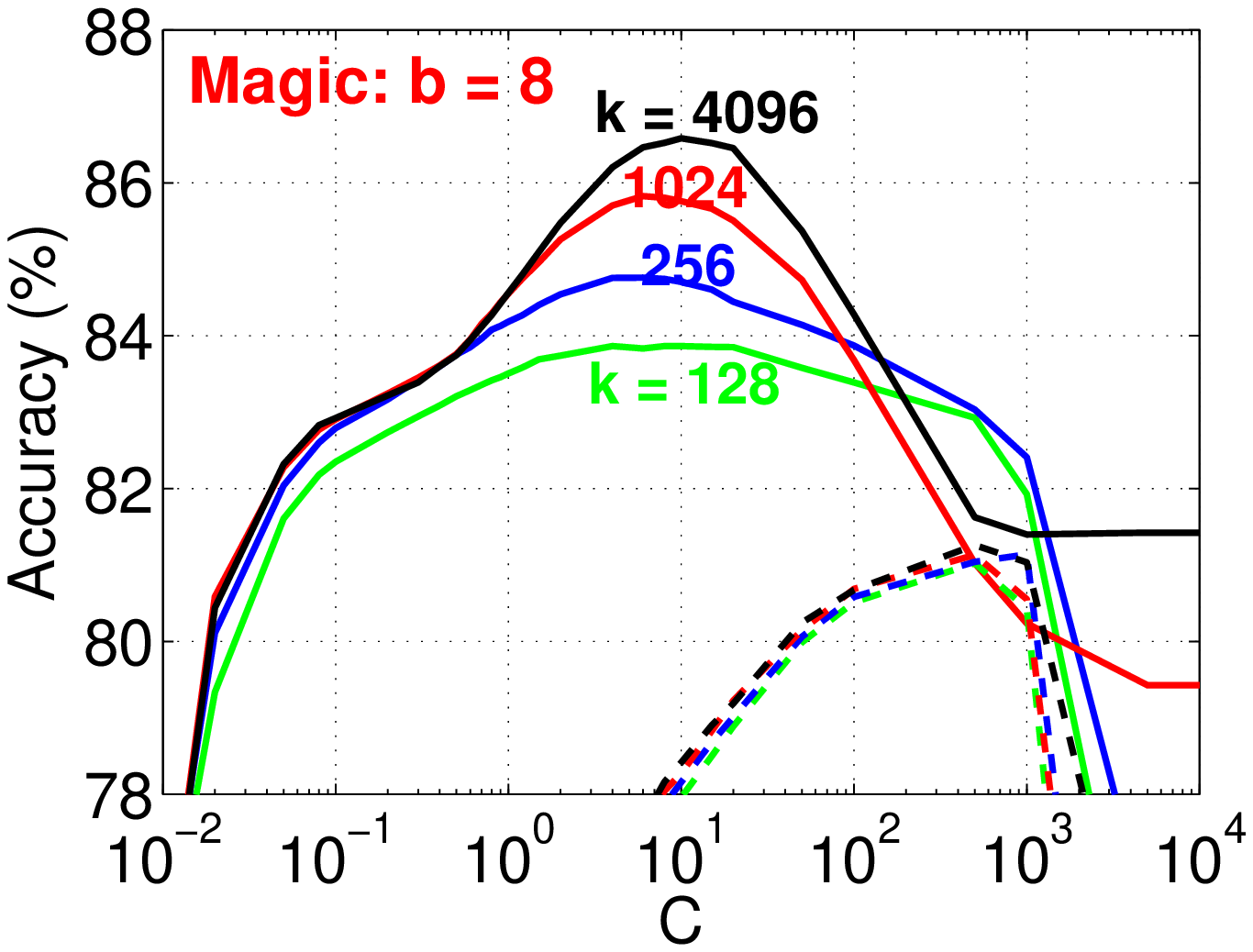}
}

\mbox{
\includegraphics[width=2.7in]{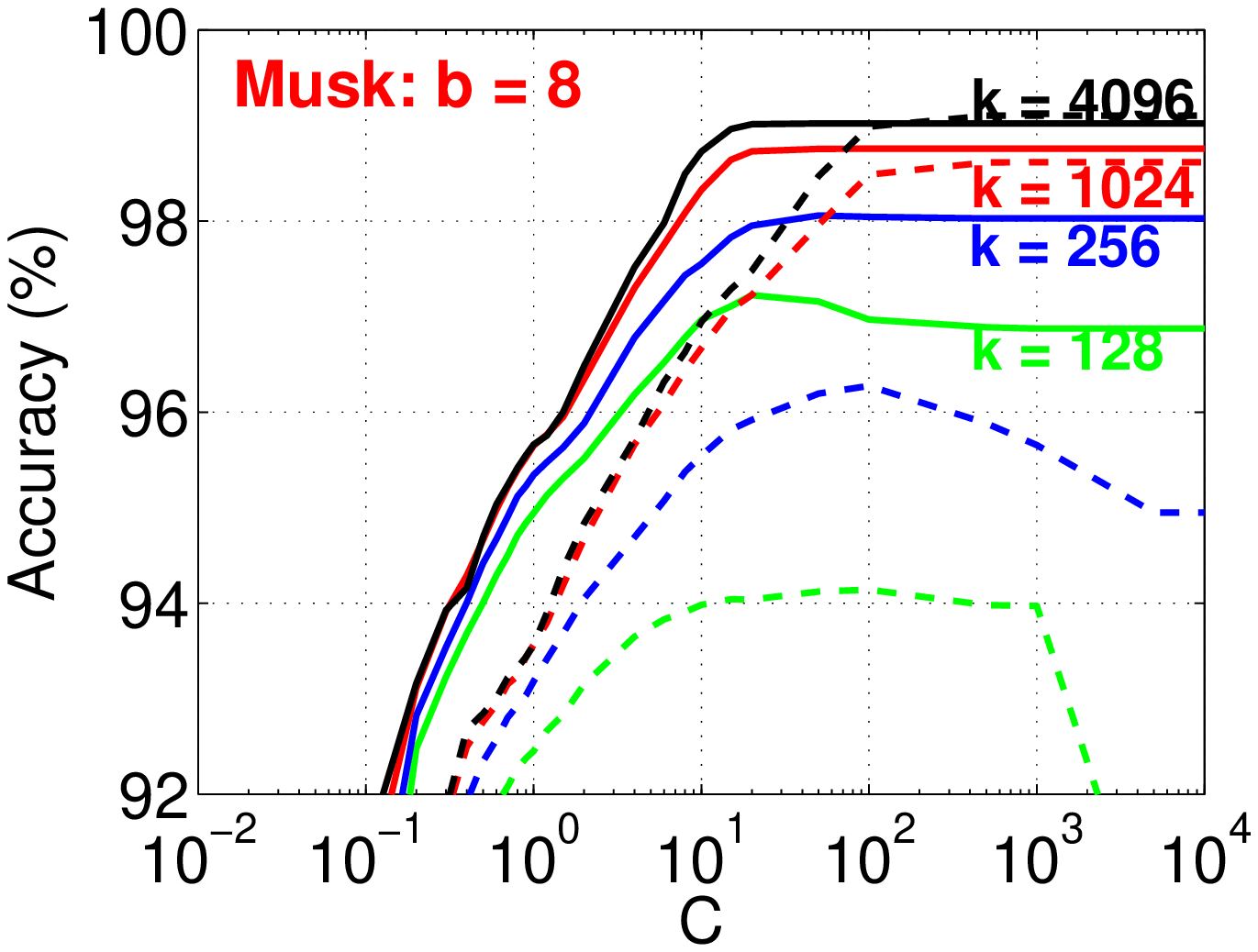}
\includegraphics[width=2.7in]{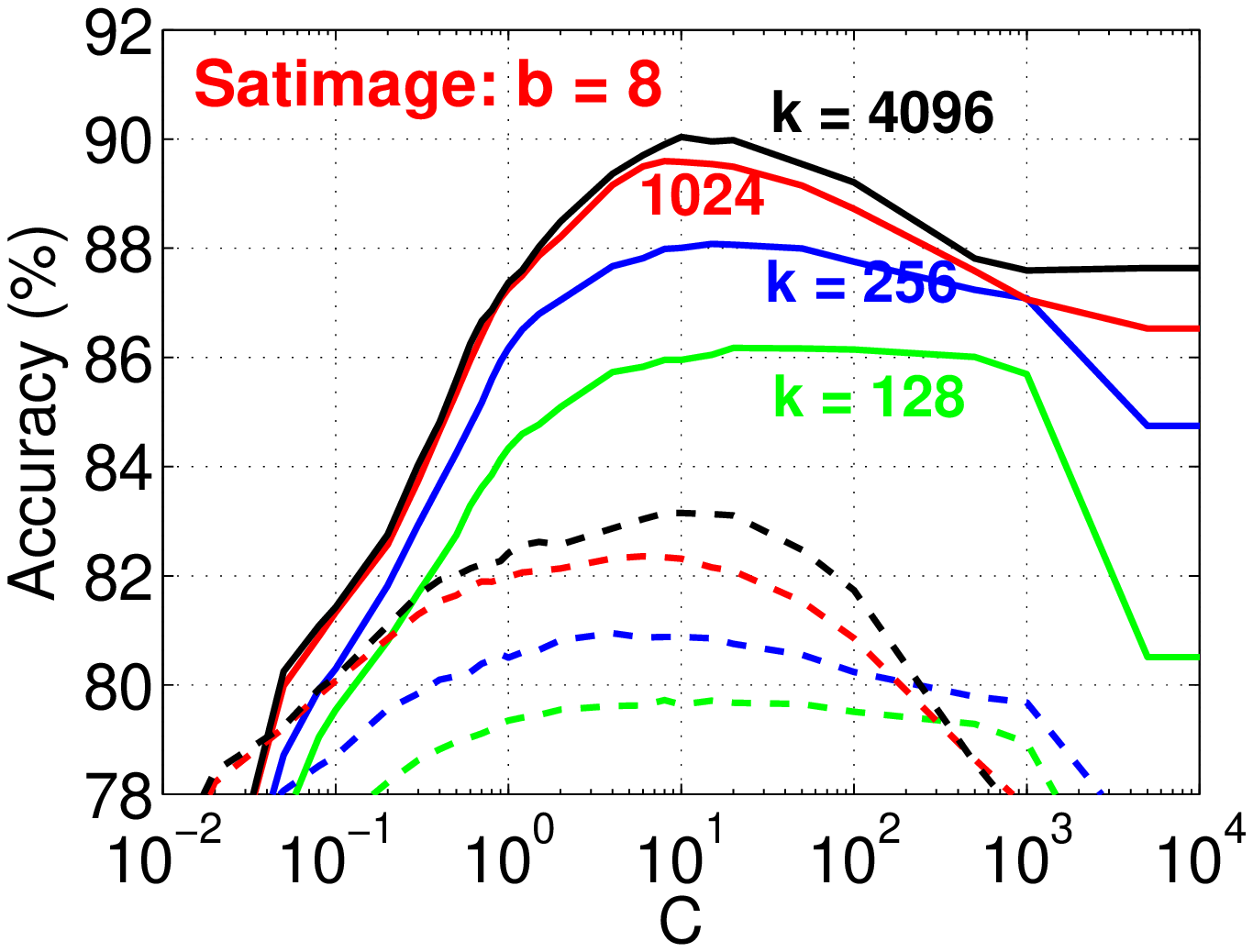}
}

\mbox{
\includegraphics[width=2.7in]{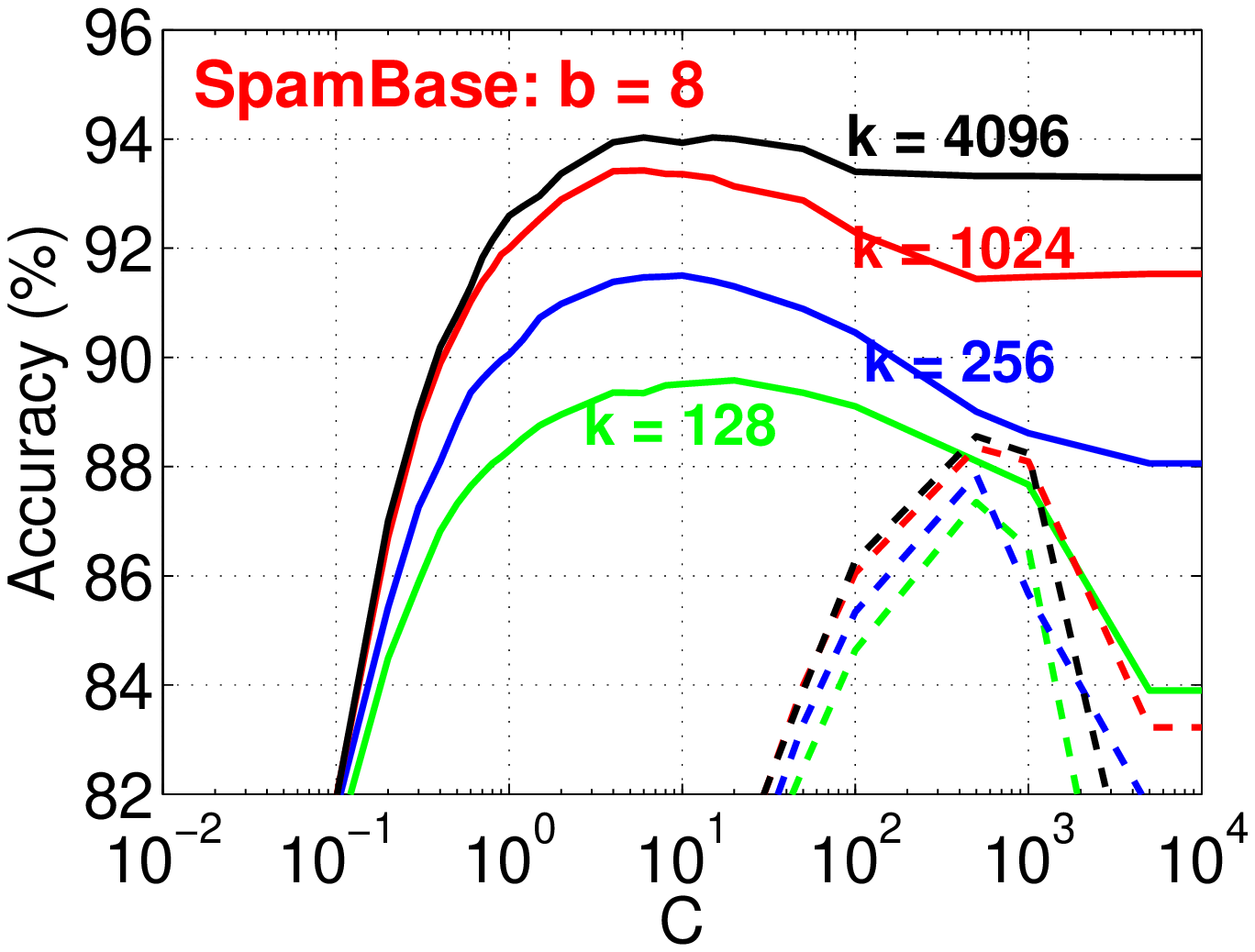}
\includegraphics[width=2.7in]{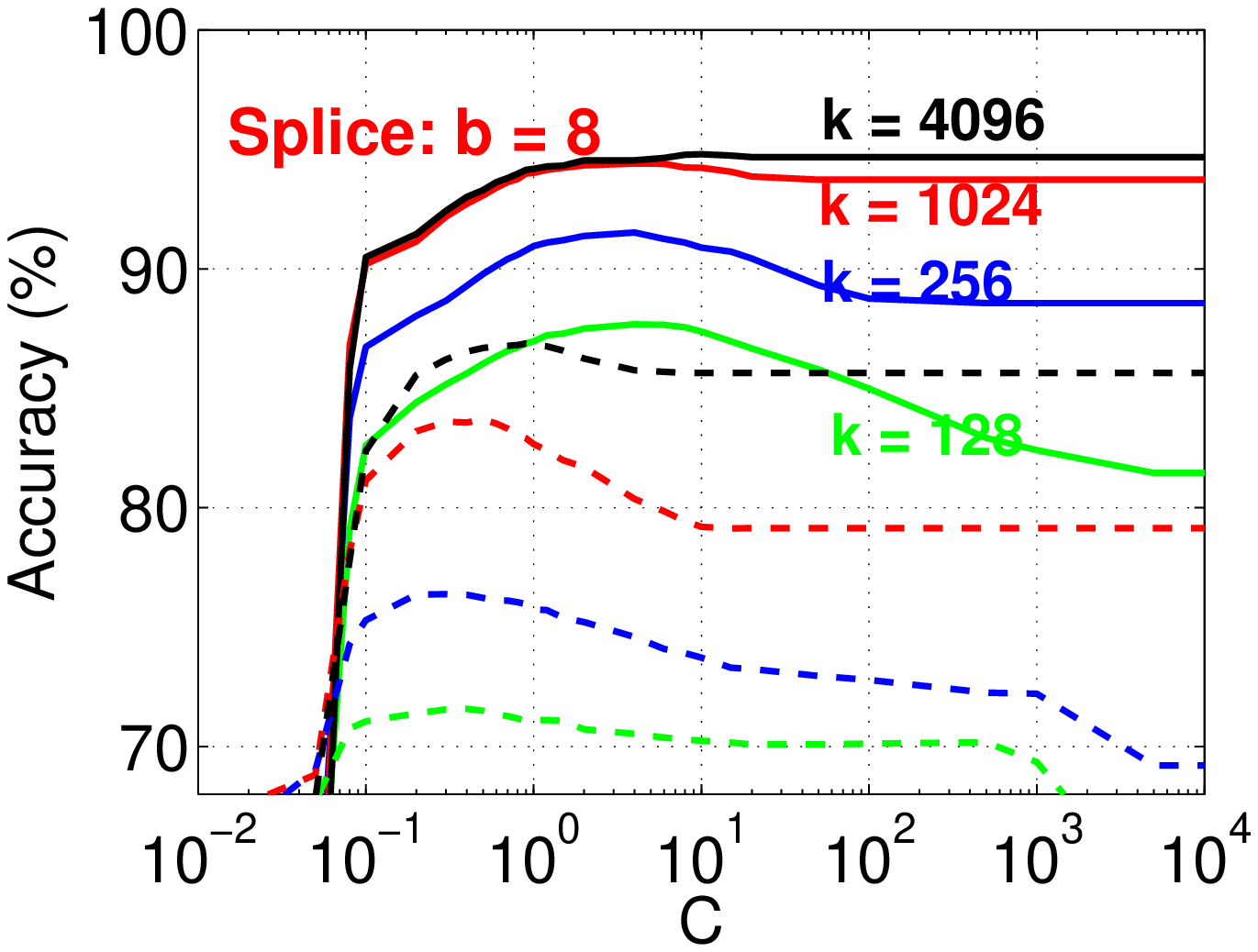}
}

\mbox{
\includegraphics[width=2.7in]{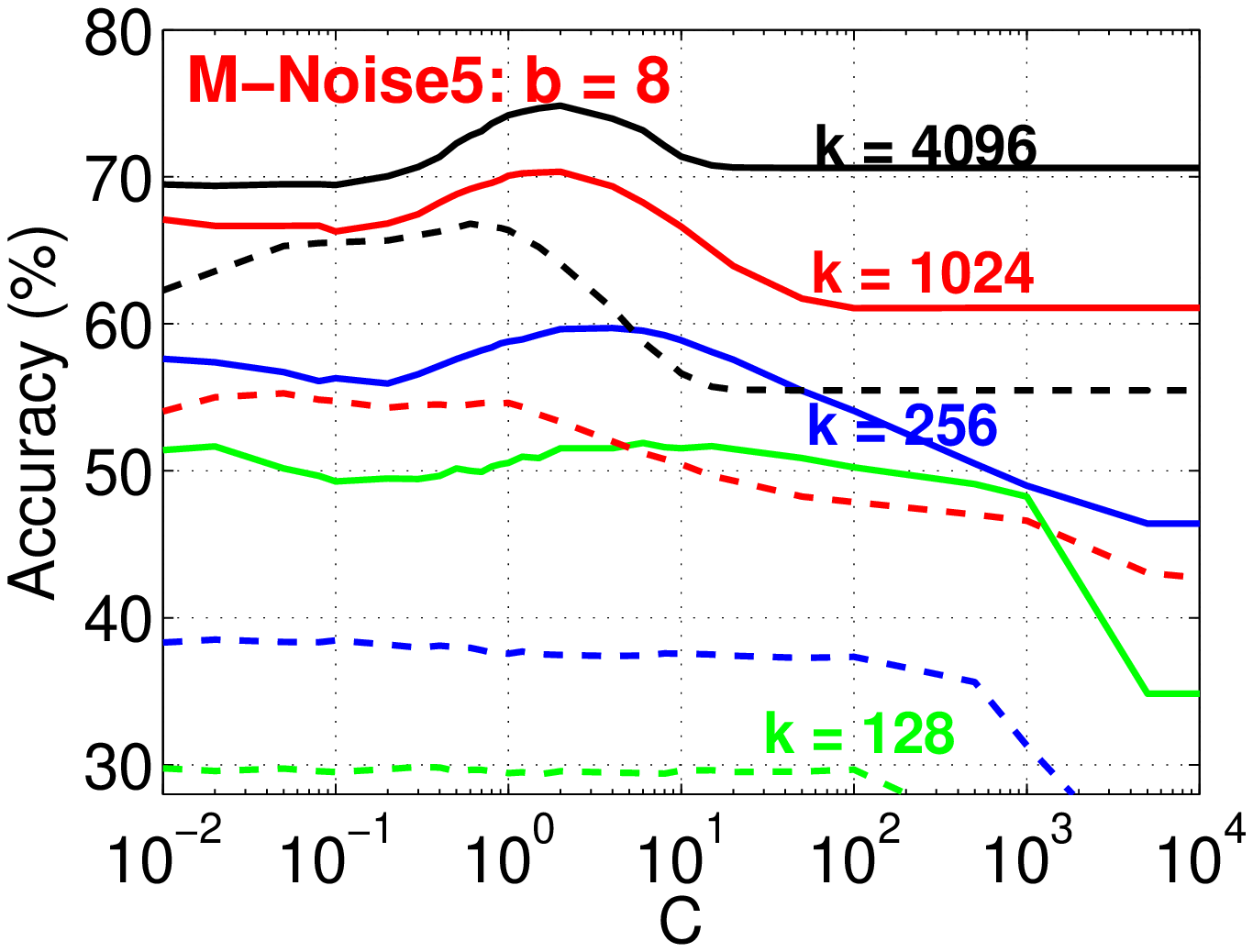}
\includegraphics[width=2.7in]{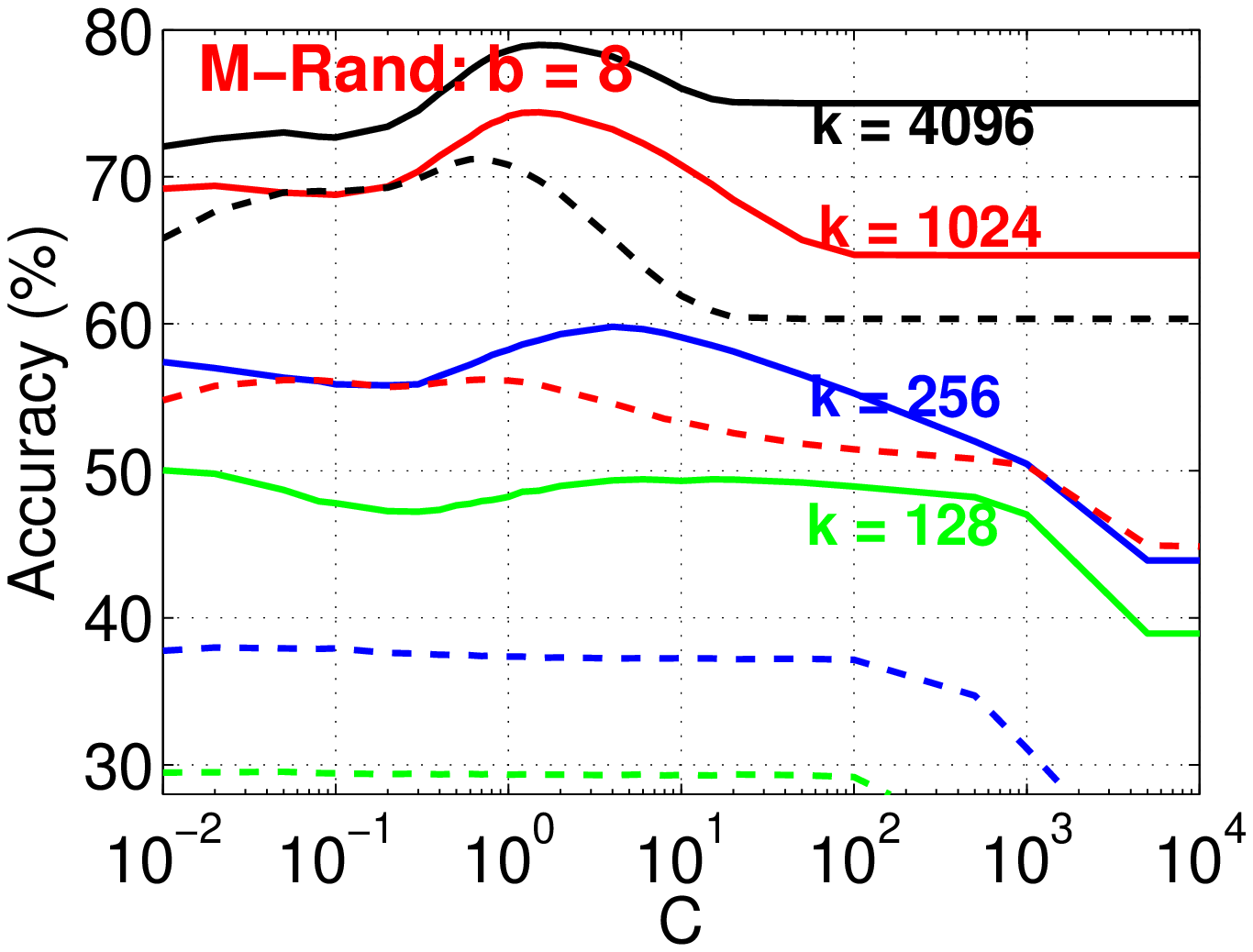}
}

\end{center}
\vspace{-0.3in}
\caption{
\textbf{More Datasets}: Test classification accuracies of the linearized GMM kernel  (solid) and  linearized RBF kernel (dashed) , using LIBLINEAR. Typically, the linearized RBF would require substantially more samples in order to reach the same accuracies as the linearized GMM kernel.
}\label{fig_OtherData}
\end{figure}

\begin{figure}[h!]
\begin{center}

\mbox{
\includegraphics[width=3in]{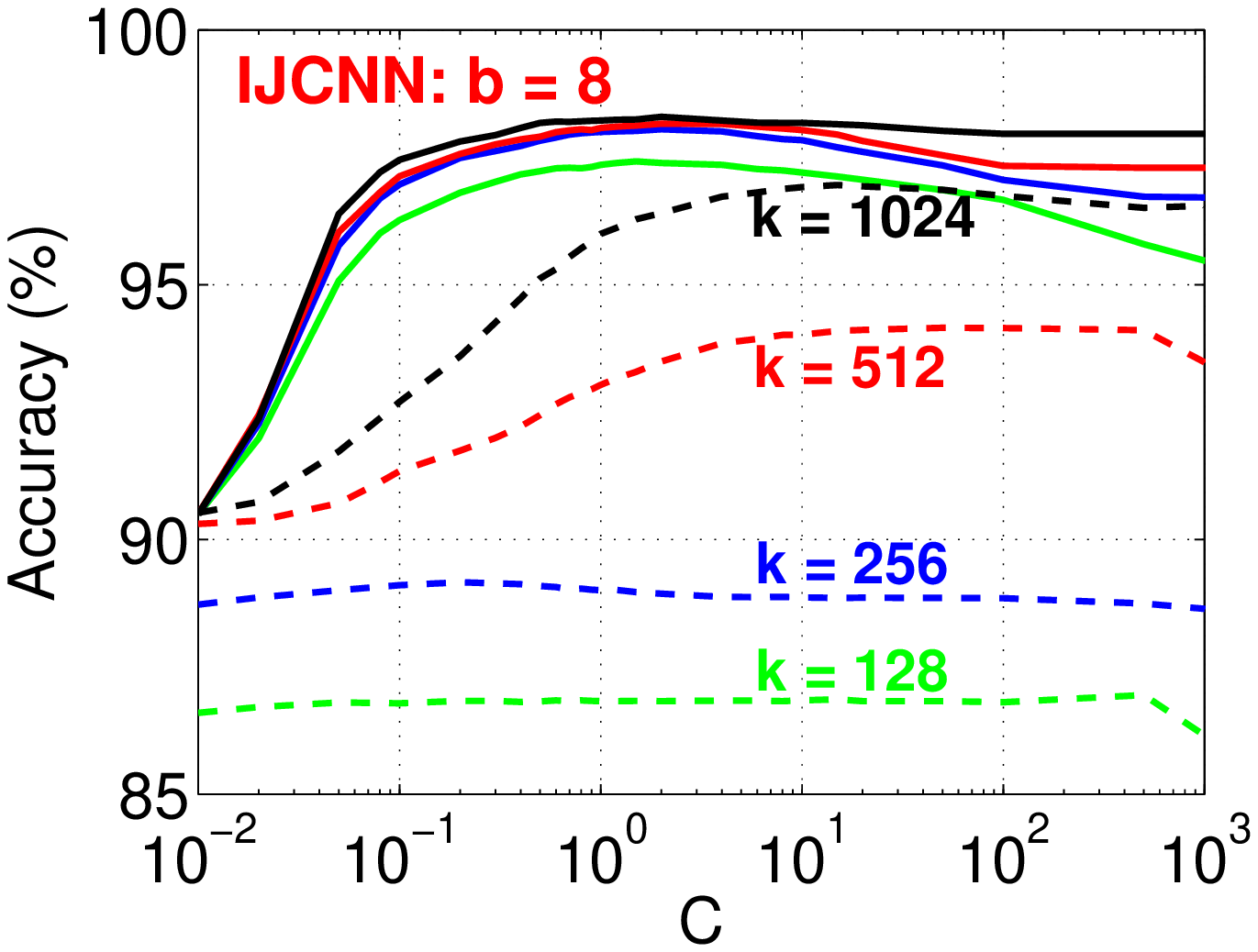}
\includegraphics[width=3in]{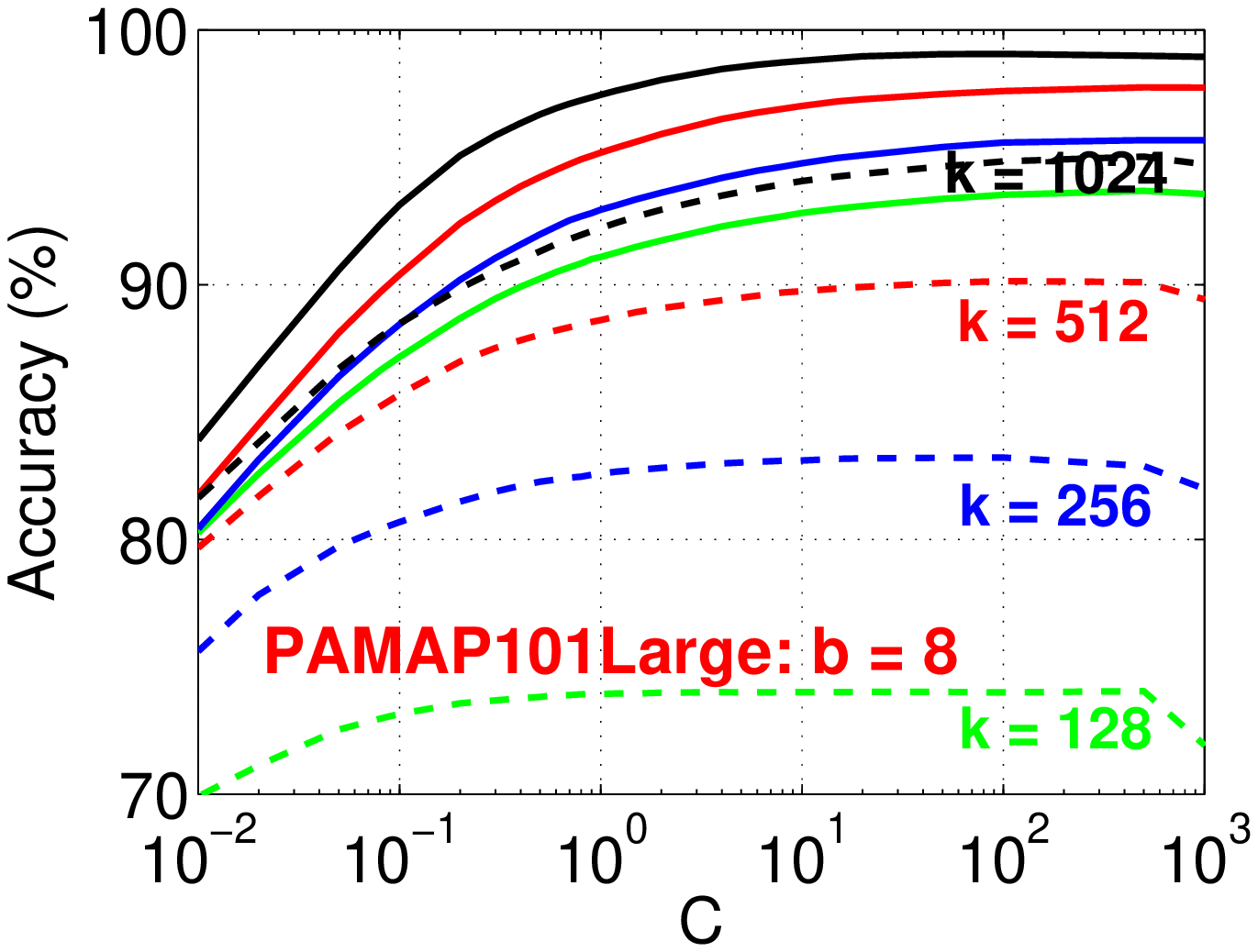}
}

\mbox{
\includegraphics[width=3in]{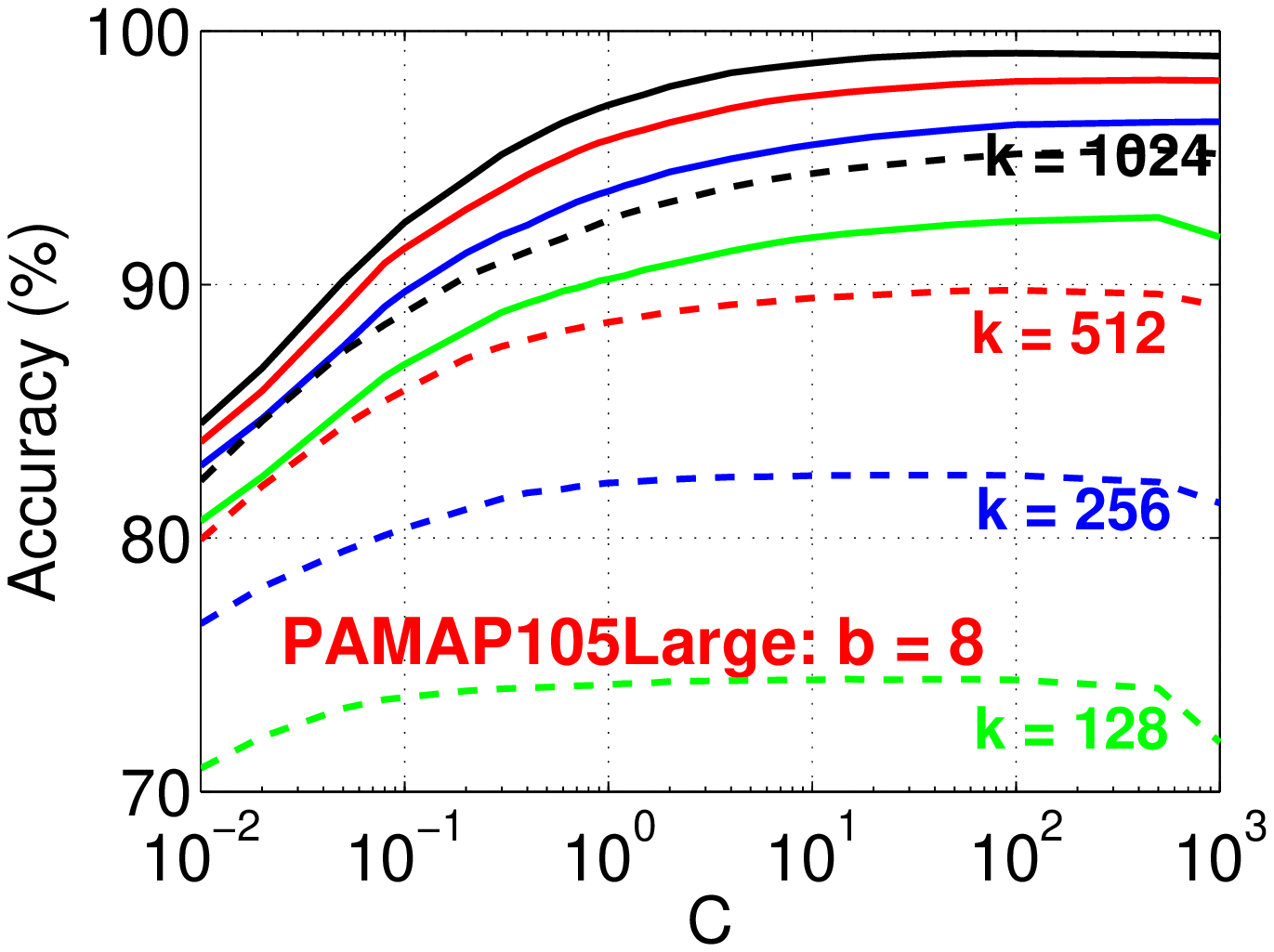}
\includegraphics[width=3in]{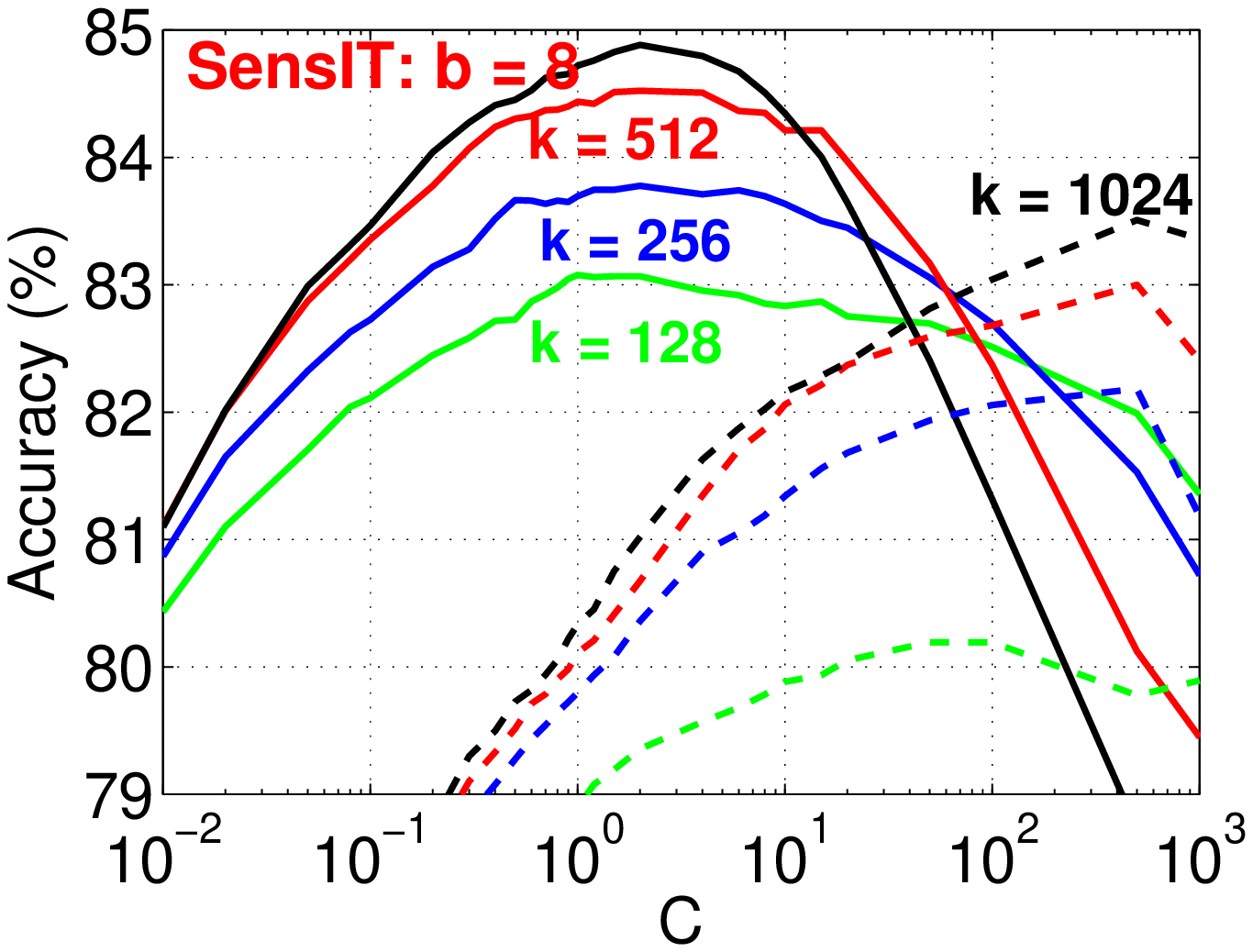}
}

\mbox{
\includegraphics[width=3in]{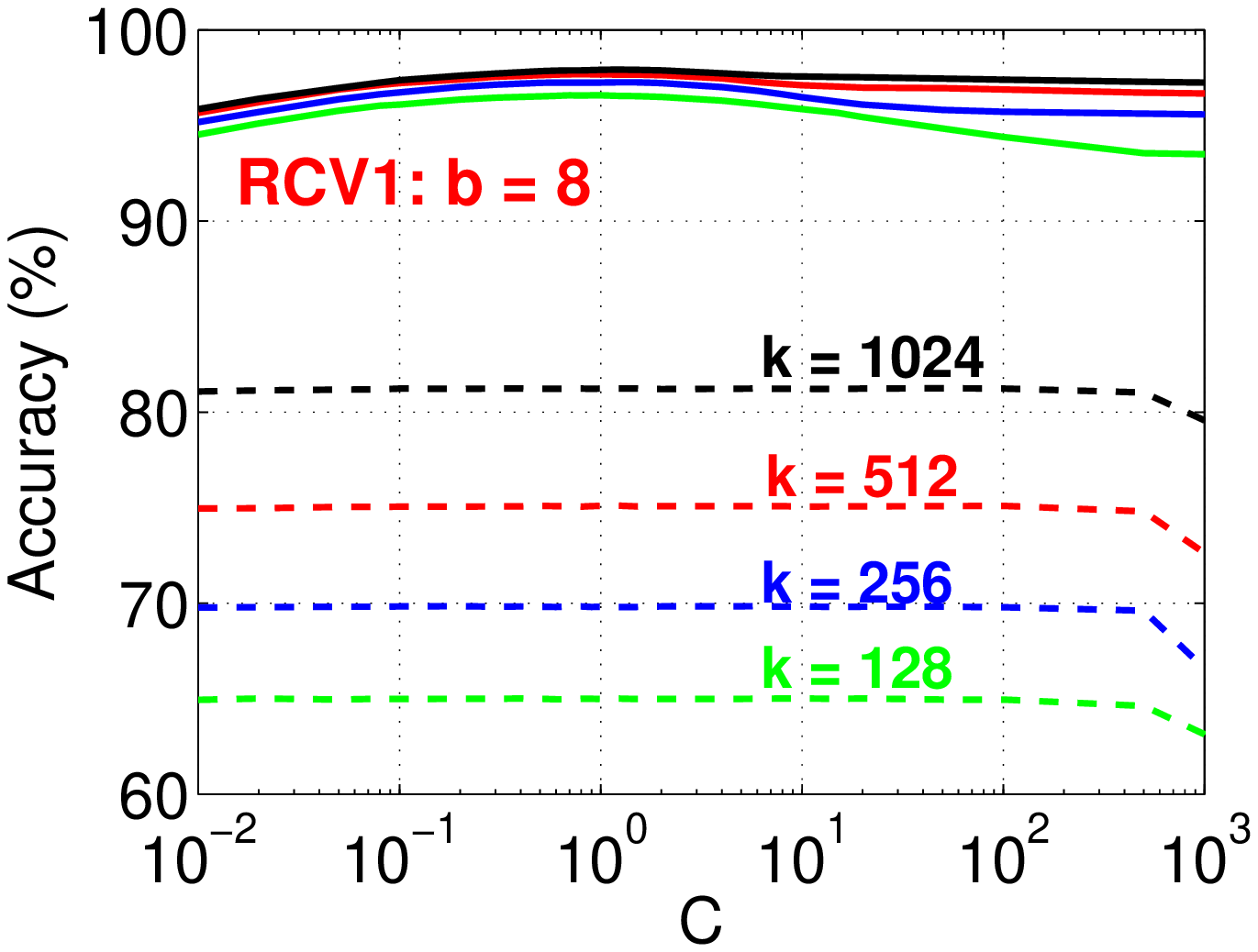}
\includegraphics[width=3in]{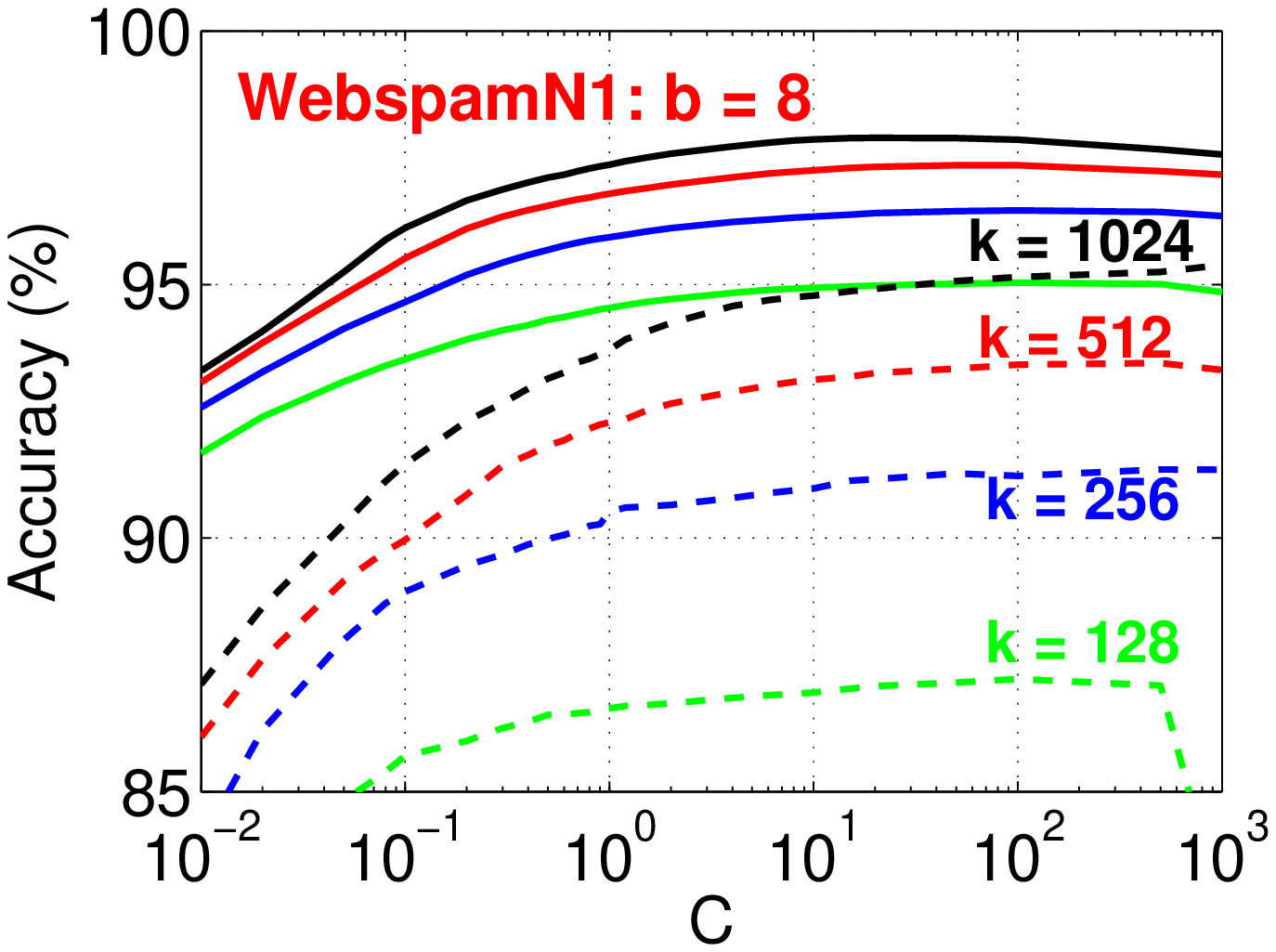}
}

\end{center}
\vspace{-0.3in}
\caption{
\textbf{Larger Datasets}: Test classification accuracies of the linearized GMM kernel with GCWS  (solid) and  linearized RBF kernel with NRFF (dashed), using LIBLINEAR, on 6 larger datasets which we can not directly compute kernel SVM classifiers. The experiments again confirm that GCWS hashing is substantially more accurate than NRFF hashing at the same sample size $k$.
}\label{fig_Larger}
\end{figure}

Figure~\ref{fig_OtherData} reports the test classification accuracies on more datasets, only for $b=8$ and $k\geq128$. Figure~\ref{fig_Larger} presents the hashing results on 6 larger datasets for which we can not directly train kernel SVMs. We report only for $b=8$ and $k$ up to 1204. All these results confirm that  linearization via GCWS works well for the GMM kernel. In contrast, the normalized random Fourier feature (NRFF) approach typically requires substantially more samples (i.e., much larger $k$). This phenomenon can  be largely explained by the theoretical results in Theorem~\ref{thm_NRFF} and Theorem~\ref{thm_g}, which conclude that GCWS hashing is more (considerably) accurate than NRFF hashing, unless the similarity is high. At high similarity, the variances of both hashing methods become very small.

\vspace{0.08in}

We should mention that the original (tuning-free) GMM kernel can be modified by introducing tuning parameters. The original GCWS algorithm can be slightly modified to linearize the new (and tunable) GMM kernel. As shown in~\cite{Report:Li_epGMM17}, on many datasets, the tunable GMM kernel can be a strong competitor compared to computationally expensive algorithms such as deep nets or trees.

\newpage\clearpage

\newpage\clearpage

\newpage\clearpage

\newpage\clearpage

\begin{figure}[h!]
\begin{center}

\mbox{
\includegraphics[width=2.3in]{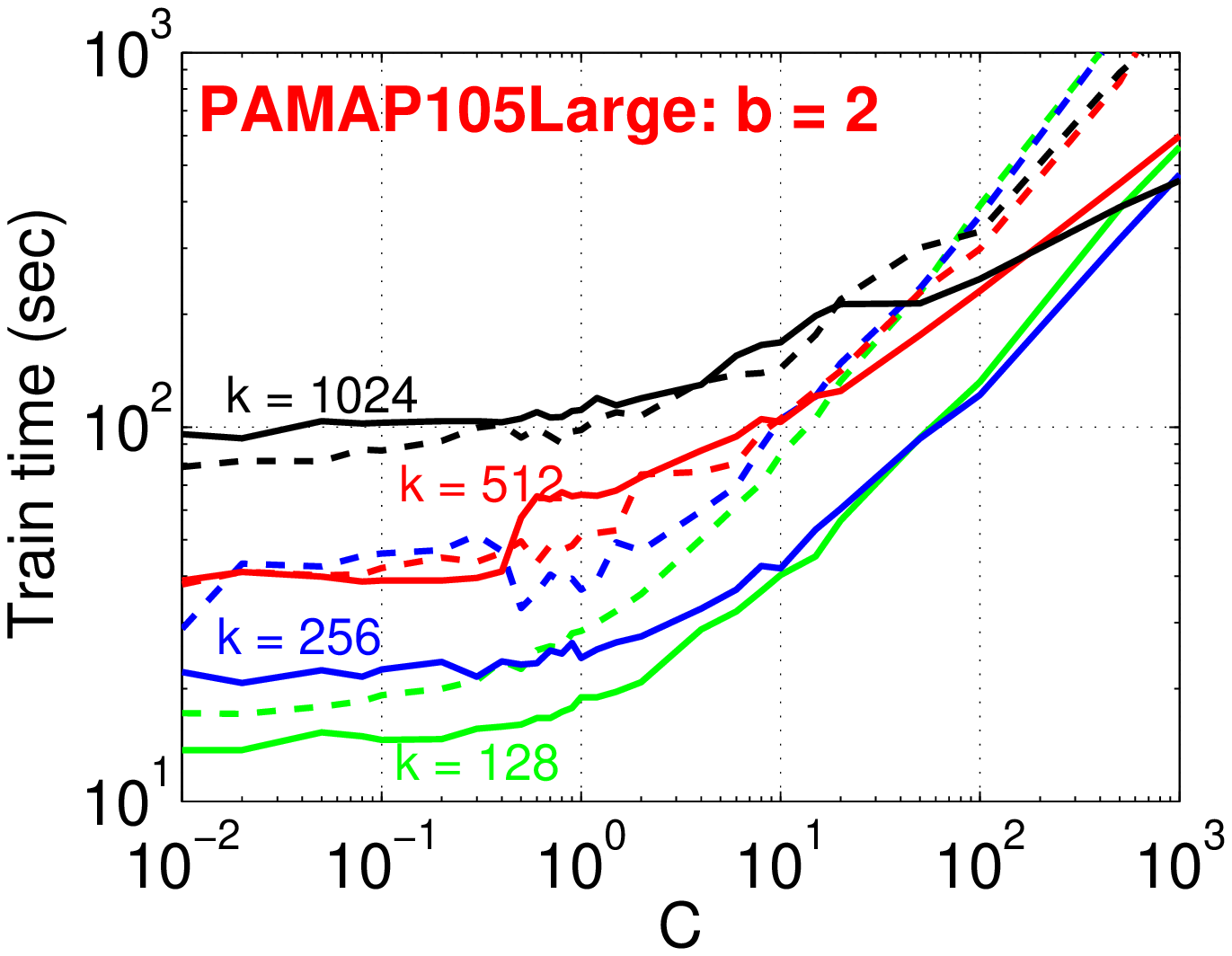}\hspace{-0.15in}
\includegraphics[width=2.3in]{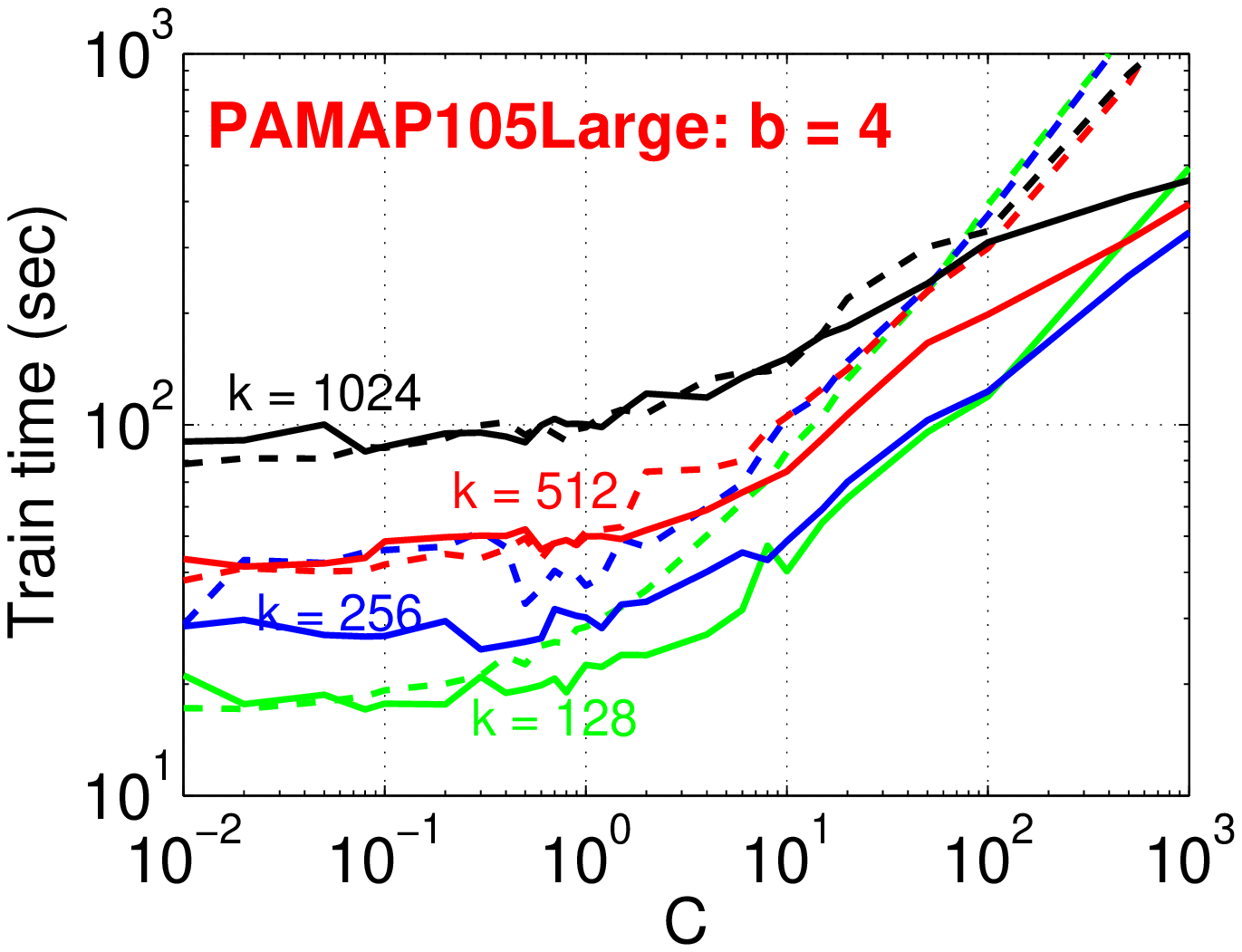}\hspace{-0.15in}
\includegraphics[width=2.3in]{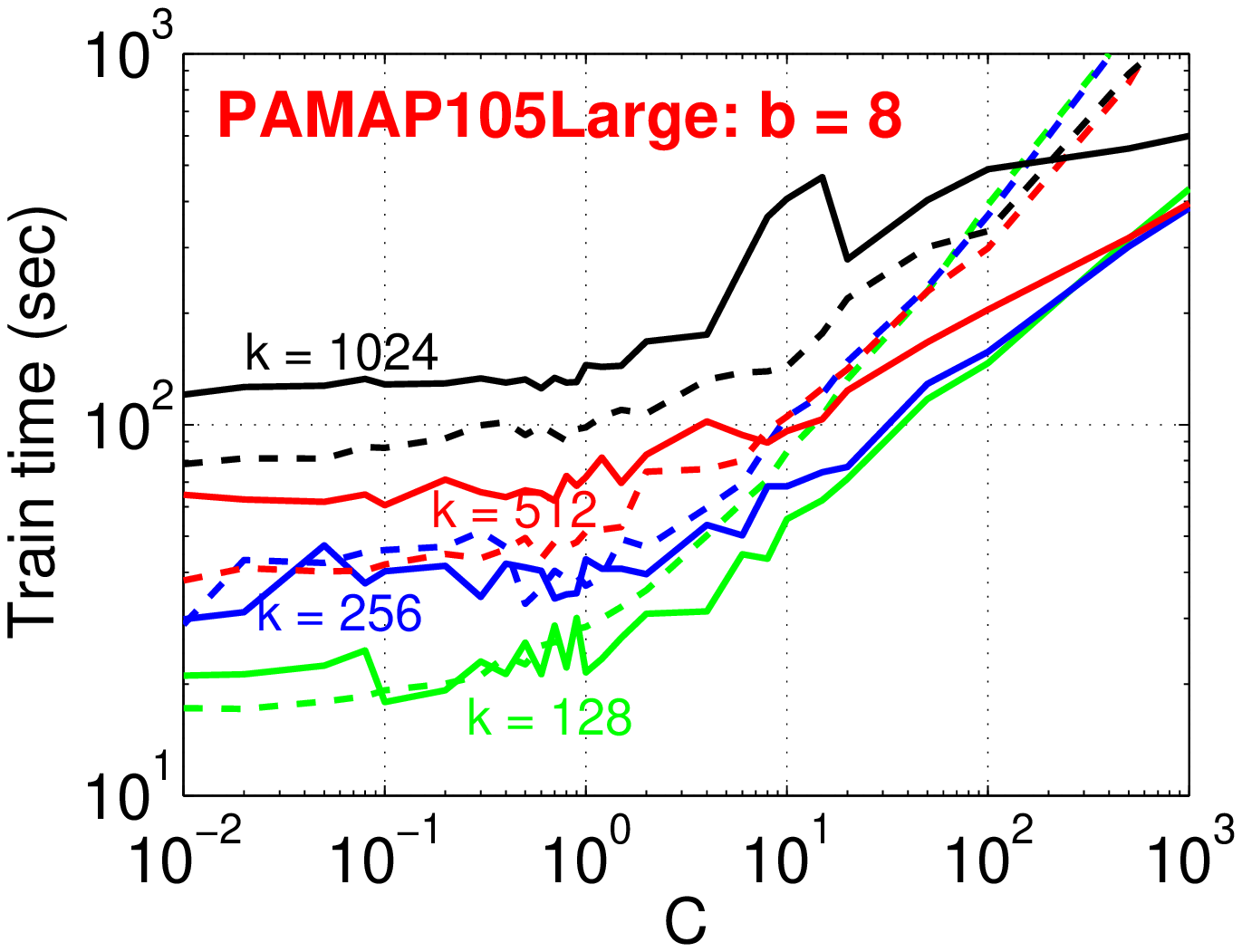}
}

\end{center}
\vspace{-0.3in}
\caption{
\textbf{PAMAP105Large}: Training times of GCWS (solid curves) and NRFF (dashed curves), for four sample sizes $k\in\{128, 256, 512, 1024\}$, and $b\in\{2, 4,8\}$.
}\label{fig_TrainTime}
\end{figure}

\noindent\textbf{Training time:}\hspace{0.1in} For linear algorithms, the training cost is largely determined by the number of nonzero entries per input data vector. In other words, at the same $k$, the training times of GCWS and NRFF will be roughly comparable. For GCWS and batch algorithms (such as LIBINEAR), a larger $b$ will  increase the training time but not much. See Figure~\ref{fig_TrainTime} for an example, which actually shows that NRFF will consume more time at high $C$ (for achieving a good accuracy). Note that, with online learning, it would be more obvious that the training time is determined by the number of nonzeros and number of epoches. For industrial practice, typically only one epoch or a few epoches are used.

\noindent\textbf{Data storage}:\hspace{0.1in} For GCWS, the storage cost per data vector is $b\times k$ while the cost for NRFF would be $k\times$ number of bits per hashed value (which might be a large value such as 32). Therefore, at the same sample size $k$, GCWS will likely need less space to store the hashed data than NRFF.

\section{Conclusion}

Large-scale machine learning has become increasingly  important in practice. For  industrial applications, it is often the case that only linear methods are affordable. It is thus practically beneficial to have methods which can provide substantially more accurate prediction results than linear methods, with no essential increase of the computation cost. The method of ``random Fourier features'' (RFF) has been a  popular tool for linearizing the radial basis function (RBF) kernel, with numerous applications in machine learning, computer vision, and beyond, e.g.,~\cite{Proc:BinaryRFF_NIPS09,Proc:Yang_NIPS12,Proc:Affandi_NIPS13,Proc:Hern_NIPS14,Proc:Dai_NIPS14,Proc:Yen_NIPS14,Proc:Hsieh_NIPS14,Proc:Shah_NIPS15,Chwialkowski_NIPS15,Richard_NIPS15}. In this paper, we rigorously prove that a simple normalization step (i.e., NRFF) can substantially improve the original RFF procedure by  reducing the estimation variance. \\

\noindent In this paper, we also propose the ``generalized min-max (GMM)'' kernel as a measure of data similarity, to effectively capture data nonlinearity. The GMM kernel can be linearized via the generalized consistent weighted sampling (GCWS). Our experimental study demonstrates that usually GCWS does not need too many samples  in order to achieve good accuracies. In particular, GCWS typically requires substantially fewer samples to reach the same accuracy as the normalized random Fourier feature (NRFF) method. This is practically  important, because the training (and testing) cost and storage cost  are  determined by the number of nonzeros (which is the number of samples in NRFF or GCWS) per data vector of the dataset. The superb empirical performance of GCWS can be largely explained by our theoretical analysis that the estimation variance of GCWS is typically much smaller than the variance of NRFF (even though NRFF has improved the original RFF).\\

\noindent By incorporating tuning parameters, \cite{Report:Li_epGMM17} demonstrated that the performance of the GMM kernel and GCWS hashing can be further improved, in some datasets remarkably so.  See~\cite{Report:Li_epGMM17} for the comparisons with deep nets and trees. Lastly, we should also mention that GCWS can be naturally applied in the context of efficient near neighbor search, due to the discrete nature of the samples, while NRFF or samples from Nystrom method can not be directly used for building hash tables.


{
\bibliographystyle{abbrv}
\bibliography{../bib/mybibfile}

}

\end{document}